\documentclass[runningheads]{llncs}
\usepackage[T1]{fontenc}
\usepackage{graphicx}
\usepackage{booktabs}
\usepackage[misc]{ifsym}
\newcommand{\corr}{(\Letter)}

\usepackage{xcolor}
\usepackage{caption}
\usepackage{subcaption}
\usepackage[flushleft]{threeparttable} 
\usepackage{threeparttablex}   
\usepackage{multirow}
\usepackage{graphicx}
\usepackage{adjustbox}
\usepackage{amssymb} 
\usepackage{makecell}
\usepackage{amsmath}
\usepackage{rotating}
\usepackage{longtable}
\usepackage{todonotes}
\usepackage{hyperref}
\usepackage{acro}
\usepackage[inline]{enumitem} 

\usepackage{orcidlink}
\usepackage[resetlabels]{multibib}

\newcites{supp}{Supplementary Material References}

\acsetup{list/display=used}  
\DeclareAcronym{unet}{
  short = U-Net,
  long  = U-Net
}
\DeclareAcronym{fcbformer}{
  short = FCBFormer,
  long  = FCBFormer
}
\DeclareAcronym{hiformer}{
  short = HiFormer,
  long  = HiFormer-B
}
\DeclareAcronym{missformer}{
  short = MISSFormer,
  long  = MISSFormer
}
\DeclareAcronym{hardnet}{
  short = HarDNet,
  long  = HarDNet-DFUS
}
\DeclareAcronym{fusegnet}{
  short = FUSegNet,
  long  = FUSegNet
}
\DeclareAcronym{segformer}{
  short = SegFormer,
  long  = SegFormer-B3
}
\DeclareAcronym{segnext}{
  short = SegNeXt,
  long  = SegNeXt-L
}
\DeclareAcronym{vwmit}{
  short = VW-MiT,
  long  = VWFormer+MiT-B3
}
\DeclareAcronym{vwconv}{
  short = VW-Conv,
  long  = VWFormer+ConvNeXt-S
}
\DeclareAcronym{vwformer}{
  short = VWFormer,
  long  = VWFormer
}
\DeclareAcronym{internimage}{
  short = InternImage,
  long  = InternImage-T
}
\DeclareAcronym{internimageUperhead}{
  short = InternImage,
  long  = InternImage-T+UPerHead
}
\DeclareAcronym{transnext}{
  short = TransNeXt,
  long  = TransNeXt-Tiny
}
\DeclareAcronym{transnextUperhead}{
  short = TransNeXt,
  long  = TransNeXt-Tiny+UPerHead
}
\DeclareAcronym{uperhead}{
  short = UPerHead,
  long  = UPerHead
}
\DeclareAcronym{dfu}{
  short = DFU,
  long  = diabetic foot ulcer
}
\DeclareAcronym{sota}{
  short = SotA,
  long  = state-of-the-art
}
\DeclareAcronym{ood}{
  short = OOD,
  long  = out-of-distribution
}
\DeclareAcronym{gp}{
  short = GP,
  long  = general-purpose
}
\DeclareAcronym{cnn}{
  short = CNN,
  long  = convolutional neural network
}
\DeclareAcronym{vit}{
  short = ViT,
  long  = vision transformer 
}
\DeclareAcronym{woundambit}{
  short = \textit{WoundAmbit},
  long  = \textit{WoundAmbit}
}
\DeclareAcronym{gmac}{
  short = GMAC,
  long  = Giga Multiply-Accumulate Operation
}
\DeclareAcronym{gradcam}{
  short = Grad-CAM,
  long  = Gradient-weighted Class Activation Mapping
}
\DeclareAcronym{hcp}{
  short = HCP,
  long  = healthcare professional
}
\DeclareAcronym{ro}{
  short = RO,
  long  = reference object
}
\DeclareAcronym{cv}{
  short = CV,
  long  = cross-validation
}
\DeclareAcronym{dl}{
  short = DL,
  long  = deep learning
}
\DeclareAcronym{gt}{
  short = GT,
  long  = ground truth
}
\DeclareAcronym{dfuc22}{
  short =  DFUC’22,
  long  = DFUC’22
}
\DeclareAcronym{fuseg}{
  short =  FUSeg,
  long  = FUSeg
}
\DeclareAcronym{aruco}{
  short =  ArUco,
  long  = ArUco
}
\DeclareAcronym{pmv}{
  short =  PMV,
  long  = pixel-wise majority vote
}

\newcommand\scalemath[2]{\scalebox{#1}{\mbox{\ensuremath{\displaystyle #2}}}}

\usepackage[autolanguage]{numprint}
\npthousandsep{,} 
\npdecimalsign{.}

\begin{document}

\title{WoundAmbit:  Bridging State-of-the-Art Semantic Segmentation and Real-World Wound Care \thanks{
This version has been accepted for publication at the Applied Data Science (ADS) track of ECML PKDD 2025 after peer review, but is not the Version of Record. The Version of Record is available online at: \href{https://doi.org/10.1007/978-3-032-06118-8_17}{10.1007/978-3-032-06118-8\_17}. 
Use of this Accepted Version is subject to the publisher’s Accepted Manuscript \href{https://www.springernature.com/gp/open-research/policies/accepted-manuscript-terms}{terms of use}.
}}

\titlerunning{WoundAmbit: Benchmarking Segmentation Models for Clinical Application}

\author{Vanessa Borst\inst{1}\orcidlink{0009-0004-7123-7934} \corr \and
Timo Dittus\inst{1}\orcidlink{0009-0008-0704-1856} \and
Tassilo Dege\inst{2}\orcidlink{0000-0001-6158-9048} \and \\
Astrid Schmieder\inst{2}\orcidlink{0000-0002-6421-9699} \and
Samuel Kounev\inst{1}\orcidlink{0000-0001-9742-2063}
}

\authorrunning{V. Borst et al.}

\institute{Julius-Maximilians-University Würzburg, 97070 Würzburg, Germany \email{\{vanessa.borst,timo.dittus,samuel.kounev\}@uni-wuerzburg.de}
\and
University Hospital of Würzburg, 97070 Würzburg, Germany \\ \email{\{Dege\_T, Schmieder\_A@\}ukw.de}
}

\maketitle             

\begin{abstract}
Chronic wounds affect a large population, particularly the elderly and diabetic patients, who often exhibit limited mobility and co-existing health conditions. Automated wound monitoring via mobile image capture can reduce in-person physician visits by enabling remote tracking of wound size. Semantic segmentation is key to this process, yet wound segmentation remains underrepresented in medical imaging research.
To address this, we benchmark state-of-the-art deep learning models from general-purpose vision, medical imaging, and top methods from public wound challenges. For a fair comparison, we standardize training, data augmentation, and evaluation, conducting cross-validation to minimize partitioning bias. We also assess real-world deployment aspects, including generalization to an out-of-distribution wound dataset, computational efficiency, and interpretability.
Additionally, we propose a reference object-based approach to convert AI-generated masks into clinically relevant wound size estimates and evaluate this, along with mask quality, for the five best architectures based on physician assessments.
Overall, the transformer-based TransNeXt showed the highest levels of 
generalizability. Despite variations in inference times, all models processed at least one image per second on the CPU, which is deemed adequate for the intended application. Interpretability analysis typically revealed prominent activations in wound regions, emphasizing focus on clinically relevant features.
Expert evaluation showed high mask approval for all analyzed models, with VWFormer and ConvNeXtS backbone performing the best. Size retrieval accuracy was similar across models, and predictions closely matched expert annotations.
Finally, we demonstrate how our AI-driven wound size estimation framework, \acs{woundambit}, is integrated into a custom telehealth system. 
Our code and supplementary material are available on \href{https://github.com/VanessaBorst/woundambit}{GitHub} and \href{https://zenodo.org/records/15673941}{Zenodo}, respectively. 

\end{abstract}

\section{Introduction}
\label{sec:intro}
%
%
%
%
Wound care is a critical aspect of healthcare, particularly for chronic wounds that require ongoing monitoring and treatment.
Current clinical practice often relies on manual wound measurements, such as estimating wound area by multiplying its longest length by its largest perpendicular width using a ruler or metric tape---although this often leads to overestimation due to irregular wound shapes~\cite{shaw2011wound}. 
Langemo et al. further reported differing interpretations among clinicians regarding how to define and measure wound length and width~\cite{langemo2008measuring}. This lack of standardization contributes to measurement subjectivity and may limit comparability between assessments.
An alternative approach involves tracing the wound on a transparent film and estimating the area using a metric grid, which may offer improved measurement reliability but still suffers from subjectivity in boundary delineation and partial cell interpretation~\cite{shaw2011wound,langemo2008measuring}.
Beyond potential inconsistencies, manual assessments are often invasive, relying on either proximity to or direct physical contact with the wound. This may cause patient discomfort and requires in-person visits with \acp{hcp}, posing logistical challenges for both patients and providers.
Automated wound segmentation offers a promising alternative, as AI-driven wound size estimation from RGB images can be integrated into various systems, particularly mobile phones with cameras. This enhances accessibility, enabling remote wound monitoring from home without requiring specialized hardware. Unlike tracing methods, AI-based approaches are non-invasive, eliminating direct wound contact.
%
%
%

This shift toward automation aligns with broader advancements in AI-driven semantic segmentation, which have mainly been shaped by the evolution of \acp{cnn} and the emergence of \acp{vit}. Thisanke et al.~\cite{thisanke2023semantic} review modern transformer-based approaches, whereas Minaee et al.~\cite{minaee2021image} focus on \ac{dl} for image segmentation excluding \acp{vit}. Recently, (multi-modal) vision foundation models, such as Segment Anything~\cite{kirillov2023segment}, have gained attention due to their large-scale pretraining and versatility across different downstream tasks~\cite{awais2025foundation}. 
Medical image segmentation has also transitioned towards \ac{dl}. Azad et al. present a comprehensive review of recent advances using \acp{vit}~\cite{azad2024advances}, while Rayed et al.~\cite{rayed2024deep} review \ac{dl} approaches more broadly.
Additionally, foundation models like Segment Anything have been adapted for medical imaging, as demonstrated by MedSAM~\cite{ma2024segment}.

In contrast, wound segmentation remains relatively underexplored, partly due to the scarcity of relevant datasets. Notable exceptions with more than \numprint{1000} wound images include \acs{fuseg}~\cite{wang2024fuseg} and DFUC~\cite{kendrick2022translating}, both of which provide annotated images of \ac{dfu} for public challenges. Additionally, Oota et al.~\cite{oota2023wsnet} provide a dataset with \numprint{2686} images of eight wound types, where annotations extend beyond the wound to include peri-wound skin areas.
Regarding existing wound segmentation techniques, early methods focused primarily on traditional feature engineering-based machine learning~\cite{wang2016area}. Over time, \ac{dl} approaches such as WSNet~\cite{oota2023wsnet}, FUSegNet~\cite{dhar2024fusegnet} and other CNN-based techniques~\cite{wang2020fully,liao2022hardnet,liu2017framework,chino2020segmenting,goyal2017fully} have emerged, alongside a few approaches for interactive wound segmentation~\cite{zhang2023interactive}. 
However, certain limitations remain:

%
%
%
%
\textbf{1. Limited Adoption of SotA Vision Models for \mbox{Wound Analysis:}}
Despite advancements in computer vision, investigations into the suitability of \ac{sota} models, particularly \acp{vit}, for wound segmentation remain limited. A 2022 survey on \ac{dl} for wound analysis~\cite{zhang2022survey} did not mention any transformer-based segmentation methods, and recent breakthroughs in \ac{gp} vision models have yet to be applied to wound analysis.

\indent \textbf{2. Lack of Consideration for Practical Deployment:}
Few studies evaluate efficiency metrics such as GMACs/FLOPs alongside segmentation performance~\cite{wang2020fully,oota2023wsnet}, and few report inference times or latency~\cite{zhang2023interactive,oota2023wsnet}. However, computational complexity is a critical factor for resource-constrained settings such as the medical sector, where GPU availability is limited. 
With few exceptions, such as the visualization of feature maps from different layers~\cite{zhang2023interactive},  explainable AI techniques are rarely applied despite their potential to improve trust among clinicians.
Lastly, model generalizability to \ac{ood} data is often overlooked, even though assessing robustness is crucial due to variability in wound types, skin tones, and uncontrolled home-based monitoring conditions such as lighting, background, and hardware.

\indent \textbf{3. Gap Between Segmentation and Clinically Relevant Wound Size:}
Existing public challenges (\acs{fuseg}, DFUC) and most wound segmentation approaches~\cite{dhar2024fusegnet,wang2020fully,liao2022hardnet,liu2017framework,goyal2017fully} do not address the conversion of segmentation masks into real-world wound size. 
Exceptions include Wang et al.~\cite{wang2016area}, who rely on a specialized imaging box, and Chairat et al.~\cite{chairat2023ai}, who use a custom calibration chart with a \acs{unet}-based model incorporating EfficientNet/MobileNetv2 encoders for wound segmentation---but do not assess size retrieval accuracy.
Chino et al.~\cite{chino2020segmenting} incorporate measurement ruler and tape detection as a third class in a \ac{unet}-based model and combine it with pixel density estimation to determine wound area. Similarly, Foltynski and Ladyzynski~\cite{foltynski2023internet} train a CNN to detect both wounds and dual calibration markers that need to be placed below and above the wound. Proprietary solutions like Swift Skin \& Wound~\cite{ramachandram2022fully} and imitoWound~\cite{imitoWound} exist but provide little insight into their calibration methods and algorithms.

\begin{figure}[ht]
    \centering
    \includegraphics[trim={0 40 0 0}, clip, width=\textwidth]{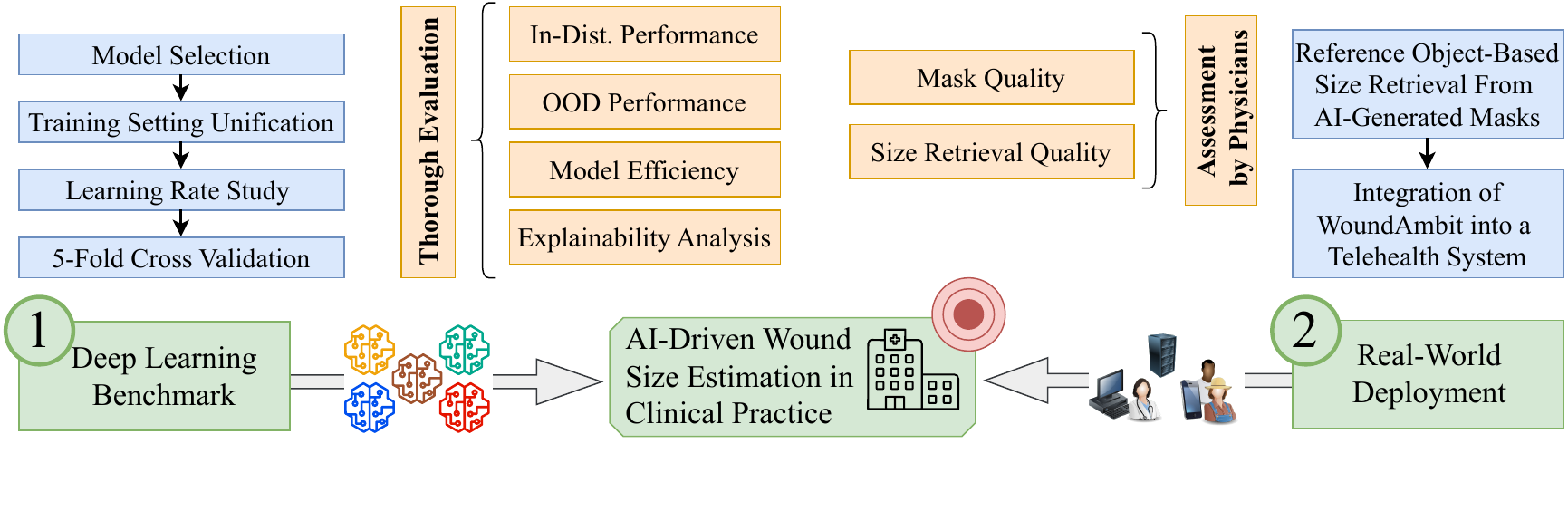}
    \caption{Schematic visualization of our \textit{WoundAmbit} approach.}
    \label{fig:approach:woundambit}
\end{figure}

%
%
%
%
To address these limitations, we introduce \acs{woundambit}, an end-to-end solution for automated wound size estimation from RGB images that bridges the gap between modern \ac{dl} and practical wound care. 
To the best of our knowledge, this study is the first to systematically transfer a diverse set of \ac{sota} \ac{gp} vision models to the wound domain and benchmark them against both medical and wound-specific segmentation methods within a unified evaluation framework. 
As summarized in Figure~\ref{fig:approach:woundambit}, our key contributions are two-fold: 

\textbf{1. Comprehensive \ac{dl} Benchmark:} 
We conduct a rigorous benchmarking study by systematically selecting 12 \ac{sota} \ac{dl} architectures from wound-specific, medical, and \ac{gp} vision models, covering a diverse range of design paradigms, including \acp{cnn}, \acp{vit}, and hybrid models. 
To ensure comparability, all models are trained under standardized conditions on publicly available data and evaluated using 5-fold \ac{cv}. In addition, our evaluation framework emphasizes clinically relevant properties by assessing generalizability on a dedicated \ac{ood} dataset. Specifically created for this work, the dataset comprises 343 wound images taken at various body sites. Moreover, we analyze computational efficiency, including trainable parameters, \acp{gmac}, and inference times on both GPU and CPU. Lastly, we investigate model interpretability using \ac{gradcam} visualizations to assess clinically relevant decision-making. 

\indent \textbf{2. Real-World Deployment:} 
To translate AI-generated wound masks into clinically meaningful size estimates, we develop and validate a \ac{ro}-based approach for precise wound surface measurement. Unlike existing methods that require neural networks to segment calibration stickers~\cite{chino2020segmenting,foltynski2023internet}, our approach leverages \acs{aruco} marker detection, making it independent of the AI algorithm used for segmentation. To evaluate reliability, we construct a dataset of 20 diverse wound images and obtain expert assessments of AI-generated mask quality from three dermatologists. Additionally, we compare AI-derived wound size estimates with physician annotations to quantify measurement accuracy. Finally, we propose a practical integration strategy for embedding AI-driven wound size estimation into a custom telemedicine framework for remote monitoring.

%
%
%
%
The paper is structured as follows: 
Section~\ref{sec:benchmark} describes the \ac{dl} benchmark (Fig.~\ref{fig:approach:woundambit}, left), covering model selection, methodology, datasets, and results. 
Section~\ref{sec:real_world_deployment} details the size retrieval process and its evaluation, along with the integration of \ac{woundambit} into a specially designed telehealth system (Fig.~\ref{fig:approach:woundambit}, right).
Finally, Section~\ref{sec:discussion_outlook} discusses key findings, and Section~\ref{sec:conclusion} concludes the study.
\section{Deep Learning Benchmark}
\label{sec:benchmark}

\subsection{Model Selection}
\label{ssec:benchmark:model_selection}
Given the large number of novel methods introduced in recent years, it is impractical to include all current \ac{sota} models in this study. 
To ensure a benchmarking process that is as representative and balanced as possible, we devise four categories, selecting several representatives from each of them as follows: 

\begin{enumerate*}[label=(\Roman*)] 
    \item \textit{\ac{dfu} segmentation}: To specifically address the challenges of wound segmentation (WS), we include the best-performing models from two publicly available \ac{dfu} segmentation challenges, as they are directly tailored to the domain of interest (\acl{hardnet}~\cite{liao2022hardnet}, \acl{fusegnet}~\cite{dhar2024fusegnet}).
    
    \item \textit{Medical segmentation}:  Medical segmentation (MS) architectures, such as \acl{unet} and its variants, have demonstrated broad applicability across diverse medical imaging tasks~\cite{siddique2021u}. To assess the potential of recent architectures that have shown effectiveness beyond wound care, we select three models with strong performance in medical image segmentation outside the wound domain (\acl{fcbformer}~\cite{sanderson2022fcbformer}, \acl{hiformer}~\cite{heidari2023hiformer}, \acl{missformer}~\cite{huang2022missformer}).
    
    \item  \textit{\ac{gp} SS models}: 
    To incorporate cutting-edge developments, we include two widely recognized \ac{gp} semantic segmentation (SS) models (\acl{segformer}~\cite{xie2021segformer} and \acl{segnext}~\cite{guo2022segnext}), along with a recent multi-scale decoder (\acl{vwformer}~\cite{yan2024vwformer}) for SS. The \acs{vwformer} model is integrated with two different backbones, MiT-B3 and ConvNeXt-S, denoted as \acs{vwmit} and \acs{vwconv}, respectively.
    \item  \textit{\ac{gp}  vision models}: 
    To account for advanced (multi-task) vision models (VM), we select two promising approaches from this category (\acl{internimage}~\cite{wang2023internimage}, \acl{transnext}~\cite{shi2024transnext}). As common in literature, we adapt them for segmentation by integrating a \acs{uperhead}~\cite{xiao2018uperhead} decoder.
    
    \item  \textit{Baseline}: Apart from these categories, we include \acs{unet}~\cite{ronneberger2015unet}, a proven and widely used model in biomedical segmentation, as a baseline.

\end{enumerate*}

Table~\ref{tab:app:selection:overview} summarizes our final selection, striking a balance between domain specificity, cross-domain variety, and methodological diversity. 
The key ideas of all architectures are briefly summarized in Section A.1, with further details available in the corresponding publications.
Overall, our selection process is guided by the following criteria:
\begin{enumerate*}[label=(\Roman*)] 
    \item \textit{Architectural diversity}: To ensure a comprehensive representation of various design paradigms, we aim for a balance between \ac{cnn}-based, \ac{vit}-based, and hybrid methods.
    \item \textit{Computational efficiency}: Given the limited resources in healthcare settings, we prioritize models that offer a good trade-off between performance and computational feasibility. 
    For \ac{gp} models, we select architectures with approximately 50–60M parameters, while wound-specific and medical models range from 30–70M parameters, mainly due to the limited availability of varied model sizes in these domains.
    \item \textit{Scientific impact and recognition}: We select models with significant visibility in their respective fields, prioritizing those published in high-impact conferences and journals. 
    \item \textit{Code availability}: To minimize the risk of implementation errors, we limit our selection to models with publicly available code.
\end{enumerate*}

\begin{table}[ht]
\caption{Overview of the selected methods besides our baseline \ac{unet}~\cite{ronneberger2015unet}.}
\label{tab:app:selection:overview}
\centering
    \begin{minipage}[t]{0.5\textwidth} 
        \centering
        \scalebox{.88}{
        \begin{threeparttable}
            \caption*{Healthcare-Related}
            \begin{tabular}{l@{\hskip 0.1in}l@{\hskip 0.1in}l@{\hskip 0.1in}lc}
                \toprule
                Cat.     & Model & Type & Size & Y. \\
                \midrule
                \multirow{3}{*}{MS}  
                    & \ac{fcbformer}~\cite{sanderson2022fcbformer}      & Hybrid & 53M & '22 \\         
                    & \acl{hiformer}~\cite{heidari2023hiformer}         & Hybrid & 30M & '23 \\                                       
                    & \acl{missformer}~\cite{huang2022missformer}     & Transf.& 43M & '23 \\                      
                \midrule
                \multirow{2}{*}{WS} 
                    & \acl{hardnet}~\cite{liao2022hardnet}          & CNN & 52M & '22 \\      
                    & \acl{fusegnet}~\cite{dhar2024fusegnet}          & CNN & 71M & '24 \\      
                \bottomrule
            \end{tabular}
            \begin{tablenotes}[para]
            \scriptsize{
            \item[1] 51M with MiT-B3/ 57M with ConvNext-S
            }
            \end{tablenotes}
        \end{threeparttable}
        }
    \end{minipage}%
    \hfill
    \begin{minipage}[t]{0.5\textwidth} 
        \centering
        \scalebox{.88}{
        \begin{threeparttable}
            \caption*{General-Purpose}
            \begin{tabular}{l@{\hskip 0.1in}l@{\hskip 0.1in}l@{\hskip 0.1in}lc}
            \toprule
            Cat. & Model & Type & Size & Y. \\
            \midrule
            \multirow{3}{*}{SS} 
            & \acl{segformer}~\cite{xie2021segformer}      & Transf. & 48M   & '21 \\                  
            & \acl{segnext}~\cite{guo2022segnext}           & CNN     & 49M   & '22 \\                  
            & $2\times$ \acl{vwformer}~\cite{yan2024vwformer}  & \multicolumn{2}{c}{-Depends-\tnote{1}}     & '24  \\               
            \midrule
            \multirow{2}{*}{VM} 
            & \acl{internimage}~\cite{wang2023internimage}\tnote{2}     & CNN     & 58M   & '23\\       
            & \acl{transnext}~\cite{shi2024transnext}\tnote{2}          & Transf. & 58M   & '24 \\      
            \bottomrule
            \end{tabular}
            \begin{tablenotes}[para]
            \scriptsize{
            \item[2] Parameter count includes \acs{uperhead} decoder
            }
            \end{tablenotes}
        \end{threeparttable}
        }
    \end{minipage}
\end{table}

\subsection{Unified Training Procedure: Methodological Details}
\label{ssec:benchmark:training_procedure}

We establish a unified benchmarking environment by standardizing the training process as follows:
To ensure an unbiased comparison while enabling the use of pre-trained weights, the deep learning methods were implemented with their architecture-specific settings, as recommended in their respective official publications and code repositories. 
Specifically, we use ImageNet-pretrained backbones for all methods. 
While maintaining key architectural parameters (e.g., layer configurations and activation functions), we standardize other training settings across all models. These include input tensor dimensions, the number of training epochs, early stopping criteria, optimizer type, loss function, and the data augmentation pipeline. Further details on preprocessing, data augmentation, and exact training configurations are provided in Section~A.2. 
To minimize biases associated with fixed learning rates, we conduct a preliminary hyperparameter tuning step for each architecture, optimizing the learning rate within a predefined search range ($10^{-4}, 5\times10^{-5}, 10^{-5}$). The best-performing learning rate for each method (see Section A.2) is then used for subsequent five-fold \ac{cv}, while all other training settings remain unified.
Overall, this ensures that performance differences arise from the intrinsic capabilities of the architectures rather than variations in training configurations, particularly data augmentation.

\subsection{Standardized Evaluation Procedure}
\label{ssec:benchmark:eval_procedure}
Our assessment strategy employs a unified evaluation framework, incorporating both traditional segmentation metrics and practical aspects essential for clinical applications. For \textit{segmentation performance}, we use metrics such as mean Intersection over Union (mIoU), Dice Similarity Coefficient (mDSC), precision (mPrc), and recall (mRec). 
These metrics evaluate performance on both our main dataset and unseen \ac{ood} data, with the latter specifically assessing the \textit{generalization capability} across diverse wound types.
In terms of \textit{model efficiency}, we report the number of trainable parameters and \acp{gmac}, along with mean inference time for GPU and CPU execution and throughput in images per second (IPS). Finally, we assess \textit{explainability} using \emph{\ac{gradcam}}-based visualizations. Further information and implementation details are available in Section~A.3.

\subsection{Datasets}
\label{ssec:benchmark:datasets}
Our experiments mainly rely on two datasets, detailed in Section~A.4. 
The first, \textit{CFU}, is used for model training, including learning rate studies and final \ac{cv}. It consists of a custom combination of the publicly available \acs{dfuc22} dataset~\cite{kendrick2022translating}, which contains \numprint{2000} annotated images, and the \ac{fuseg}’21 challenge dataset~\cite{wang2024fuseg}, with \numprint{1010} labeled images. 
After removing duplicates and highly similar images, \numprint{2887} unique images remain for our experiments. Additionally, we use the \acs{dfuc22} test set for external validation of our models via the challenge’s live leaderboard.
The second dataset, denoted as \textit{out-of-distribution (OOD)}, was collected at the University Hospital of Würzburg with approval from the local ethics committee. It comprises 343 expert-annotated wound images from various anatomical sites, extending beyond foot ulcers. To better reflect real-world clinical conditions, where patient-acquired images often lack standardized imaging protocols, certain wounds were intentionally captured multiple times from varying distances, angles, and perspectives. 
Notably, \textit{\ac{ood}} is used exclusively for evaluation, not for training. Due to privacy regulations, it remains confidential; however, selected examples are shared at \href{https://github.com/VanessaBorst/woundambit}{GitHub} with written consent.
\subsection{Performance on CFU Dataset and \ac{dfuc22} Live Leaderboard}
\label{ssec:benchmark:eval:dfuc}
Table~\ref{tab:eval:classification_performance} presents the 5-fold \ac{cv} results on CFU alongside each model's performance on the \ac{dfuc22} live leaderboard. 
For the latter, instead of selecting the best checkpoint from individual folds, segmentation masks are generated using \acp{pmv} across all five instances of each architecture.
This approach mitigates overfitting to individual training folds while leveraging the collective strengths of multiple trained instances. 

On CFU, \ac{transnext} demonstrated the highest performance, achieving a mIoU of 79.8 and an mDSC of 88.7. \ac{segnext} followed closely with a mIoU of 79.5 and an mDSC of 88.6, exhibiting strong consistency across the \ac{cv} folds. 
With similarly low inter-fold variability, \ac{segformer} ranked third, achieving a mIoU of 78.9 and an mDSC of 88.2.
%
On the \ac{dfuc22} leaderboard, \ac{transnext} also achieved the highest performance among our models, with a mIoU of 62.8 and an mDSC of 73.0, slightly surpassing \ac{segnext} and \ac{vwconv}---the latter ranking third despite its moderate performance on CFU. 
Notably, \ac{transnext} ranked 6th out of 60 (top 10\%) based on the best submission per participant despite no dataset-specific optimization, which highlights its strong out-of-the-box performance.
In contrast, \ac{unet} and most medical approaches, except \ac{fcbformer}, showed lower segmentation performance and ranked further down on both CFU and \ac{dfuc22}. 
\begin{table*}[ht!]
    \centering
    \caption[]{5-fold CV (mean±SD, \%) and ensemble \ac{dfuc22} leaderboard scores.}
    \label{tab:eval:classification_performance}
   \begin{adjustbox}{max width=0.8\textwidth}
        \begin{tabular} {l@{\hskip 0.2in}
        c@{\hskip 0.1in}c@{\hskip 0.1in}c@{\hskip 0.1in}c@{\hskip 0.2in}
        cc}
        \toprule
         \multirow{2}{*}{\textbf{Type}} 
         & \multicolumn{4}{c}{Avg. CFU} 
         & \multicolumn{2}{c}{\ac{dfuc22}}  \\ 
         \cmidrule(lr{20pt}){2-5}\cmidrule{6-7}
          & \multicolumn{1}{c}{mIoU$\downarrow$} &  mDSC &  mPrc & mRec & mIoU & mDSC \\ \midrule\midrule
           \ac{transnext}                    & \textbf{79.8±1.4} & \textbf{88.7±0.8} & \textbf{90.9±0.4} & \textbf{86.7±1.8} & \textbf{62.8} & \textbf{73.0} \\
           \ac{segnext}                      & 79.5±0.7 & 88.6±0.4 & 90.7±0.6 & 86.6±0.6 & 62.1 & 72.3 \\
           \ac{segformer}                    & 78.9±0.8 & 88.2±0.5 & 90.4±0.7 & 86.1±1.4 & 61.8 & 72.1 \\
           \ac{fcbformer}                    & 78.6±1.5 & 88.0±0.9 & 90.6±0.6 & 85.6±1.9 & 62.0 & 72.2 \\
           \ac{internimage}                  & 78.5±0.9 & 88.0±0.5 & 89.8±1.5 & 86.2±0.6 & 61.7 & 72.0 \\
           \acs{vwmit}                        & 78.5±1.7 & 87.9±1.0 & 90.2±1.1 & 85.8±2.2 & 61.7 & 72.0 \\
           \ac{vwconv}                       & 78.4±0.5 & 87.9±0.3 & 90.1±1.9 & 85.9±2.0 & 62.0 & 72.3 \\
           \ac{fusegnet}                     & 78.0±1.3 & 87.6±0.8 & 90.6±0.7 & 84.9±1.3 & 61.3 & 71.6 \\
           \ac{hardnet}                      & 76.9±1.4 & 86.9±0.9 & 88.3±2.4 & 85.8±2.5 & 60.4 & 70.8 \\
           \ac{unet}                         & 74.1±0.9 & 85.1±0.6 & 88.1±1.1 & 82.5±1.5 & 57.6 & 68.0 \\
           \ac{hiformer}                     & 73.8±1.8 & 84.9±1.2 & 87.6±1.6 & 82.5±1.7 & 57.7 & 68.2 \\
           \ac{missformer}                   & 70.0±2.7 & 82.3±1.8 & 85.8±3.4 & 79.3±3.3 & 55.6 & 66.4 \\
        \bottomrule
        \end{tabular}
\end{adjustbox}
\end{table*}
\subsection{Performance on Out-of-Distribution (OOD) Data}
\label{ssec:benchmark:eval:OOD}
Table~\ref{tab:eval:OOD:performance} presents the segmentation performance on the unseen \ac{ood} dataset, which includes previously unobserved anatomical regions, such as the head and breast, thereby assessing model generalization to domain shifts.
In addition to reporting the average and standard deviation (SD) across the five \ac{cv} models per architecture, we again provide \ac{pmv} results to ensure a more stable and unbiased evaluation. Notably, for all models, the \ac{pmv} performance surpasses the average performance of the best models from individual folds in both mIoU and mDSC, further confirming its robustness.
\ac{transnext} demonstrated the strongest generalization, achieving the highest mIoU (79.4) and mDSC (88.5) in the \ac{pmv} setting, followed closely by \ac{internimage} (mIoU 78.8, mDSC 88.1) and \ac{vwmit} (mIoU 78.0, mDSC 87.6). The performance drop relative to the in-distribution CFU data was minimal ($\leq$0.5pp mIoU, $\leq$ 0.3pp mDSC) for the top three models. \ac{internimage} even showed a 0.3pp improvement in mIoU, indicating strong adaptability to novel wound types.
\ac{segformer}, \ac{vwconv}, and \ac{hardnet} also maintained competitive performance relative to CFU ($\sim$1pp mIoU drop). \ac{fcbformer} achieved the highest absolute \ac{pmv} scores among the medical architectures (mIoU 76.5, mDSC 86.7). In contrast, the remaining medical models, except for \ac{hardnet}, exhibited substantial performance declines (4–7.9pp mIoU), with \ac{hiformer} dropping below 68\% mIoU. Notably, \ac{segnext}, despite its strong in-distribution and \ac{dfuc22} performance, saw a sharp decline of 3.7pp in mIoU.
\begin{table*}[ht!]
    \centering
    \caption[]{Segmentation performance (mean±SD, \%) on the OOD dataset using the five trained CV models, with CFU mean IoU and DSC for reference.}
    \label{tab:eval:OOD:performance}
   \begin{adjustbox}{max width=1\textwidth}
        \begin{tabular} {l@{\hskip 0.1in}
        c@{\hskip 0.05in}c@{\hskip 0.1in}
        c@{\hskip 0.05in}c@{\hskip 0.05in}c@{\hskip 0.05in}c@{\hskip 0.05in}
        cccc}
        \toprule
         \multirow{2}{*}{\textbf{Type}} 
         & \multicolumn{2}{c}{Avg. CFU}  
         & \multicolumn{4}{c}{Avg. OOD}  
         & \multicolumn{4}{c}{Maj. vote OOD}    \\ 
         \cmidrule(lr){2-3}\cmidrule(lr{10pt}){4-7}\cmidrule(){8-11}
          &  \multicolumn{1}{c}{mIoU} &  mDSC & \multicolumn{1}{c}{mIoU} &  mDSC &  mPrc & mRec & \multicolumn{1}{c}{mIoU$\downarrow$} &  mDSC &  mPrc & mRec \\ \midrule\midrule
           \ac{transnext}     & \textbf{79.8} & \textbf{88.7}      & \textbf{78.3±1.7} & \textbf{87.8±1.1} & 93.6±0.2 & \textbf{82.7±1.9 }     & \textbf{79.4} & \textbf{88.5} & 94.3 & \textbf{83.4} \\
           \ac{internimage}   & 78.5 & 88.0      & 76.0±3.2 & 86.3±2.1 & 92.4±0.6 & 81.1±3.6      & 78.8 & 88.1 & 93.7 & 83.2 \\
           \ac{vwmit}         & 78.5 & 87.9      & 75.7±1.6 & 86.2±1.0 & 92.8±0.5 & 80.5±2.0      & 78.0 & 87.6 & 93.7 & 82.3 \\
           \ac{segformer}     & 78.9 & 88.2      & 76.3±1.2 & 86.5±0.7 & 93.1±0.6 & 80.8±1.4      & 77.7 & 87.5 & 94.1 & 81.7 \\
           \ac{vwconv}        & 78.4 & 87.9      & 74.9±2.7 & 85.7±1.8 & 93.1±0.8 & 79.4±3.5      & 77.2 & 87.1 & 94.4 & 80.9 \\
           \ac{fcbformer}     & 78.6 & 88.0      & 74.0±1.9 & 85.1±1.3 & 93.2±1.0 & 78.3±2.7      & 76.5 & 86.7 & 94.5 & 80.0 \\
           \ac{hardnet}       & 76.9 & 86.9      & 73.0±2.6 & 84.4±1.7 & 91.1±2.1 & 78.7±4.1      & 75.9 & 86.3 & 93.1 & 80.4 \\
           \ac{segnext}       & 79.5 & 88.6      & 74.5±1.7 & 85.4±1.1 & 93.4±0.3 & 78.6±2.0      & 75.8 & 86.2 & 94.4 & 79.4 \\
           \ac{fusegnet}      & 78.0 & 87.6      & 71.5±3.2 & 83.4±2.2 & \textbf{94.2±0.8} & 74.9±3.9      & 74.0 & 85.1 & \textbf{95.7} & 76.5 \\
           \ac{unet}          & 74.1 & 85.1      & 65.6±2.7 & 79.2±2.0 & 91.9±0.6 & 69.7±3.3      & 67.5 & 80.6 & 93.3 & 70.9 \\
           \ac{missformer}    & 70.0 & 82.3      & 64.3±2.6 & 78.2±2.0 & 88.7±1.6 & 70.1±3.5      & 66.0 & 79.5 & 90.2 & 71.1 \\
           \ac{hiformer}      & 73.8 & 84.9      & 63.1±5.1 & 77.3±3.9 & 90.8±2.0 & 67.7±6.3      & 65.9 & 79.4 & 92.4 & 69.6 \\
        \bottomrule
        \end{tabular}
\end{adjustbox}
\end{table*}
\subsection{Model Efficiency Analysis}
\label{ssec:benchmark:eval:efficiency}
We report key computational metrics in Table~\ref{tab:eval:efficiency:comparison}. 
Among all architectures, \ac{fusegnet} has the highest number of trainable parameters (71M), while \ac{hiformer} and \ac{unet} are the smallest. In terms of \acp{gmac}, \ac{missformer} is the most efficient, whereas \ac{unet}, and, especially, \ac{transnext} and \ac{internimage} exhibit higher computational complexity.
Interestingly, parameter count and GMACs do not always directly correspond to inference time and throughput (TP). For instance, despite having similar parameter counts to most models (50–60M), \ac{transnext} and \ac{fcbformer} exhibit considerably higher inference times and lower TP, making them the slowest on both GPU and CPU. In contrast, \ac{unet}, despite having the third-highest GMAC count, achieves the fastest GPU inference time, highlighting the influence of architectural design and optimizations on computational efficiency.
The results reveal a trade-off between efficiency and segmentation performance. While models such as \ac{unet}, \ac{missformer}, and \ac{hiformer} excel in inference speed on both GPU and CPU, advanced vision models like \ac{transnext} and \ac{internimage} offer superior segmentation performance at a moderate computational cost. Notably, even \ac{transnext}, the slowest model, maintains a throughput of approximately one IPS on the CPU and up to 24 IPS on the GPU.

\begin{table}[htb]
    \centering
    \small
    \caption[Model Efficiency Analysis]{Evaluation of model efficiency, with OOD maj. vote mIoU for reference.}
    \label{tab:eval:efficiency:comparison}
    \begin{adjustbox}{max width=0.95\textwidth}
        \begin{threeparttable}
            \begin{tabular}{l@{\hskip 0.15in}
            c@{\hskip 0.15in}
            c@{\hskip 0.15in}
            c@{\hskip 0.15in}
            c@{\hskip 0.1in}c@{\hskip 0.15in}
            c@{\hskip 0.1in}c@{\hskip 0.15in}
            }
                \toprule
                \multirow{2}{*}{\textbf{Type}} & OOD & Params & \multirow{2}{*}{GMACs}  & \multicolumn{2}{c}{$\varnothing$ Inference Time ± SD (in ms)} & \multicolumn{2}{c}{TP (in IPS)} \\ 
                \cmidrule(lr{10pt}){5-6}\cmidrule(lr){7-8}
                 & mIoU$\downarrow$ & (in M) & & GPU & CPU & GPU & CPU \\ 
                \midrule
                \ac{transnext}    &  \textbf{79.4} &   57.74 &    238.41     &   41.58 ± 0.10  & 1012.75 ± 243.11           & 24.05          & 0.99  \\ 
                \ac{internimage}  &  78.8 &            58.37 &    234.23     &   28.71 ± 0.29  &   692.35 ± 42.90           & 34.83          & 1.44  \\ 
                \ac{vwmit}        &  78.0 &            51.42 &     62.79     &   25.65 ± 0.11  &   439.98 ± 55.75           & 38.98          & 2.27  \\ 
                \ac{segformer}    &  77.7 &            47.22 &     71.18     &   23.54 ± 0.08  &   429.89 ± 17.95           & 42.48          & 2.33  \\ 
                \ac{vwconv}       &  77.2 &            57.00 &     77.43     &   25.55 ± 0.03  &   339.08 ± 38.59           & 39.13          & 2.95  \\ 
                \ac{fcbformer}    &  76.5 &            52.96 &    149.93     &   51.27 ± 0.09  &   875.13 ± 41.01           & 19.51          & 1.14  \\ 
                \ac{hardnet}      &  75.9 &            51.06 &    138.92     &   38.01 ± 0.16  &   669.29 ± 17.98           & 26.31          & 1.49  \\ 
                \ac{segnext}      &  75.8 &            48.77 &     64.18     &   29.50 ± 0.42  &   616.49 ± 59.53           & 33.90          & 1.62  \\ 
                \ac{fusegnet}     &  74.0 &            70.97 &     35.51     &   32.99 ± 0.13  &   526.02 ± 69.17           & 30.31          & 1.90  \\ 
                \ac{unet}         &  67.5 &            31.03 &    192.67     &   \textbf{13.79 ± 0.02}  &   307.96 ± 15.82  & \textbf{72.50} & 3.25  \\ 
                \ac{missformer}   &  66.0 &            42.46 & \textbf{7.16} &   17.78 ± 0.15  &   \textbf{203.07 ± 19.05}  & 56.24          &\textbf{ 4.92}  \\ 
                \ac{hiformer}     &  65.9 &   \textbf{25.51} &     11.51     &   14.09 ± 0.13  &   251.57 ± 19.80           & 70.96          & 3.98 \\
                \bottomrule
            \end{tabular}
        \end{threeparttable}
    \end{adjustbox}
\end{table}

\subsection{Explainability Insights}
\label{ssec:benchmark:eval:XAI}

\begin{figure}[!ht]
    \centering
    \begin{turn}{-90}
        \begin{adjustbox}{max height=\textwidth, max width=\textheight, keepaspectratio}
        \begin{turn}{90}\hspace{1.2cm} 1 \hspace{1.1cm} 2 \hspace{1.2cm} 3 \hspace{1cm} 4 \hspace{0.8cm} 5 \hspace{0.7cm}6 \hspace{0.9cm} 7 \hspace{1.2cm} 8 \hspace{1cm} 9 \hspace{0.7cm} 10 \hspace{0.7cm} 11 \hspace{0.5cm} 12\end{turn}
            \begin{subfigure}{0.165\textwidth}
                \centering
                \includegraphics[trim={80 120 70 140}, clip, width=1.5cm, angle=180]{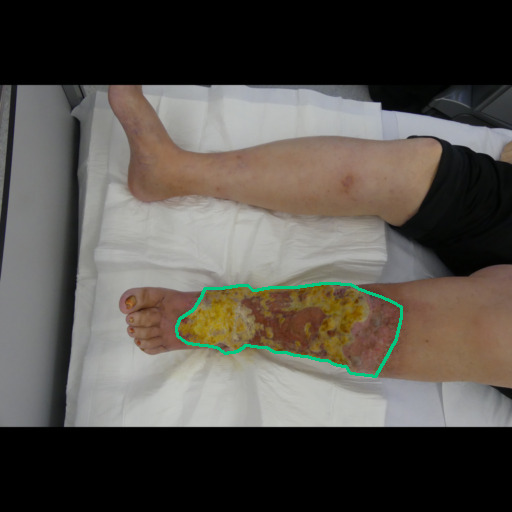}\\[4pt]
                \includegraphics[trim={80 165 0 85}, clip, width=1.5cm, angle=180]{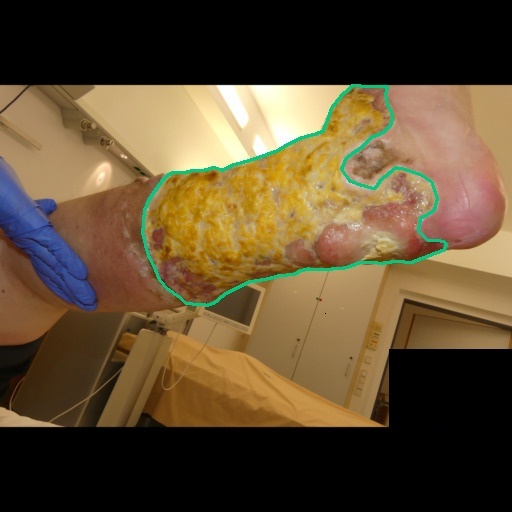}\\[4pt]
                \includegraphics[trim={30 65 0 65}, clip, width=1.5cm, angle=180]{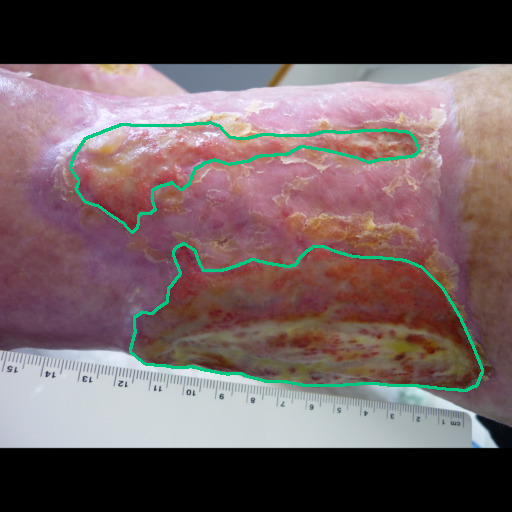}\\[4pt]
                \includegraphics[trim={0 90 0 90}, clip, width=1.5cm, angle=180]{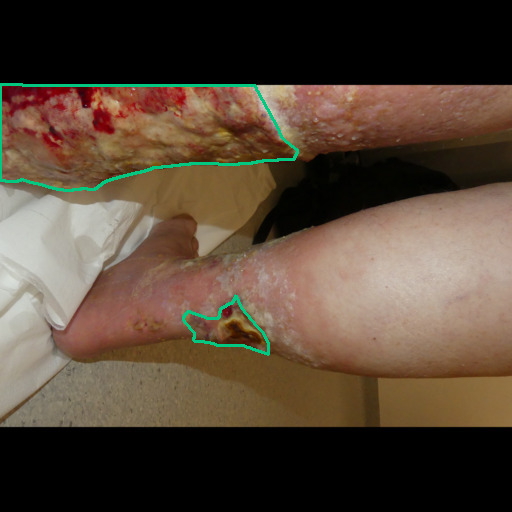}\\[4pt]
                \includegraphics[trim={150 90 60 85}, clip, height=1.5cm, angle=90]{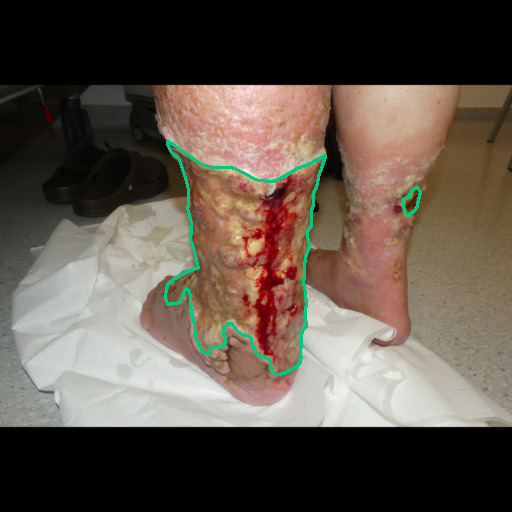}\\[4pt]
                \includegraphics[trim={110 90 120 90}, clip, height=1.5cm, angle=90]{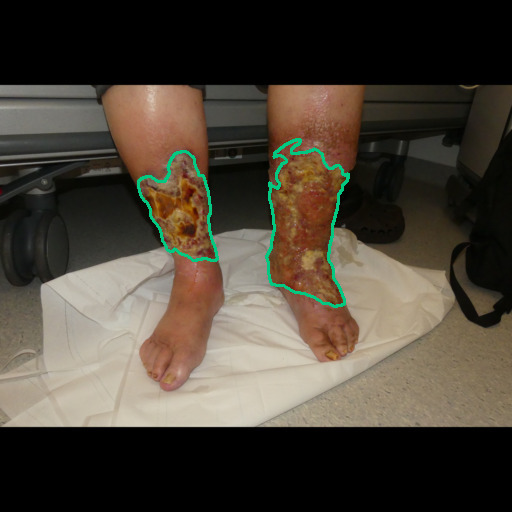}\\[4pt]
                \includegraphics[trim={90 0 90 0}, clip, height=1.5cm, angle=90]{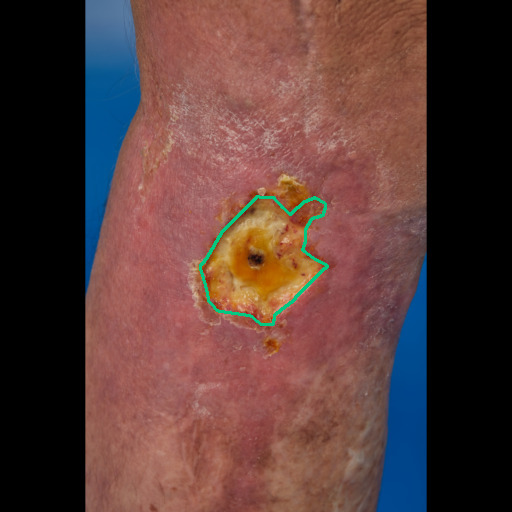}\\[4pt]
                \includegraphics[trim={220 105 100 65}, clip, height=1.5cm, angle=90]{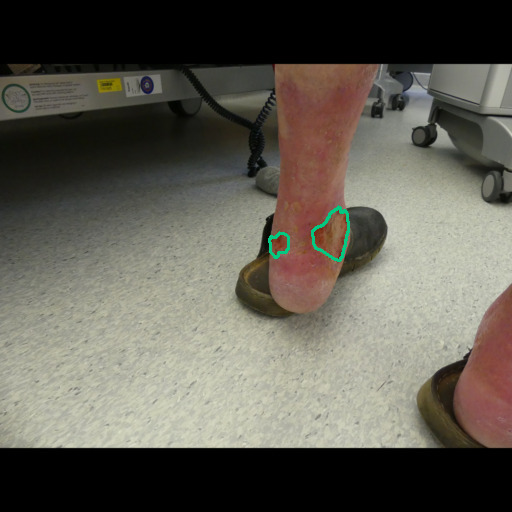}\\[4pt]
                \includegraphics[trim={110 0 85 40}, clip, height=1.5cm, angle=90]{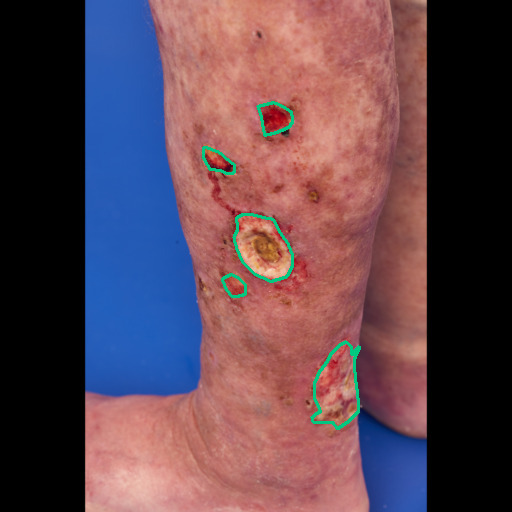}\\[4pt]
                \includegraphics[trim={160 115 90 130}, clip, height=1.5cm, angle=90]{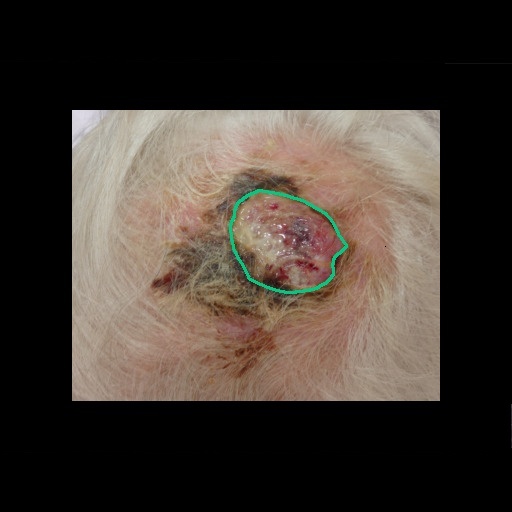}\\[4pt]
                \includegraphics[trim={160 90 20 90}, clip, height=1.5cm, angle=90]{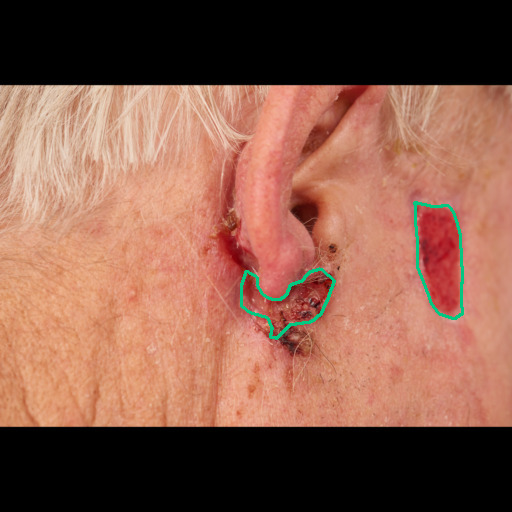}\\[4pt]
                \includegraphics[trim={64 10 64 0}, clip, height=1.5cm, angle=90]{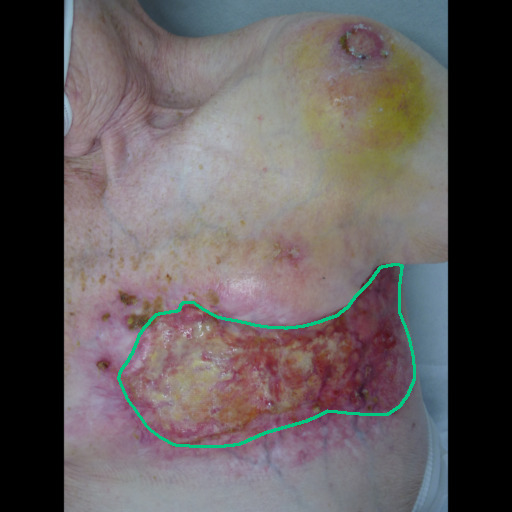}\\[4pt]
                \caption*{\scriptsize GT}
            \end{subfigure}
            \hspace{-0.6cm}
            \begin{subfigure}{0.165\textwidth}
                \centering
                \includegraphics[trim={80 120 70 140}, clip, width=1.5cm, angle=180]{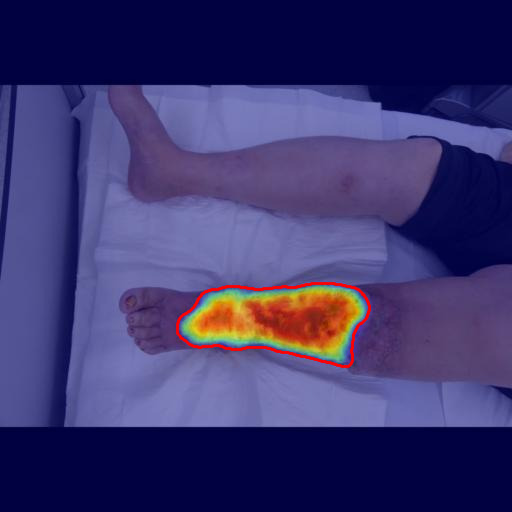}\\[4pt]
                \includegraphics[trim={80 165 0 85}, clip, width=1.5cm, angle=180]{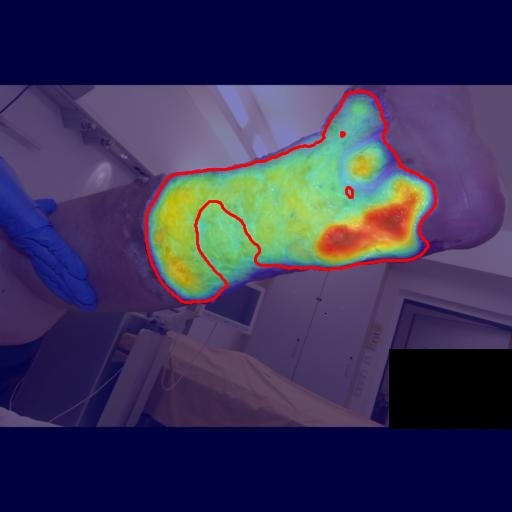}\\[4pt]
                \includegraphics[trim={30 65 0 65}, clip, width=1.5cm, angle=180]{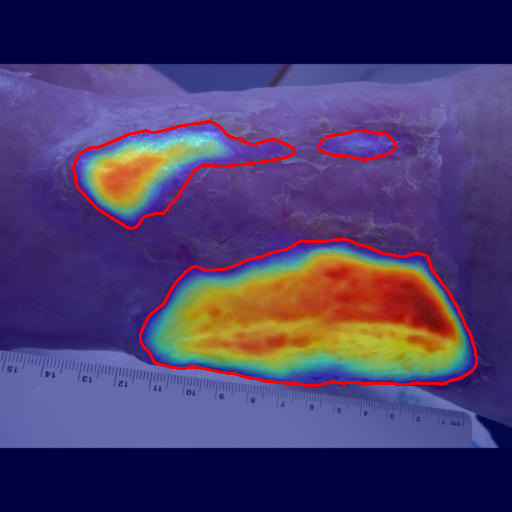}\\[4pt]
                \includegraphics[trim={0 90 0 90}, clip, width=1.5cm, angle=180]{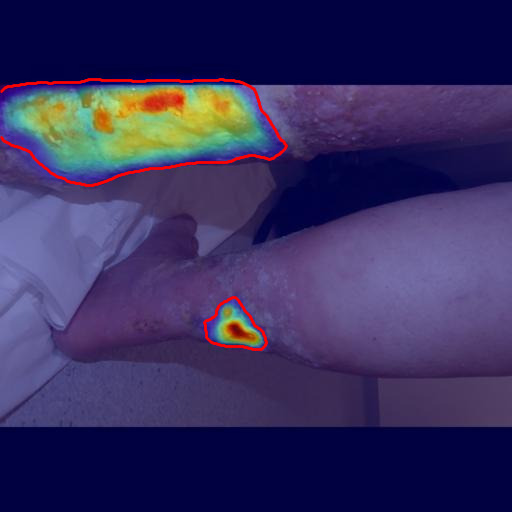}\\[4pt]
                \includegraphics[trim={150 90 60 85}, clip, height=1.5cm, angle=90]{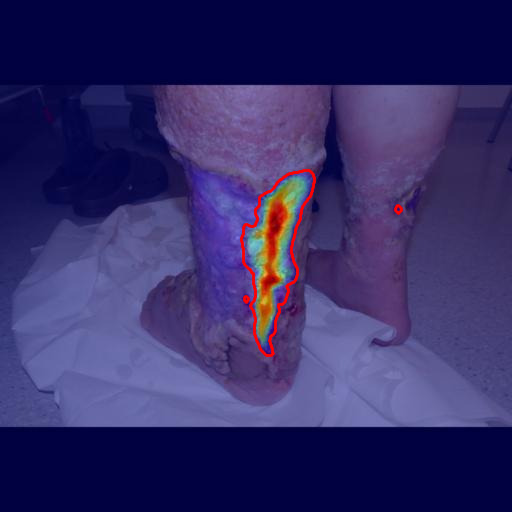}\\[4pt]
                \includegraphics[trim={110 90 120 90}, clip, height=1.5cm, angle=90]{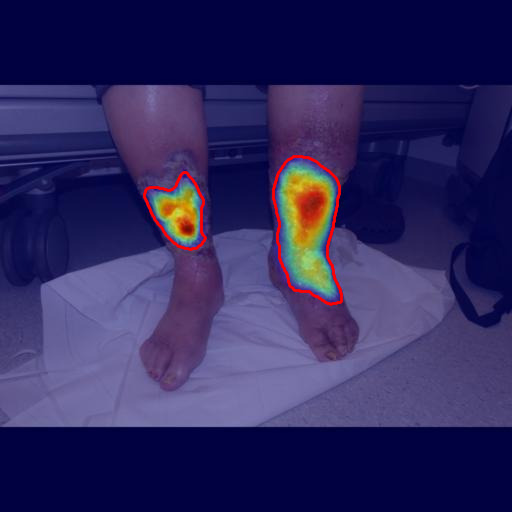}\\[4pt]
                \includegraphics[trim={90 0 90 0}, clip, height=1.5cm, angle=90]{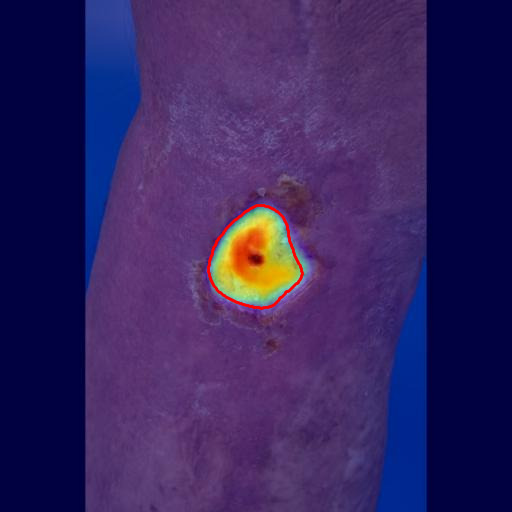}\\[4pt]
                \includegraphics[trim={220 105 100 65}, clip, height=1.5cm, angle=90]{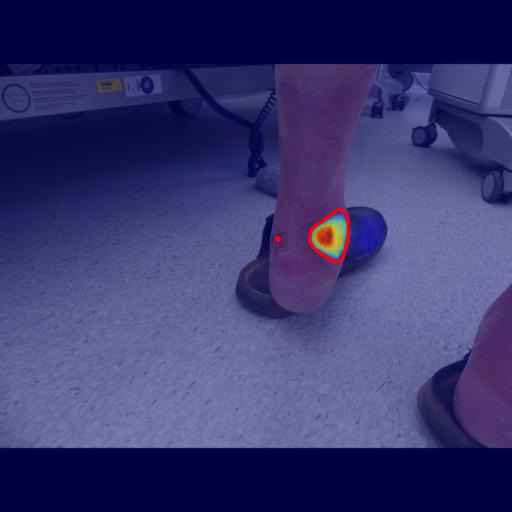}\\[4pt]
                \includegraphics[trim={110 0 85 40}, clip, height=1.5cm, angle=90]{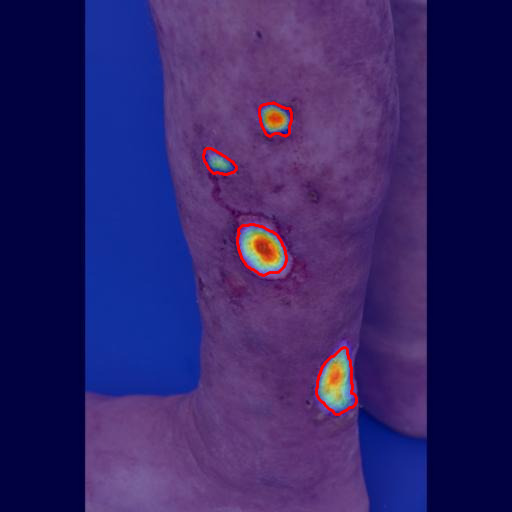}\\[4pt]
                \includegraphics[trim={160 115 90 130}, clip, height=1.5cm, angle=90]{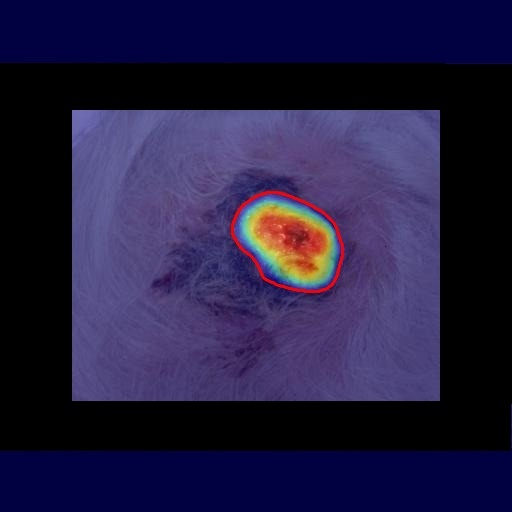}\\[4pt]
                \includegraphics[trim={160 90 20 90}, clip, height=1.5cm, angle=90]{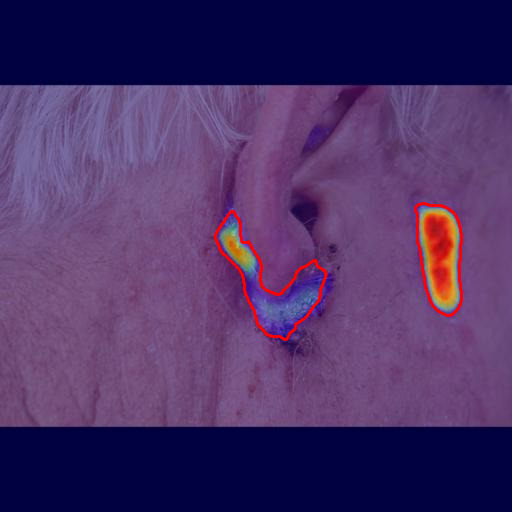}\\[4pt]
                \includegraphics[trim={64 10 64 0}, clip, height=1.5cm, angle=90]{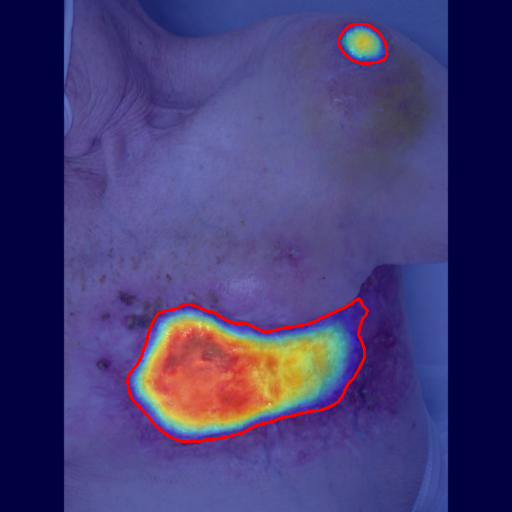}\\[4pt]
                \caption*{\scriptsize TransN.}
            \end{subfigure}
            \hspace{-0.6cm}
            \begin{subfigure}{0.165\textwidth}
                \centering
                \includegraphics[trim={80 120 70 140}, clip, width=1.5cm, angle=180]{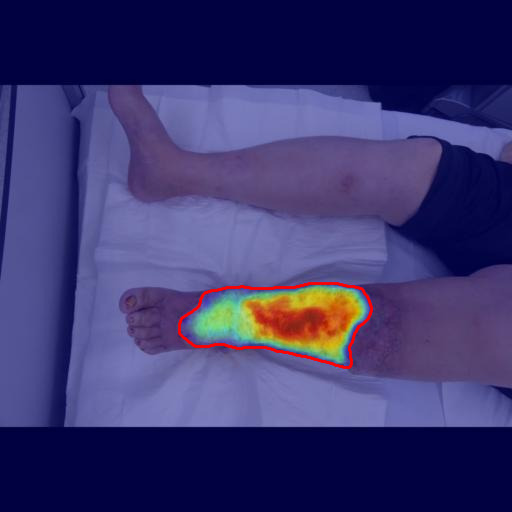}\\[4pt]
                \includegraphics[trim={80 165 0 85}, clip, width=1.5cm, angle=180]{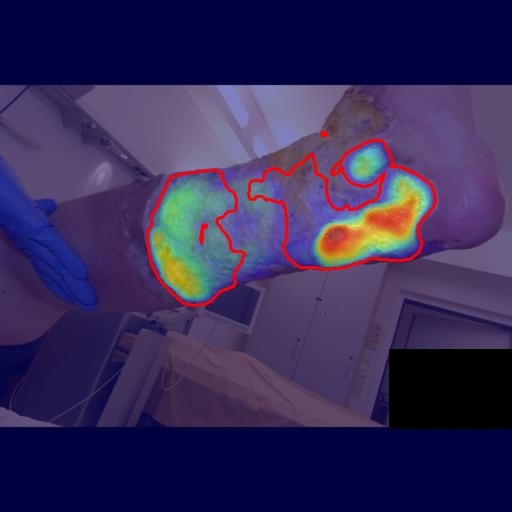}\\[4pt]
                \includegraphics[trim={30 65 0 65}, clip, width=1.5cm, angle=180]{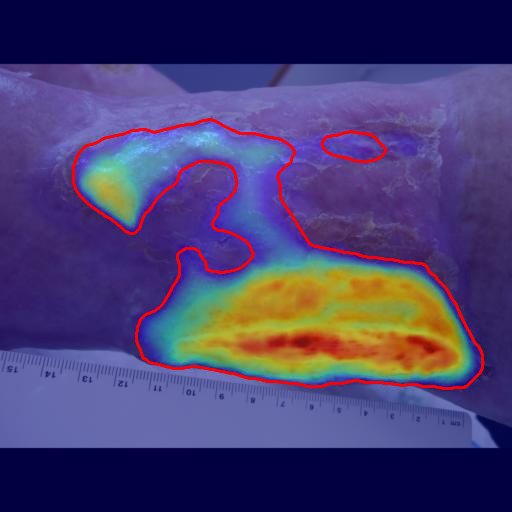}\\[4pt]
                \includegraphics[trim={0 90 0 90}, clip, width=1.5cm, angle=180]{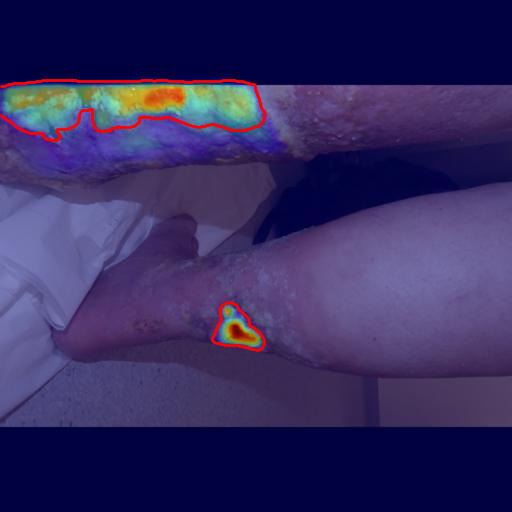}\\[4pt]
                \includegraphics[trim={150 90 60 85}, clip, height=1.5cm, angle=90]{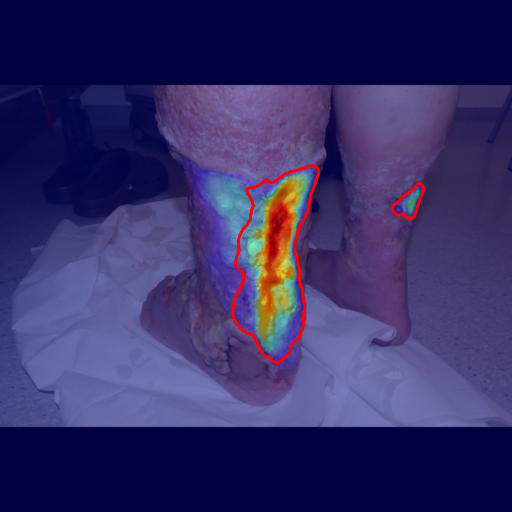}\\[4pt]
                \includegraphics[trim={110 90 120 90}, clip, height=1.5cm, angle=90]{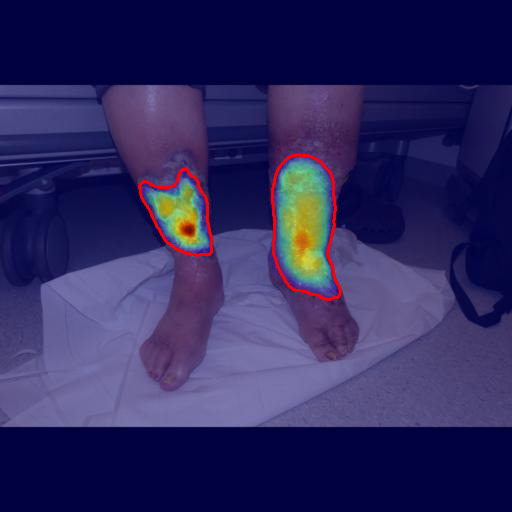}\\[4pt]
                \includegraphics[trim={90 0 90 0}, clip, height=1.5cm, angle=90]{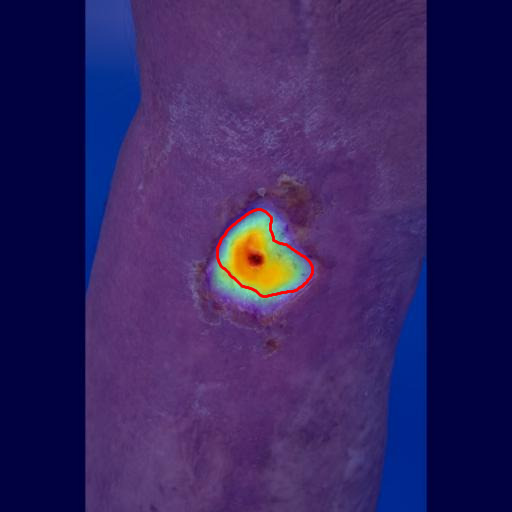}\\[4pt]
                \includegraphics[trim={220 105 100 65}, clip, height=1.5cm, angle=90]{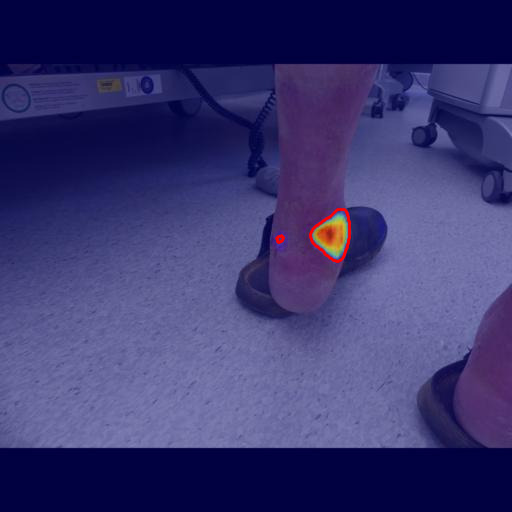}\\[4pt]
                \includegraphics[trim={110 0 85 40}, clip, height=1.5cm, angle=90]{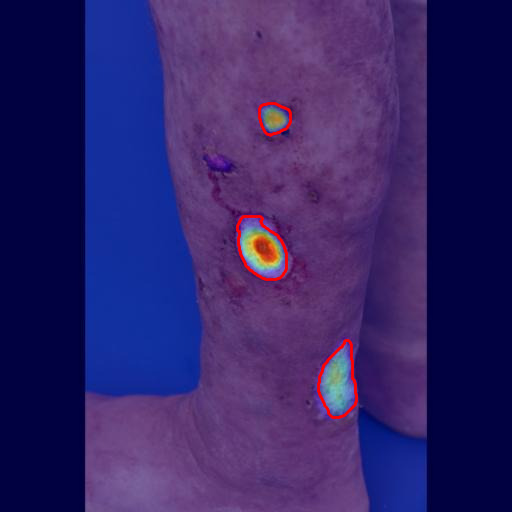}\\[4pt]
                \includegraphics[trim={160 115 90 130}, clip, height=1.5cm, angle=90]{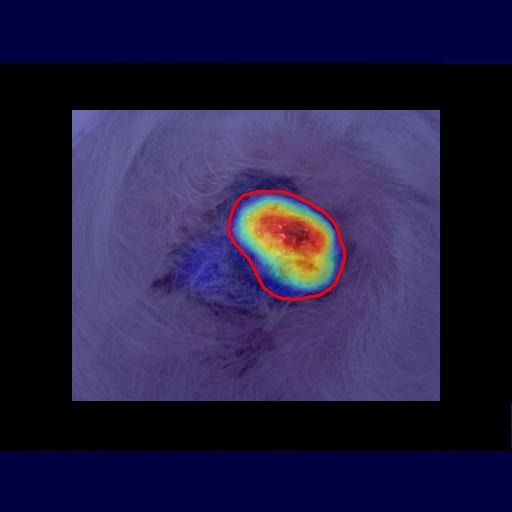}\\[4pt]
                \includegraphics[trim={160 90 20 90}, clip, height=1.5cm, angle=90]{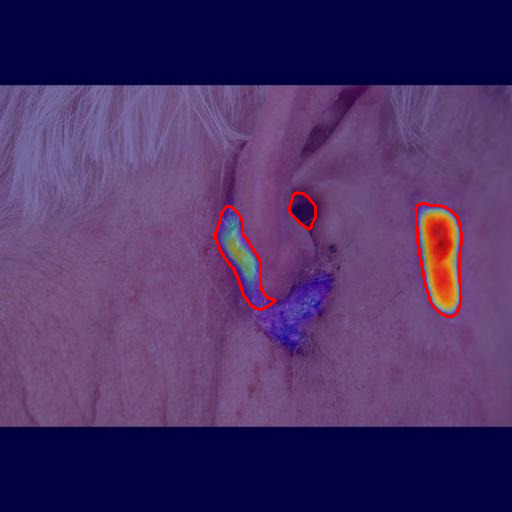}\\[4pt]
                \includegraphics[trim={64 10 64 0}, clip, height=1.5cm, angle=90]{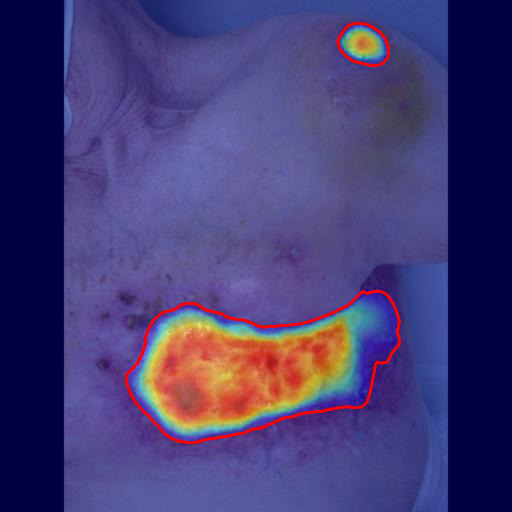}\\[4pt]
                \caption*{\scriptsize InternIm.}
            \end{subfigure}
            \hspace{-0.6cm}
            \begin{subfigure}{0.165\textwidth}
                \centering
                \includegraphics[trim={80 120 70 140}, clip, width=1.5cm, angle=180]{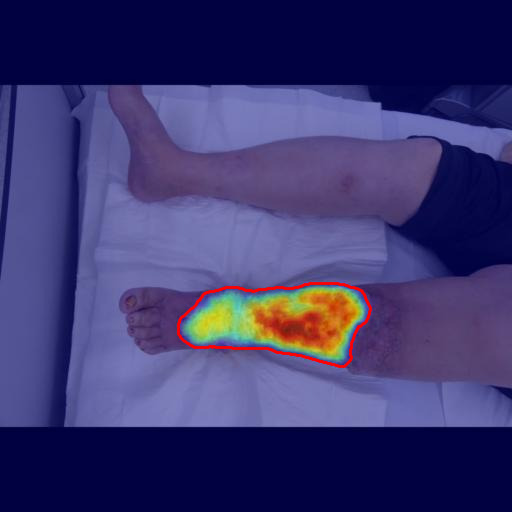}\\[4pt]
                \includegraphics[trim={80 165 0 85}, clip, width=1.5cm, angle=180]{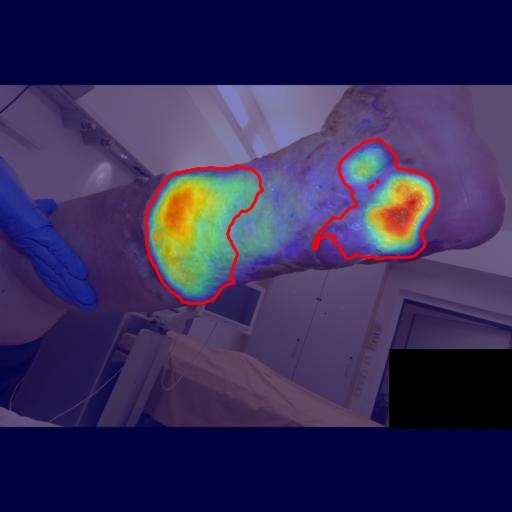}\\[4pt]
                \includegraphics[trim={30 65 0 65}, clip, width=1.5cm, angle=180]{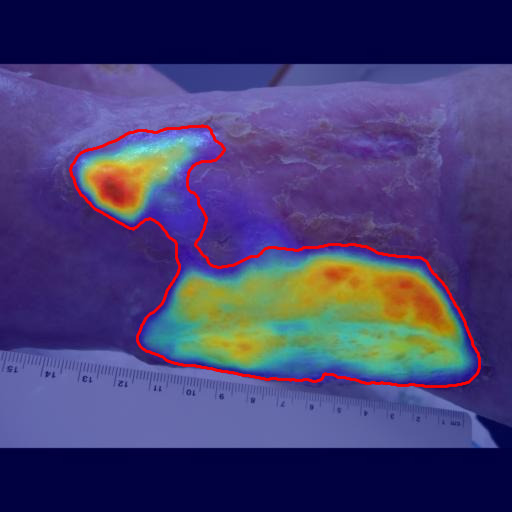}\\[4pt]
                \includegraphics[trim={0 90 0 90}, clip, width=1.5cm, angle=180]{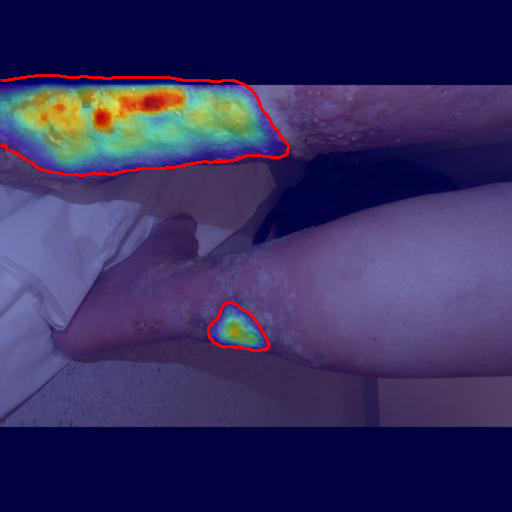}\\[4pt]
                \includegraphics[trim={150 90 60 85}, clip, height=1.5cm, angle=90]{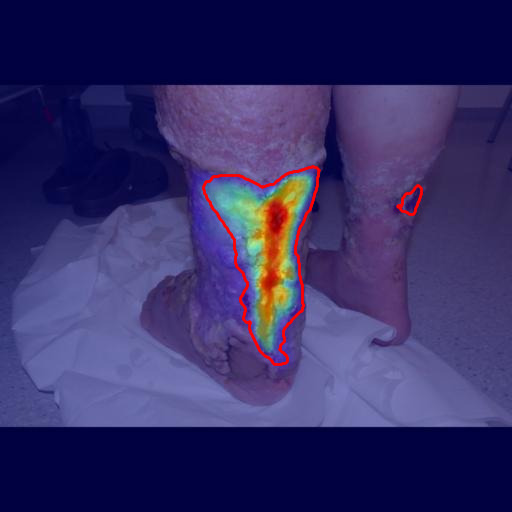}\\[4pt]
                \includegraphics[trim={110 90 120 90}, clip, height=1.5cm, angle=90]{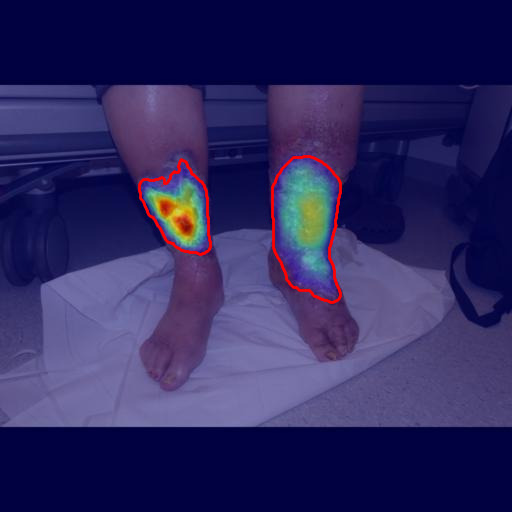}\\[4pt]
                \includegraphics[trim={90 0 90 0}, clip, height=1.5cm, angle=90]{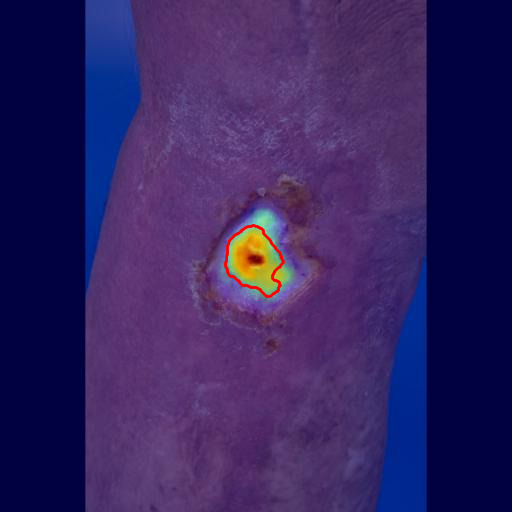}\\[4pt]
                \includegraphics[trim={220 105 100 65}, clip, height=1.5cm, angle=90]{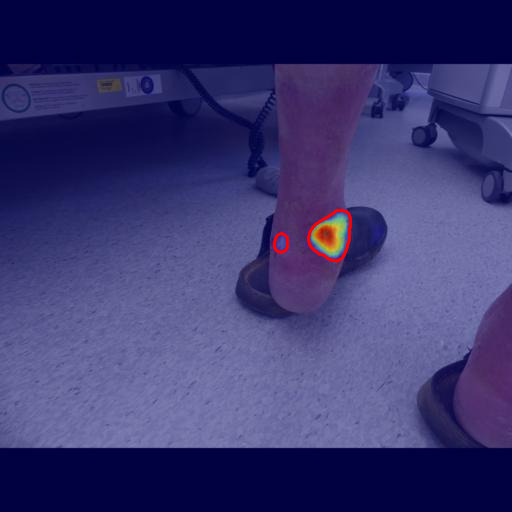}\\[4pt]
                \includegraphics[trim={110 0 85 40}, clip, height=1.5cm, angle=90]{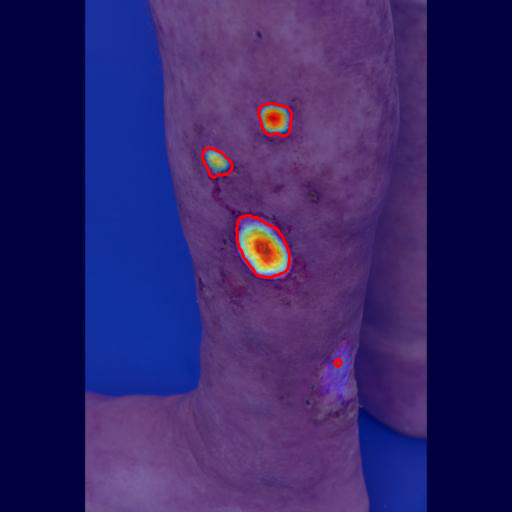}\\[4pt]
                \includegraphics[trim={160 115 90 130}, clip, height=1.5cm, angle=90]{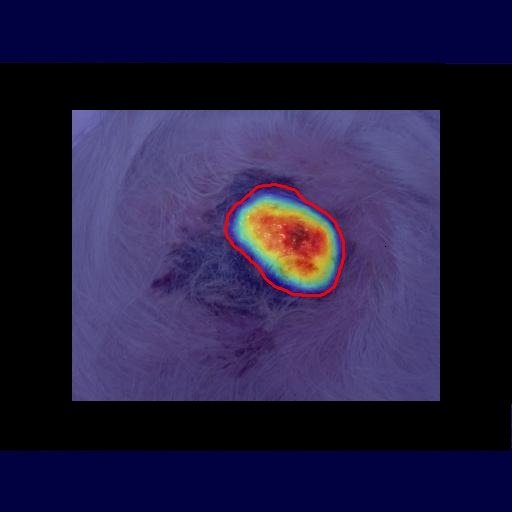}\\[4pt]
                \includegraphics[trim={160 90 20 90}, clip, height=1.5cm, angle=90]{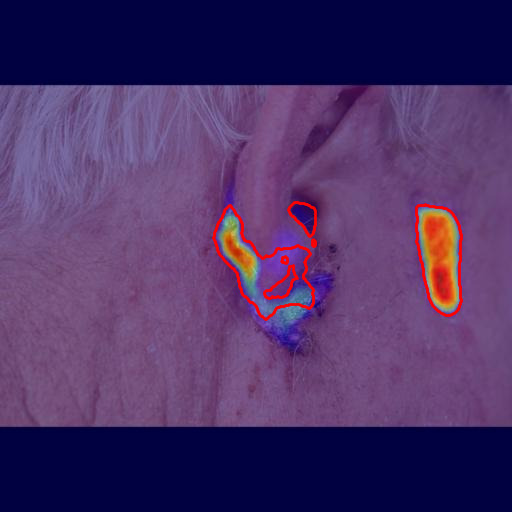}\\[4pt]
                \includegraphics[trim={64 10 64 0}, clip, height=1.5cm, angle=90]{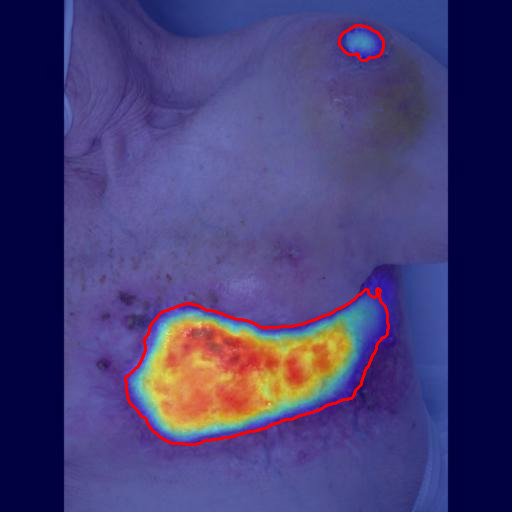}\\[4pt]
                \caption*{\scriptsize VW-MiT}
            \end{subfigure}
            \hspace{-0.6cm}
            \begin{subfigure}{0.165\textwidth}
                \centering
                \includegraphics[trim={80 120 70 140}, clip, width=1.5cm, angle=180]{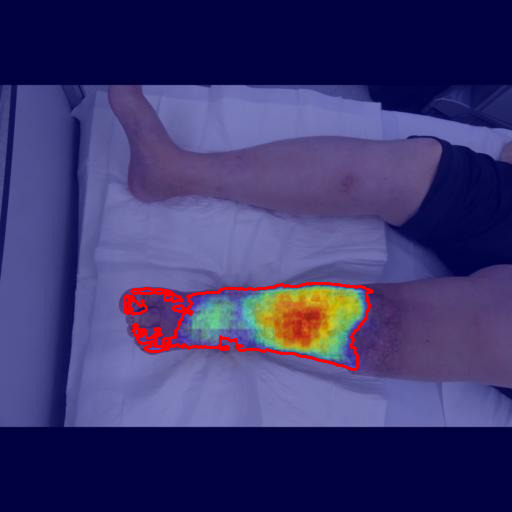}\\[4pt]
                \includegraphics[trim={80 165 0 85}, clip, width=1.5cm, angle=180]{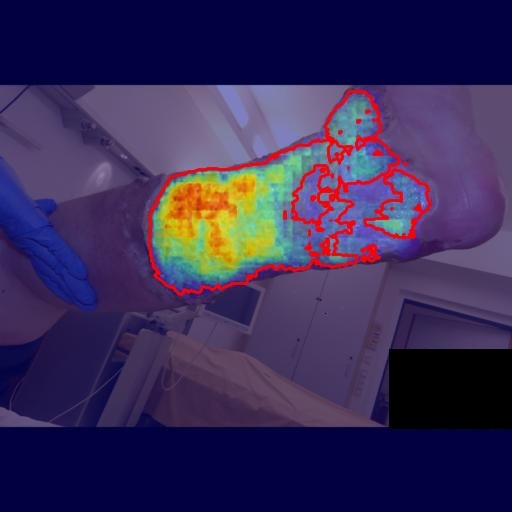}\\[4pt]
                \includegraphics[trim={30 65 0 65}, clip, width=1.5cm, angle=180]{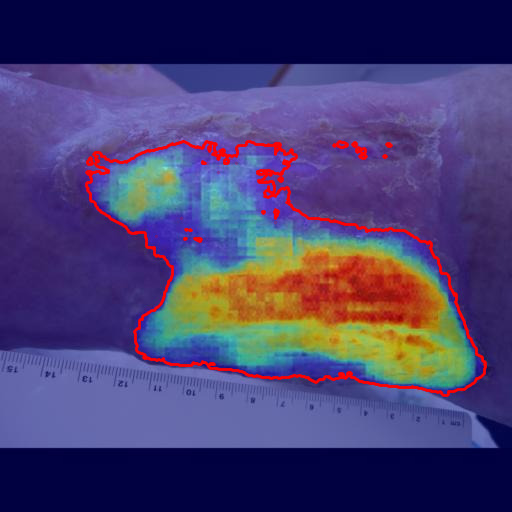}\\[4pt]
                \includegraphics[trim={0 90 0 90}, clip, width=1.5cm, angle=180]{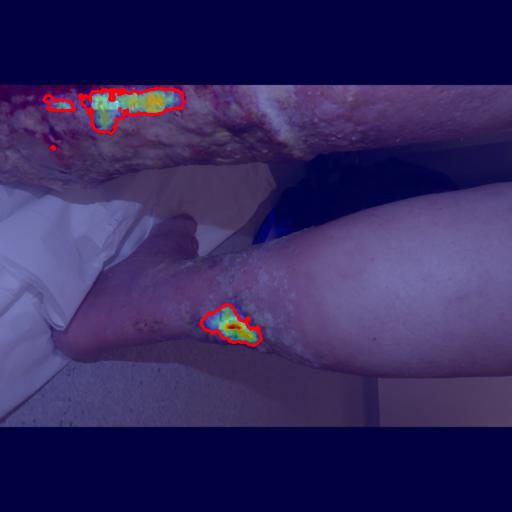}\\[4pt]
                \includegraphics[trim={150 90 60 85}, clip, height=1.5cm, angle=90]{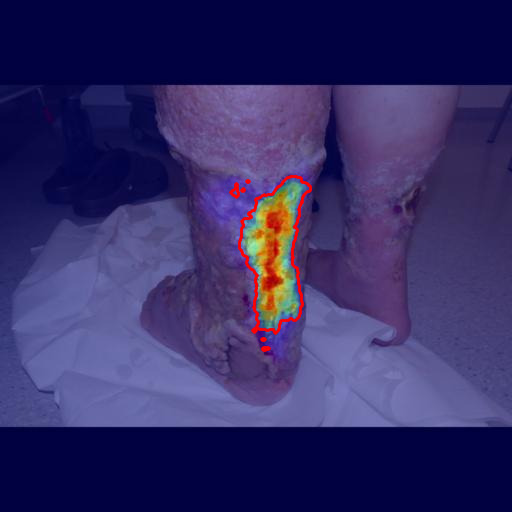}\\[4pt]
                \includegraphics[trim={110 90 120 90}, clip, height=1.5cm, angle=90]{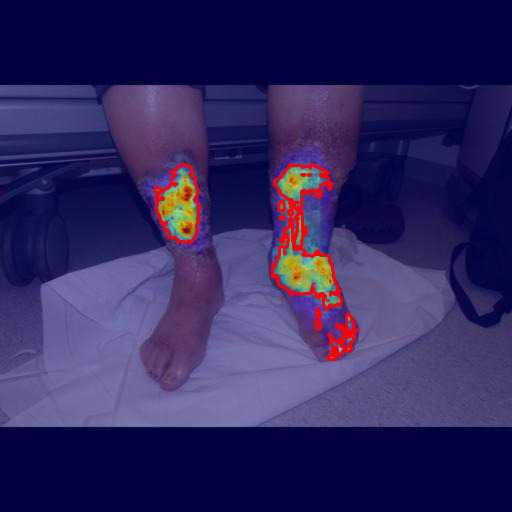}\\[4pt]
                \includegraphics[trim={90 0 90 0}, clip, height=1.5cm, angle=90]{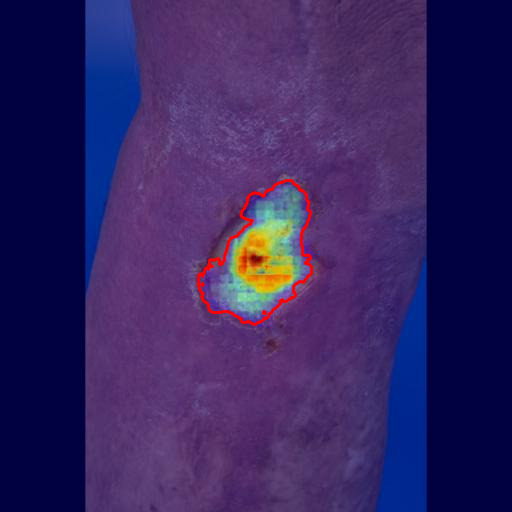}\\[4pt]
                \includegraphics[trim={220 105 100 65}, clip, height=1.5cm, angle=90]{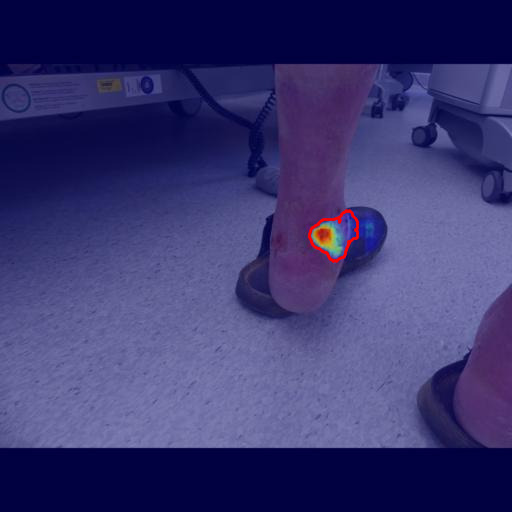}\\[4pt]
                \includegraphics[trim={110 0 85 40}, clip, height=1.5cm, angle=90]{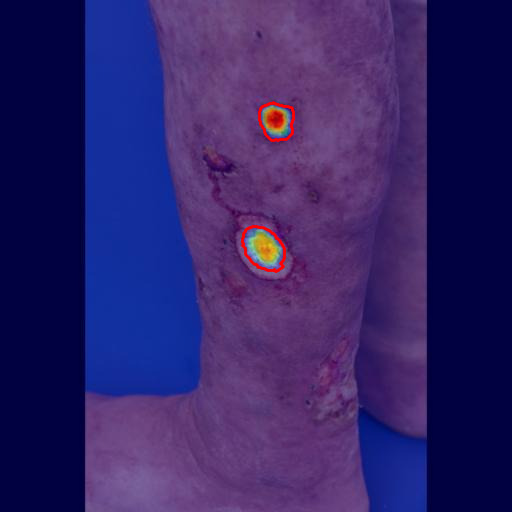}\\[4pt]
                \includegraphics[trim={160 115 90 130}, clip, height=1.5cm, angle=90]{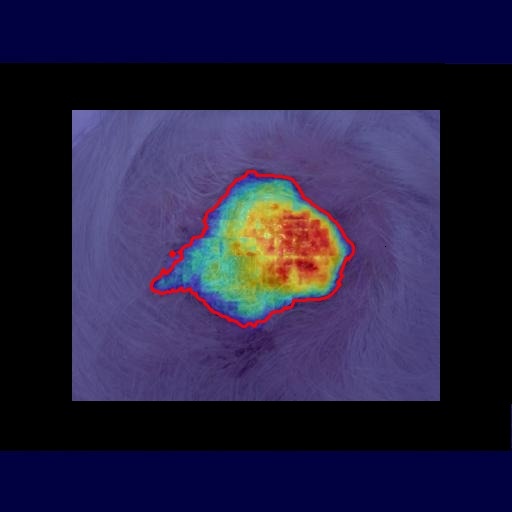}\\[4pt]
                \includegraphics[trim={160 90 20 90}, clip, height=1.5cm, angle=90]{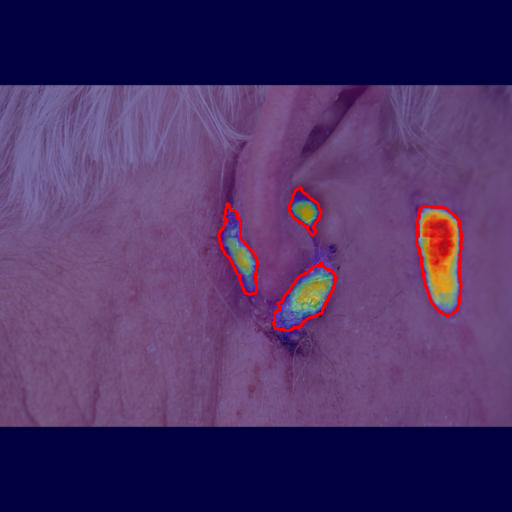}\\[4pt]
                \includegraphics[trim={64 10 64 0}, clip, height=1.5cm, angle=90]{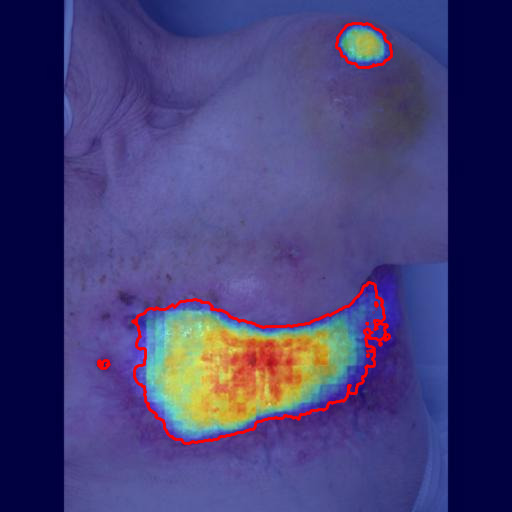}\\[4pt]
                \caption*{\scriptsize MISSForm.}
            \end{subfigure}
            \hspace{-0.6cm}
            \begin{subfigure}{0.165\textwidth}
                \centering
                \includegraphics[trim={80 120 70 140}, clip, width=1.5cm, angle=180]{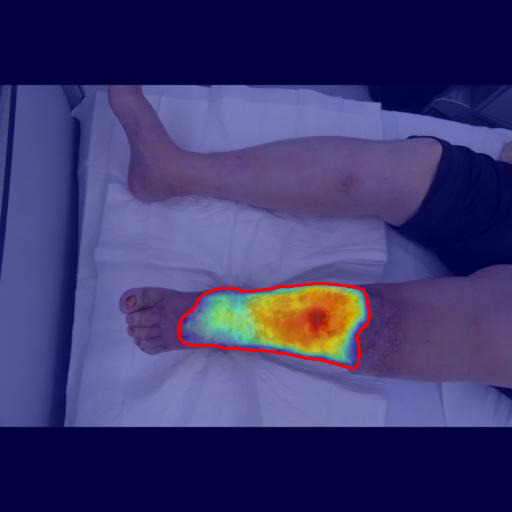}\\[4pt]
                \includegraphics[trim={80 165 0 85}, clip, width=1.5cm, angle=180]{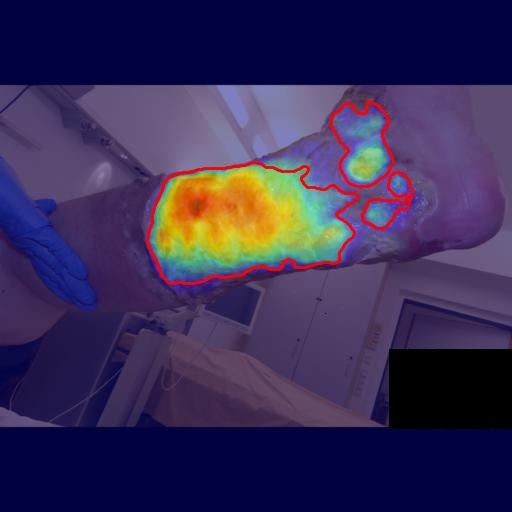}\\[4pt]
                \includegraphics[trim={30 65 0 65}, clip, width=1.5cm, angle=180]{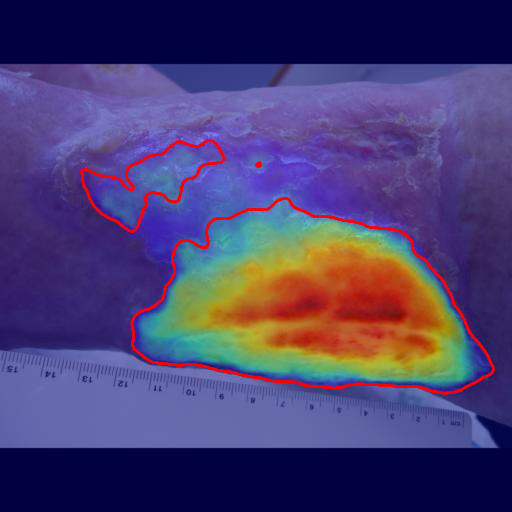}\\[4pt]
                \includegraphics[trim={0 90 0 90}, clip, width=1.5cm, angle=180]{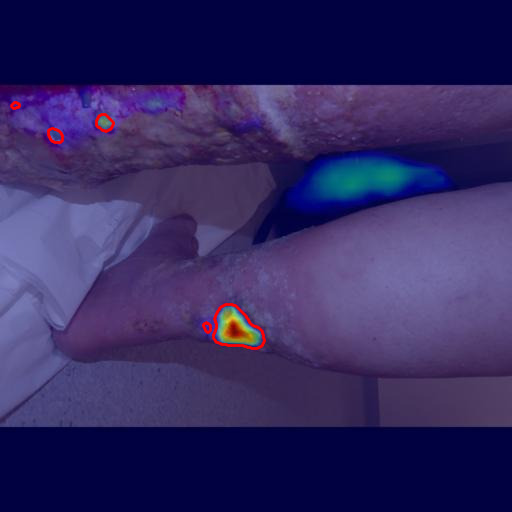}\\[4pt]
                \includegraphics[trim={150 90 60 85}, clip, height=1.5cm, angle=90]{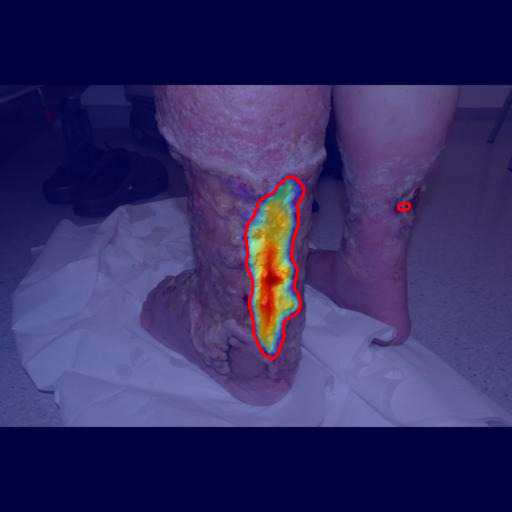}\\[4pt]
                \includegraphics[trim={110 90 120 90}, clip, height=1.5cm, angle=90]{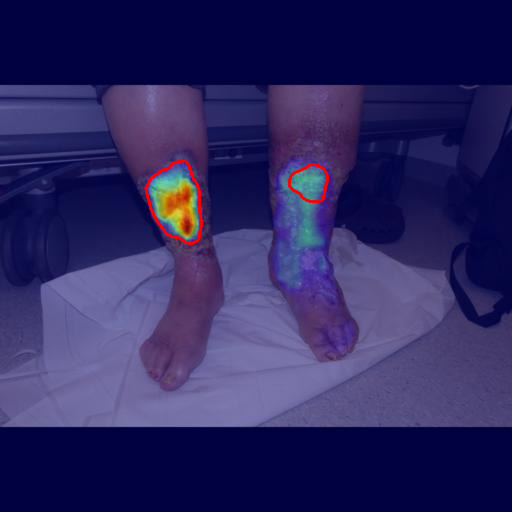}\\[4pt]
                \includegraphics[trim={90 0 90 0}, clip, height=1.5cm, angle=90]{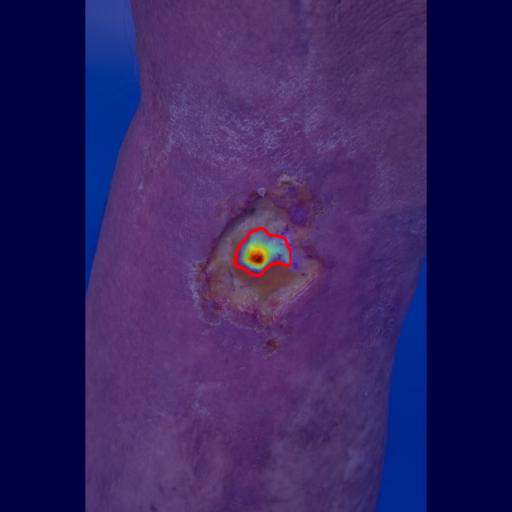}\\[4pt]
                \includegraphics[trim={220 105 100 65}, clip, height=1.5cm, angle=90]{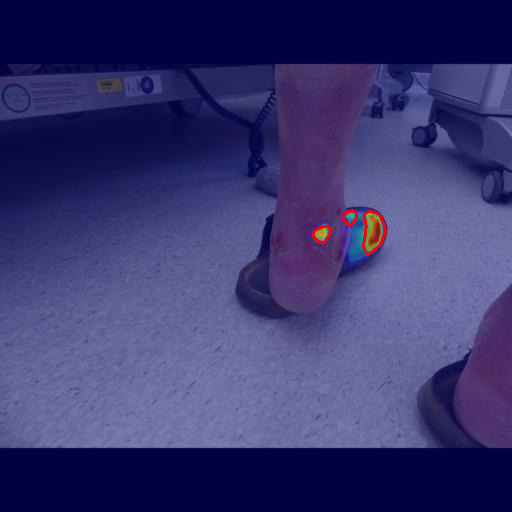}\\[4pt]
                \includegraphics[trim={110 0 85 40}, clip, height=1.5cm, angle=90]{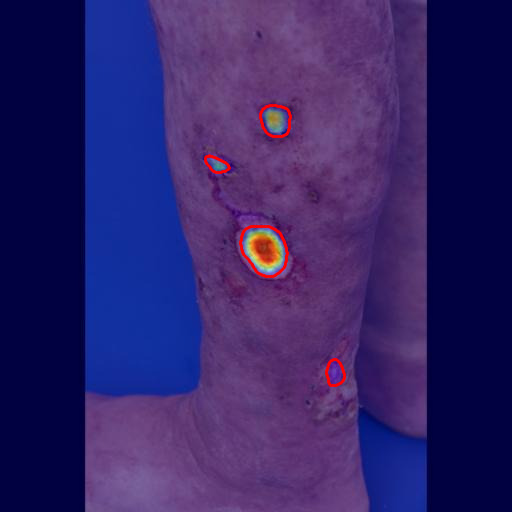}\\[4pt]
                \includegraphics[trim={160 115 90 130}, clip, height=1.5cm, angle=90]{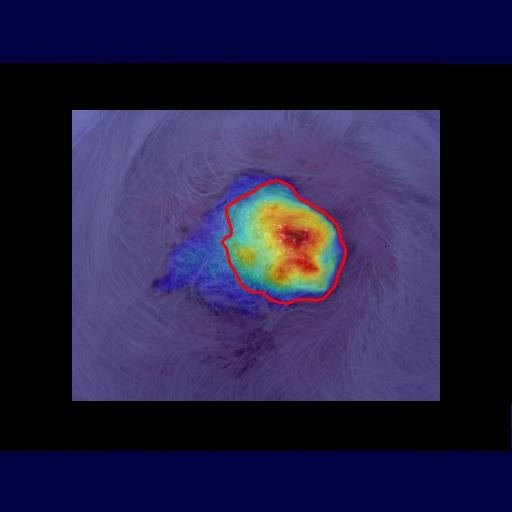}\\[4pt]
                \includegraphics[trim={160 90 20 90}, clip, height=1.5cm, angle=90]{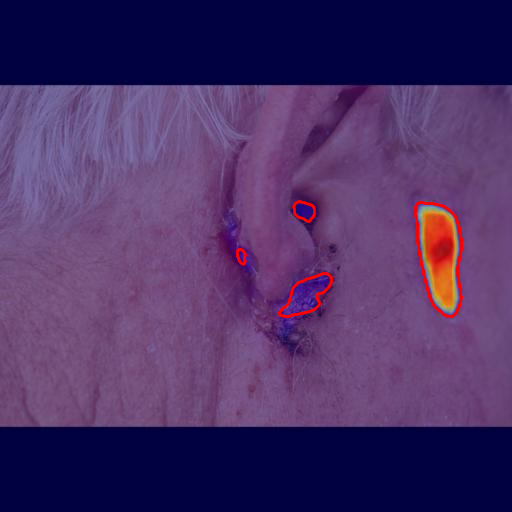}\\[4pt]
                \includegraphics[trim={64 10 64 0}, clip, height=1.5cm, angle=90]{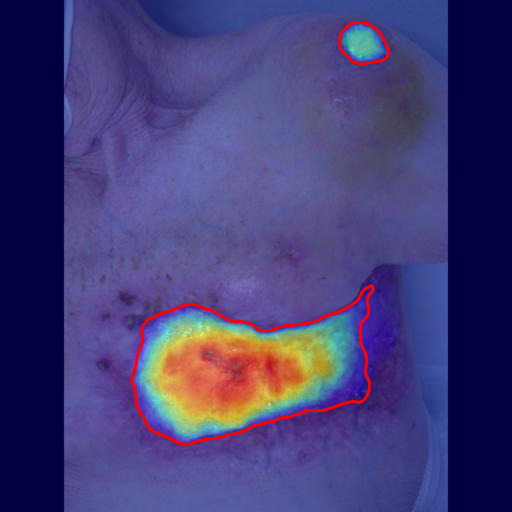}\\[4pt]
                \caption*{\scriptsize HiFormer}
            \end{subfigure}
        \end{adjustbox}
    \end{turn}
    \caption{\textit{\ac{gradcam}} for exemplary OOD images, alongside their GT annotation.}
    \label{fig:eval:grad_cam}
\end{figure}

To assess the decision-making of our models, we applied \textit{\ac{gradcam}} to generate visual explanations of predictions. Figure~\ref{fig:eval:grad_cam} presents visualizations for the top three and two lowest-performing models from Table~\ref{tab:eval:OOD:performance}, along with the predicted segmentation masks (red) for selected OOD images. Additional comparisons across architectures and further examples of different body localization, wound size, and color variations are provided in Section~A.5.

By comparing heatmaps with \ac{gt} masks (green), we qualitatively assess model reliability, focusing on whether activations align with clinically relevant wound features rather than artifacts. Across all models, strong activations were frequently observed in wound regions, aligning well with \ac{gt} masks. 
Additionally, non-wound regions often exhibited minimal to no activation, suggesting effective prioritization of pathological features. 
Qualitative comparisons indicate that higher-performing models from Table~\ref{tab:eval:OOD:performance} tend to generate sharper heatmap boundaries and achieve more precise wound localization, whereas lower-performing models exhibit weaker or more diffuse activations (3, 7). 
In some cases, non-wound objects (e.g., shoes, 5) were misidentified as wounds, while certain wound regions (6, 7, 8), including those near image edges (9), were not fully captured, especially by lower-performing models. Additionally, weaker models occasionally generated pixelated or irregular mask predictions, as indicated by frayed or poorly defined red contours (7, 11, 12).
While not necessarily generalizable, these qualitative insights align with quantitative performance metrics and enhance interpretability for clinicians, allowing for a deeper understanding of model behavior and fostering greater trust in AI systems.

\section{Real-World Deployment}
\label{sec:real_world_deployment}

\subsection{Automated Size Retrieval from AI-Generated Masks}

\begin{figure}[ht]
    \centering
    \begin{subfigure}{.4\textwidth}
        \centering
        \raisebox{0.1cm}{\includegraphics[width=\linewidth]{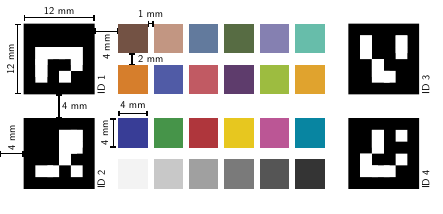}}
        \caption{Reference object (RO)}
        \label{fig:size_retrieval:sub1}
    \end{subfigure}%
    \begin{subfigure}{.4\textwidth}
        \centering
        \begin{minipage}{0.5\textwidth}
            \centering
            \includegraphics[width=0.95\linewidth, trim={0 0 0 0},clip]{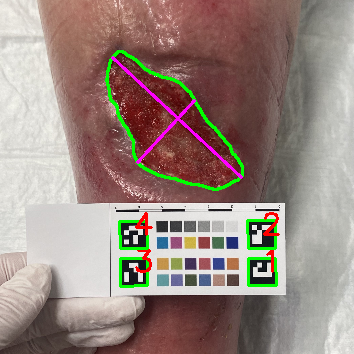}
        \end{minipage}%
        \begin{minipage}{0.5\textwidth}
            \centering
            \includegraphics[width=0.95\linewidth, trim={0 0 0 0},clip]{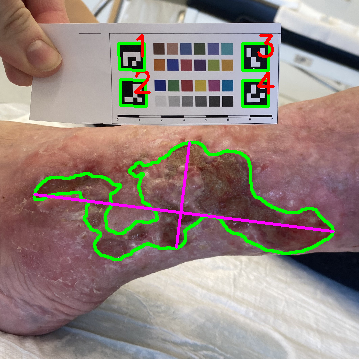}
        \end{minipage}
        \caption{Correct RO placement}
        \label{fig:size_retrieval:sub2}
    \end{subfigure}%
    \begin{subfigure}{.2\textwidth}
        \centering
        \includegraphics[width=0.95\linewidth, trim={0 0 0 0},clip]{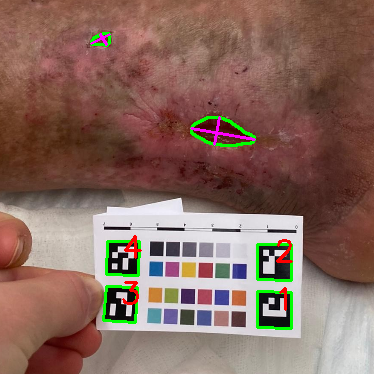}
        \caption{Neg. example}
        \label{fig:size_retrieval:sub3}
    \end{subfigure}%
    \caption{Visualization of the size retrieval process for exemplar real-world patients.}
    \label{fig:size_retrieval}
\end{figure}

\textbf{Reference Object (RO) Design}. 
Based on Chairat et al.~\cite{chairat2023ai}, we developed a custom \ac{ro} with slight adjustments to size and layout (see Fig.~\ref{fig:size_retrieval:sub1}).
Our \ac{ro} is placed next to the wound and features four \ac{aruco} markers, each with a 12 mm side length and a 4 mm margin of white space to facilitate accurate detection. 
To enhance functionality, we included a Macbeth color chart with 24 color patches between the markers, aligned such that the top and bottom edges of the patch groups align with the marker corners to simplify automated extraction.

\textbf{Benefits.}
Unlike existing methods~\cite{chino2020segmenting,foltynski2023internet}, our \ac{aruco}-based \ac{ro}, with its known dimensions, enables precise object size estimation independent of AI algorithms. Notably, this approach eliminates the need for including \ac{ro} detection as an additional class in model training. 
For future applications, the Macbeth color chart enables color calibration and wound color assessment. In environments with few natural reference points (e.g., human skin), the fixed design helps estimate camera position and rotation, which could be used for real-time patient guidance during image capture (e.g., adjusting distance or angle).

\textbf{Size Retrieval Process.}
We detect the \ac{aruco} markers using OpenCV's \texttt{ArucoDetector}. 
To estimate the pixel-to-millimeter (px/mm) ratio, we design a robust approach that utilizes multiple \ac{aruco} markers whenever possible, computing the ratio using predefined real-world distances between specific marker pairs.
By calculating the mean px/mm ratio across all available pairs, we reduce the impact of perspective distortion and measurement noise. Moreover, this approach enables more stable estimation, even in cases where some markers are partially occluded or missing. 
If only one marker is detected, we estimate the px/mm ratio by averaging its width and height.
%
Afterward, we convert segmented wound areas into square millimeters using this ratio.
In addition, we determine the wound's height and width following standard clinical measurement practices~\cite{langemo2008measuring}. We identify the longest diagonal of each mask contour (with at least seven points). Then, a perpendicular vector is computed and moved incrementally along the diagonal. At each step, we measure the distance between intersections of the perpendicular line and the contour, selecting the two points with the greatest separation to define the second diagonal, representing wound width.
Figure~\ref{fig:size_retrieval} demonstrates the size retrieval process for three representative patients. Green contours indicate detected markers and wound regions, while pink lines represent the diagonals approximating wound height and width. Red numbers show the retrieved marker IDs. 
Notably, accurate wound size estimation requires the \ac{ro} to be placed as close as possible to the wound and, crucially, within the same plane to minimize distortion. In contrast, Subfigure~\ref{fig:size_retrieval:sub3} shows a failure case where the wound is identified correctly, but its size is underestimated due to the markers being positioned closer to the camera than the wound.

\textbf{Evaluation.}
To evaluate performance, we collect $N=20$ diverse wound images using our \ac{ro} and an iPhone 11. The top five methods, selected based on \ac{pmv} mIoU scores from the OOD dataset, are assessed on these images using \acp{pmv} per architecture. Additionally, an ensemble of all $5\times5=25$ models is constructed.
For qualitative evaluation, $|D|=3$ dermatologists independently review AI-generated masks. A total of $120$ image-mask pairs ($N \times 6 \text{ AI variants}$) are presented in random order, with each mask rated as either ``good'' or ``bad''. Physicians also estimate wound sizes and select the best mask per image. Figure~B.1 provides screenshots of the annotation tool.
We define the following for evaluation: t: 
The \textit{Clinical Mask Approval (CMA)} measures how often physicians rate the masks as ``good'' across all images and evaluators. The \textit{Expert Choice Rate (ECR)} quantifies the proportion, averaged across all images and dermatologists, where a model's mask is selected as the best of six for a given image.
\begin{minipage}[b]{0.35\textwidth}
    \begin{equation}
        \scalemath{0.85}{\textbf{CMA} = \frac{\text{\# Good}}{|D| \times N} \times 100}
    \end{equation}
\end{minipage}
\hfill
\begin{minipage}[b]{0.6\textwidth}
    \begin{equation}
        \scalemath{0.85}{\textbf{ECR} = \frac{\text{\# Times } \text{ selected as best}}{|D| \times N} \times 100\%}
    \end{equation}
\end{minipage}\\

Since physicians provide separate height and width estimates instead of direct area measurements, we use these as proxies for size estimation accuracy. The \ac{gt} height and width for image $i$ are defined as the \textbf{mean} of all $|D|$ physician estimates, denoted as $H_{GT}^{(i)}$ and $W_{GT}^{(i)}$.
To assess size retrieval quality, we employ the Mean Absolute Error (MAE) and Mean Absolute Percentage Error (MAPE). For each dimension $X \in \{H, W\}$, these metrics quantify the deviation between model predictions $X_{pred}^{(i)}$ and the expert-derived GT $X_{GT}^{(i)}$:

\begin{minipage}[b]{0.4\textwidth}
    \begin{equation}
        \scalemath{0.85}{\textbf{MAE} = \frac{1}{N} \sum_{i=1}^{N} | X_{pred}^{(i)} - X_{GT}^{(i)}|}
    \end{equation}
\end{minipage}
\hfill
\begin{minipage}[b]{0.55\textwidth}
    \begin{equation}
        \scalemath{0.85}{\textbf{MAPE} = \frac{1}{N} \sum_{i=1}^{N} \frac{| X_{pred}^{(i)} - X_{GT}^{(i)}|}{X_{GT}^{(i)}} \times 100\%}
    \end{equation}
\end{minipage}\\

However, size estimation in medical imaging is inherently subjective, leading to significant variability among physicians~\cite{jorgensen2016methods}.
To ensure a meaningful evaluation, 
we quantify inter-rater variability using a relative deviation metric. We exclude any image $i$ where the following deviation exceeds a threshold of 0.5 in either height or width annotations, with $X \in {H, W}$ representing the dimension, and $X_{d_j}^{(i)}$ denoting the size annotation provided by physician $d_j$ for image $i$:
\begin{equation}
        \scalemath{0.85}{\textbf{Relative Deviation} = \frac{\text{max}\big(X_{d_1}^{(i)}, X_{d_2}^{(i)}, X_{d_3}^{(i)}\big) - \text{min}\big(X_{d_1}^{(i)}, X_{d_2}^{(i)}, X_{d_3}^{(i)}\big)} {\text{median}\big(X_{d_1}^{(i)}, X_{d_2}^{(i)}, X_{d_3}^{(i)}\big)}}
\end{equation}

\begin{figure}[ht]
    \centering
    \begin{subfigure}{.13\textwidth}
        \centering
        \includegraphics[width=\linewidth]{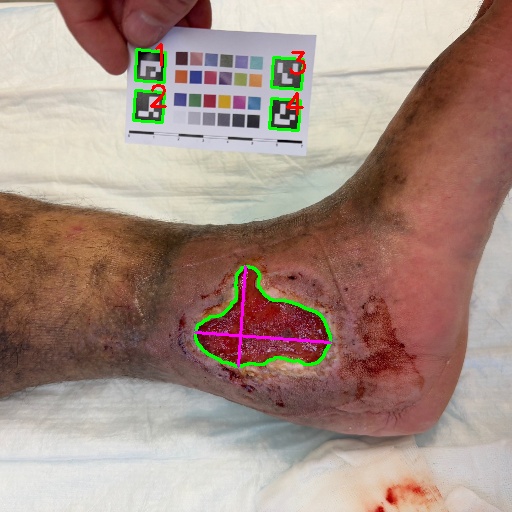}
    \end{subfigure}%
    \hfill
    \begin{subfigure}{.13\textwidth}
        \centering
        \includegraphics[width=\linewidth]{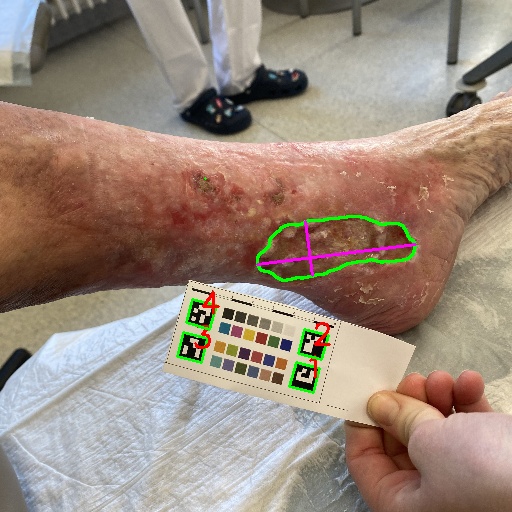}
    \end{subfigure}%
    \hfill
    \begin{subfigure}{.13\textwidth}
        \centering
        \includegraphics[width=\linewidth]{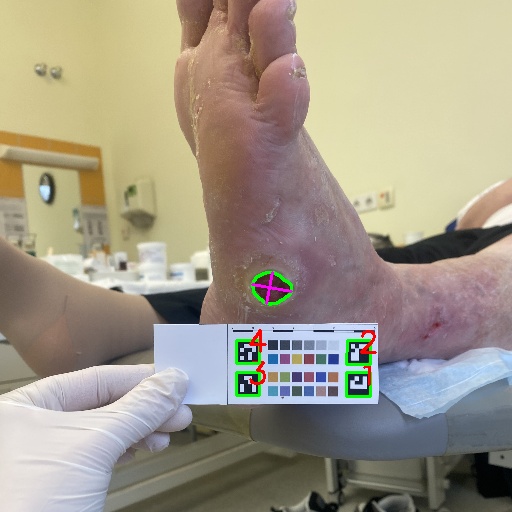}
    \end{subfigure}%
    \hfill
    \begin{subfigure}{.13\textwidth}
        \centering
        \includegraphics[width=\linewidth]{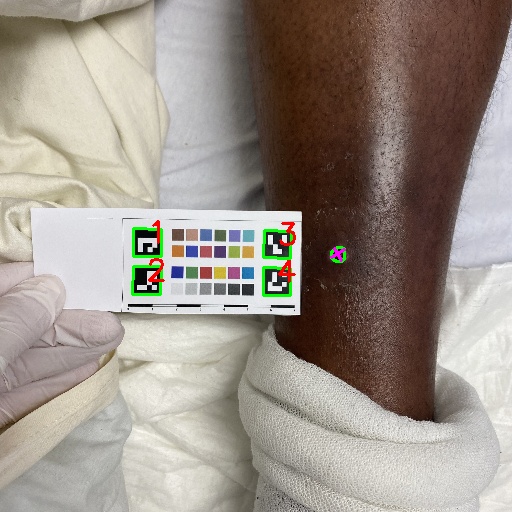}
    \end{subfigure}%
    \hfill
    \begin{subfigure}{.13\textwidth}
        \centering
        \includegraphics[width=\linewidth]{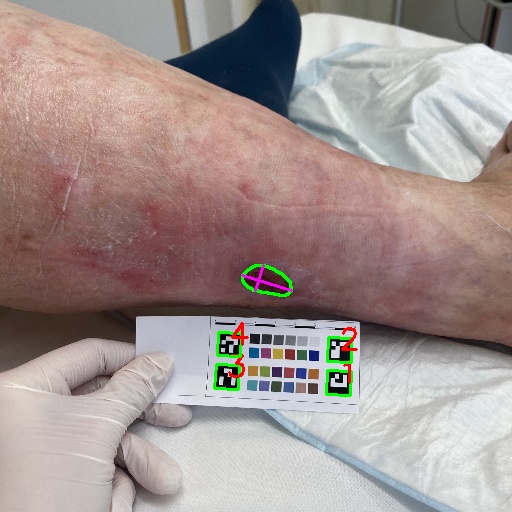}
    \end{subfigure}%
    \hfill
    \begin{subfigure}{.13\textwidth}
        \centering
        \includegraphics[width=\linewidth]{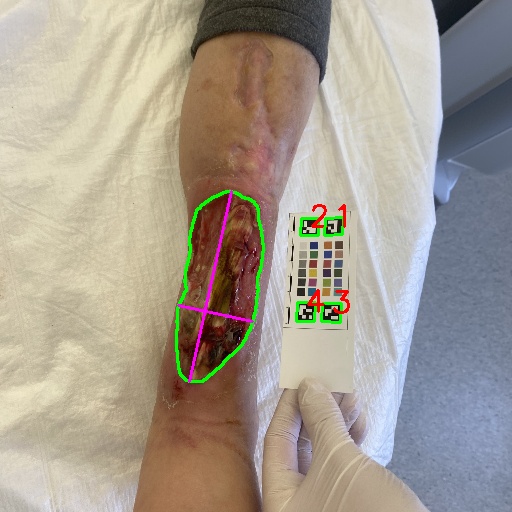}
    \end{subfigure}%
    \hfill
    \begin{subfigure}{.13\textwidth}
        \centering
        \includegraphics[width=\linewidth]{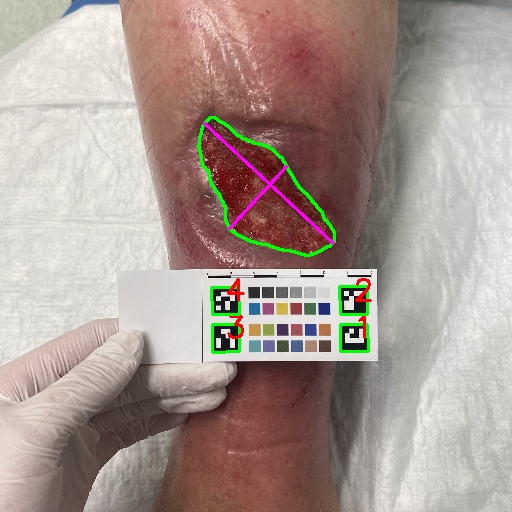}
    \end{subfigure}%
    \caption{Overview of the seven wounds with consistent expert size estimates.}
    \label{fig:size_retrieval:consistently:annotated:images}
\end{figure}

After removing inconsistently annotated images, we use the remaining seven for error calculations (see Table~\ref{tab:eval:size_retrieval:rel_dev:0.5}). Figure~\ref{fig:size_retrieval:consistently:annotated:images} confirms their variability in wound shape, size, and skin tone, ensuring a representative subset for size retrieval evaluation.
We also report the mean predicted height and width (MPH, MPW) with SD across the $N=7$ images, alongside the average wound dimensions estimated by the $|D|$ physicians.
This analyzes model consistency across wound types and retrieval capabilities relative to expert estimates. 
For completeness, Table~B.1 includes the masks, raw size predictions (including area), and expert estimates for all 20 images, which are also shared at \href{https://github.com/VanessaBorst/woundambit}{GitHub} with written consent.

The CMA ranged from 83.3\% to 86.7\% across models, reflecting high mask quality and strong physician approval. ECR varied notably, with \ac{vwconv} achieving the highest approval rate (35.0\%) and \ac{transnext} and the Ensemble the lowest (8.3\%).
For size retrieval, all models performed similarly, with MPH and MPW closely aligning with the \ac{gt}. On average, height estimates were slightly overestimated, while width was slightly underestimated. Nevertheless, consistently low MAE ($\leq$4.4 mm height, $\leq$3.2 mm width) and MAPE ($\leq$14.1\% height, $\leq$16.2\% width) indicate robust size retrieval across wound sizes.

\begin{table}[htb]
    \centering
    \small
    \caption[]{Expert-based mask and size retrieval assessment.}
    \label{tab:eval:size_retrieval:rel_dev:0.5}
    \begin{adjustbox}{max width=0.95\textwidth}
        \begin{threeparttable}
            \begin{tabular}{l@{\hskip 0.2in}
            l@{\hskip 0.1in}c@{\hskip 0.2in}
            c@{\hskip 0.1in}c@{\hskip 0.1in}c@{\hskip 0.2in}
            c@{\hskip 0.1in}c@{\hskip 0.1in}c}
            \toprule
            \multirow{4}{*}{\textbf{Model}} & \multicolumn{2}{c}{Mask Quality} & \multicolumn{6}{c}{Size Retrieval Quality ($N=7$)}   \\
            \cmidrule(lr){4-9}
             & \multicolumn{2}{c}{($N=20$)}
             & \multicolumn{3}{c}{Height} & \multicolumn{3}{c}{Width} \\
            \cmidrule(r){2-3}\cmidrule(r){4-6}\cmidrule(lr){7-9}
                & CMA\tnote{1} & ECR\tnote{1} & MPH \tnote{2,3} & MAE\tnote{2} & MAPE\tnote{1}  & MPW \tnote{2,4} & MAE\tnote{2} & MAPE\tnote{1}\\
            \midrule
            \ac{transnext}           & 85.0   &  8.3            & \textbf{54.6 ± 41.0} & 4.4 & 12.3 & 27.2 ± 15.6 & 3.2 & 15.8 \\
            \ac{internimage}         & \textbf{86.7}    & 15.0  & 54.9 ± 40.9 & 3.9 & 12.2 & \textbf{26.6} ± 15.0 & 3.2 & 16.2 \\
            \ac{vwmit}               & \textbf{86.7}    & 13.3  & 55.4 ± 41.5 & 4.3 & 14.1 & 27.7 ± 16.0 & \textbf{2.3} & 13.3 \\
            \ac{segformer}           & 83.3   &  13.3           & 54.8 ± 41.2 & \textbf{3.8} & 12.0 & 27.1 ± 15.8 & 2.6 & \textbf{13.2} \\
            \ac{vwconv}              & 85.0   &  \textbf{35.0}  & 54.8 ± 41.6 & 4.0 & \textbf{9.9}  & \textbf{26.6} ± 15.8 & 3.2 & 14.4 \\             
            \midrule
            Ensemble                 & \textbf{86.7}   &  8.3   & 55.0 ± 41.3 & 4.1 & 12.2 & 27.0 ± 15.7 & 2.9 & 15.1 \\
            \bottomrule
            \end{tabular}
            \begin{tablenotes}[para]
                \small
                \item[1] In \%
                \item[2] In mm
                \hspace{1.4cm}
                \item[3] Mean $H_{GT}$: 54.3 ± 41.6
                \hspace{0.5cm}
                \item[4] Mean $W_{GT}$: 29.2 ± 16.3
            \end{tablenotes}
        \end{threeparttable}
    \end{adjustbox}
\end{table}

\subsection{Seamless Integration into a Custom Telehealth System}
\begin{figure}[ht]
    \centering
    \includegraphics[width=0.85\textwidth]{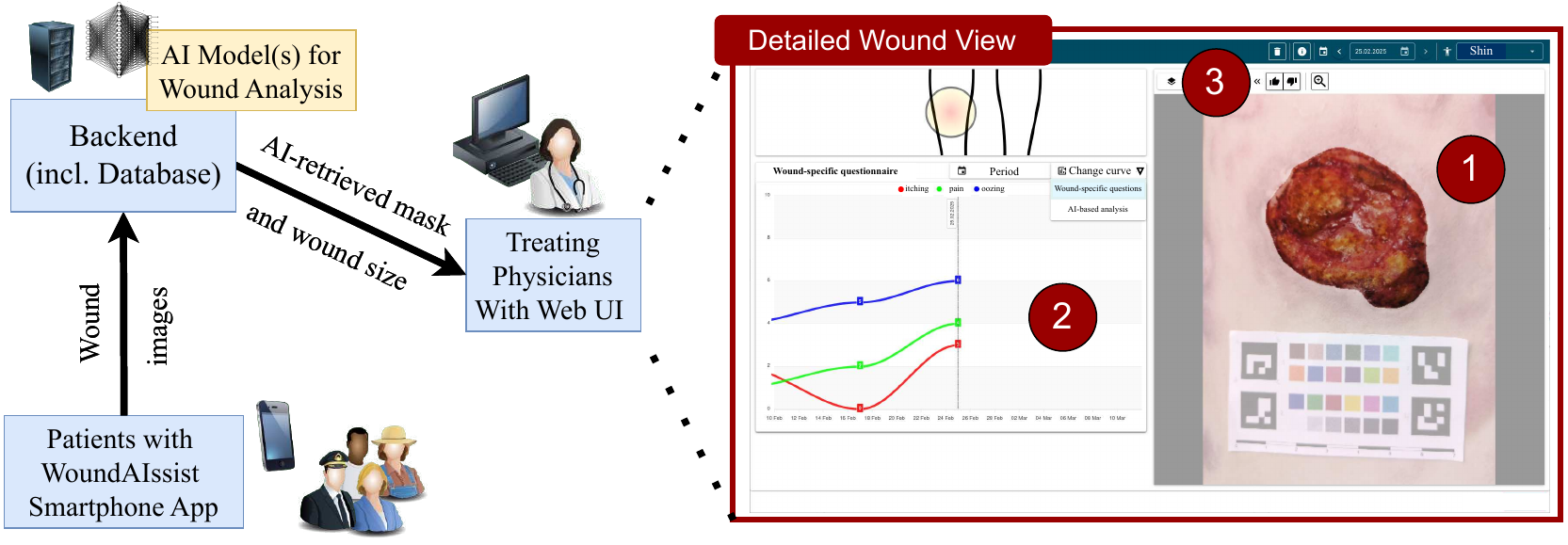}
    \caption{Simplified visualization of the overall telehealth system.}
    \label{fig:dashboard}
\end{figure}

We integrated the AI module, which demonstrated superior performance in terms of ECR, \ac{vwconv}, and wound size retrieval, into a custom-developed telehealth platform to enable future evaluation of its impact on patient outcomes.
The system architecture (see Figure~\ref{fig:dashboard}) consists of three core components: (1) the \textit{WoundAIssist} mobile application~\cite{borst2025woundaissist}, (2) a web interface for physicians, and (3) a backend with a dedicated microservice for AI processing.
Among other features, patients use our purpose-built app, \textit{WoundAIssist}, to capture and upload wound images at regular intervals, accompanied by self-reported data on wound-specific factors (e.g., pain, itching, oozing) and overall well-being (e.g., mood, activity impact, quality of life)~\cite{borst2025woundaissist}. \ac{hcp} use a web-based dashboard for remote patient monitoring, providing a concise summary of essential patient information. The dashboard also includes a detailed wound view (Fig.~\ref{fig:dashboard}, right, translated), which displays the selected wound images alongside the AI-predicted wound area (\textcircled{\raisebox{-0.9pt}{1}}). Additionally, the dashboard features trajectory curves for all patient-reported wound scores and AI-derived wound size progression (\textcircled{\raisebox{-0.9pt}{2}}).
To support continuous model refinement, the system incorporates a feedback mechanism, allowing clinicians to assess the segmentation masks by a simple approval process (\textcircled{\raisebox{-0.9pt}{3}}).

The system is currently deployed on a secure, GDPR-compliant cloud server and is actively used as part of an ongoing longitudinal clinical study. Approximately 30 patients are expected to participate, each using the \textit{WoundAIssist} app on their personal smartphones over six months. This long-term clinical trial aims to empirically assess the effectiveness of the overall telehealth solution, with a focus on patient adherence and wound care outcomes. It also enables the collection of real-world images captured by patients, which provide a valuable basis for both validating the performance of the proposed AI-based wound size estimation and informing future improvements to enhance its robustness.
\section{Discussion and Outlook}
\label{sec:discussion_outlook}
This study provides a comprehensive benchmark of various \ac{dl} models for wound segmentation, addressing real-world challenges in clinical deployment and telehealth applications.
Interestingly, many of the top-performing models achieved comparable performance on the evaluated datasets, suggesting that binary wound segmentation may be a relatively straightforward task for advanced deep learning methods. This phenomenon may be attributed to the distinctive visual characteristics exhibited by wounds in conventional samples, which enable modern architectures to extract relevant features effectively.

Although no clear trend emerged between CNN- and ViT-based models, the results on both the in-distribution CFU and the \ac{dfuc22} live leaderboard indicate the superior performance of modern \ac{gp} architectures, such as \ac{transnext}, over specialized medical models. Similarly, the top five models on the \ac{ood} dataset were \ac{gp} architectures, further emphasizing the superiority of contemporary SS and vision models over specialized medical techniques in handling OOD data. This enhanced performance is further supported by visual analysis of Grad-CAM heatmaps.
Despite the higher computational demands of the top-performing models, inference times remained within practical limits. With a minimum throughput of at least one image per second on the CPU, all models are considered suitable for clinical use and remote wound size monitoring, given the manageable data volumes.  
Expert evaluations of mask quality and automated size estimates from the top five models yielded promising results, with high mask approval rates and size estimates closely aligning with dermatologist annotations. While no architecture clearly outperformed the others in size retrieval, \ac{vwconv} was most favored by experts for best mask selection.

While our wound size estimation framework, \ac{woundambit}, holds great potential, 
it also presents limitations and highlights directions for future research:

\textbf{RO Placement and Robustness Analysis Using Real-World Data.}
The evaluation primarily used publicly available and \ac{ood} images captured in controlled hospital settings. To improve model robustness and assess size estimation in telemedicine contexts, it is essential to include images from home environments, particularly those taken by patients using smartphones. 
These images exhibit greater variability in lighting, background, and image quality, which may introduce practical challenges, such as suboptimal RO placement or partial occlusion of ArUco markers.
Accurate size estimation requires the RO to be approximately coplanar with the wound surface; however, the current system lacks a mechanism to guide patients in placing the RO correctly.
Incorrect positioning can cause perspective distortion and reduced measurement accuracy. Future versions could incorporate real-time user guidance, such as augmented reality feedback, to assist patients during image capture.
%
Moreover, our evaluation included only a single image with partially occluded ArUco markers, limiting our ability to quantify the impact of incomplete or failed marker detection. 
Future work should involve systematic stress testing to assess how partial occlusions, non-coplanar RO placement, or complete detection failure affect size estimation accuracy. To improve robustness in such scenarios, alternative strategies should be explored to maintain functionality.  
The patient-captured images from our ongoing longitudinal clinical study can provide a diverse and realistic dataset to evaluate and improve model robustness under suboptimal, real-world conditions.

\textbf{Model Selection.}
While efforts were made to select a representative set of models, the evaluation is constrained by the chosen architectures. The three lowest-performing models -- \ac{unet}, \ac{missformer}, and \ac{hiformer} -- are notably smaller in terms of parameters, which may introduce a slight bias. However, the general trend favoring the GP SS and GP vision models remains consistent, even regarding the larger medical networks.

\textbf{Over-Reliance on Technology.}
From a clinical perspective, \ac{woundambit} serves as a decision support system, not a replacement for physicians. While it provides automated wound size tracking, clinicians continue to review both the original wound image and the AI-generated output, ensuring that clinical decisions are not based solely on AI predictions. Nevertheless, the potential risk of over-reliance on technology should be further explored in future studies. 

\textbf{Clinical Validation.}
Ultimately, clinical validation of \ac{woundambit} is crucial to confirm its effectiveness in practice beyond the scope of technical evaluation. The ongoing longitudinal study will provide critical insight, and follow-up analyses should explore patient adherence over extended periods as well as the system’s overall impact on wound management outcomes and patient well-being.

\section{Conclusion}
\label{sec:conclusion}
This work advances automated wound segmentation by benchmarking 12 \ac{sota} \ac{dl} methods from \ac{gp} vision, medical segmentation, and wound care. We address challenges such as domain variability, computational constraints, and model interpretability. Our evaluation on both in-distribution and \ac{ood} datasets demonstrates that modern GP models, especially \ac{transnext} and \ac{vwformer}, exhibit strong generalization and remain computationally feasible for clinical use.
While our AI-driven wound size estimation framework, \ac{woundambit}, shows strong clinical promise, future work will focus on expanding the dataset with smartphone images from diverse patient demographics and conditions for model finetuning. Additionally, integrating \ac{woundambit} with existing clinical systems and assessing patient-reported outcomes in clinical trials will be crucial for evaluating its real-world impact. Lastly, incorporating the \ac{ood} dataset and prospectively collected smartphone images into the training is expected to improve model performance further.
Overall, this work represents an important step toward bridging recent advances in computer vision with real-world wound care, offering the potential to improve remote wound size monitoring and enhance patient care.

\begin{credits}
\subsubsection{\ackname} 
We thank Luisa Deutzmann for her support in creating the 20-image dataset for size retrieval evaluation and Caroline Glatzel for her contributions to manual size annotation and assessment of AI-generated masks.

\subsubsection{Ethics} 
The authors declare no conflicts of interest. 
Ethical approval for this study was waived by the Ethics Committee of the Medical Faculty of the University of Würzburg. All patients provided informed consent for the publication of their photographs.

\subsubsection{Supplementary Materials}
The trained k-fold models can be accessed via Zenodo at \href{https://zenodo.org/records/15123640}{10.5281/zenodo.15123640}.
The complete source code is archived on Zenodo at \href{https://zenodo.org/records/15682317}{10.5281/zenodo.15682317} and actively maintained on \href{https://github.com/VanessaBorst/woundambit}{GitHub}.
Additional materials 
are provided in the supplementary document available at \href{https://zenodo.org/records/15673941}{10.5281/zenodo.15673941}.

\end{credits}

%
%
%

\bibliographystyle{splncs04}
\bibliography{woundambit}

\appendix
\title{WoundAmbit: Bridging State-of-the-Art Semantic Segmentation and Real-World Wound Care\\- Supplementary Material -}
\titlerunning{WoundAmbit: Supplementary Material}
\author{Vanessa Borst\inst{1}\orcidlink{0009-0004-7123-7934} \corr \and
Timo Dittus\inst{1}\orcidlink{0009-0008-0704-1856} \and
Tassilo Dege\inst{2}\orcidlink{0000-0001-6158-9048} \and \\
Astrid Schmieder\inst{2}\orcidlink{0000-0002-6421-9699} \and
Samuel Kounev\inst{1}\orcidlink{0000-0001-9742-2063}
}
\authorrunning{V. Borst et al.}

\institute{Julius-Maximilians-University Würzburg, 97070 Würzburg, Germany 
\and
University Hospital of Würzburg, 97070 Würzburg, Germany 
}

\maketitle
\counterwithin{figure}{section}
\counterwithin{table}{section}
\counterwithin{equation}{section}
\section{Deep Learning Benchmark}

\subsection{Architecture-Specific Details and Deviations}

As mentioned in the main paper, all architectures are implemented with their architecture-specific settings, as recommended in their respective official publications and code repositories.
However, for the sake of completeness, we briefly detail the core ideas and some architecture-specific remarks below:\\

\textbf{\acl{hardnet}~\citesupp{appendix_liao2022hardnet}.} 
\acl{hardnet} was the winner of the \acs{dfuc22} challenge. It uses the HarDNet-MSEG model~\citesupp{appendix_huang2021hardnet} as a basis. The authors enhance the model, among others, by modifying each HarDBlock in the encoder with a novel HarDBlockV2 and replacing the Receptive Field Block (RFB) modules in the decoder with that of the Lawin Transformer. Additionally, the architecture incorporates a Squeeze-and-Excitation (SE) attention module after each block output to exploit multi-scale information better. During training, they employ deep supervision to calculate multiple losses from different stages of the network.\\

\textbf{\acs{fusegnet}~\citesupp{appendix_dhar2024fusegnet}.} 
\ac{fusegnet}, or more precisely, an ensemble of five cross-fold instances (x-FUSegNet), achieved the best performance in the \acs{fuseg}'21 challenge, aggregating predictions by averaging the outputs of the five constituent models. 
\ac{fusegnet} employs a pre-trained EfficientNet-b7 encoder, paired with a decoder that incorporates modified spatial and channel SE attention (scSE) modules, referred to as parallel scSE (P-scSE), at the middle of each decoder stage. This novel attention mechanism combines both additive and max-out scSE, except for the last stage, where max-out scSE is bypassed due to the reduced number of feature maps.

\textit{Notably, we report 71.0M parameters for \acs{fusegnet}, as measured using the \texttt{calflops} library, for consistency with other methods. In contrast, the original publication reports 64.9M parameters, a value obtained using \texttt{torchsummary} with the same model code.}\\

\newpage
\textbf{\acs{fcbformer}~\citesupp{appendix_sanderson2022fcbformer}.} 
The \ac{fcbformer} was originally proposed for polyp segmentation. Its encoder consists of two complementary branches: a Transformer Branch (TB) and a Fully Convolutional Branch (FCB).
The TB is based on the SSFormer architecture~\citesupp{appendix_wang2022ssformer}, employing an unmodified Pyramid Vision Transformer (PVTv2-B3)~\citesupp{appendix_wang2022pvtv2} as its backbone and integrating an enhanced local emphasis module within the progressive locality decoder.
The FCB comprises a series of residual blocks (RBs), each containing group normalization layers, SiLU activation functions, and convolutional layers. A skip connection is applied within each RB, and the overall structure follows a U-Net-style design, with skip connections between the encoder and decoder layers.
The final prediction head generates segmentation maps by concatenating the upsampled features from the TB with the output from the FCB, producing a full-resolution output.\\

\textbf{\acs{hiformer}~\citesupp{appendix_heidari2023hiformer}.} 
\ac{hiformer} is a resource-efficient medical segmentation model that bridges CNNs and vision transformers by incorporating both to extract local and global contextual feature representations, respectively. Specifically, its hybrid encoder comprises two hierarchical components: a CNN (ResNet50 by default, including in HiFormer-B) and a Swin Transformer. These are connected through a Double-Level Fusion (DLF) module embedded within the skip connections of the encoder-decoder structure, enabling effective fusion of local and global features while maintaining feature consistency. The decoder, inspired by the Semantic FPN~\citesupp{appendix_kirillov2019panoptic}, combines multi-scale features to produce a unified mask.

\textit{In preliminary experiments, we evaluated both HiFormer-B and HiFormer-L, selecting HiFormerB for our final benchmark due to its higher mIoU performance on the validation split of our main dataset while maintaining a lower number of trainable parameters.}\\

\textbf{\acs{missformer}~\citesupp{appendix_huang2022missformer}.} 
\ac{missformer} is a hierarchical transformer-based encoder-decoder architecture tailored for medical image segmentation. It introduces a modified feed-forward layer, termed ReMix-FFN, within each transformer block to more effectively integrate local context and global dependencies. The hierarchical encoder is composed of multiple stages, each combining transformer blocks with ReMix-FFN and patch merging layers for progressive feature extraction. Similarly, the decoder utilizes transformer blocks with ReMix-FFN and patch expanding layers to restore spatial resolution.
To enhance multi-scale feature fusion and bridge the encoder and decoder, MISSFormer incorporates a ReMixed Transformer Context Bridge between them, which integrates local and global correlations across feature scales.

\textit{
In their original study, the authors report that \ac{missformer}, when trained from scratch, outperforms state-of-the-art methods pre-trained on ImageNet. In our evaluation, however, we observed that pre-training still led to improved validation mIoU performance. We also compared internal input sizes of 224 and 512, selecting 224 for the final benchmark due to its comparable mIoU performance with considerably reduced resource demands.}\\

\textbf{\acs{segformer}~\citesupp{appendix_xie2021segformer}.} 
The \ac{segformer} is a transformer-based architecture developed for semantic segmentation (SS). It utilizes a lightweight decoder consisting solely of multi-layer perceptron (MLP) layers, which produces segmentation maps at one-quarter of the input resolution without relying on computationally intensive components. These maps are subsequently upsampled using bilinear interpolation.
The decoder integrates multi-scale features extracted from different stages of a hierarchical, position-encoding-free transformer encoder referred to as the Mix Transformer (MiT). This encoder effectively captures both high-resolution coarse features and low-resolution fine-grained details.
Rather than employing traditional positional encodings, MiT incorporates novel Mix-FFN layers within each transformer block. These layers mix a $3\times3$ convolution and an MLP into each feed-forward network (FFN). In addition to Mix-FFN, each transformer block integrates efficient self-attention mechanisms and overlapping patch merging.
By omitting absolute positional encodings, MiT eliminates the need for positional embedding interpolation during inference on images with resolutions that differ from those seen during training, thereby enabling robust segmentation performance across arbitrary input sizes. \\

\textbf{\acs{segnext}~\citesupp{appendix_guo2022segnext}.} 
\ac{segnext} is a CNN architecture for SS that employs a novel Multi-Scale Convolutional Attention Network (MSCAN) as its encoder and a Hamburger module~\citesupp{appendix_geng2021attention} as its decoder. 
The encoder employs a four-stage pyramid structure, where each stage comprises a downsampling block followed by stacked ViT-inspired building blocks. Notably, these blocks replace self-attention with Multi-Scale Convolutional Attention (MSCA) modules and use batch normalization instead of layer normalization.
Each MSCA module contains three components: a depth-wise convolution for local information aggregation, multi-branch depth-wise strip convolutions for capturing multi-scale context, and a $1\times1$ convolution for modeling channel relationships, producing attention weights for reweighting the input.
The decoder aggregates features from the last three encoder stages and applies a Hamburger module for extracting global information. This module models context discovery as a low-rank completion problem, solved via matrix decomposition, which enhances computational efficiency.\\

\textbf{\acs{vwformer}~\citesupp{appendix_yan2024vwformer}.} 
\ac{vwformer} is a novel multi-scale decoder architecture designed for SS. It employs Varying Window Attention (VWA) and multiple MLP modules to learn improved multi-scale representations. 
Specifically, the decoder processes multi-level feature maps from the last three encoder stages through a multi-layer aggregation using MLPs, followed by VWA-based multi-scale representation learning, and concludes with low-level feature enhancement via an additional set of MLPs.
The core idea of VWA is to keep the query positioned on a local window but enlarge the context window size to encompass a broader surrounding area, thereby varying the receptive field of the query. 
To mitigate the increased computational cost typically introduced by enlarged context windows without compromising segmentation performance, the authors introduce pre-scaling, densely overlapping patch embedding (DOPE), and copy-shift padding (CSP), the latter of which is designed to prevent attention collapse.
In their original study, the authors evaluate \ac{vwformer} with different hierarchical backbones, including ConvNeXt, Swin Transformer, and the MiT encoder from \ac{segformer}.

\textit{As the corresponding GitHub repository did not provide the exact configurations used for the MiT combination, an exact replication of their settings in our benchmark was not feasible. For our evaluation, we set the \texttt{nheads} parameter to one and enabled the \texttt{short\_cut} parameter.}\\

\textbf{\acs{internimage}~\citesupp{appendix_wang2023internimage}.} 
\ac{internimage} is a large-scale CNN-based vision foundation model that adopts DCNv3---an enhanced deformable convolution (DCN) with a fixed kernel size of $3\times3$---as its core operator. 
These dynamic and sparse convolutions enable the model to learn effective receptive fields, including long-range dependencies, and to perform adaptive spatial aggregation in a manner similar to ViTs. Crucially, the sampling offsets and modulation scalars are adaptively adjusted according to the input data.
The basic units of \acs{internimage} resemble transformer-style blocks, integrating components such as layer normalization, FFNs, and GELU activation, in addition to DCNv3.
\ac{internimage} adopts a hierarchical architecture to extract multi-scale feature maps. This hierarchy is constructed using stacked basic units, preceded by a convolutional stem, and connected via intermediate convolutional downsampling layers across four stages. 
To systematically define model variants, the authors propose a rule-based configuration scheme governed by four hyperparameters, along with scaling rules to construct models of varying capacities.\\

\textbf{\acs{transnext}~\citesupp{appendix_shi2024transnext}.} 
\ac{transnext} introduces a novel visual backbone that integrates Aggregated Attention as a token mixer and a Convolutional Gated Linear Unit (GLU) as an advanced channel mixer.
Aggregated Attention, inspired by human foveal vision and continuous eye movement, is an attention mechanism that operates on a per-pixel basis. 
It enables fine-grained perception through attention to local neighboring features, while a parallel path performs coarse-grained attention on spatially downsampled features to add global perception.
To further enrich the attention dynamics, the model introduces learnable tokens that interact with standard queries and keys. This design diversifies the generation of affinity matrices beyond simple query-key similarity, enabling the integration of multiple attention mechanisms within a single attention layer. 
The Convolutional GLU module bridges the gap between GLU and traditional SE mechanisms.
It enables each token to compute channel attention based on features from its nearest neighbors, thereby improving local feature modeling and enhancing model robustness. 
Overall, the \ac{transnext} architecture follows a hierarchical design with four stages and uses overlapping patch embeddings similar to those in PVTv2~\citesupp{appendix_wang2022pvtv2}. While Convolutional GLU with GELU activation is consistently applied as the channel mixer across all stages, Aggregated Attention is employed only in the first three. In the final stage, a modified form of multi-head self-attention, aligned with PVTv2, is used instead.\\

\subsection{Details on the Unified Training Settings}

\subsubsection{Preprocessing and Data Augmentation.} 
All images were resized to 512×512 pixels and normalized for both training and evaluation. Annotated masks were thresholded to eliminate out-of-class pixels. To improve model robustness, we applied data augmentation to each \emph{training} sample, including Gaussian blur (kernel size 25, $\sigma \sim U[0.001, 2.0]$), random affine transforms (translations $\leq 25\%$, rotations $\leq 180^\circ$, scaling 75--125\%, shear $\leq 17.5^\circ$), and random horizontal/vertical flips (each with 50\% probability).
Additionally, photometric distortions were introduced, modifying brightness (87.5--112.5\%), contrast (50-150\%), saturation (50-150\%), and hue ($\le 5\% $), applied with a 50\% probability. \\

\subsubsection{Implementation.}
All architectures were implemented in Python 3.11 using PyTorch 2.0.1. Model training was performed on a server equipped with two partitioned NVIDIA A100 GPUs, utilizing two 42\,GB VRAM and two 21\,GB VRAM instances. The training was conducted end-to-end on the CFU dataset using the AdamW optimizer and the Reflected Exponential (REX)~\citesupp{appendix_chen2022rex} learning rate scheduler, with a maximum of 120 epochs for parameter tuning and 200 epochs for \ac{cv}. Binary cross-entropy (BCE) loss was employed, and a batch size of eight was selected as the largest feasible size across all methods, given hardware constraints.
A preliminary learning rate search explored $[10^{-4}, 5 \times 10^{-5}, 10^{-5}]$, identifying $10^{-4}$ as preferable for most methods based on validation IoU and $5 \times 10^{-5}$ as yielding superior performance for \acs{hiformer}, \acs{segformer}, \acs{segnext}, \acs{vwmit}, and \acs{transnext}.

During \ac{cv}, early stopping was applied with patience of 20 epochs and a minimum IoU improvement threshold of 0.001 on the validation set, saving the best-performing model weights for final test evaluation. While some architectures may benefit from extended training, we adopted early stopping with a generous patience of 20 epochs to mitigate overfitting and enhance generalization.\\

\textit{For all architectures, we enabled full determinism mode in the configuration file; however, some models include components for which no fully deterministic implementation is available. Except for \acs{segnext}, which exhibited convergence issues under mixed precision, all architectures were trained using NVIDIA’s mixed precision mode to reduce memory consumption and computational load without impacting performance~\citesupp{appendix_micikevicius2018mixed}.}

\newpage
\subsection{Details on the Uniform Evaluation Procedure}
\label{appendix:ssec:benchmark:eval_procedure}

\noindent \textbf{Segmentation Performance.} We use established metrics, including mean Intersection over Union (mIoU), Dice Similarity Coefficient (mDSC), precision (mPrc), and recall (mRec). The metrics are computed excluding the background class, using \emph{micro-averaging}, which aggregates pixel counts across all samples within a given subset of the dataset before calculating the metric values.\\

\noindent \textbf{Generalization.} We assess each model’s generalization capability using unseen \ac{ood} data. This aspect is particularly critical for remote patient monitoring, where wound images are captured using various mobile devices under diverse environmental conditions.\\

\noindent \textbf{Model Efficiency.} 
To evaluate model complexity, we report the number of trainable parameters as an indicator of model capacity and computational demands, along with \acp{gmac} to quantify the arithmetic operations required for a single forward pass. Both metrics are computed using the \texttt{calflops} tool~\citesupp{appendix_calflops2023}.

For efficiency analysis, we measure inference time separately for GPU and CPU execution, alongside throughput in images per second (IPS). Inference times are recorded with a batch size of one, averaged over \numprint{3000} images following a warm-up of 10 images per model. Experiments are conducted on an NVIDIA A100 GPU (40 GB partition) and an AMD EPYC 7763 64-core CPU. To ensure realistic computational loads, we use actual wound images 
, randomly sampling 300 images from both our primary and the \ac{ood} datasets. 
Inference times are measured for each of the five \ac{cv} models per architecture, resulting in a total of $300 \times 5 \times 2 = \numprint{3000}$ images.\\

\noindent \textbf{Explainability.}
To improve model interpretability, we utilize \emph{Grad-CAM}-based visualizations to identify the image regions most influential in segmentation decisions. This method provides insights into the decision-making process, allowing us to assess whether the models correctly focus on wound regions while minimizing attention to non-wound areas and background artifacts. Heatmaps are generated using a slightly modified version of the \texttt{M3d-CAM} PyTorch library~\citesupp{appendix_m3dCAM2020} to integrate it into our framework. We use the \texttt{layer='auto'} setting, which automatically selects the last layer suitable for extracting attention maps, and visualize the averaged heatmaps across the five \ac{cv} models.

\newpage
\subsection{Datasets}
\label{appendix:ssec:benchmark:datasets}
\textbf{CFU.}
We conduct experiments using a custom \textit{Combined Foot Ulcers (CFU)} dataset, created by merging the publicly available \textit{Foot Ulcer Segmentation Challenge 2021 (\acs{fuseg})}~\citesupp{appendix_wang2024fuseg} and \textit{Diabetic Foot Ulcer Challenge 2022 (DFUC 2022)}~\citesupp{appendix_kendrick2022translating,appendix_yap2024diabetic} datasets. \acs{fuseg} contains \numprint{1210} images (512×512 pixels), while DFUC 2022 includes \numprint{4000} images (640×480 pixels). 
Due to private test set annotations, we exclude the test sets of both datasets (200 from \acs{fuseg}; \numprint{2000} from DFUC 2022). Moreover, we remove duplicates, identifying them as such by checking for identical raw bytes or using perceptual hashing with a Hamming distance of $\le 11$. After eliminating 123 redundant image-mask pairs, the final CFU dataset comprises \numprint{2887} unique foot ulcer samples.
For our experiments, CFU is randomly split into training (60\%), validation (20\%), and test (20\%) sets for the learning rate study, and into five equally sized folds for \ac{cv}.\\

\textbf{DFUC'22 Test Set.}
To externally validate our models, we submit predictions for \numprint{2000} test images to the challenge’s leaderboard, ensuring standardized evaluation. This is done once, without optimization for leaderboard metrics.\\

\textbf{Out-of-Distribution Dataset (OOD).}
Our OOD dataset was collected at the University Hospital of Würzburg, 
following approval from the local ethics committee. Most images have a resolution of $2672 \times 4000$ pixels. They include a diverse range of chronic wounds from various body locations and, thus, extend beyond \acl{dfu}. A total of 343 images were manually annotated by dermatologists for this study with ImageJ, and annotations were stored as JPG files alongside the originals. 
While no duplicates exist, some wounds were intentionally captured multiple times from different distances, angles, and perspectives to reflect real-world conditions, where patient-acquired mobile images lack standardized setups. This variability allows for a robust evaluation of model generalization.
Importantly, this dataset is used solely for evaluating the models trained on the CFU dataset; no additional training is performed. Due to privacy regulations, the majority of the OOD dataset remains confidential, as written consent for public sharing was not obtained from all patients. 
\subsection{Further Explainability Insights}

Below, we present Grad-CAM visualizations for various wounds across different body sites, complementing those in the main paper, to provide deeper insights into the models' decision-making process.

\begin{figure}[htb!]
    \centering
    \begin{turn}{-90}
        \begin{adjustbox}{max height=\textwidth, max width=\textheight, keepaspectratio}
            \begin{subfigure}{0.165\textwidth}
                \centering
                \includegraphics[trim={80 120 70 140}, clip, width=1.5cm, angle=180]{images/XAI/GT/ulc0387_04012022_3_GT.jpg}\\[4pt]
                \includegraphics[trim={80 165 0 85}, clip, width=1.5cm, angle=180]{images/XAI/GT/ulc0390_04012022_6_GT.jpg}\\[4pt]
                \includegraphics[trim={30 65 0 65}, clip, width=1.5cm, angle=180]{images/XAI/GT/ulc0098_07072022_5_GT.jpg}\\[4pt]
                \includegraphics[trim={0 90 0 90}, clip, width=1.5cm, angle=180]{images/XAI/GT/ulc0411_17032022_16_GT.jpg}\\[4pt]
                \includegraphics[trim={150 90 60 85}, clip, height=1.5cm, angle=90]{images/XAI/GT/ulc0406_17032022_11_GT.jpg}\\[4pt]
                \includegraphics[trim={110 90 120 90}, clip, height=1.5cm, angle=90]{images/XAI/GT/ulc0396_17032022_1_GT.jpg}\\[4pt]
                \includegraphics[trim={90 0 90 0}, clip, height=1.5cm, angle=90]{images/XAI/GT/ulc0119_03112022_3_GT.jpg}\\[4pt]
                \includegraphics[trim={220 105 100 65}, clip, height=1.5cm, angle=90]{images/XAI/GT/ulc0355_05072022_4_GT.jpg}\\[4pt]
                \includegraphics[trim={110 0 85 40}, clip, height=1.5cm, angle=90]{images/XAI/GT/ulc0059_11072022_5_GT.jpg}\\[4pt]
                \includegraphics[trim={160 115 90 130}, clip, height=1.5cm, angle=90]{images/XAI/GT/ulc0343_18032022_1_GT.jpg}\\[4pt]
                \includegraphics[trim={160 90 20 90}, clip, height=1.5cm, angle=90]{images/XAI/GT/ulc0245_31052022_6_GT.jpg}\\[4pt]
                \includegraphics[trim={64 10 64 0}, clip, height=1.5cm, angle=90]{images/XAI/GT/ulc0048_01082022_3_GT.jpg}\\[4pt]
                \caption*{\scriptsize GT}
            \end{subfigure}
            \hspace{-0.6cm}
            \begin{subfigure}{0.165\textwidth}
                \centering
                \includegraphics[trim={80 120 70 140}, clip, width=1.5cm, angle=180]{images/XAI/Overlays/ulc0387_04012022_3_TransNeXt_overlay.jpg}\\[4pt]
                \includegraphics[trim={80 165 0 85}, clip, width=1.5cm, angle=180]{images/XAI/Overlays/ulc0390_04012022_6_TransNeXt_overlay.jpg}\\[4pt]
                \includegraphics[trim={30 65 0 65}, clip, width=1.5cm, angle=180]{images/XAI/Overlays/ulc0098_07072022_5_TransNeXt_overlay.jpg}\\[4pt]
                \includegraphics[trim={0 90 0 90}, clip, width=1.5cm, angle=180]{images/XAI/Overlays/ulc0411_17032022_16_TransNeXt_overlay.jpg}\\[4pt]
                \includegraphics[trim={150 90 60 85}, clip, height=1.5cm, angle=90]{images/XAI/Overlays/ulc0406_17032022_11_TransNeXt_overlay.jpg}\\[4pt]
                \includegraphics[trim={110 90 120 90}, clip, height=1.5cm, angle=90]{images/XAI/Overlays/ulc0396_17032022_1_TransNeXt_overlay.jpg}\\[4pt]
                \includegraphics[trim={90 0 90 0}, clip, height=1.5cm, angle=90]{images/XAI/Overlays/ulc0119_03112022_3_TransNeXt_overlay.jpg}\\[4pt]
                \includegraphics[trim={220 105 100 65}, clip, height=1.5cm, angle=90]{images/XAI/Overlays/ulc0355_05072022_4_TransNeXt_overlay.jpg}\\[4pt]
                \includegraphics[trim={110 0 85 40}, clip, height=1.5cm, angle=90]{images/XAI/Overlays/ulc0059_11072022_5_TransNeXt_overlay.jpg}\\[4pt]
                \includegraphics[trim={160 115 90 130}, clip, height=1.5cm, angle=90]{images/XAI/Overlays/ulc0343_18032022_1_TransNeXt_overlay.jpg}\\[4pt]
                \includegraphics[trim={160 90 20 90}, clip, height=1.5cm, angle=90]{images/XAI/Overlays/ulc0245_31052022_6_TransNeXt_overlay.jpg}\\[4pt]
                \includegraphics[trim={64 10 64 0}, clip, height=1.5cm, angle=90]{images/XAI/Overlays/ulc0048_01082022_3_TransNeXt_overlay.jpg}\\[4pt]
                \caption*{\scriptsize TransN.}
            \end{subfigure}
            \hspace{-0.6cm}
            \begin{subfigure}{0.165\textwidth}
                \centering
                \includegraphics[trim={80 120 70 140}, clip, width=1.5cm, angle=180]{images/XAI/Overlays/ulc0387_04012022_3_InternImage_overlay.jpg}\\[4pt]
                \includegraphics[trim={80 165 0 85}, clip, width=1.5cm, angle=180]{images/XAI/Overlays/ulc0390_04012022_6_InternImage_overlay.jpg}\\[4pt]
                \includegraphics[trim={30 65 0 65}, clip, width=1.5cm, angle=180]{images/XAI/Overlays/ulc0098_07072022_5_InternImage_overlay.jpg}\\[4pt]
                \includegraphics[trim={0 90 0 90}, clip, width=1.5cm, angle=180]{images/XAI/Overlays/ulc0411_17032022_16_InternImage_overlay.jpg}\\[4pt]
                \includegraphics[trim={150 90 60 85}, clip, height=1.5cm, angle=90]{images/XAI/Overlays/ulc0406_17032022_11_InternImage_overlay.jpg}\\[4pt]
                \includegraphics[trim={110 90 120 90}, clip, height=1.5cm, angle=90]{images/XAI/Overlays/ulc0396_17032022_1_InternImage_overlay.jpg}\\[4pt]
                \includegraphics[trim={90 0 90 0}, clip, height=1.5cm, angle=90]{images/XAI/Overlays/ulc0119_03112022_3_InternImage_overlay.jpg}\\[4pt]
                \includegraphics[trim={220 105 100 65}, clip, height=1.5cm, angle=90]{images/XAI/Overlays/ulc0355_05072022_4_InternImage_overlay.jpg}\\[4pt]
                \includegraphics[trim={110 0 85 40}, clip, height=1.5cm, angle=90]{images/XAI/Overlays/ulc0059_11072022_5_InternImage_overlay.jpg}\\[4pt]
                \includegraphics[trim={160 115 90 130}, clip, height=1.5cm, angle=90]{images/XAI/Overlays/ulc0343_18032022_1_InternImage_overlay.jpg}\\[4pt]
                \includegraphics[trim={160 90 20 90}, clip, height=1.5cm, angle=90]{images/XAI/Overlays/ulc0245_31052022_6_InternImage_overlay.jpg}\\[4pt]
                \includegraphics[trim={64 10 64 0}, clip, height=1.5cm, angle=90]{images/XAI/Overlays/ulc0048_01082022_3_InternImage_overlay.jpg}\\[4pt]
                \caption*{\scriptsize InternIm.}
            \end{subfigure}
            \hspace{-0.6cm}
            \begin{subfigure}{0.165\textwidth}
                \centering
                \includegraphics[trim={80 120 70 140}, clip, width=1.5cm, angle=180]{images/XAI/Overlays/ulc0387_04012022_3_VWFormerMiTB3_overlay.jpg}\\[4pt]
                \includegraphics[trim={80 165 0 85}, clip, width=1.5cm, angle=180]{images/XAI/Overlays/ulc0390_04012022_6_VWFormerMiTB3_overlay.jpg}\\[4pt]
                \includegraphics[trim={30 65 0 65}, clip, width=1.5cm, angle=180]{images/XAI/Overlays/ulc0098_07072022_5_VWFormerMiTB3_overlay.jpg}\\[4pt]
                \includegraphics[trim={0 90 0 90}, clip, width=1.5cm, angle=180]{images/XAI/Overlays/ulc0411_17032022_16_VWFormerMiTB3_overlay.jpg}\\[4pt]
                \includegraphics[trim={150 90 60 85}, clip, height=1.5cm, angle=90]{images/XAI/Overlays/ulc0406_17032022_11_VWFormerMiTB3_overlay.jpg}\\[4pt]
                \includegraphics[trim={110 90 120 90}, clip, height=1.5cm, angle=90]{images/XAI/Overlays/ulc0396_17032022_1_VWFormerMiTB3_overlay.jpg}\\[4pt]
                \includegraphics[trim={90 0 90 0}, clip, height=1.5cm, angle=90]{images/XAI/Overlays/ulc0119_03112022_3_VWFormerMiTB3_overlay.jpg}\\[4pt]
                \includegraphics[trim={220 105 100 65}, clip, height=1.5cm, angle=90]{images/XAI/Overlays/ulc0355_05072022_4_VWFormerMiTB3_overlay.jpg}\\[4pt]
                \includegraphics[trim={110 0 85 40}, clip, height=1.5cm, angle=90]{images/XAI/Overlays/ulc0059_11072022_5_VWFormerMiTB3_overlay.jpg}\\[4pt]
                \includegraphics[trim={160 115 90 130}, clip, height=1.5cm, angle=90]{images/XAI/Overlays/ulc0343_18032022_1_VWFormerMiTB3_overlay.jpg}\\[4pt]
                \includegraphics[trim={160 90 20 90}, clip, height=1.5cm, angle=90]{images/XAI/Overlays/ulc0245_31052022_6_VWFormerMiTB3_overlay.jpg}\\[4pt]
                \includegraphics[trim={64 10 64 0}, clip, height=1.5cm, angle=90]{images/XAI/Overlays/ulc0048_01082022_3_VWFormerMiTB3_overlay.jpg}\\[4pt]
                \caption*{\scriptsize VW-MiT}
            \end{subfigure}
            \hspace{-0.6cm}
            \begin{subfigure}{0.165\textwidth}
                \centering
                \includegraphics[trim={80 120 70 140}, clip, width=1.5cm, angle=180]{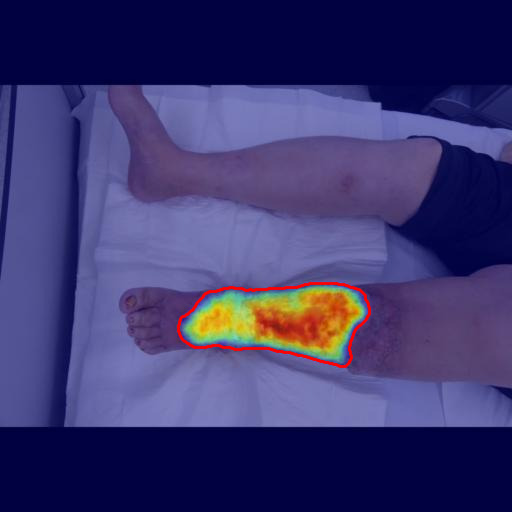}\\[4pt]
                \includegraphics[trim={80 165 0 85}, clip, width=1.5cm, angle=180]{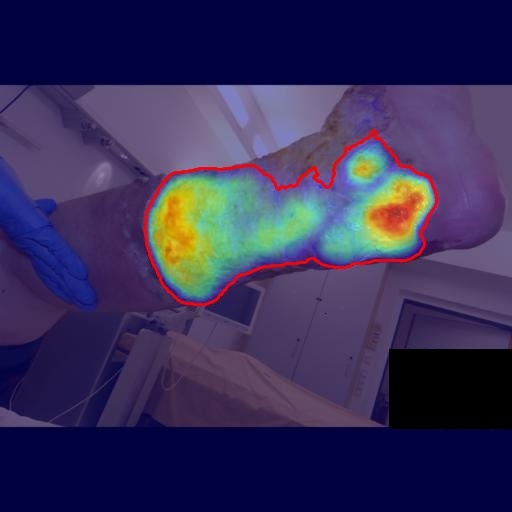}\\[4pt]
                \includegraphics[trim={30 65 0 65}, clip, width=1.5cm, angle=180]{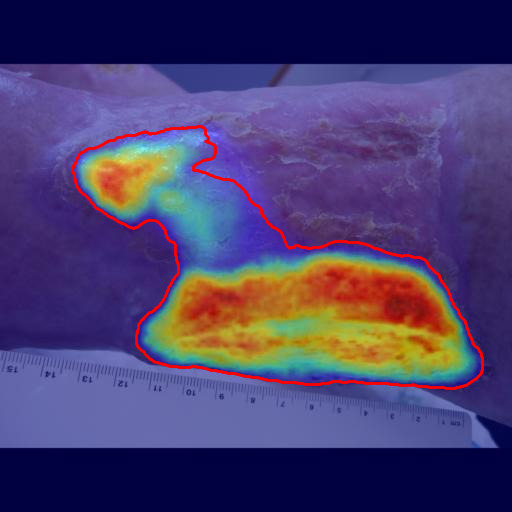}\\[4pt]
                \includegraphics[trim={0 90 0 90}, clip, width=1.5cm, angle=180]{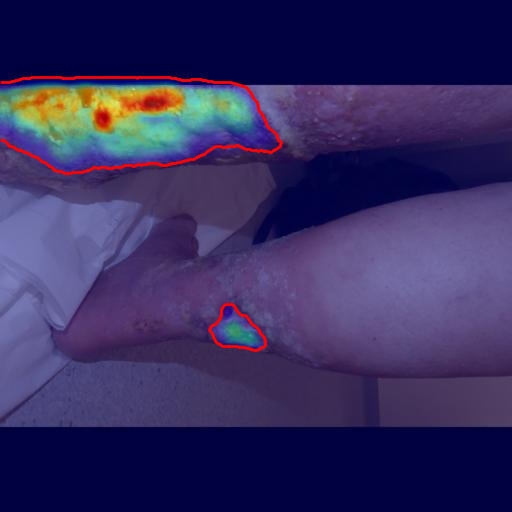}\\[4pt]
                \includegraphics[trim={150 90 60 85}, clip, height=1.5cm, angle=90]{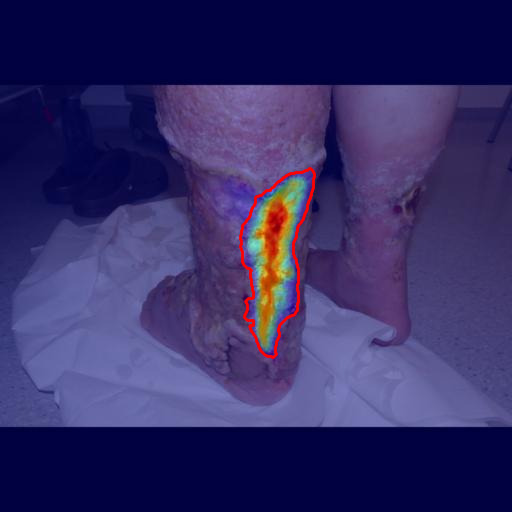}\\[4pt]
                \includegraphics[trim={110 90 120 90}, clip, height=1.5cm, angle=90]{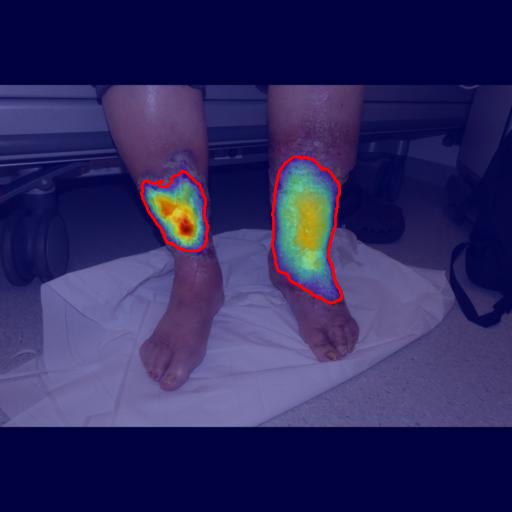}\\[4pt]
                \includegraphics[trim={90 0 90 0}, clip, height=1.5cm, angle=90]{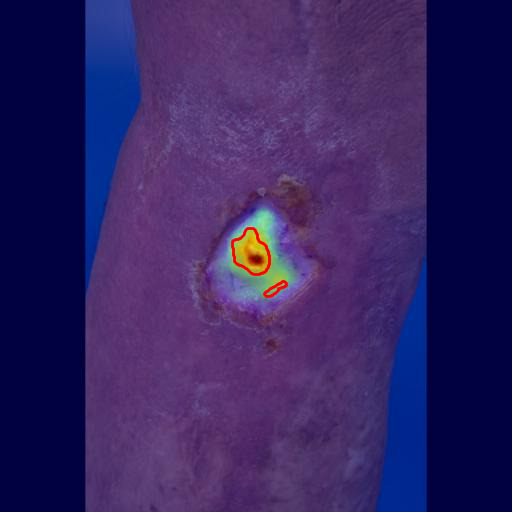}\\[4pt]
                \includegraphics[trim={220 105 100 65}, clip, height=1.5cm, angle=90]{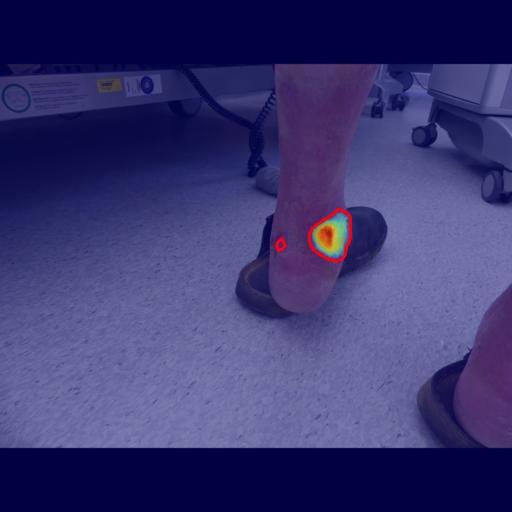}\\[4pt]
                \includegraphics[trim={110 0 85 40}, clip, height=1.5cm, angle=90]{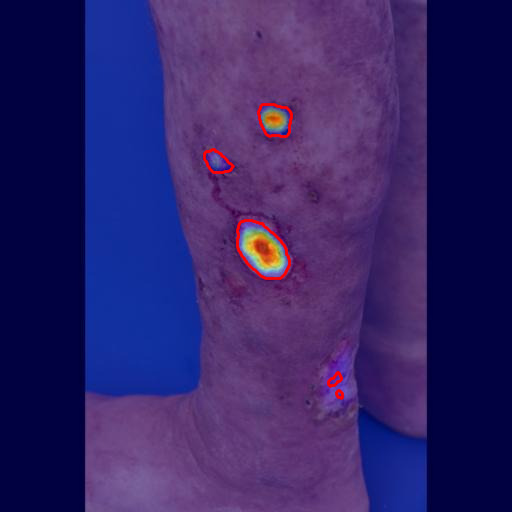}\\[4pt]
                \includegraphics[trim={160 115 90 130}, clip, height=1.5cm, angle=90]{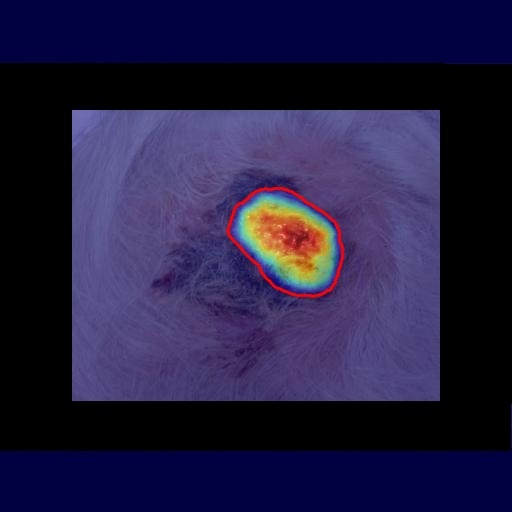}\\[4pt]
                \includegraphics[trim={160 90 20 90}, clip, height=1.5cm, angle=90]{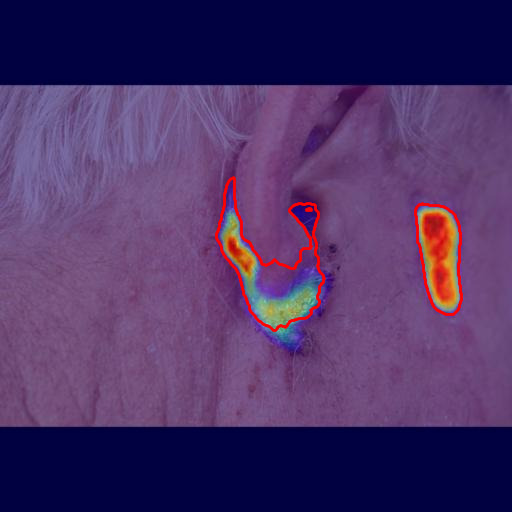}\\[4pt]
                \includegraphics[trim={64 10 64 0}, clip, height=1.5cm, angle=90]{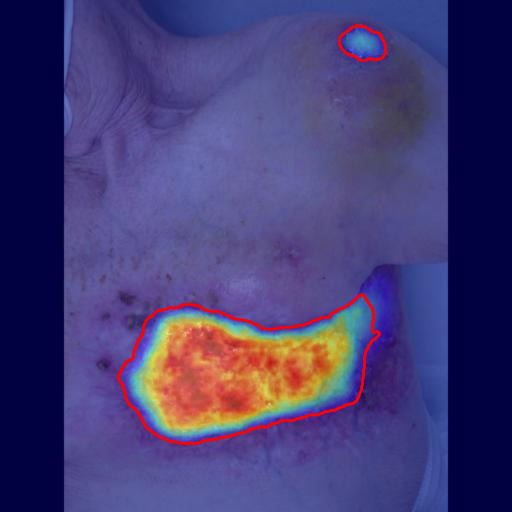}\\[4pt]
                \caption*{\scriptsize SegForm.}
            \end{subfigure}
            \hspace{-0.6cm}
            \begin{subfigure}{0.165\textwidth}
                \centering
                \includegraphics[trim={80 120 70 140}, clip, width=1.5cm, angle=180]{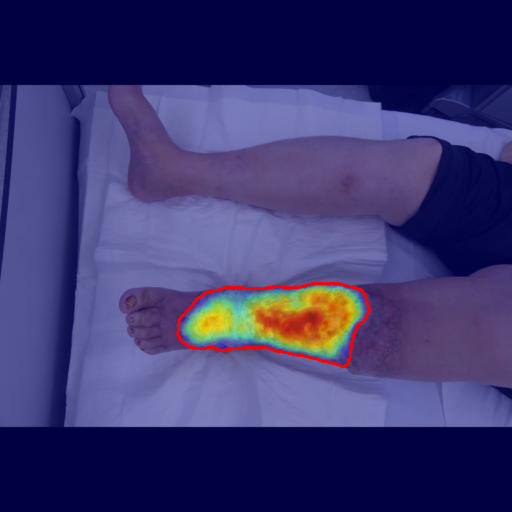}\\[4pt]
                \includegraphics[trim={80 165 0 85}, clip, width=1.5cm, angle=180]{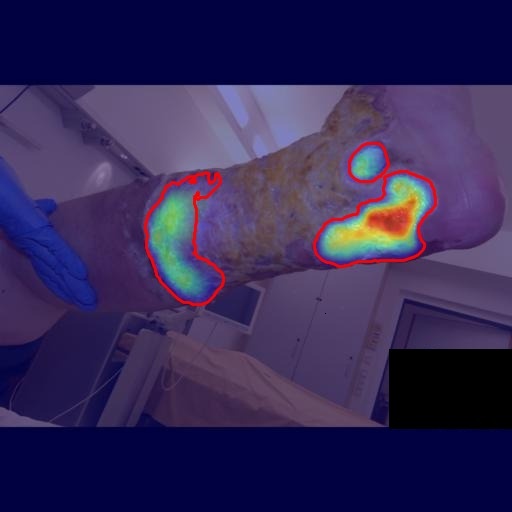}\\[4pt]
                \includegraphics[trim={30 65 0 65}, clip, width=1.5cm, angle=180]{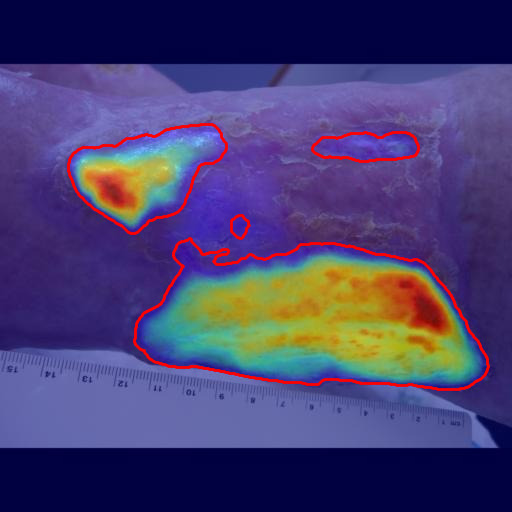}\\[4pt]
                \includegraphics[trim={0 90 0 90}, clip, width=1.5cm, angle=180]{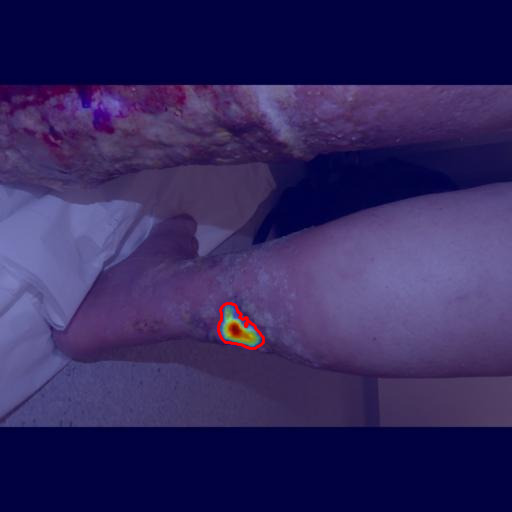}\\[4pt]
                \includegraphics[trim={150 90 60 85}, clip, height=1.5cm, angle=90]{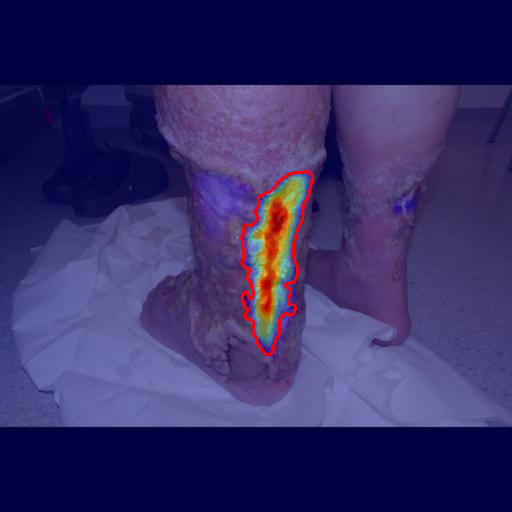}\\[4pt]
                \includegraphics[trim={110 90 120 90}, clip, height=1.5cm, angle=90]{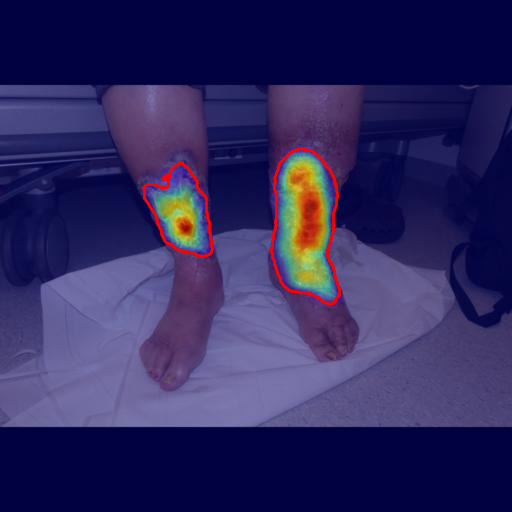}\\[4pt]
                \includegraphics[trim={90 0 90 0}, clip, height=1.5cm, angle=90]{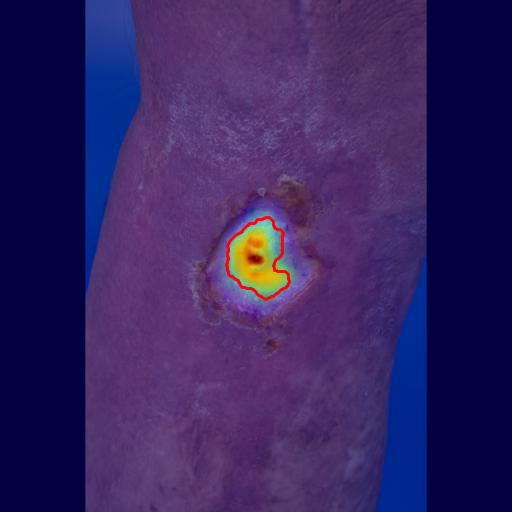}\\[4pt]
                \includegraphics[trim={220 105 100 65}, clip, height=1.5cm, angle=90]{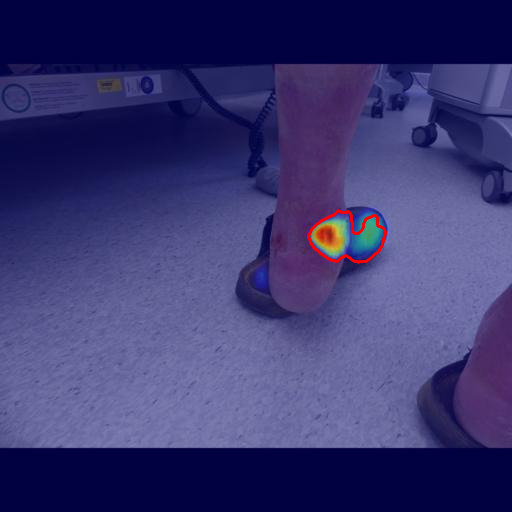}\\[4pt]
                \includegraphics[trim={110 0 85 40}, clip, height=1.5cm, angle=90]{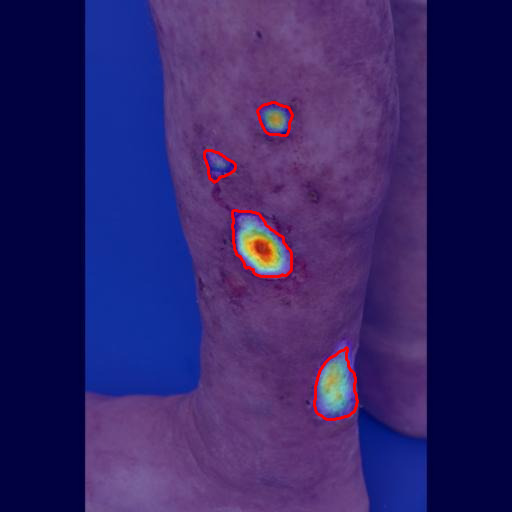}\\[4pt]
                \includegraphics[trim={160 115 90 130}, clip, height=1.5cm, angle=90]{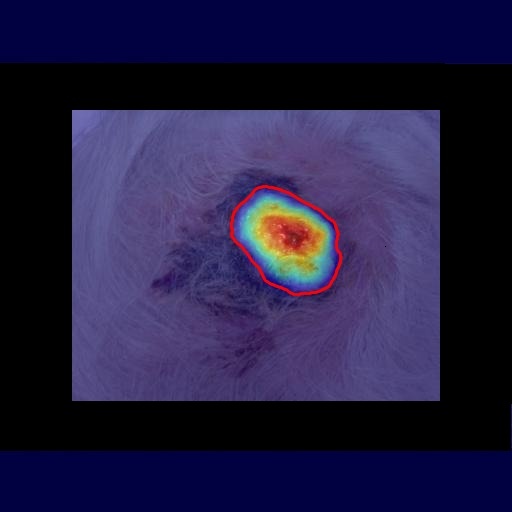}\\[4pt]
                \includegraphics[trim={160 90 20 90}, clip, height=1.5cm, angle=90]{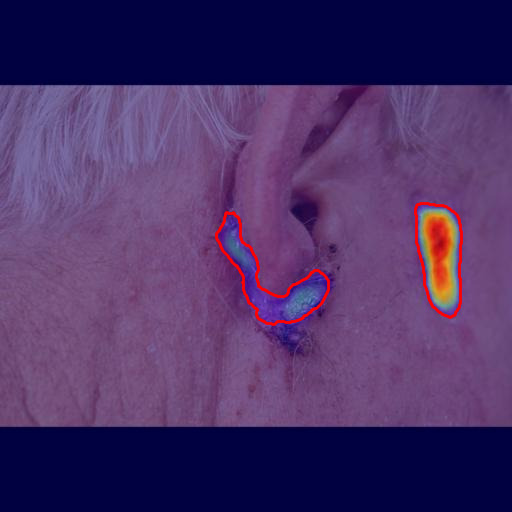}\\[4pt]
                \includegraphics[trim={64 10 64 0}, clip, height=1.5cm, angle=90]{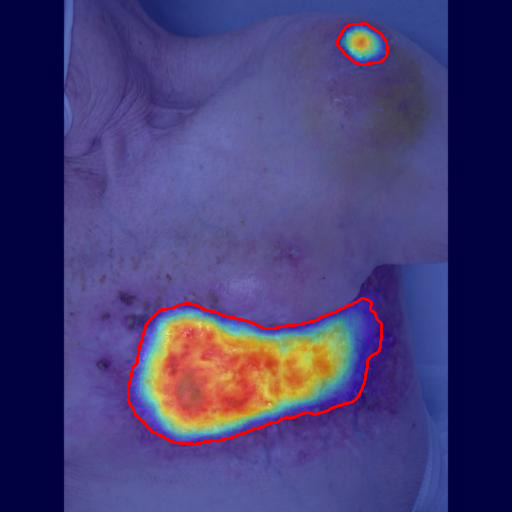}\\[4pt]
                \caption*{\scriptsize VW-Conv}
            \end{subfigure}
            \hspace{-0.6cm}
            \begin{subfigure}{0.165\textwidth}
                \centering
                \includegraphics[trim={80 120 70 140}, clip, width=1.5cm, angle=180]{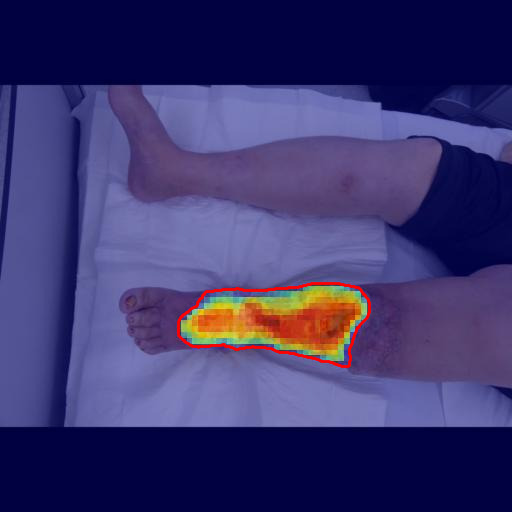}\\[4pt]
                \includegraphics[trim={80 165 0 85}, clip, width=1.5cm, angle=180]{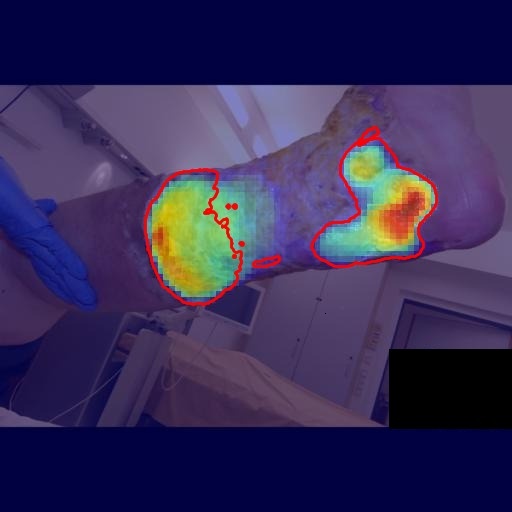}\\[4pt]
                \includegraphics[trim={30 65 0 65}, clip, width=1.5cm, angle=180]{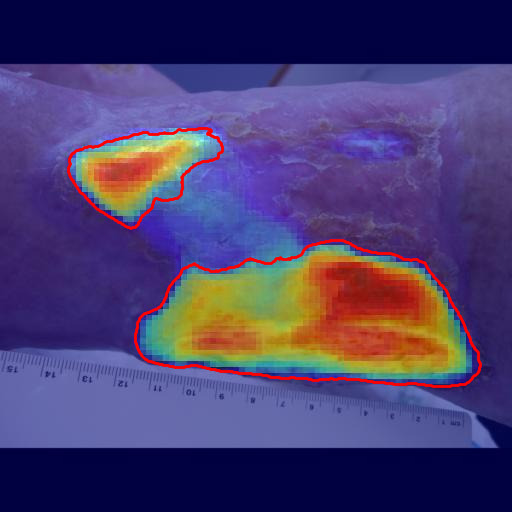}\\[4pt]
                \includegraphics[trim={0 90 0 90}, clip, width=1.5cm, angle=180]{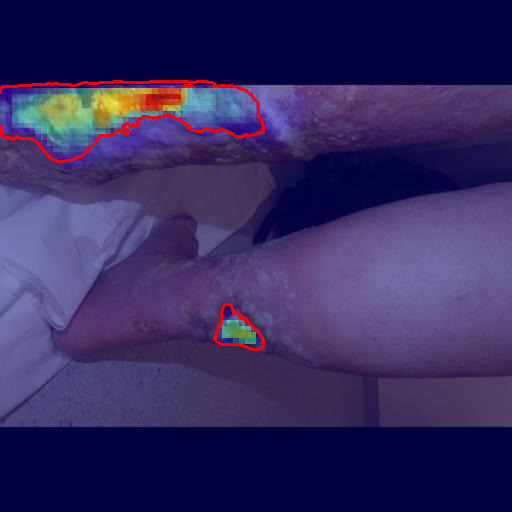}\\[4pt]
                \includegraphics[trim={150 90 60 85}, clip, height=1.5cm, angle=90]{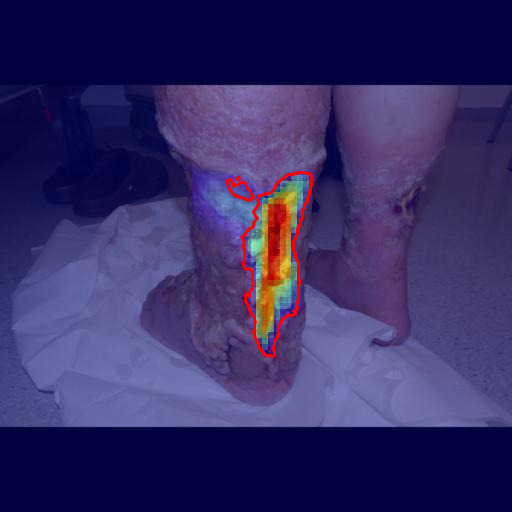}\\[4pt]
                \includegraphics[trim={110 90 120 90}, clip, height=1.5cm, angle=90]{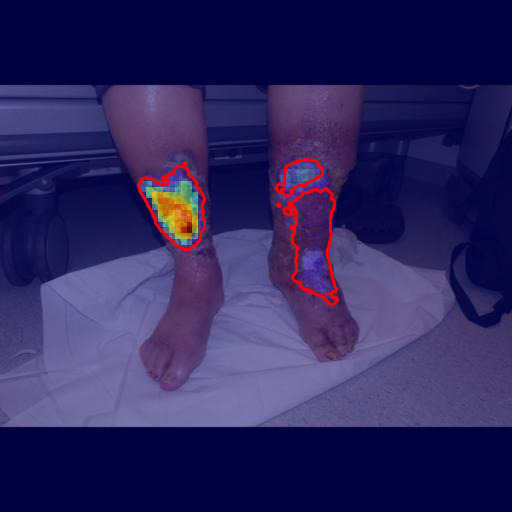}\\[4pt]
                \includegraphics[trim={90 0 90 0}, clip, height=1.5cm, angle=90]{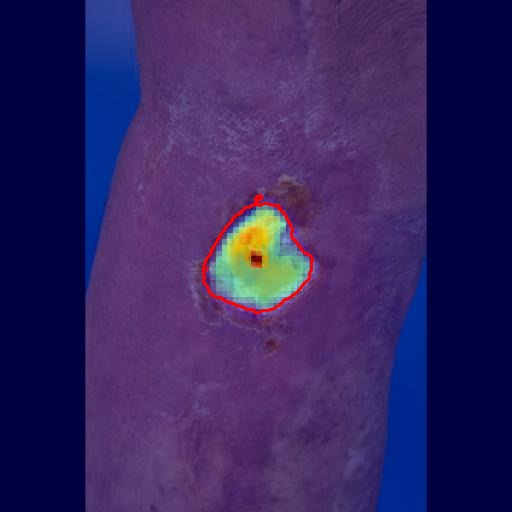}\\[4pt]
                \includegraphics[trim={220 105 100 65}, clip, height=1.5cm, angle=90]{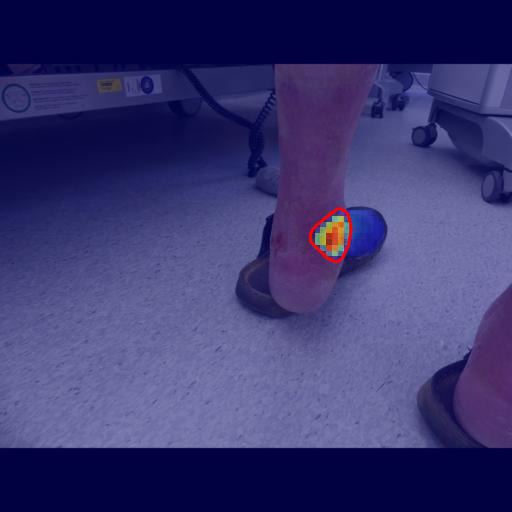}\\[4pt]
                \includegraphics[trim={110 0 85 40}, clip, height=1.5cm, angle=90]{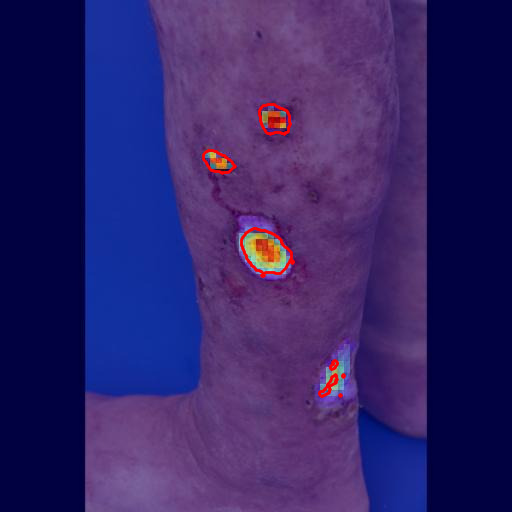}\\[4pt]
                \includegraphics[trim={160 115 90 130}, clip, height=1.5cm, angle=90]{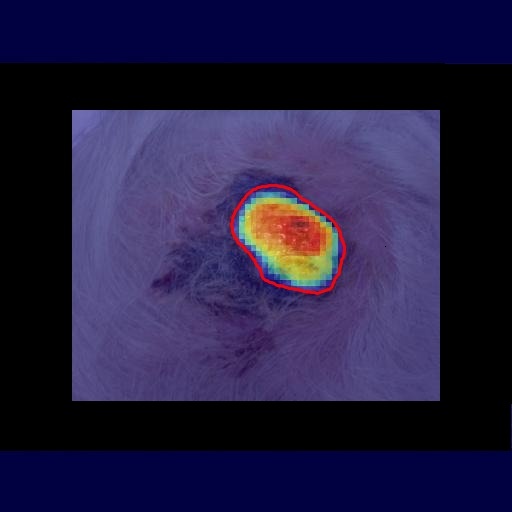}\\[4pt]
                \includegraphics[trim={160 90 20 90}, clip, height=1.5cm, angle=90]{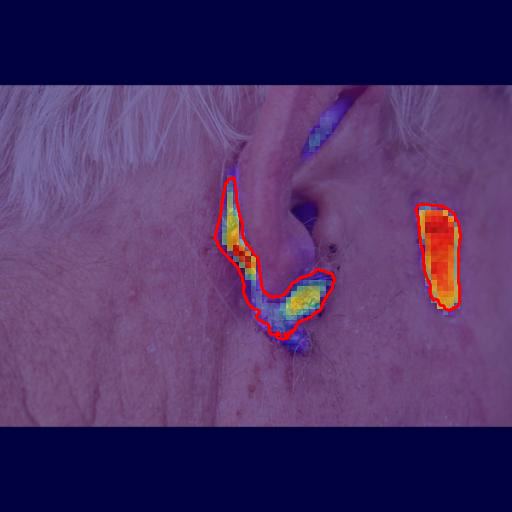}\\[4pt]
                \includegraphics[trim={64 10 64 0}, clip, height=1.5cm, angle=90]{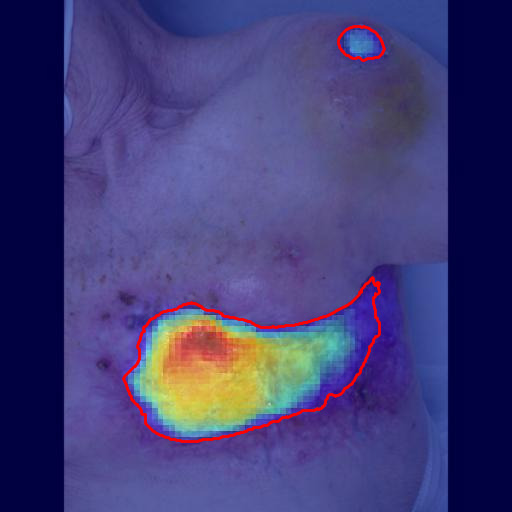}\\[4pt]
                \caption*{\scriptsize FCBForm.}
            \end{subfigure}
            \hspace{-0.6cm}
            \begin{subfigure}{0.165\textwidth}
                \centering
                \includegraphics[trim={80 120 70 140}, clip, width=1.5cm, angle=180]{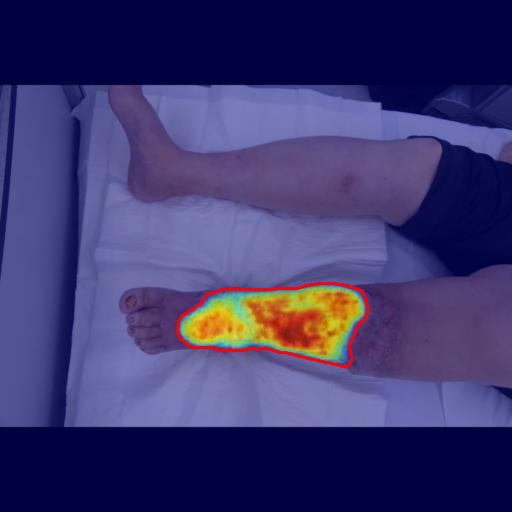}\\[4pt]
                \includegraphics[trim={80 165 0 85}, clip, width=1.5cm, angle=180]{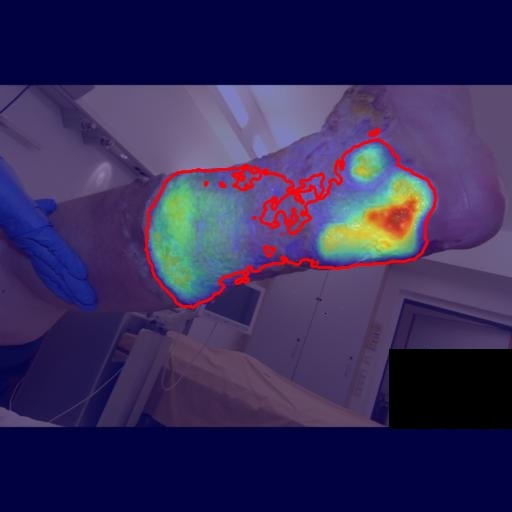}\\[4pt]
                \includegraphics[trim={30 65 0 65}, clip, width=1.5cm, angle=180]{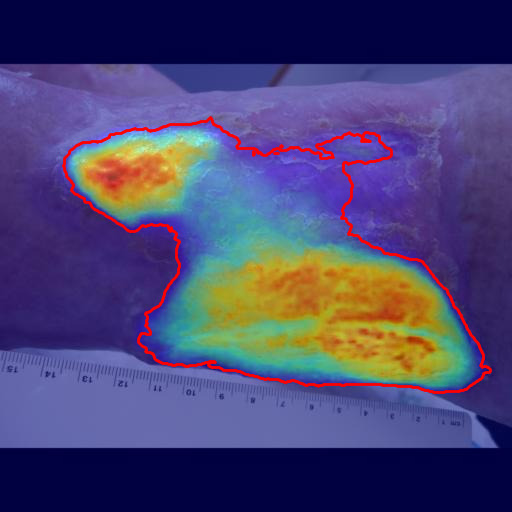}\\[4pt]
                \includegraphics[trim={0 90 0 90}, clip, width=1.5cm, angle=180]{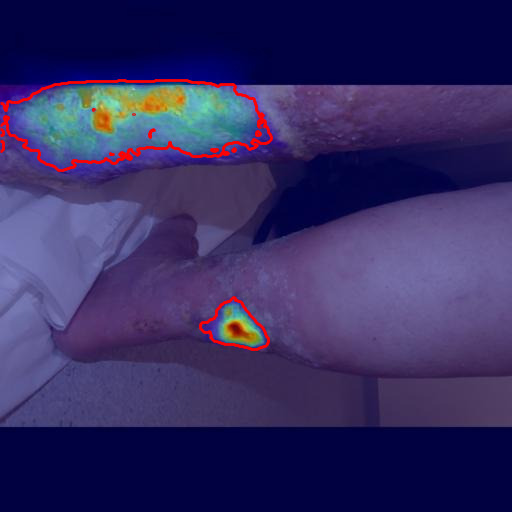}\\[4pt]
                \includegraphics[trim={150 90 60 85}, clip, height=1.5cm, angle=90]{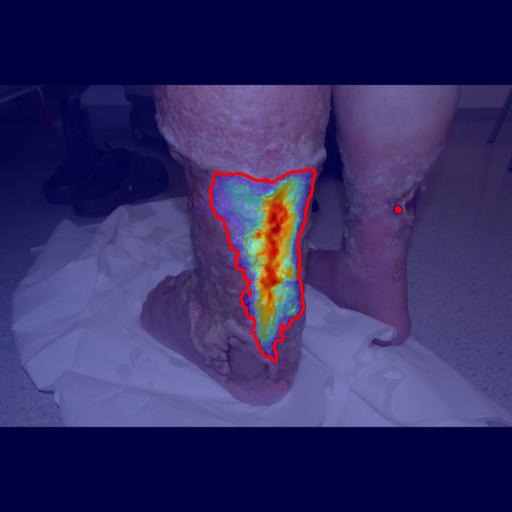}\\[4pt]
                \includegraphics[trim={110 90 120 90}, clip, height=1.5cm, angle=90]{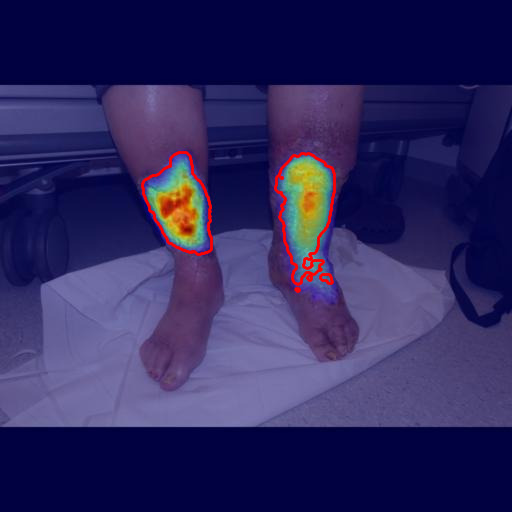}\\[4pt]
                \includegraphics[trim={90 0 90 0}, clip, height=1.5cm, angle=90]{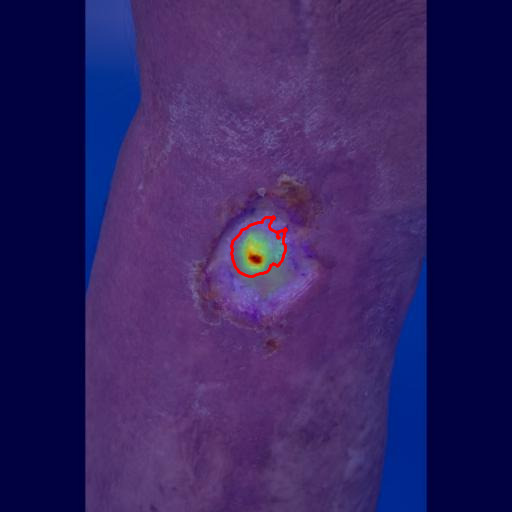}\\[4pt]
                \includegraphics[trim={220 105 100 65}, clip, height=1.5cm, angle=90]{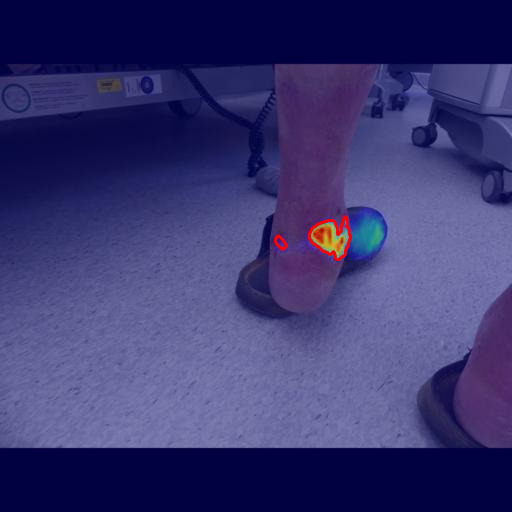}\\[4pt]
                \includegraphics[trim={110 0 85 40}, clip, height=1.5cm, angle=90]{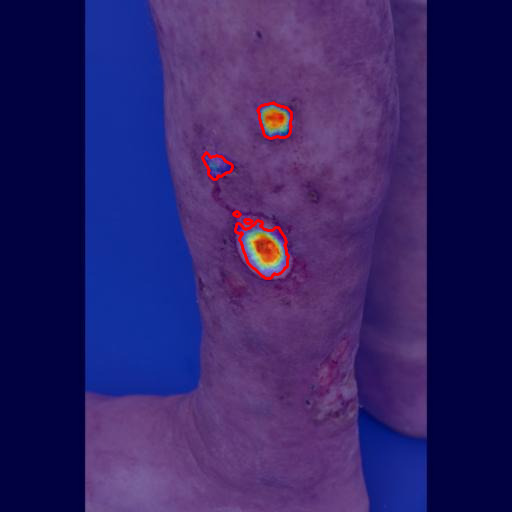}\\[4pt]
                \includegraphics[trim={160 115 90 130}, clip, height=1.5cm, angle=90]{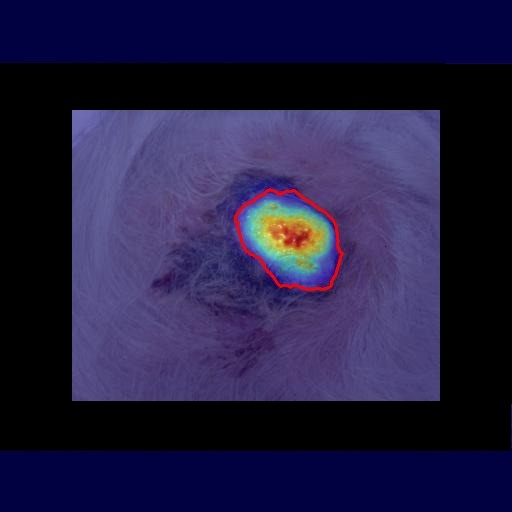}\\[4pt]
                \includegraphics[trim={160 90 20 90}, clip, height=1.5cm, angle=90]{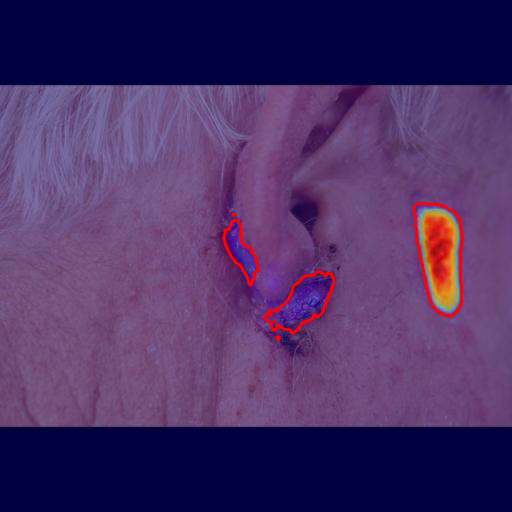}\\[4pt]
                \includegraphics[trim={64 10 64 0}, clip, height=1.5cm, angle=90]{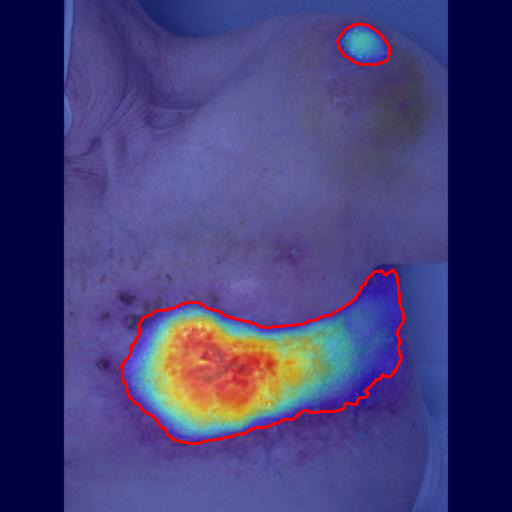}\\[4pt]
                \caption*{\scriptsize HarDNet}
            \end{subfigure}
            \hspace{-0.6cm}
            \begin{subfigure}{0.165\textwidth}
                \centering
                \includegraphics[trim={80 120 70 140}, clip, width=1.5cm, angle=180]{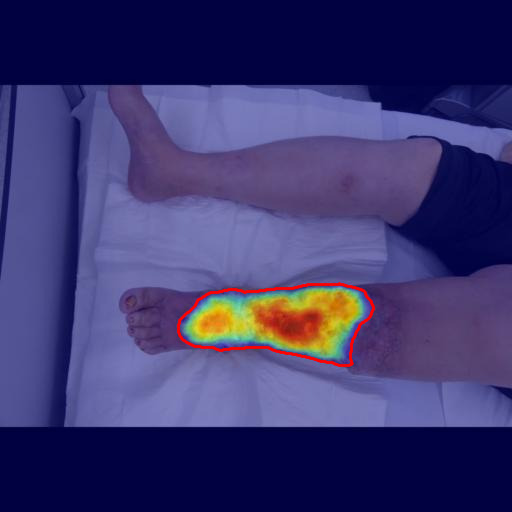}\\[4pt]
                \includegraphics[trim={80 165 0 85}, clip, width=1.5cm, angle=180]{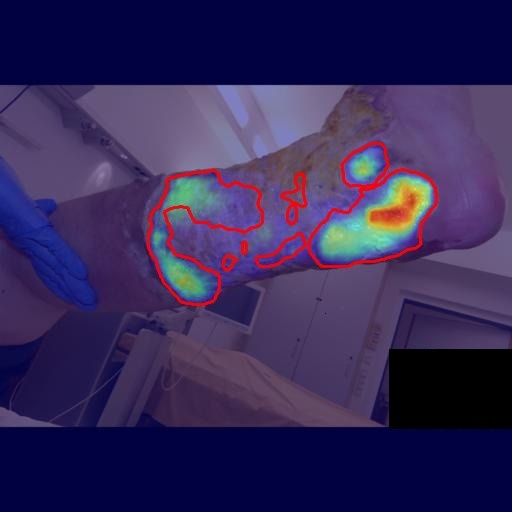}\\[4pt]
                \includegraphics[trim={30 65 0 65}, clip, width=1.5cm, angle=180]{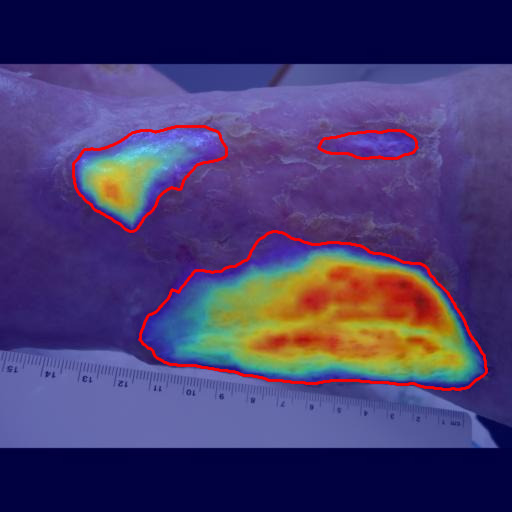}\\[4pt]
                \includegraphics[trim={0 90 0 90}, clip, width=1.5cm, angle=180]{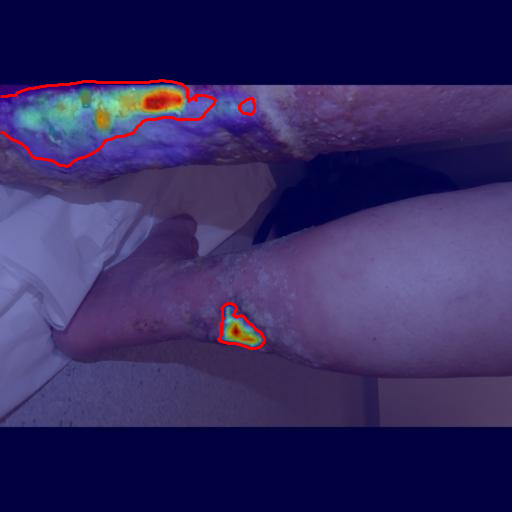}\\[4pt]
                \includegraphics[trim={150 90 60 85}, clip, height=1.5cm, angle=90]{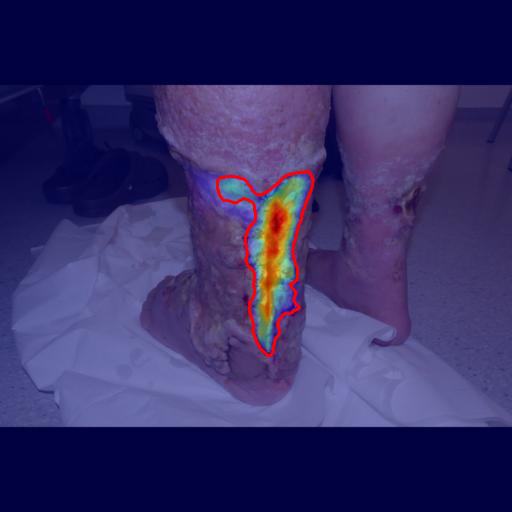}\\[4pt]
                \includegraphics[trim={110 90 120 90}, clip, height=1.5cm, angle=90]{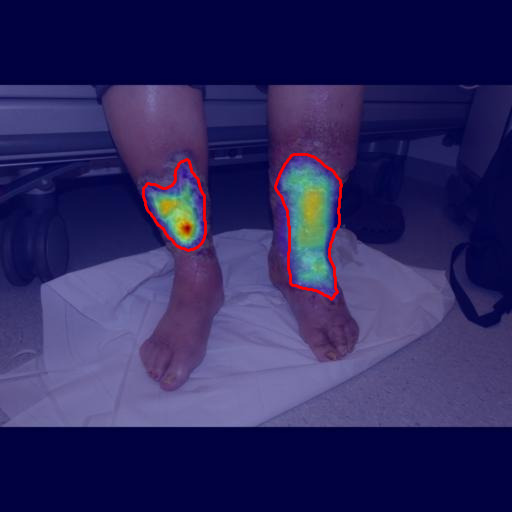}\\[4pt]
                \includegraphics[trim={90 0 90 0}, clip, height=1.5cm, angle=90]{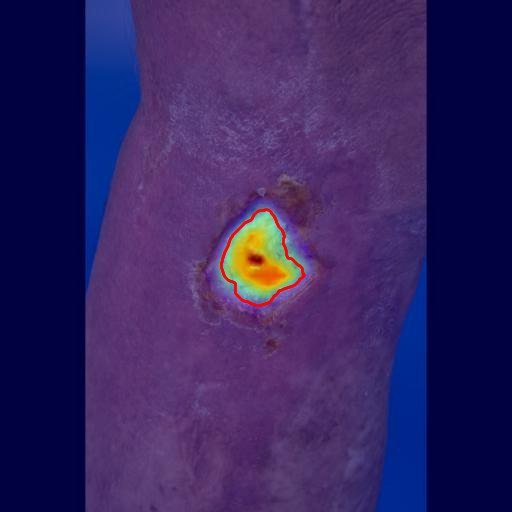}\\[4pt]
                \includegraphics[trim={220 105 100 65}, clip, height=1.5cm, angle=90]{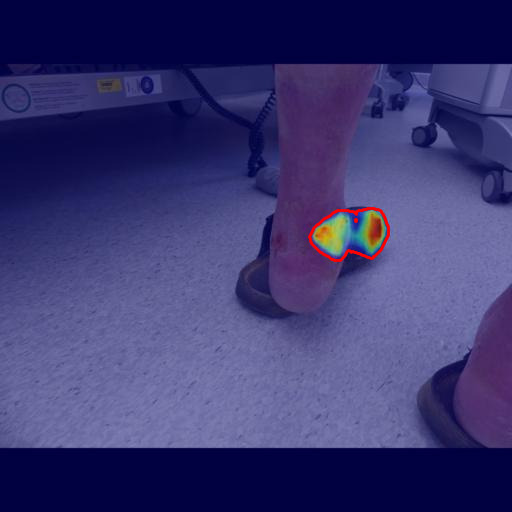}\\[4pt]
                \includegraphics[trim={110 0 85 40}, clip, height=1.5cm, angle=90]{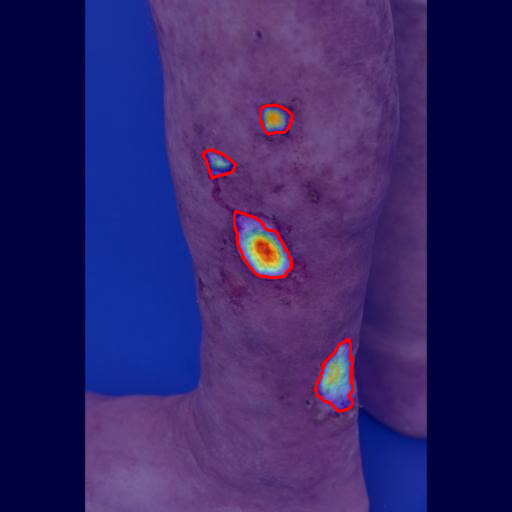}\\[4pt]
                \includegraphics[trim={160 115 90 130}, clip, height=1.5cm, angle=90]{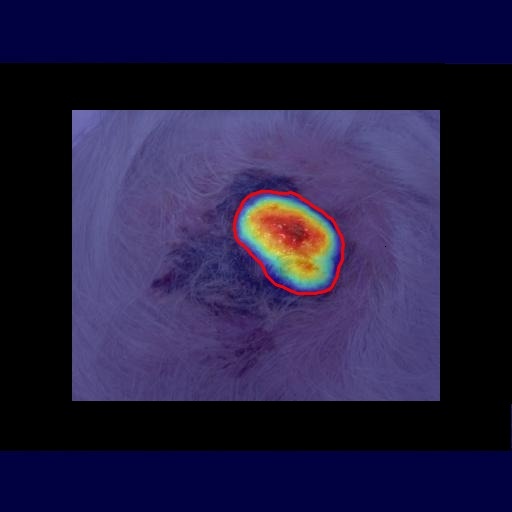}\\[4pt]
                \includegraphics[trim={160 90 20 90}, clip, height=1.5cm, angle=90]{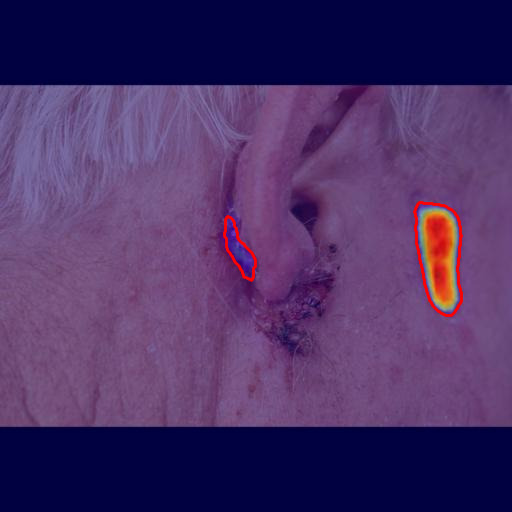}\\[4pt]
                \includegraphics[trim={64 10 64 0}, clip, height=1.5cm, angle=90]{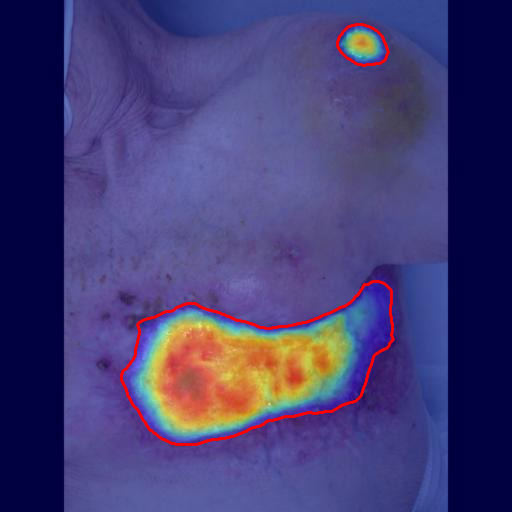}\\[4pt]
                \caption*{\scriptsize SegNeXt}
            \end{subfigure}
            \hspace{-0.6cm}
            \begin{subfigure}{0.165\textwidth}
                \centering
                \includegraphics[trim={80 120 70 140}, clip, width=1.5cm, angle=180]{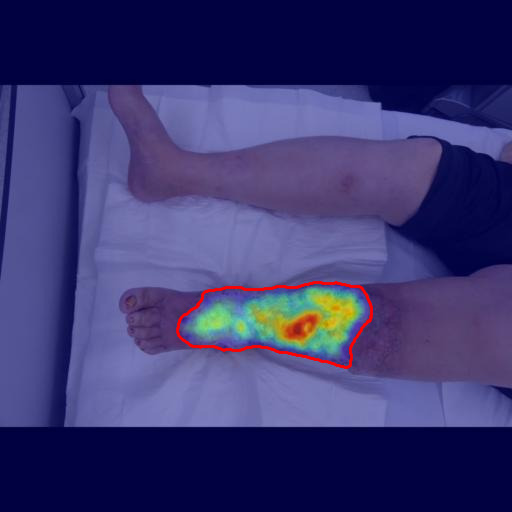}\\[4pt]
                \includegraphics[trim={80 165 0 85}, clip, width=1.5cm, angle=180]{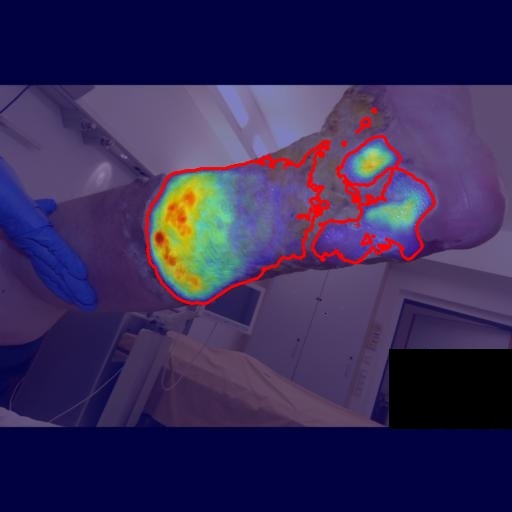}\\[4pt]
                \includegraphics[trim={30 65 0 65}, clip, width=1.5cm, angle=180]{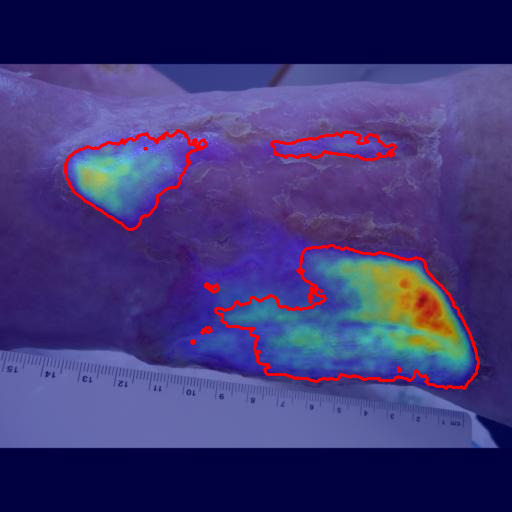}\\[4pt]
                \includegraphics[trim={0 90 0 90}, clip, width=1.5cm, angle=180]{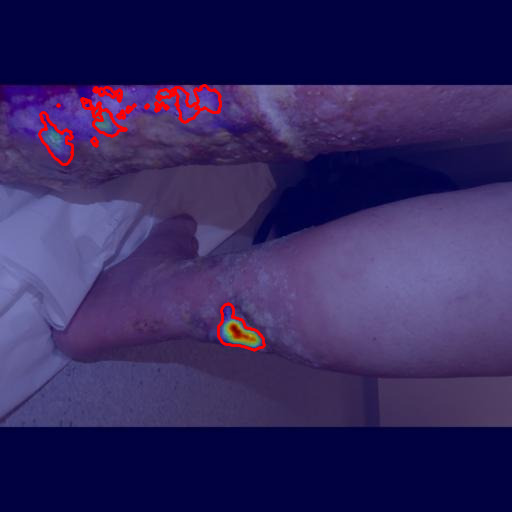}\\[4pt]
                \includegraphics[trim={150 90 60 85}, clip, height=1.5cm, angle=90]{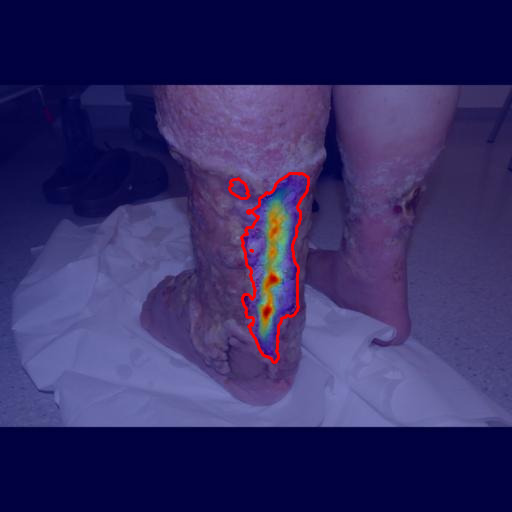}\\[4pt]
                \includegraphics[trim={110 90 120 90}, clip, height=1.5cm, angle=90]{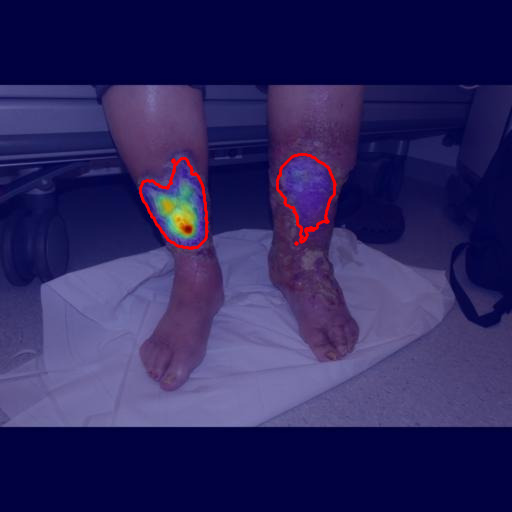}\\[4pt]
                \includegraphics[trim={90 0 90 0}, clip, height=1.5cm, angle=90]{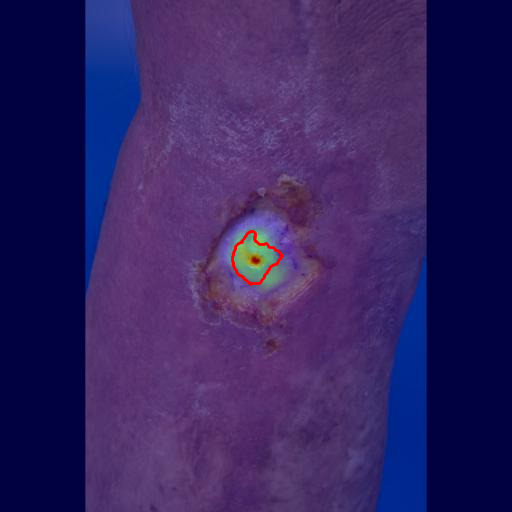}\\[4pt]
                \includegraphics[trim={220 105 100 65}, clip, height=1.5cm, angle=90]{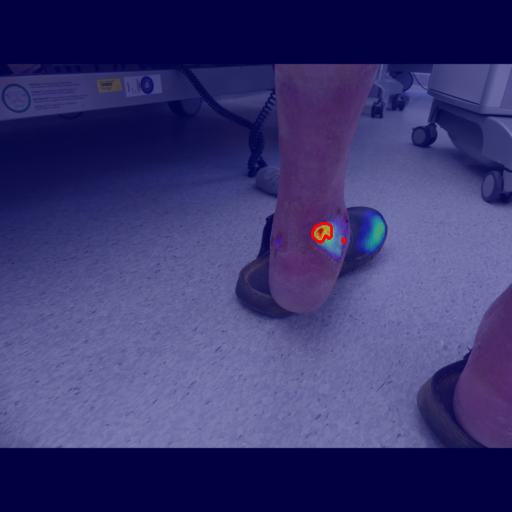}\\[4pt]
                \includegraphics[trim={110 0 85 40}, clip, height=1.5cm, angle=90]{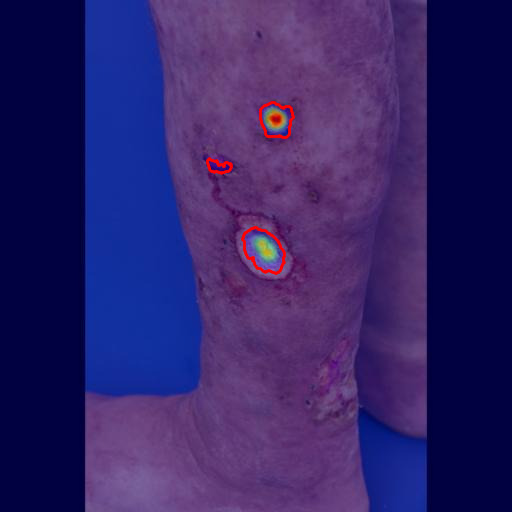}\\[4pt]
                \includegraphics[trim={160 115 90 130}, clip, height=1.5cm, angle=90]{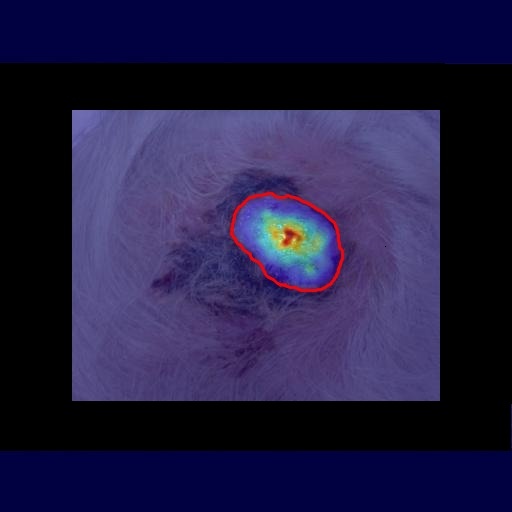}\\[4pt]
                \includegraphics[trim={160 90 20 90}, clip, height=1.5cm, angle=90]{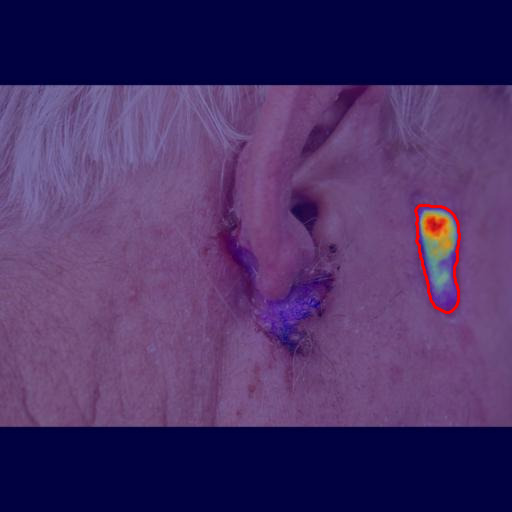}\\[4pt]
                \includegraphics[trim={64 10 64 0}, clip, height=1.5cm, angle=90]{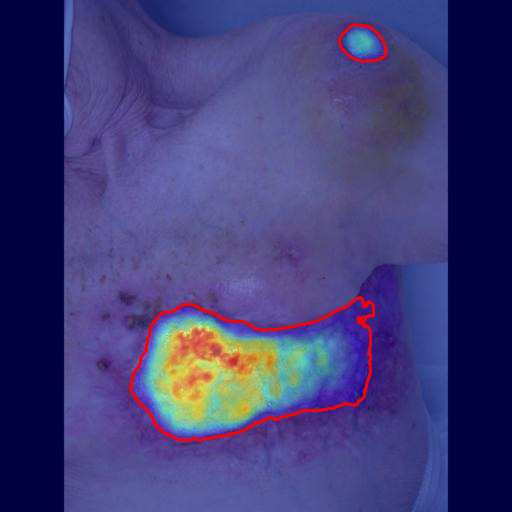}\\[4pt]
                \caption*{\scriptsize FuSegN.}
            \end{subfigure}
            \hspace{-0.6cm}
            \begin{subfigure}{0.165\textwidth}
                \centering
                \includegraphics[trim={80 120 70 140}, clip, width=1.5cm, angle=180]{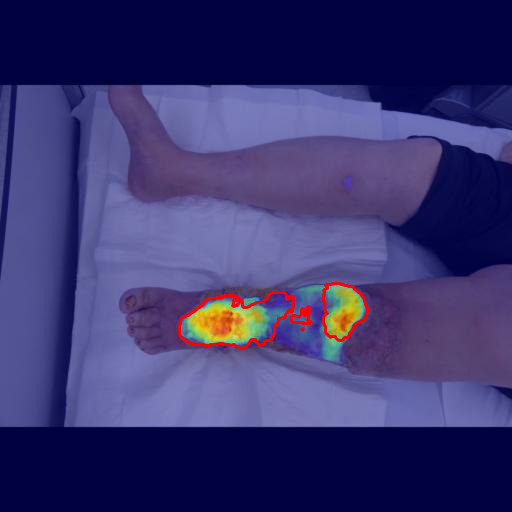}\\[4pt]
                \includegraphics[trim={80 165 0 85}, clip, width=1.5cm, angle=180]{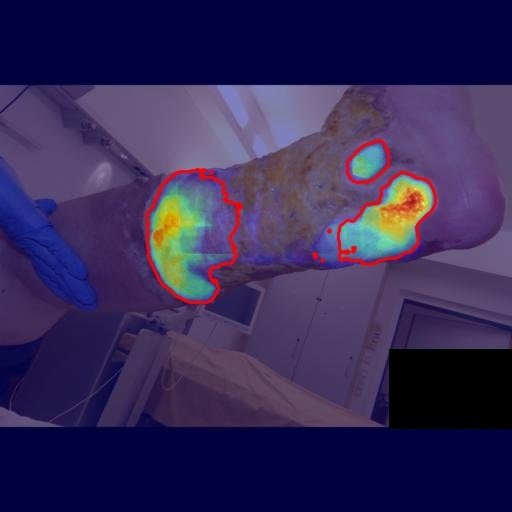}\\[4pt]
                \includegraphics[trim={30 65 0 65}, clip, width=1.5cm, angle=180]{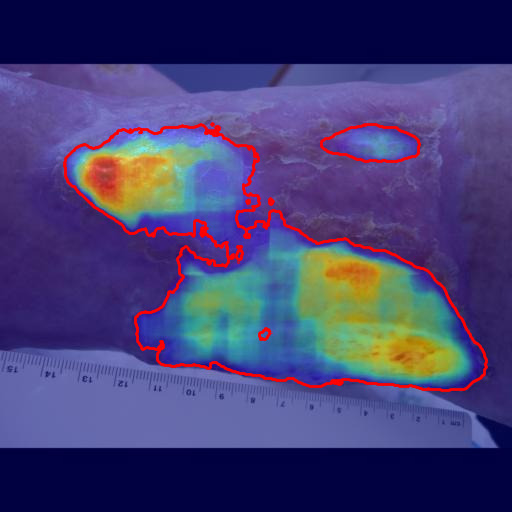}\\[4pt]
                \includegraphics[trim={0 90 0 90}, clip, width=1.5cm, angle=180]{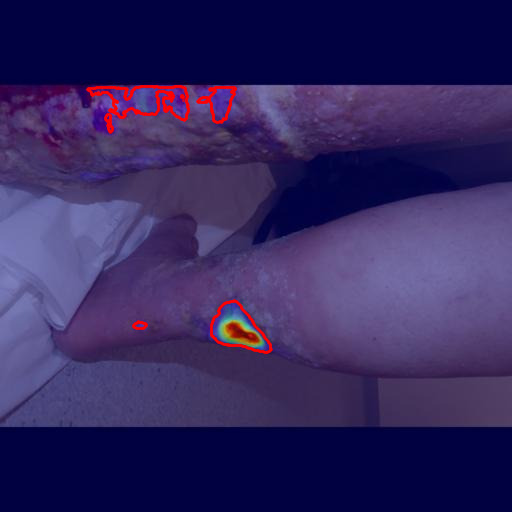}\\[4pt]
                \includegraphics[trim={150 90 60 85}, clip, height=1.5cm, angle=90]{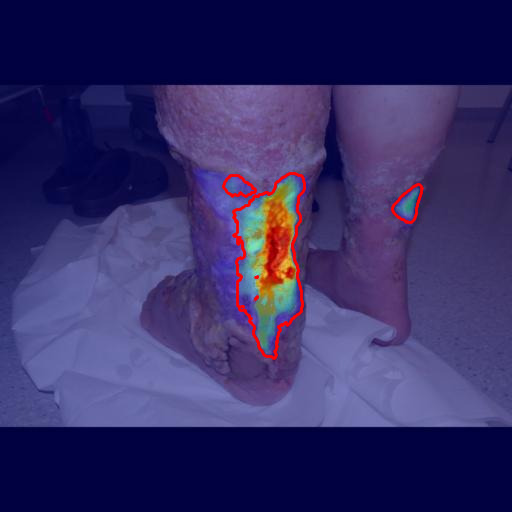}\\[4pt]
                \includegraphics[trim={110 90 120 90}, clip, height=1.5cm, angle=90]{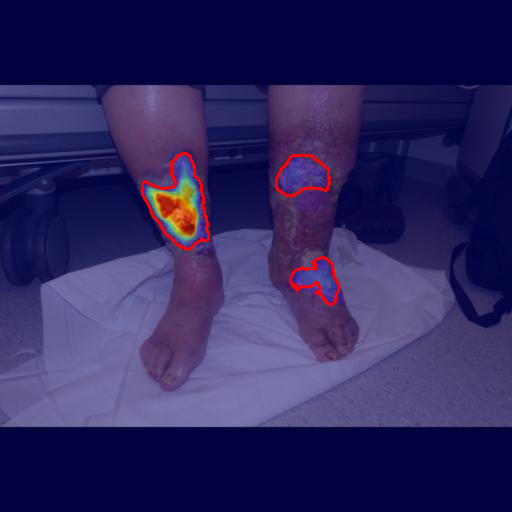}\\[4pt]
                \includegraphics[trim={90 0 90 0}, clip, height=1.5cm, angle=90]{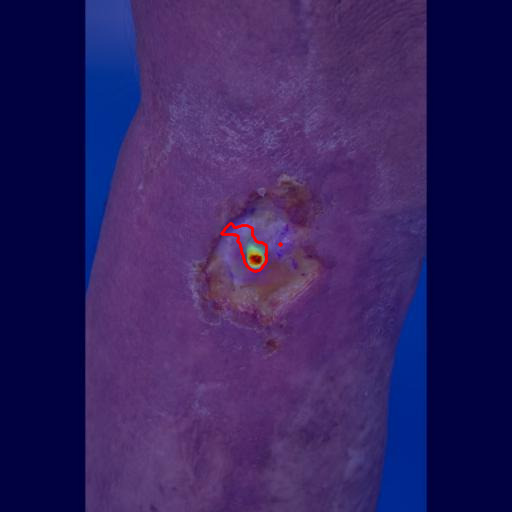}\\[4pt]
                \includegraphics[trim={220 105 100 65}, clip, height=1.5cm, angle=90]{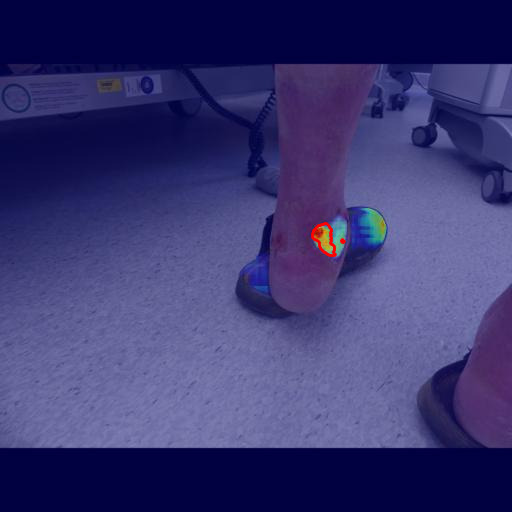}\\[4pt]
                \includegraphics[trim={110 0 85 40}, clip, height=1.5cm, angle=90]{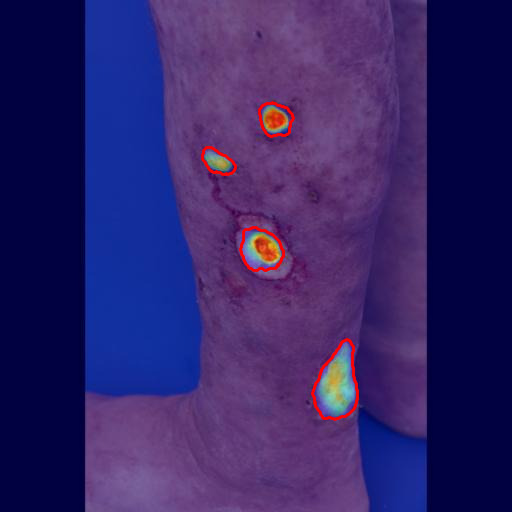}\\[4pt]
                \includegraphics[trim={160 115 90 130}, clip, height=1.5cm, angle=90]{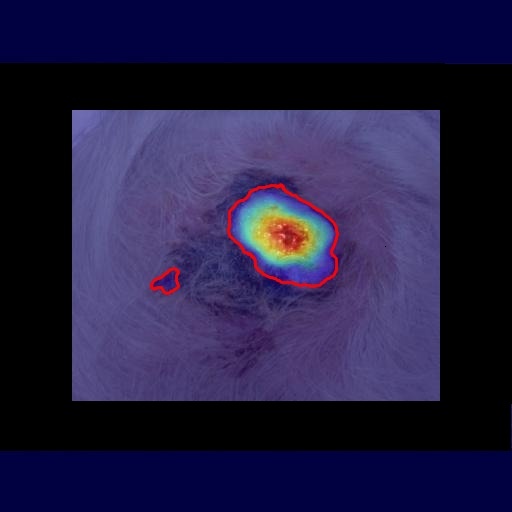}\\[4pt]
                \includegraphics[trim={160 90 20 90}, clip, height=1.5cm, angle=90]{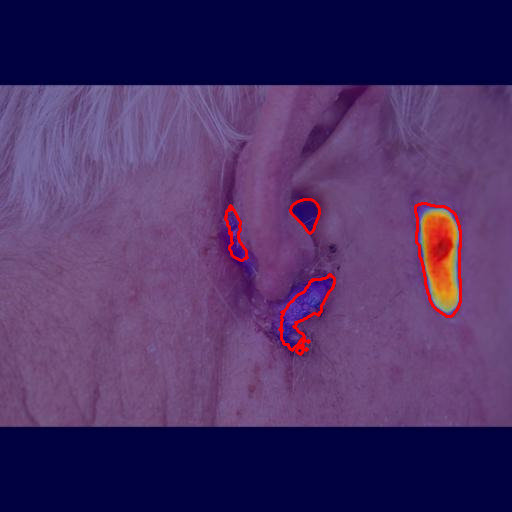}\\[4pt]
                \includegraphics[trim={64 10 64 0}, clip, height=1.5cm, angle=90]{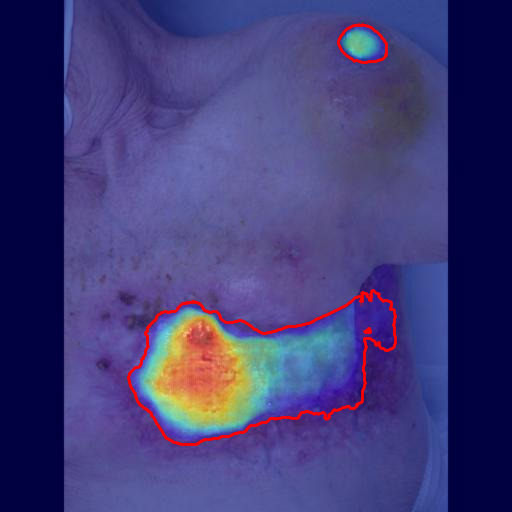}\\[4pt]
                \caption*{\scriptsize U-Net}
            \end{subfigure}
            \hspace{-0.6cm}
            \begin{subfigure}{0.165\textwidth}
                \centering
                \includegraphics[trim={80 120 70 140}, clip, width=1.5cm, angle=180]{images/XAI/Overlays/ulc0387_04012022_3_MISSFormer_overlay.jpg}\\[4pt]
                \includegraphics[trim={80 165 0 85}, clip, width=1.5cm, angle=180]{images/XAI/Overlays/ulc0390_04012022_6_MISSFormer_overlay.jpg}\\[4pt]
                \includegraphics[trim={30 65 0 65}, clip, width=1.5cm, angle=180]{images/XAI/Overlays/ulc0098_07072022_5_MISSFormer_overlay.jpg}\\[4pt]
                \includegraphics[trim={0 90 0 90}, clip, width=1.5cm, angle=180]{images/XAI/Overlays/ulc0411_17032022_16_MISSFormer_overlay.jpg}\\[4pt]
                \includegraphics[trim={150 90 60 85}, clip, height=1.5cm, angle=90]{images/XAI/Overlays/ulc0406_17032022_11_MISSFormer_overlay.jpg}\\[4pt]
                \includegraphics[trim={110 90 120 90}, clip, height=1.5cm, angle=90]{images/XAI/Overlays/ulc0396_17032022_1_MISSFormer_overlay.jpg}\\[4pt]
                \includegraphics[trim={90 0 90 0}, clip, height=1.5cm, angle=90]{images/XAI/Overlays/ulc0119_03112022_3_MISSFormer_overlay.jpg}\\[4pt]
                \includegraphics[trim={220 105 100 65}, clip, height=1.5cm, angle=90]{images/XAI/Overlays/ulc0355_05072022_4_MISSFormer_overlay.jpg}\\[4pt]
                \includegraphics[trim={110 0 85 40}, clip, height=1.5cm, angle=90]{images/XAI/Overlays/ulc0059_11072022_5_MISSFormer_overlay.jpg}\\[4pt]
                \includegraphics[trim={160 115 90 130}, clip, height=1.5cm, angle=90]{images/XAI/Overlays/ulc0343_18032022_1_MISSFormer_overlay.jpg}\\[4pt]
                \includegraphics[trim={160 90 20 90}, clip, height=1.5cm, angle=90]{images/XAI/Overlays/ulc0245_31052022_6_MISSFormer_overlay.jpg}\\[4pt]
                \includegraphics[trim={64 10 64 0}, clip, height=1.5cm, angle=90]{images/XAI/Overlays/ulc0048_01082022_3_MISSFormer_overlay.jpg}\\[4pt]
                \caption*{\scriptsize MISSForm.}
            \end{subfigure}
            \hspace{-0.6cm}
            \begin{subfigure}{0.165\textwidth}
                \centering
                \includegraphics[trim={80 120 70 140}, clip, width=1.5cm, angle=180]{images/XAI/Overlays/ulc0387_04012022_3_HiFormerB_overlay.jpg}\\[4pt]
                \includegraphics[trim={80 165 0 85}, clip, width=1.5cm, angle=180]{images/XAI/Overlays/ulc0390_04012022_6_HiFormerB_overlay.jpg}\\[4pt]
                \includegraphics[trim={30 65 0 65}, clip, width=1.5cm, angle=180]{images/XAI/Overlays/ulc0098_07072022_5_HiFormerB_overlay.jpg}\\[4pt]
                \includegraphics[trim={0 90 0 90}, clip, width=1.5cm, angle=180]{images/XAI/Overlays/ulc0411_17032022_16_HiFormerB_overlay.jpg}\\[4pt]
                \includegraphics[trim={150 90 60 85}, clip, height=1.5cm, angle=90]{images/XAI/Overlays/ulc0406_17032022_11_HiFormerB_overlay.jpg}\\[4pt]
                \includegraphics[trim={110 90 120 90}, clip, height=1.5cm, angle=90]{images/XAI/Overlays/ulc0396_17032022_1_HiFormerB_overlay.jpg}\\[4pt]
                \includegraphics[trim={90 0 90 0}, clip, height=1.5cm, angle=90]{images/XAI/Overlays/ulc0119_03112022_3_HiFormerB_overlay.jpg}\\[4pt]
                \includegraphics[trim={220 105 100 65}, clip, height=1.5cm, angle=90]{images/XAI/Overlays/ulc0355_05072022_4_HiFormerB_overlay.jpg}\\[4pt]
                \includegraphics[trim={110 0 85 40}, clip, height=1.5cm, angle=90]{images/XAI/Overlays/ulc0059_11072022_5_HiFormerB_overlay.jpg}\\[4pt]
                \includegraphics[trim={160 115 90 130}, clip, height=1.5cm, angle=90]{images/XAI/Overlays/ulc0343_18032022_1_HiFormerB_overlay.jpg}\\[4pt]
                \includegraphics[trim={160 90 20 90}, clip, height=1.5cm, angle=90]{images/XAI/Overlays/ulc0245_31052022_6_HiFormerB_overlay.jpg}\\[4pt]
                \includegraphics[trim={64 10 64 0}, clip, height=1.5cm, angle=90]{images/XAI/Overlays/ulc0048_01082022_3_HiFormerB_overlay.jpg}\\[4pt]
                \caption*{\scriptsize HiFormer}
            \end{subfigure}
        \end{adjustbox}
    \end{turn}
    \caption{Grad-CAM for exemplary OOD images, alongside their GT annotation.}
    \label{fig:eval:grad_cam:all_models}
\end{figure}

\begin{figure}[htb!]
    \centering
    \begin{turn}{-90}
        \begin{adjustbox}{max height=\textwidth, max width=\textheight, keepaspectratio}
            \begin{subfigure}{0.165\textwidth}
                \centering
                \includegraphics[trim={0 85 0 85}, clip, width=1.5cm, angle=180]{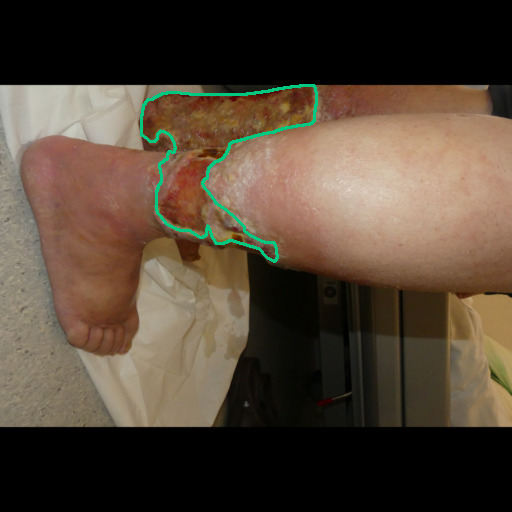}\\[4pt]
                \includegraphics[trim={40 90 40 90}, clip, width=1.5cm, angle=180]{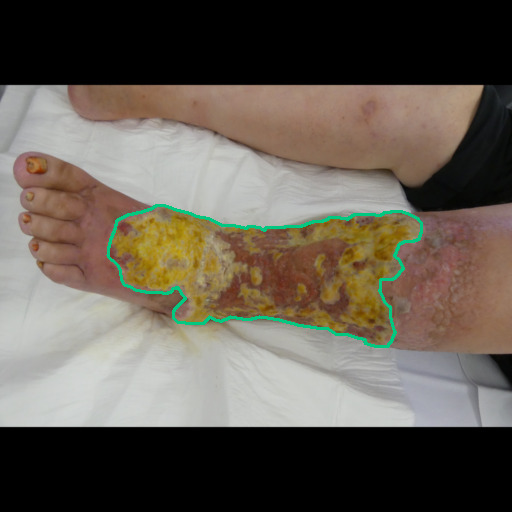}\\[4pt]
                \includegraphics[trim={50 120 90 120}, clip, width=1.5cm, angle=0]{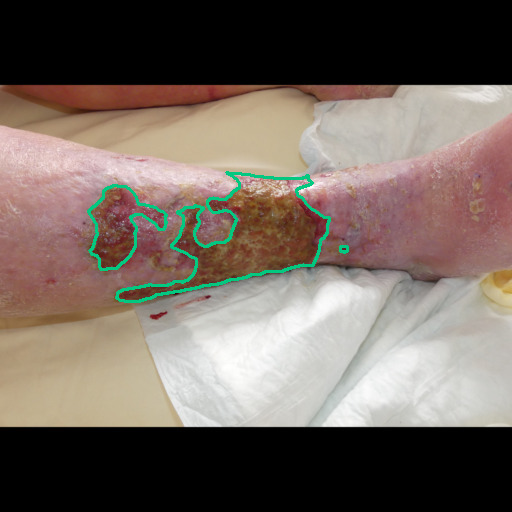}\\[4pt]
                \includegraphics[trim={85 0 85 0}, clip, height=1.5cm, angle=90]{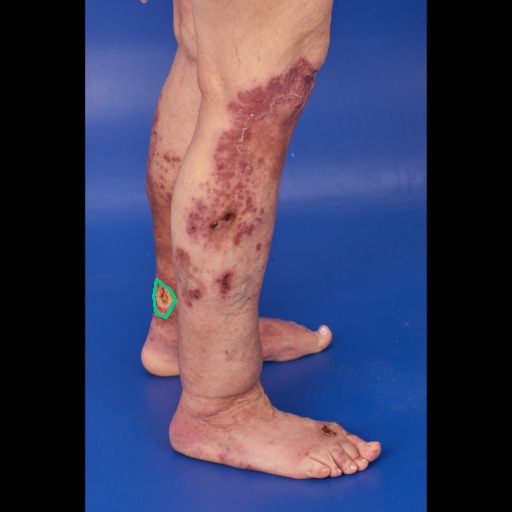}\\[4pt]
                \includegraphics[trim={85 0 85 50}, clip, height=1.5cm, angle=90]{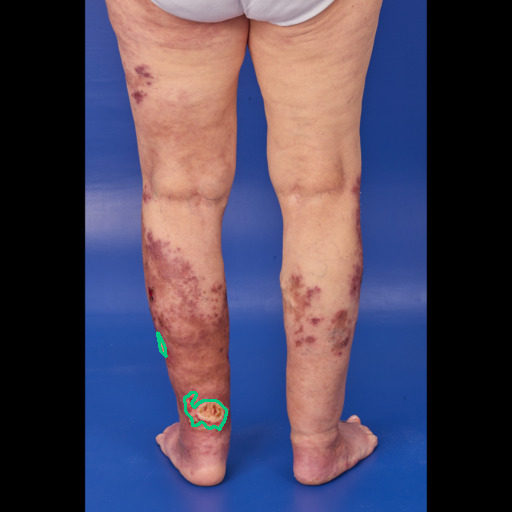}\\[4pt]
                \includegraphics[trim={140 90 130 150}, clip, height=1.5cm, angle=90]{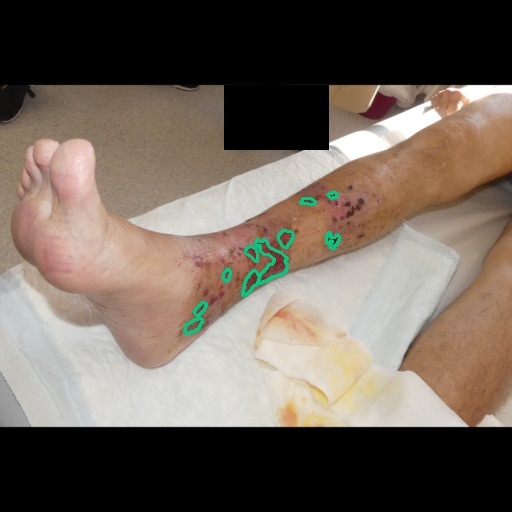}\\[4pt] 
                \includegraphics[trim={100 90 100 90}, clip, height=1.5cm, angle=90]{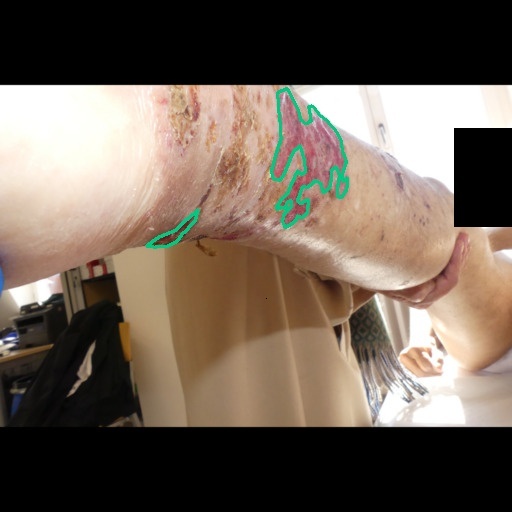}\\[4pt]
                \includegraphics[trim={110 90 100 90}, clip, height=1.5cm, angle=90]{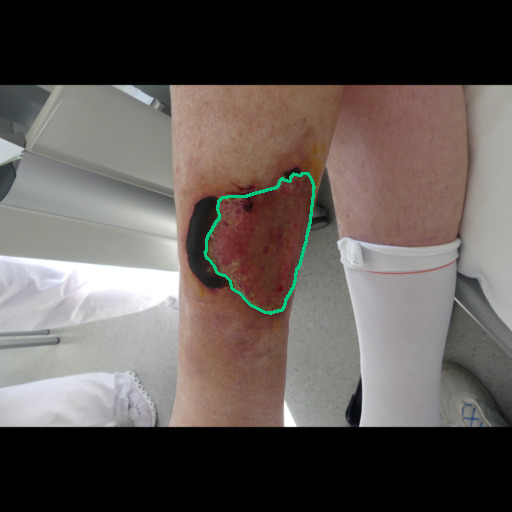}\\[4pt]
                \includegraphics[trim={130 90 145 90}, clip, height=1.5cm, angle=90]{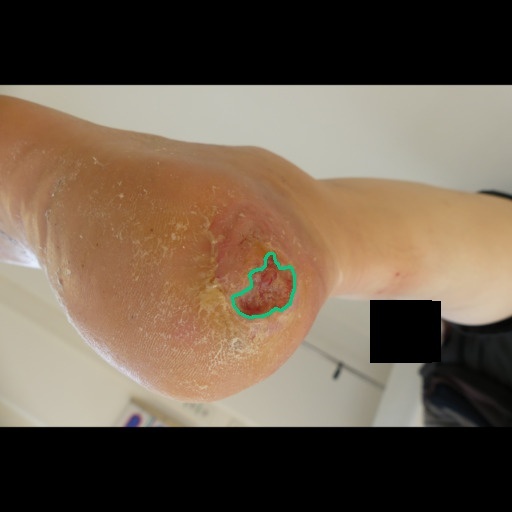}\\[4pt]
                \includegraphics[trim={80 90 120 90}, clip, height=1.5cm, angle=90]{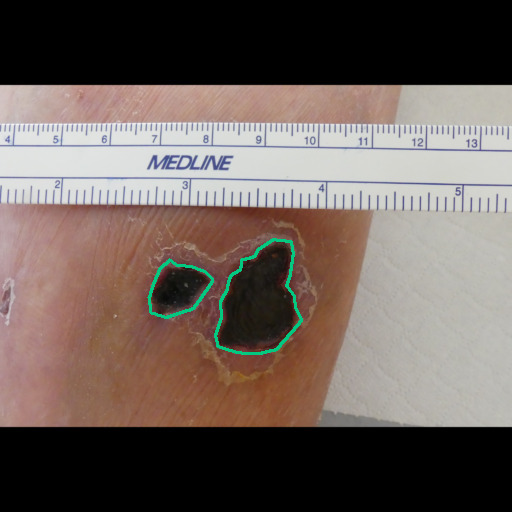}\\[4pt]
                \includegraphics[trim={110 0 110 140}, clip, height=1.5cm, angle=90]{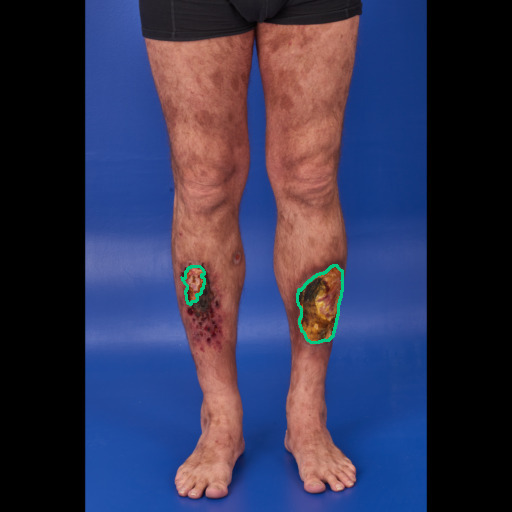}\\[4pt]
                \caption*{\scriptsize GT}
            \end{subfigure}
            \hspace{-0.6cm}
            \begin{subfigure}{0.165\textwidth}
                \centering
                \includegraphics[trim={0 85 0 85}, clip, width=1.5cm, angle=180]{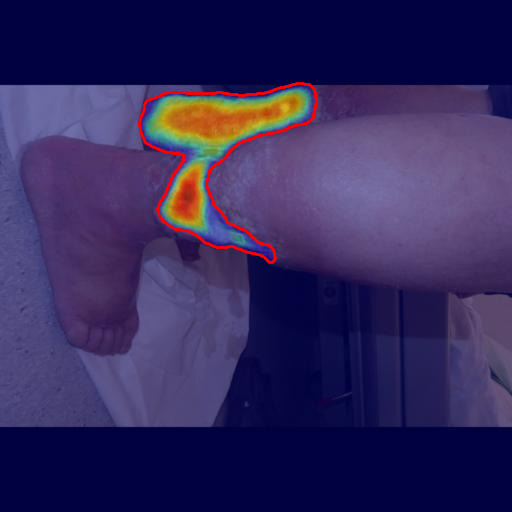}\\[4pt]
                \includegraphics[trim={40 90 40 90}, clip, width=1.5cm, angle=180]{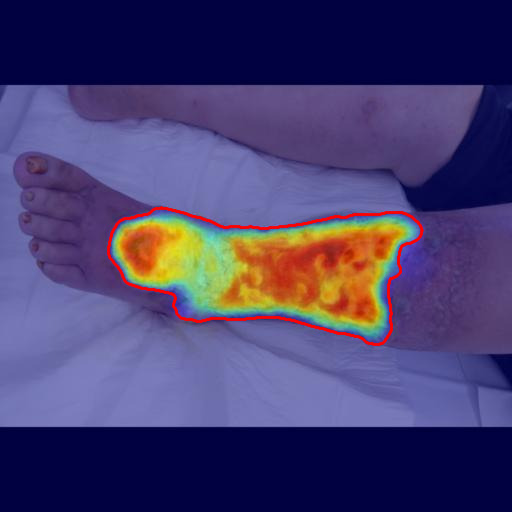}\\[4pt]
                \includegraphics[trim={50 120 90 120}, clip, width=1.5cm, angle=0]{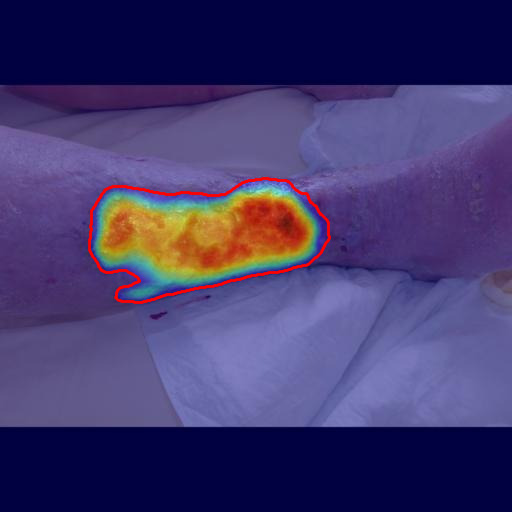}\\[4pt]
                \includegraphics[trim={85 0 85 0}, clip, height=1.5cm, angle=90]{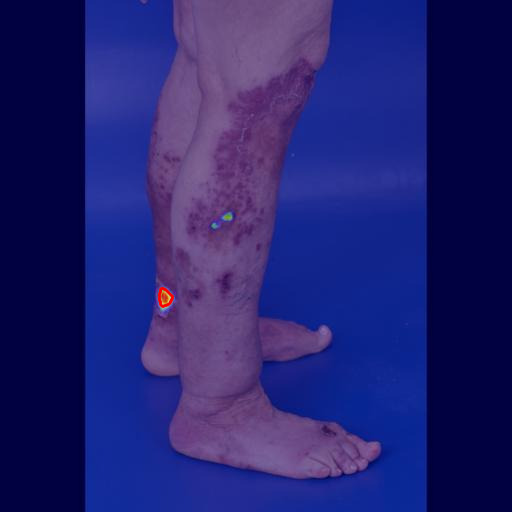}\\[4pt]
                \includegraphics[trim={85 0 85 50}, clip, height=1.5cm, angle=90]{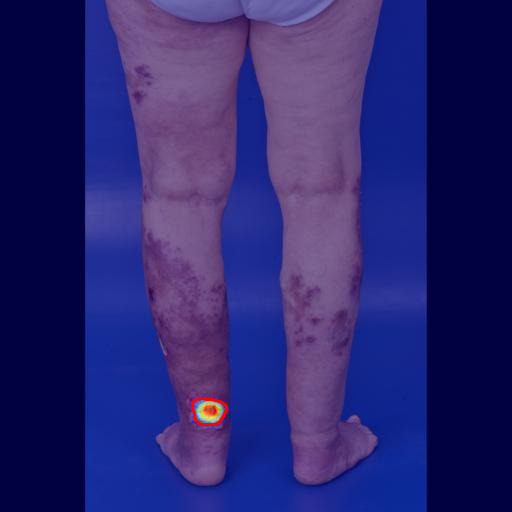}\\[4pt]
                \includegraphics[trim={140 90 130 150}, clip, height=1.5cm, angle=90]{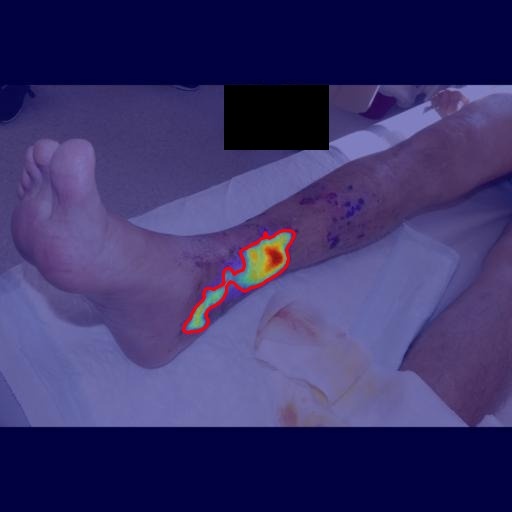}\\[4pt]
                \includegraphics[trim={100 90 100 90}, clip, height=1.5cm, angle=90]{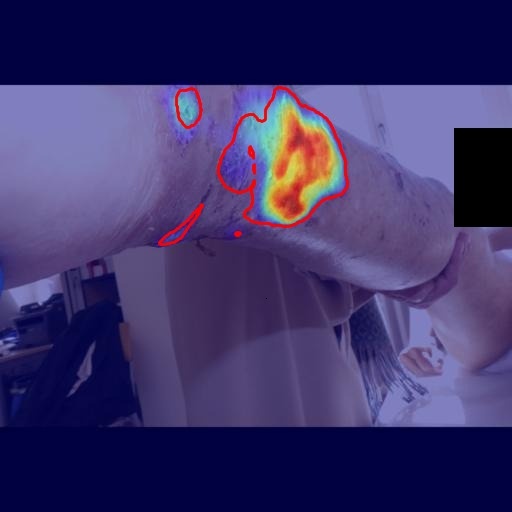}\\[4pt]
                \includegraphics[trim={110 90 100 90}, clip, height=1.5cm, angle=90]{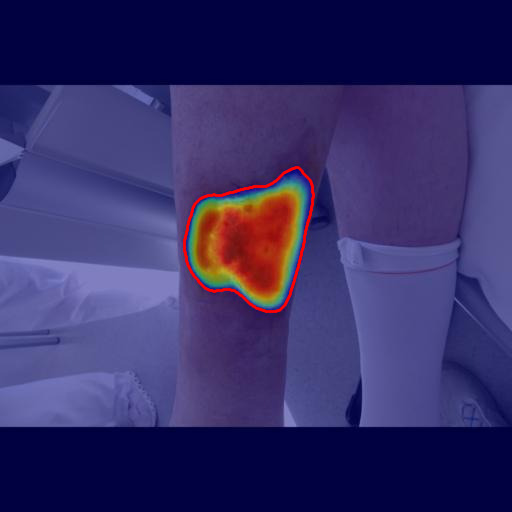}\\[4pt]
                \includegraphics[trim={130 90 145 90}, clip, height=1.5cm, angle=90]{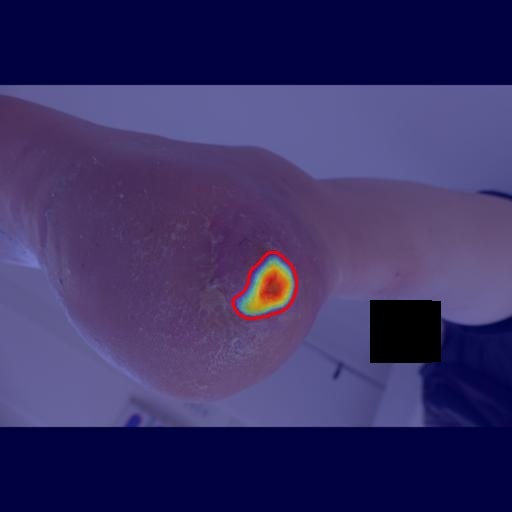}\\[4pt]
                \includegraphics[trim={80 90 120 90}, clip, height=1.5cm, angle=90]{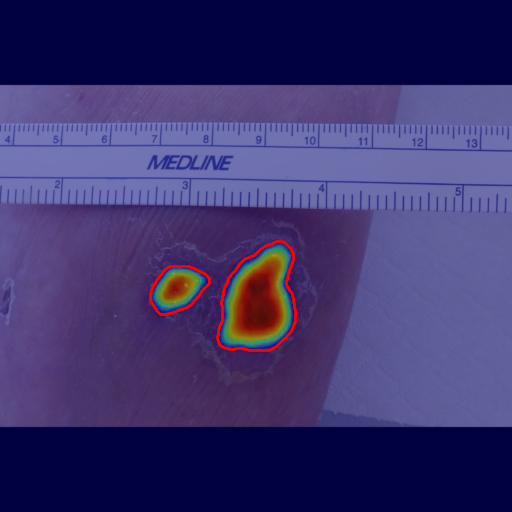}\\[4pt]
                \includegraphics[trim={110 0 110 140}, clip, height=1.5cm, angle=90]{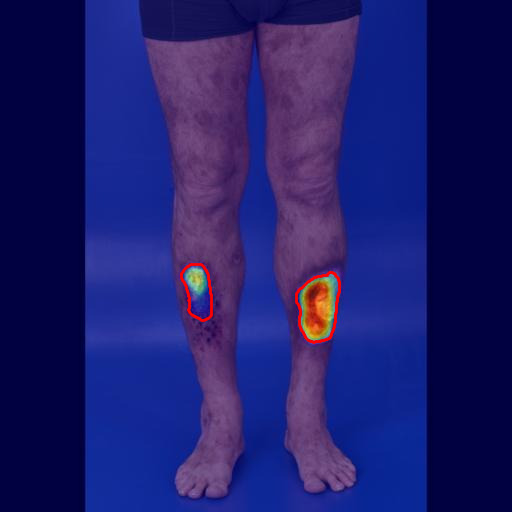}\\[4pt]
                \caption*{\scriptsize TransN.}
            \end{subfigure}
            \hspace{-0.6cm}
            \begin{subfigure}{0.165\textwidth}
                \centering
                \includegraphics[trim={0 85 0 85}, clip, width=1.5cm, angle=180]{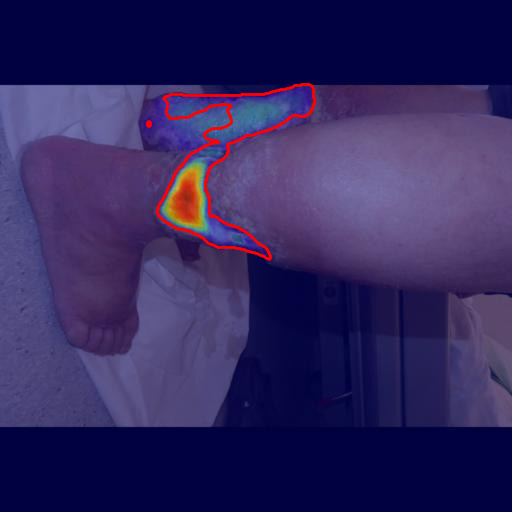}\\[4pt]
                \includegraphics[trim={40 90 40 90}, clip, width=1.5cm, angle=180]{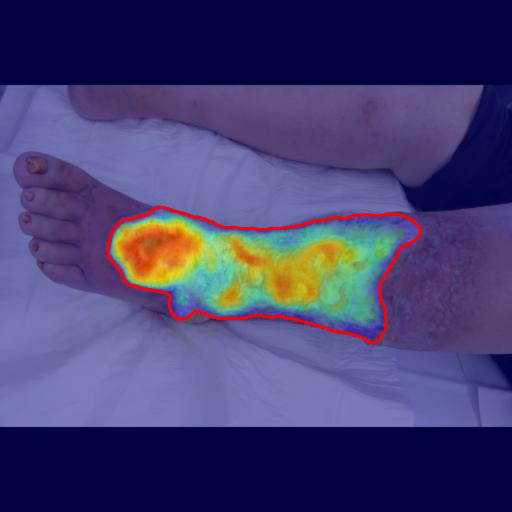}\\[4pt]
                \includegraphics[trim={50 120 90 120}, clip, width=1.5cm, angle=0]{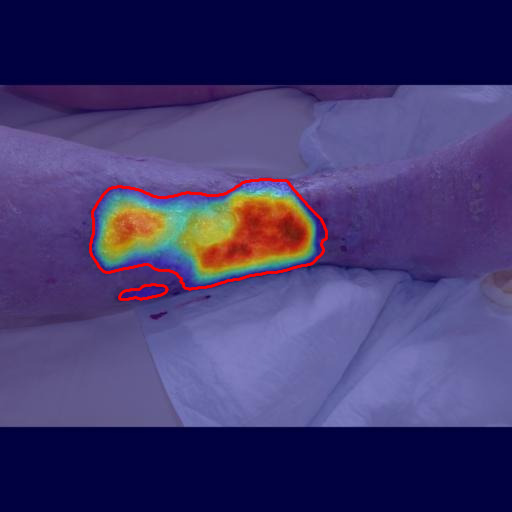}\\[4pt]
                \includegraphics[trim={85 0 85 0}, clip, height=1.5cm, angle=90]{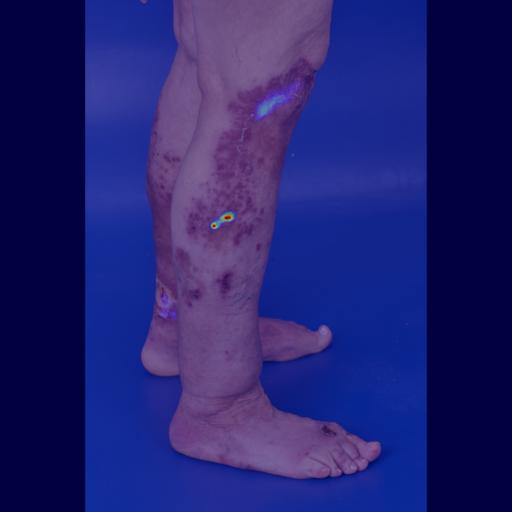}\\[4pt]
                \includegraphics[trim={85 0 85 50}, clip, height=1.5cm, angle=90]{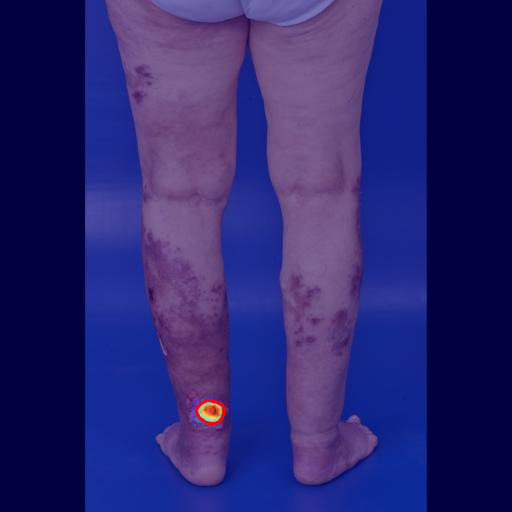}\\[4pt]
                \includegraphics[trim={140 90 130 150}, clip, height=1.5cm, angle=90]{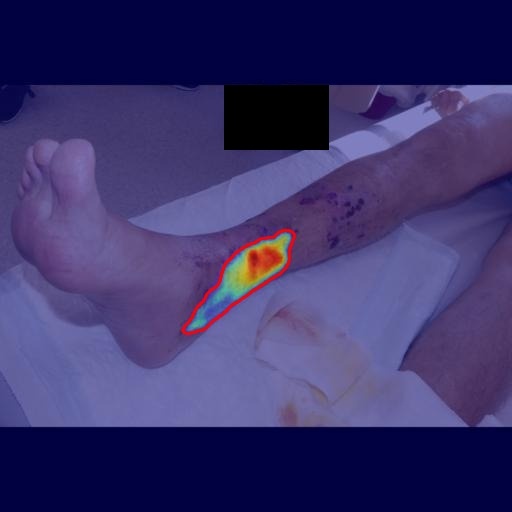}\\[4pt]
                \includegraphics[trim={100 90 100 90}, clip, height=1.5cm, angle=90]{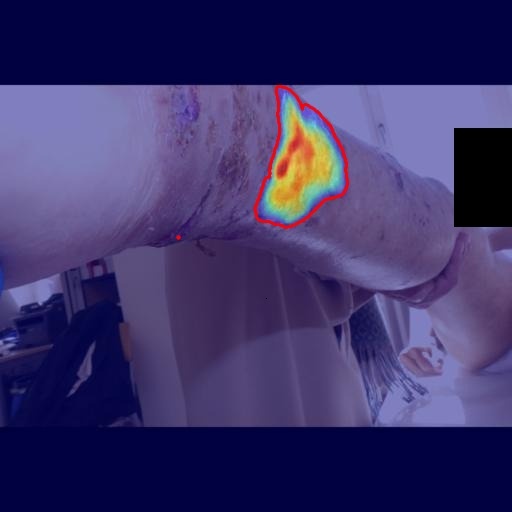}\\[4pt]
                \includegraphics[trim={110 90 100 90}, clip, height=1.5cm, angle=90]{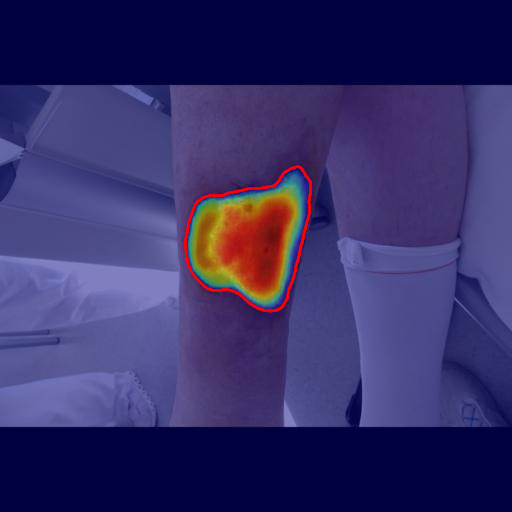}\\[4pt]
                \includegraphics[trim={130 90 145 90}, clip, height=1.5cm, angle=90]{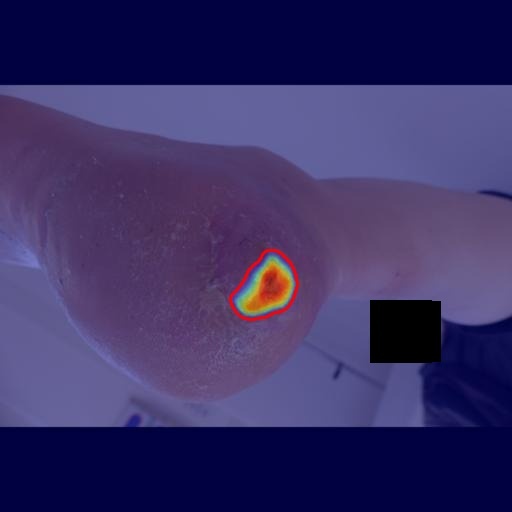}\\[4pt]
                \includegraphics[trim={80 90 120 90}, clip, height=1.5cm, angle=90]{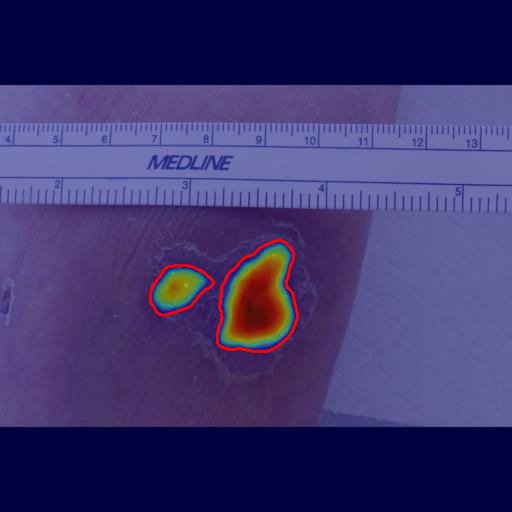}\\[4pt]
                \includegraphics[trim={110 0 110 140}, clip, height=1.5cm, angle=90]{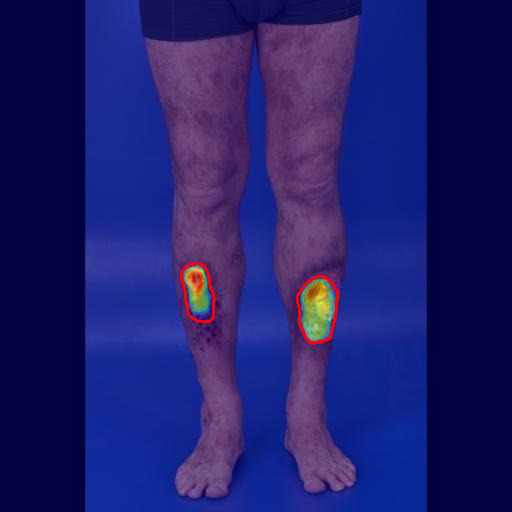}\\[4pt]
                \caption*{\scriptsize InternIm.}
            \end{subfigure}
            \hspace{-0.6cm}
            \begin{subfigure}{0.165\textwidth}
                \centering
                \includegraphics[trim={0 85 0 85}, clip, width=1.5cm, angle=180]{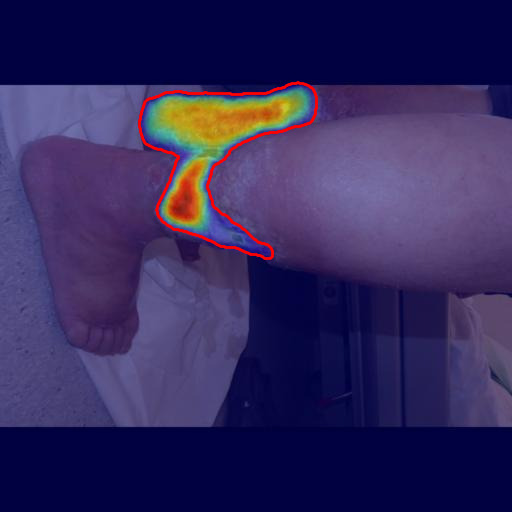}\\[4pt]
                \includegraphics[trim={40 90 40 90}, clip, width=1.5cm, angle=180]{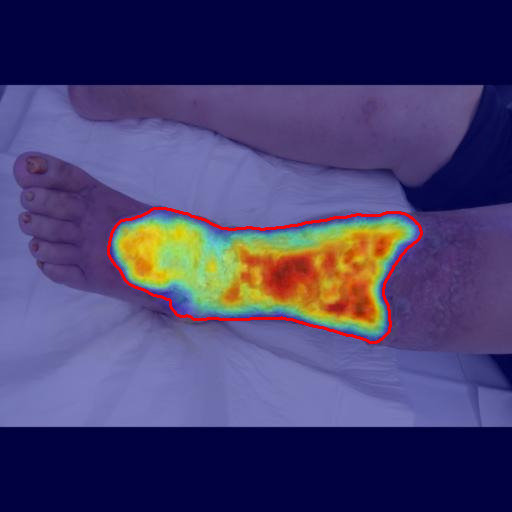}\\[4pt]
                \includegraphics[trim={50 120 90 120}, clip, width=1.5cm, angle=0]{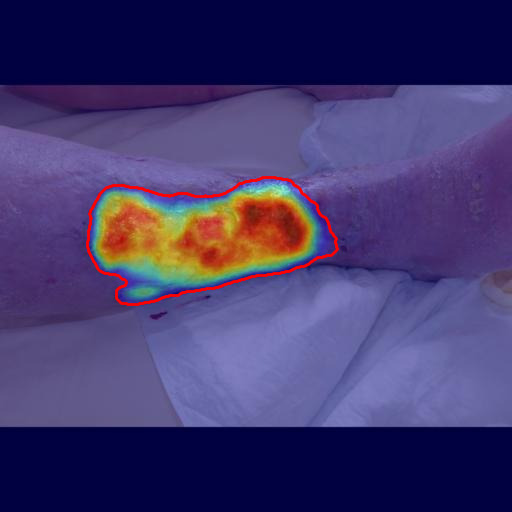}\\[4pt]
                \includegraphics[trim={85 0 85 0}, clip, height=1.5cm, angle=90]{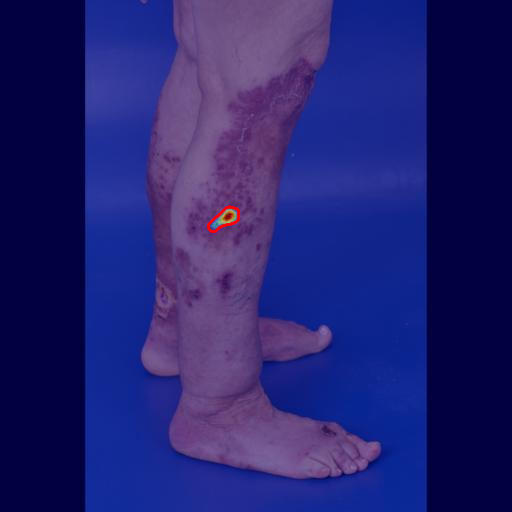}\\[4pt]
                \includegraphics[trim={85 0 85 50}, clip, height=1.5cm, angle=90]{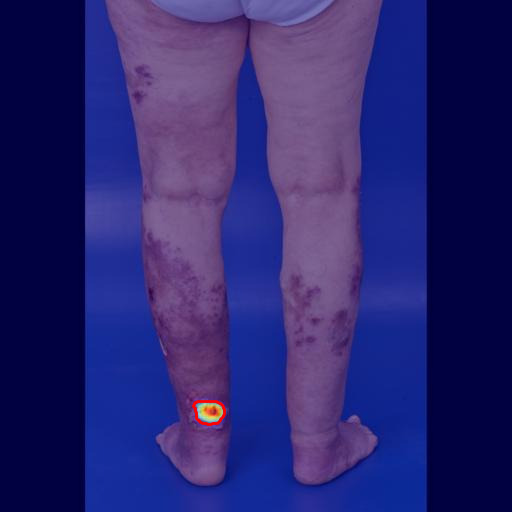}\\[4pt]
                \includegraphics[trim={140 90 130 150}, clip, height=1.5cm, angle=90]{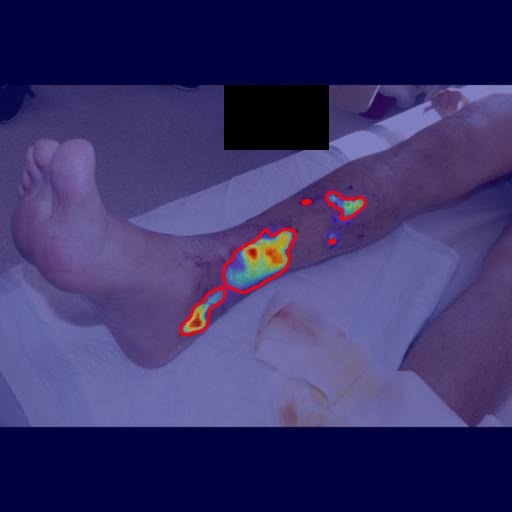}\\[4pt]
                \includegraphics[trim={100 90 100 90}, clip, height=1.5cm, angle=90]{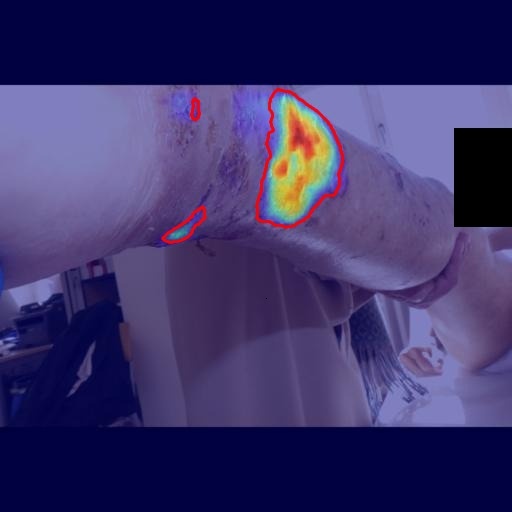}\\[4pt]
                \includegraphics[trim={110 90 100 90}, clip, height=1.5cm, angle=90]{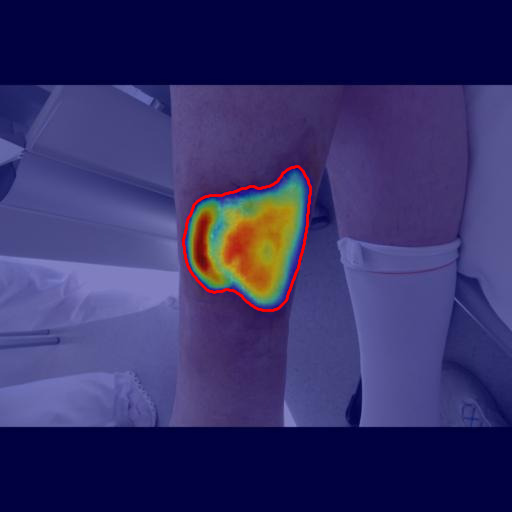}\\[4pt]
                \includegraphics[trim={130 90 145 90}, clip, height=1.5cm, angle=90]{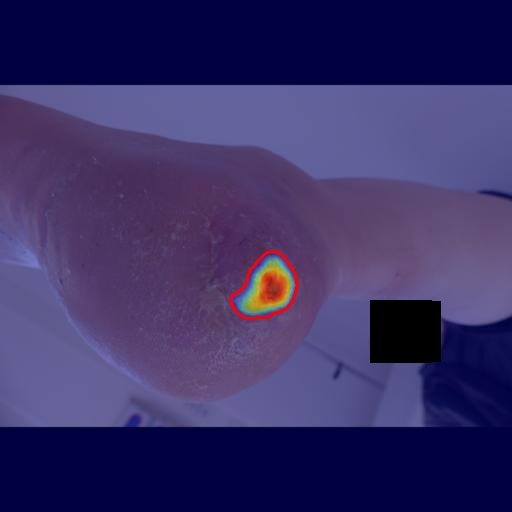}\\[4pt]
                \includegraphics[trim={80 90 120 90}, clip, height=1.5cm, angle=90]{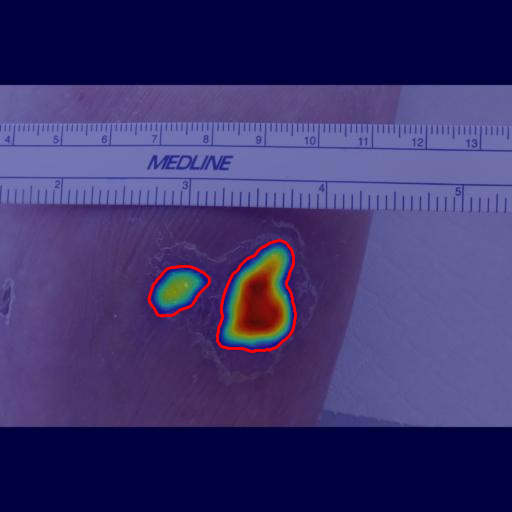}\\[4pt]
                \includegraphics[trim={110 0 110 140}, clip, height=1.5cm, angle=90]{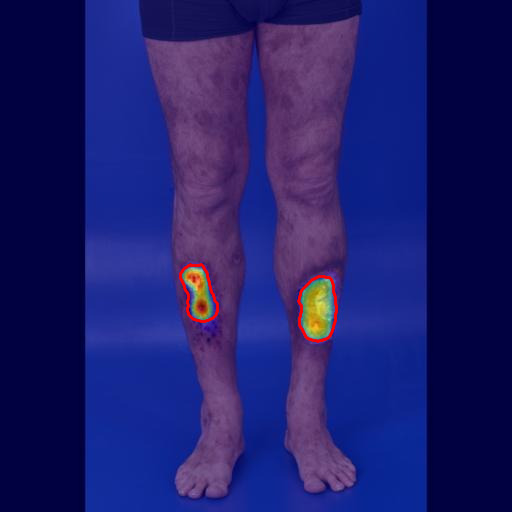}\\[4pt]
                \caption*{\scriptsize VW-MiT}
            \end{subfigure}
            \hspace{-0.6cm}
            \begin{subfigure}{0.165\textwidth}
                \centering
                \includegraphics[trim={0 85 0 85}, clip, width=1.5cm, angle=180]{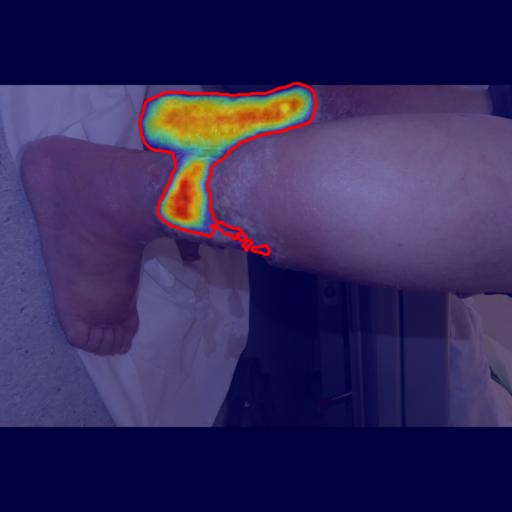}\\[4pt]
                \includegraphics[trim={40 90 40 90}, clip, width=1.5cm, angle=180]{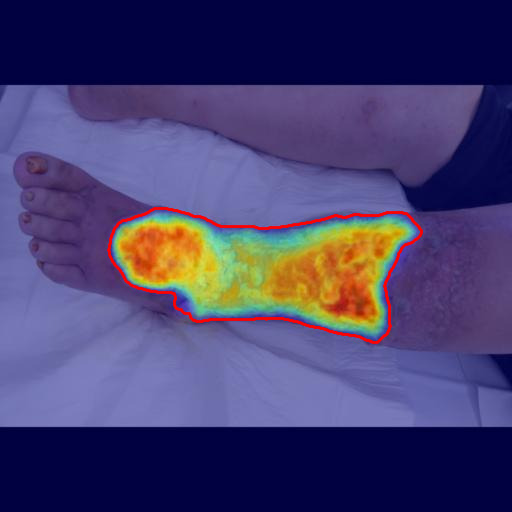}\\[4pt]
                \includegraphics[trim={50 120 90 120}, clip, width=1.5cm, angle=0]{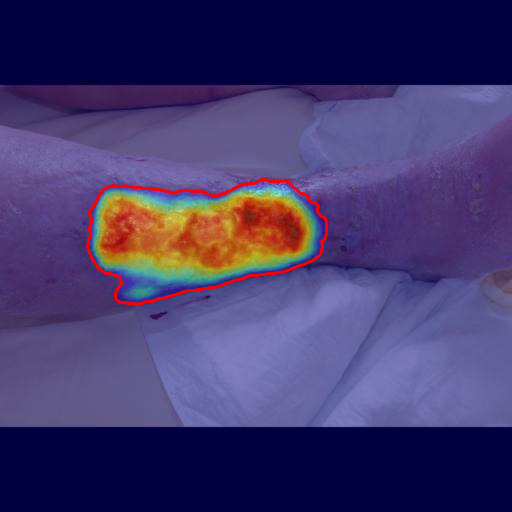}\\[4pt]
                \includegraphics[trim={85 0 85 0}, clip, height=1.5cm, angle=90]{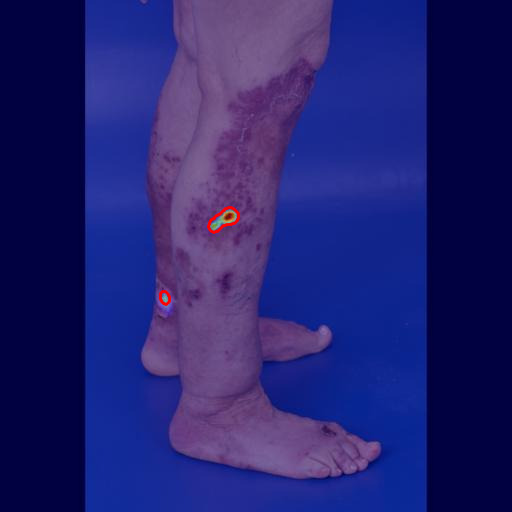}\\[4pt]
                \includegraphics[trim={85 0 85 50}, clip, height=1.5cm, angle=90]{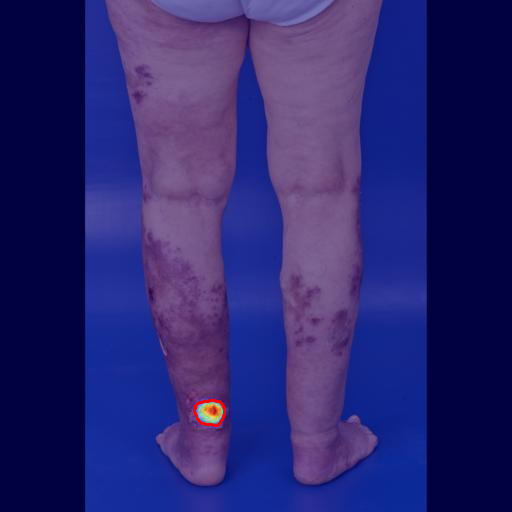}\\[4pt]
                \includegraphics[trim={140 90 130 150}, clip, height=1.5cm, angle=90]{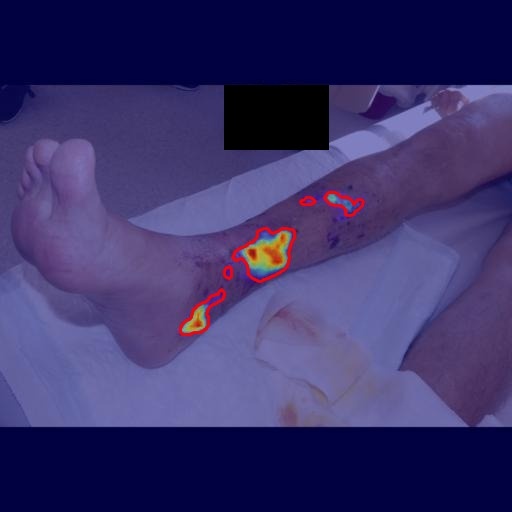}\\[4pt]
                \includegraphics[trim={100 90 100 90}, clip, height=1.5cm, angle=90]{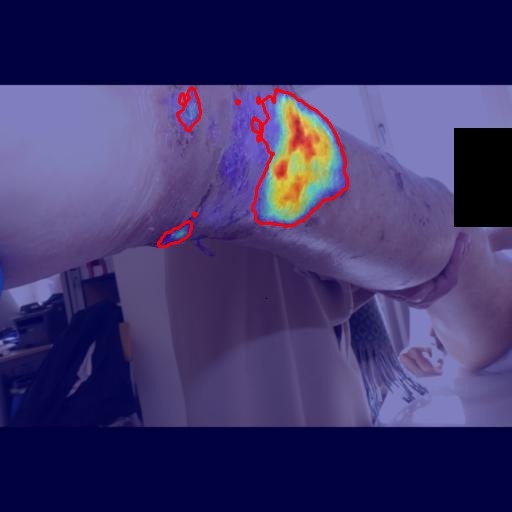}\\[4pt]
                \includegraphics[trim={110 90 100 90}, clip, height=1.5cm, angle=90]{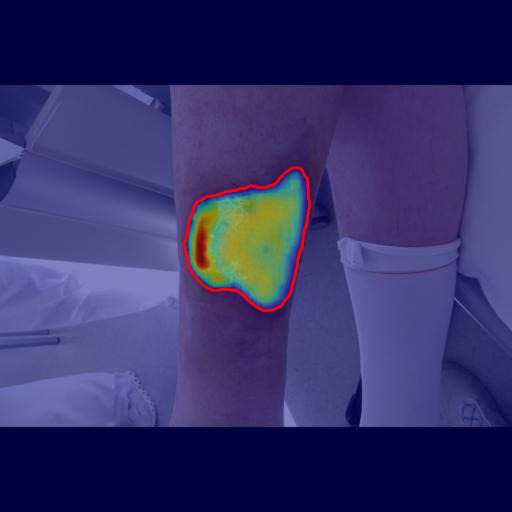}\\[4pt]
                \includegraphics[trim={130 90 145 90}, clip, height=1.5cm, angle=90]{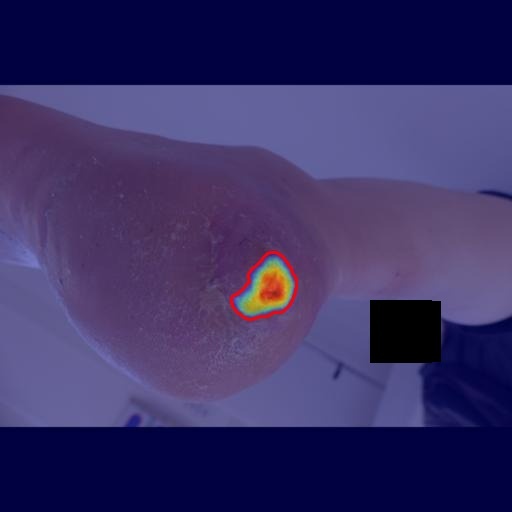}\\[4pt]
                \includegraphics[trim={80 90 120 90}, clip, height=1.5cm, angle=90]{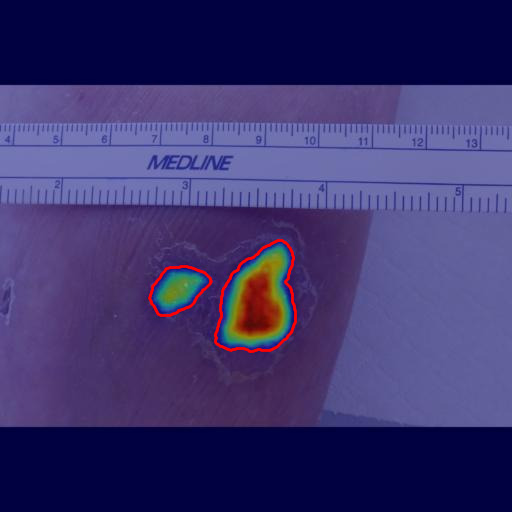}\\[4pt]
                \includegraphics[trim={110 0 110 140}, clip, height=1.5cm, angle=90]{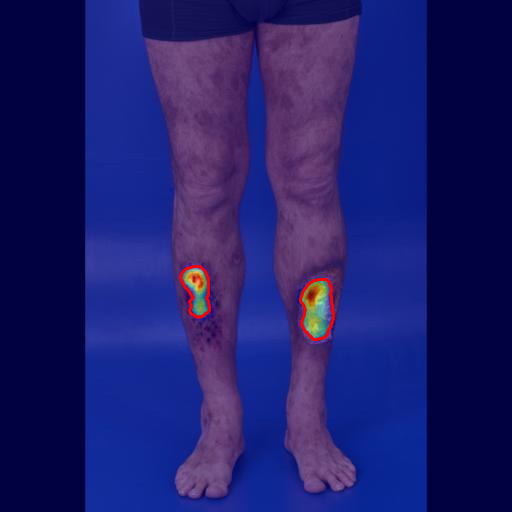}\\[4pt]
                \caption*{\scriptsize SegForm.}
            \end{subfigure}
            \hspace{-0.6cm}
            \begin{subfigure}{0.165\textwidth}
                \centering
                \includegraphics[trim={0 85 0 85}, clip, width=1.5cm, angle=180]{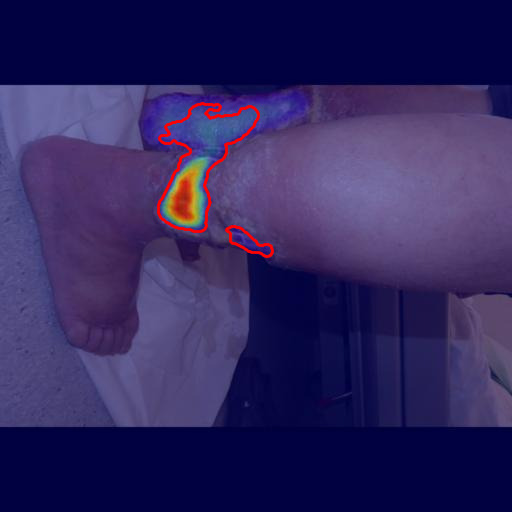}\\[4pt]
                \includegraphics[trim={40 90 40 90}, clip, width=1.5cm, angle=180]{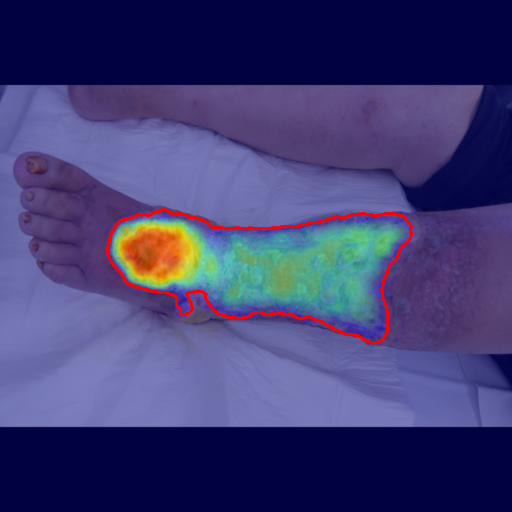}\\[4pt]
                \includegraphics[trim={50 120 90 120}, clip, width=1.5cm, angle=0]{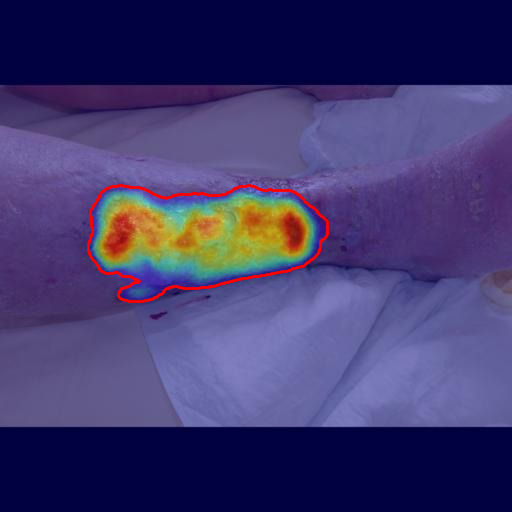}\\[4pt]
                \includegraphics[trim={85 0 85 0}, clip, height=1.5cm, angle=90]{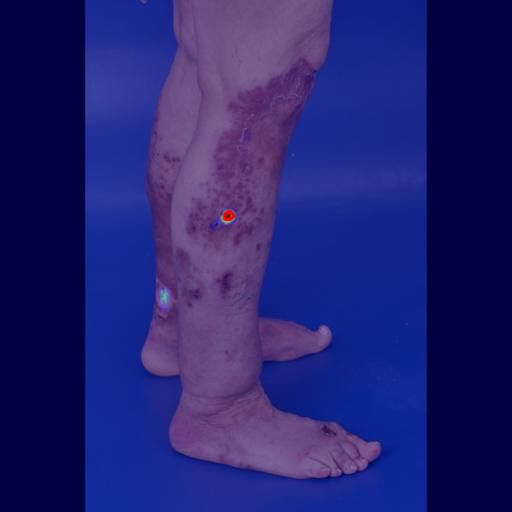}\\[4pt]
                \includegraphics[trim={85 0 85 50}, clip, height=1.5cm, angle=90]{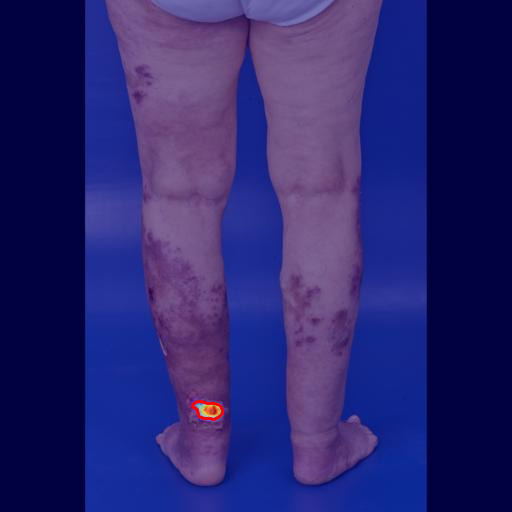}\\[4pt]
                \includegraphics[trim={140 90 130 150}, clip, height=1.5cm, angle=90]{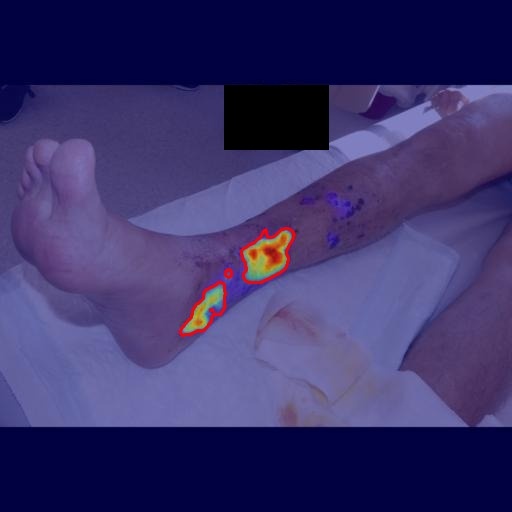}\\[4pt]
                \includegraphics[trim={100 90 100 90}, clip, height=1.5cm, angle=90]{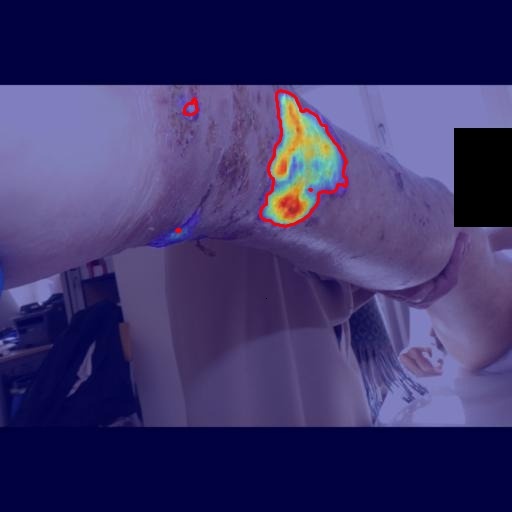}\\[4pt]
                \includegraphics[trim={110 90 100 90}, clip, height=1.5cm, angle=90]{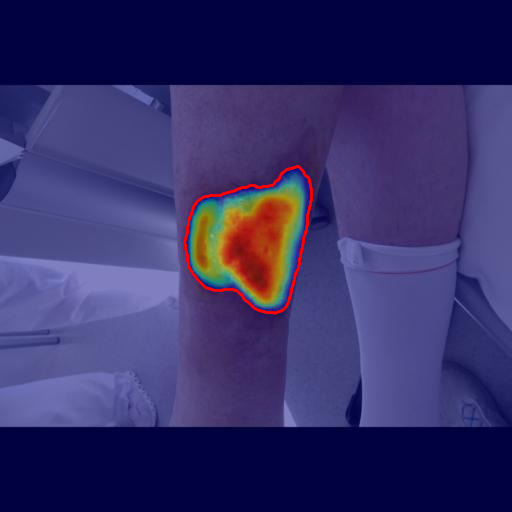}\\[4pt]
                \includegraphics[trim={130 90 145 90}, clip, height=1.5cm, angle=90]{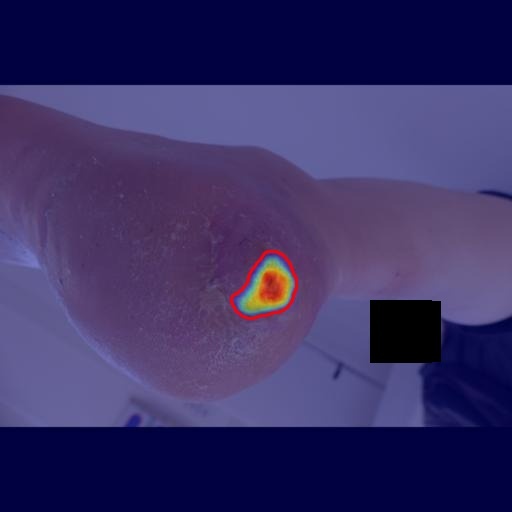}\\[4pt]
                \includegraphics[trim={80 90 120 90}, clip, height=1.5cm, angle=90]{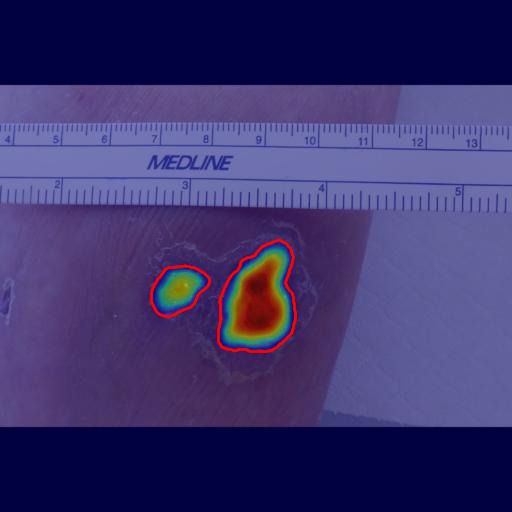}\\[4pt]
                \includegraphics[trim={110 0 110 140}, clip, height=1.5cm, angle=90]{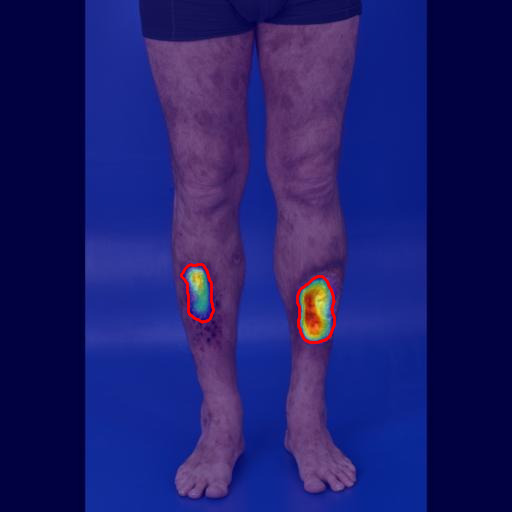}\\[4pt]
                \caption*{\scriptsize VW-Conv}
            \end{subfigure}
            \hspace{-0.6cm}
            \begin{subfigure}{0.165\textwidth}
                \centering
                \includegraphics[trim={0 85 0 85}, clip, width=1.5cm, angle=180]{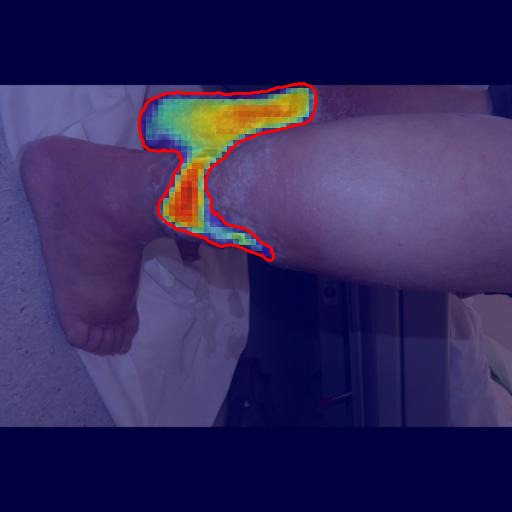}\\[4pt]
                \includegraphics[trim={40 90 40 90}, clip, width=1.5cm, angle=180]{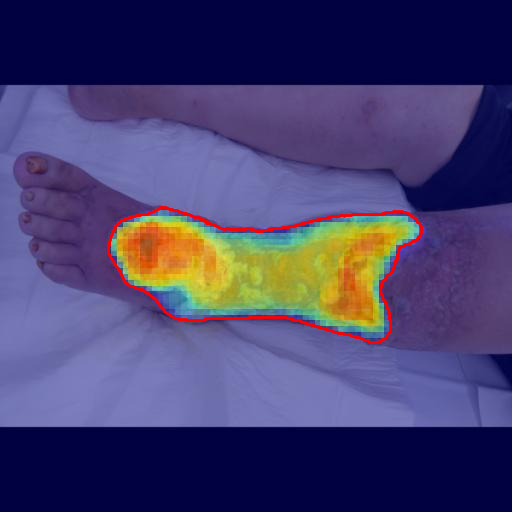}\\[4pt]
                \includegraphics[trim={50 120 90 120}, clip, width=1.5cm, angle=0]{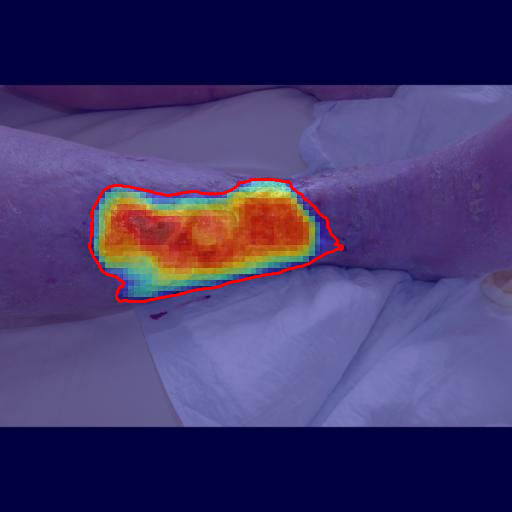}\\[4pt]
                \includegraphics[trim={85 0 85 0}, clip, height=1.5cm, angle=90]{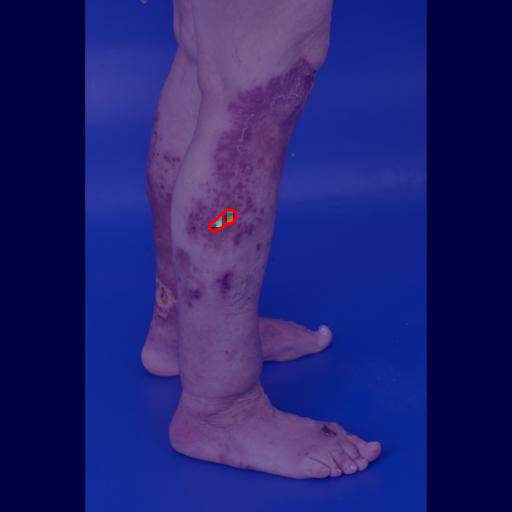}\\[4pt]
                \includegraphics[trim={85 0 85 50}, clip, height=1.5cm, angle=90]{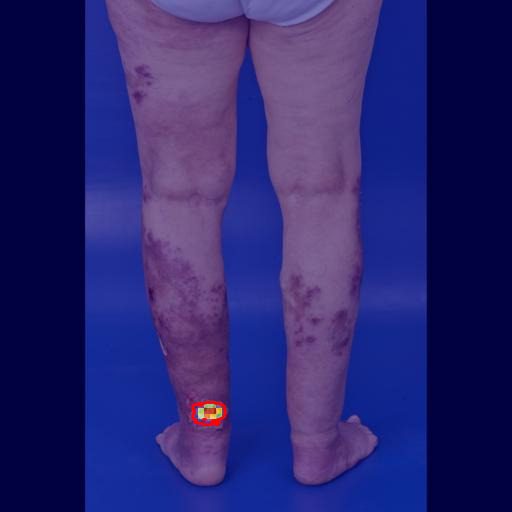}\\[4pt]
                \includegraphics[trim={140 90 130 150}, clip, height=1.5cm, angle=90]{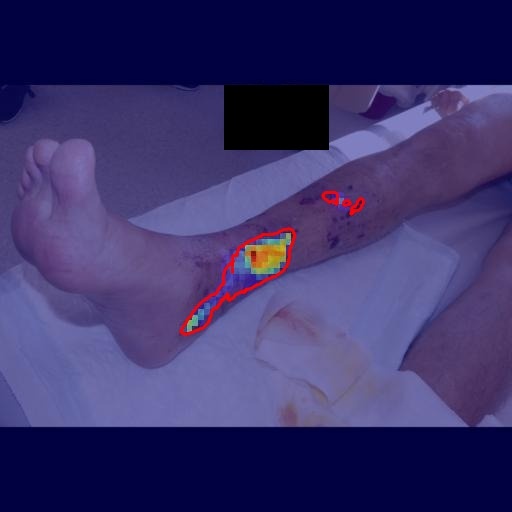}\\[4pt]
                \includegraphics[trim={100 90 100 90}, clip, height=1.5cm, angle=90]{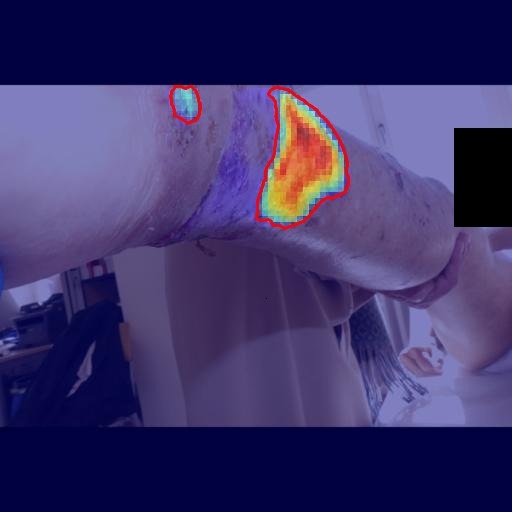}\\[4pt]
                \includegraphics[trim={110 90 100 90}, clip, height=1.5cm, angle=90]{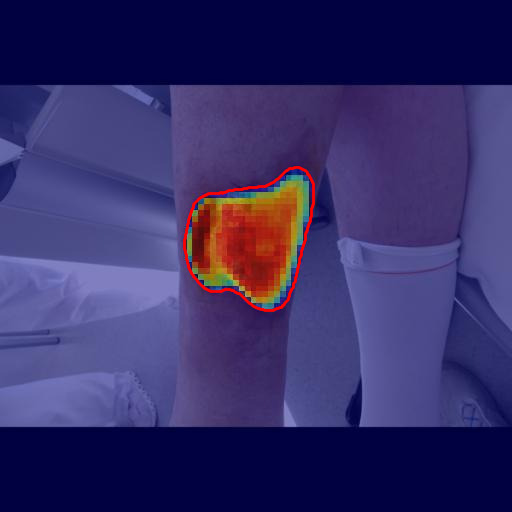}\\[4pt]
                \includegraphics[trim={130 90 145 90}, clip, height=1.5cm, angle=90]{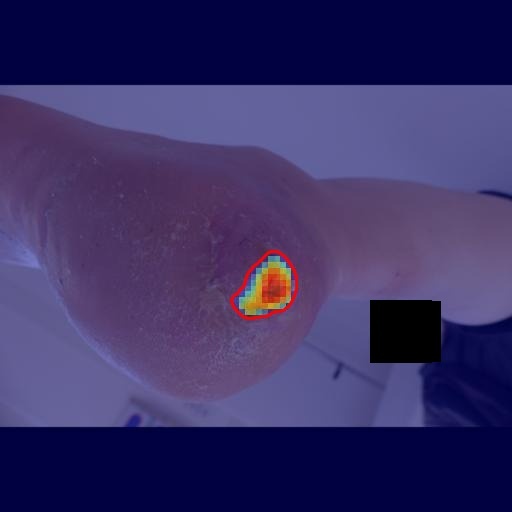}\\[4pt]
                \includegraphics[trim={80 90 120 90}, clip, height=1.5cm, angle=90]{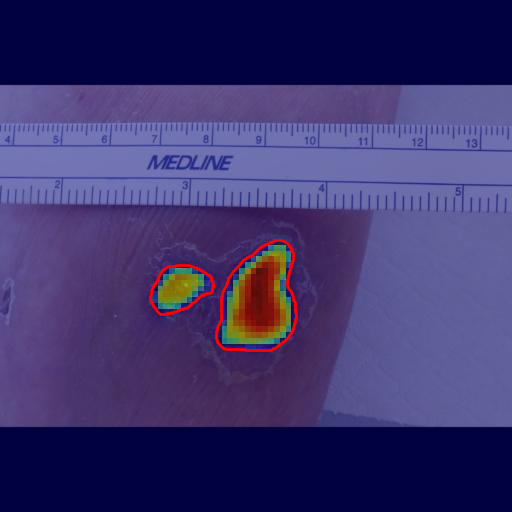}\\[4pt]
                \includegraphics[trim={110 0 110 140}, clip, height=1.5cm, angle=90]{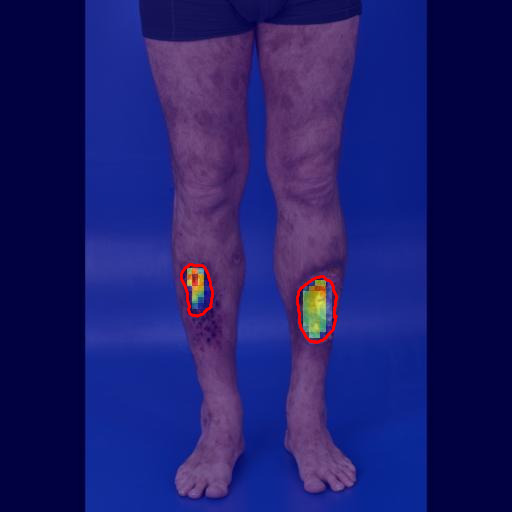}\\[4pt]
                \caption*{\scriptsize FCBForm.}
            \end{subfigure}
            \hspace{-0.6cm}
            \begin{subfigure}{0.165\textwidth}
                \centering
                \includegraphics[trim={0 85 0 85}, clip, width=1.5cm, angle=180]{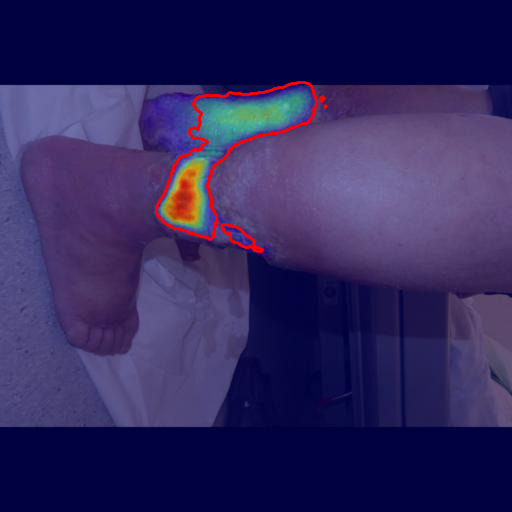}\\[4pt]
                \includegraphics[trim={40 90 40 90}, clip, width=1.5cm, angle=180]{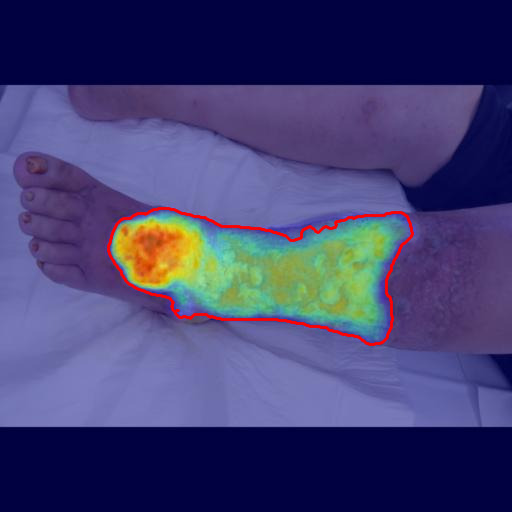}\\[4pt]
                \includegraphics[trim={50 120 90 120}, clip, width=1.5cm, angle=0]{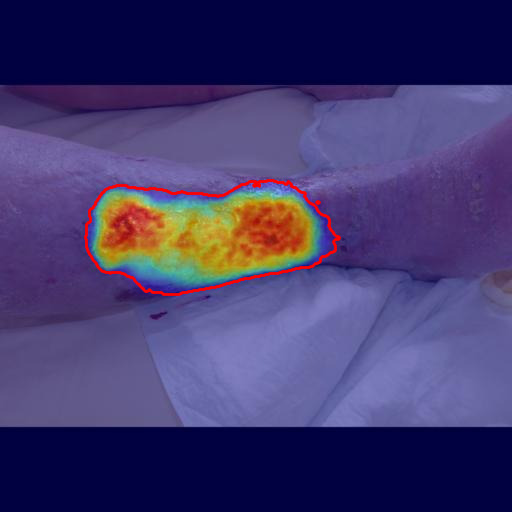}\\[4pt]
                \includegraphics[trim={85 0 85 0}, clip, height=1.5cm, angle=90]{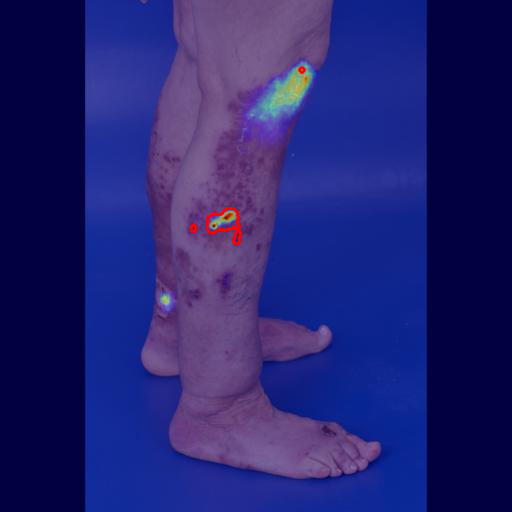}\\[4pt]
                \includegraphics[trim={85 0 85 50}, clip, height=1.5cm, angle=90]{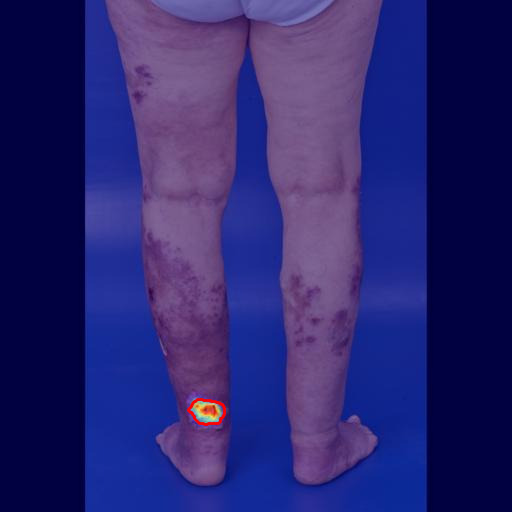}\\[4pt]
                \includegraphics[trim={140 90 130 150}, clip, height=1.5cm, angle=90]{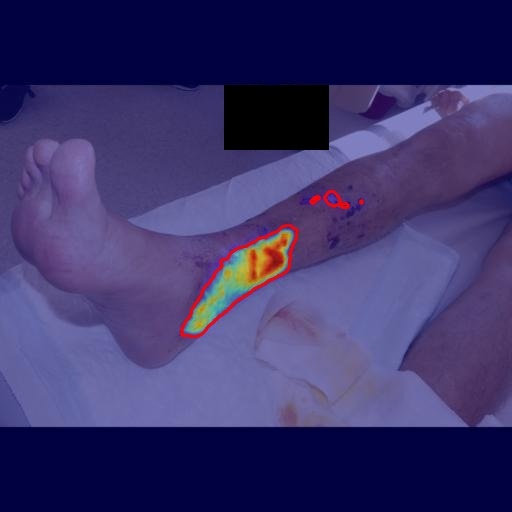}\\[4pt]
                \includegraphics[trim={100 90 100 90}, clip, height=1.5cm, angle=90]{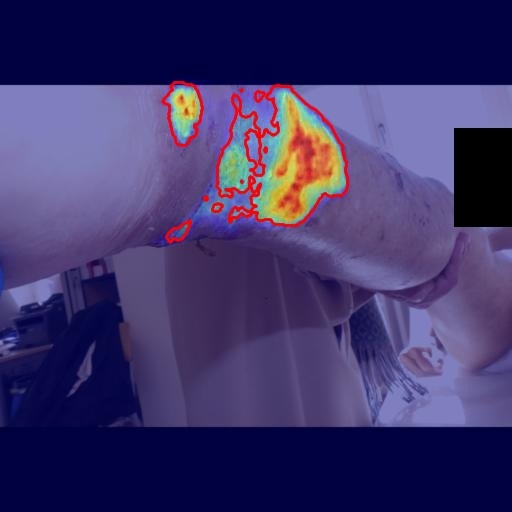}\\[4pt]
                \includegraphics[trim={110 90 100 90}, clip, height=1.5cm, angle=90]{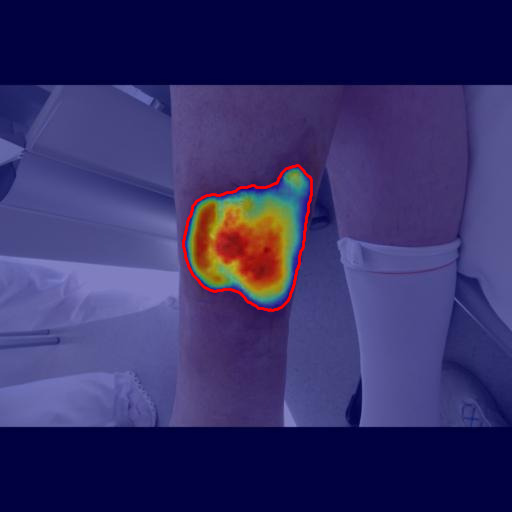}\\[4pt]
                \includegraphics[trim={130 90 145 90}, clip, height=1.5cm, angle=90]{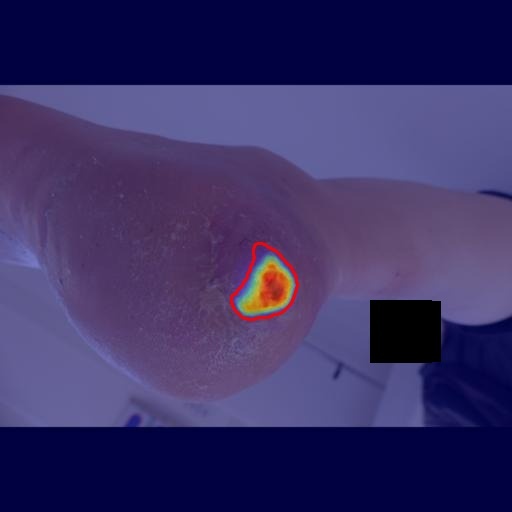}\\[4pt]
                \includegraphics[trim={80 90 120 90}, clip, height=1.5cm, angle=90]{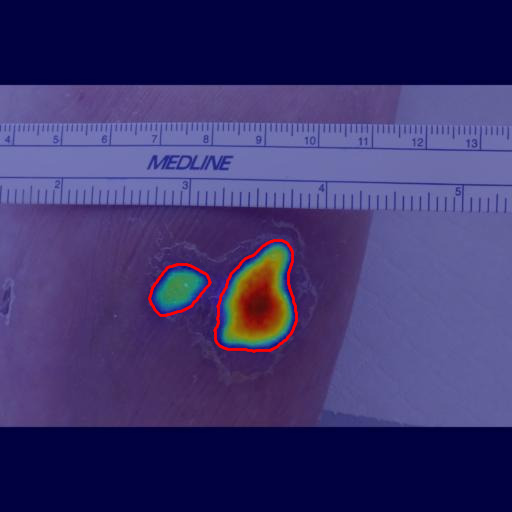}\\[4pt]
                \includegraphics[trim={110 0 110 140}, clip, height=1.5cm, angle=90]{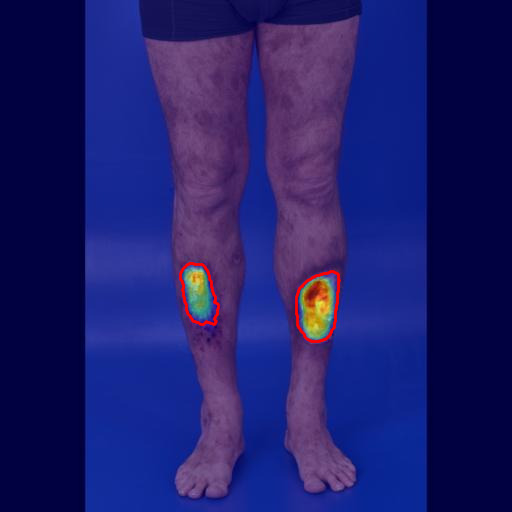}\\[4pt]
                \caption*{\scriptsize HarDNet}
            \end{subfigure}
            \hspace{-0.6cm}
            \begin{subfigure}{0.165\textwidth}
                \centering
                \includegraphics[trim={0 85 0 85}, clip, width=1.5cm, angle=180]{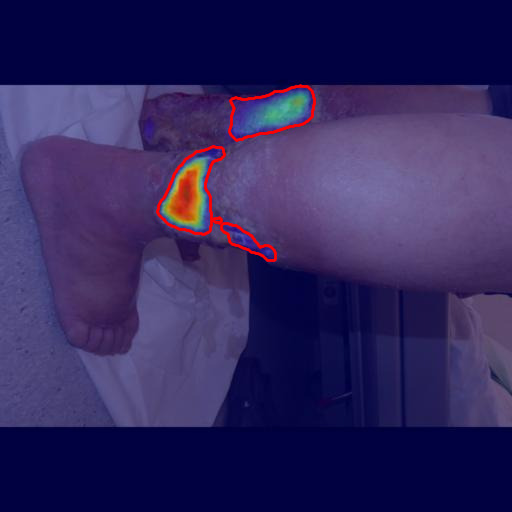}\\[4pt]
                \includegraphics[trim={40 90 40 90}, clip, width=1.5cm, angle=180]{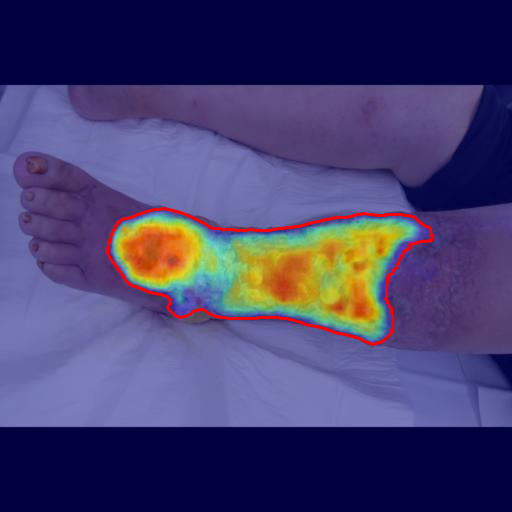}\\[4pt]
                \includegraphics[trim={50 120 90 120}, clip, width=1.5cm, angle=0]{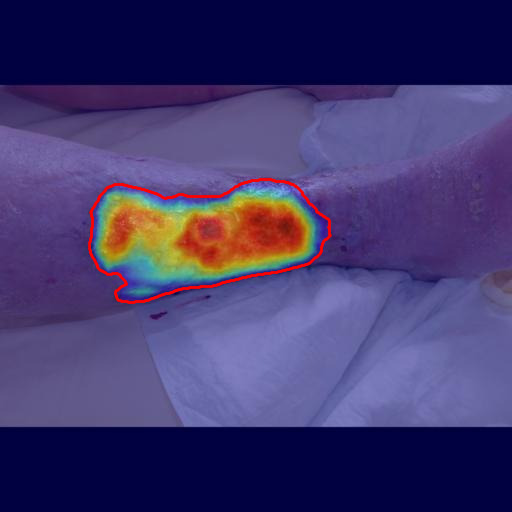}\\[4pt]
                \includegraphics[trim={85 0 85 0}, clip, height=1.5cm, angle=90]{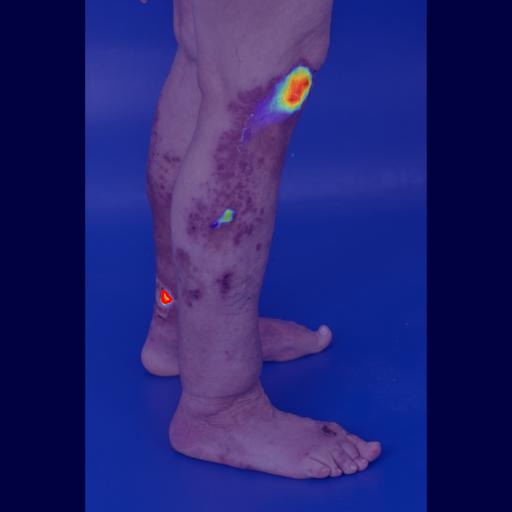}\\[4pt]
                \includegraphics[trim={85 0 85 50}, clip, height=1.5cm, angle=90]{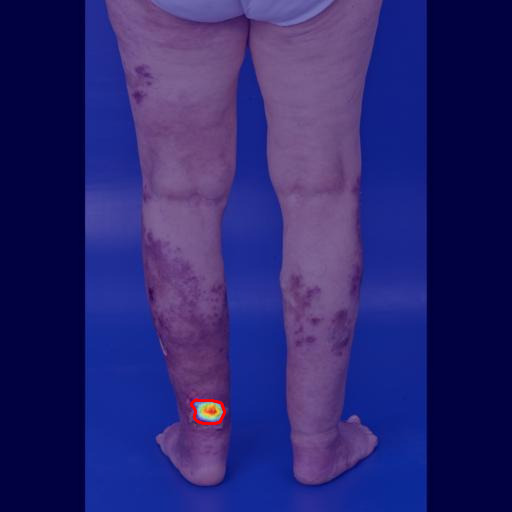}\\[4pt]
                \includegraphics[trim={140 90 130 150}, clip, height=1.5cm, angle=90]{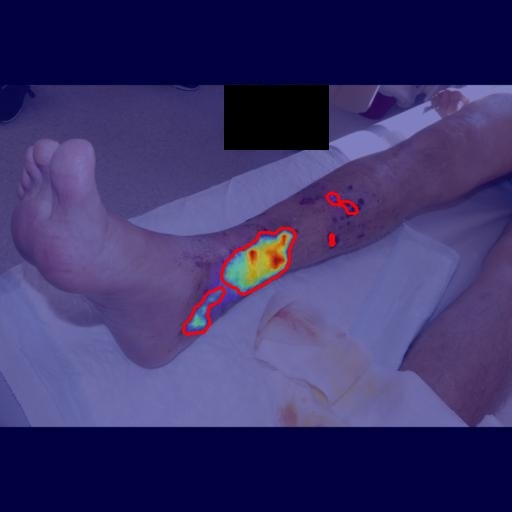}\\[4pt]
                \includegraphics[trim={100 90 100 90}, clip, height=1.5cm, angle=90]{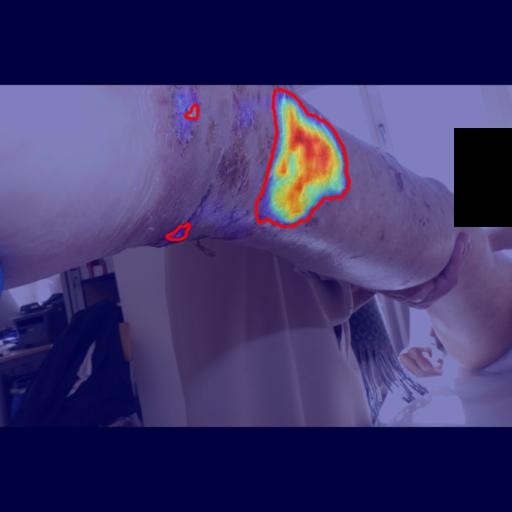}\\[4pt]
                \includegraphics[trim={110 90 100 90}, clip, height=1.5cm, angle=90]{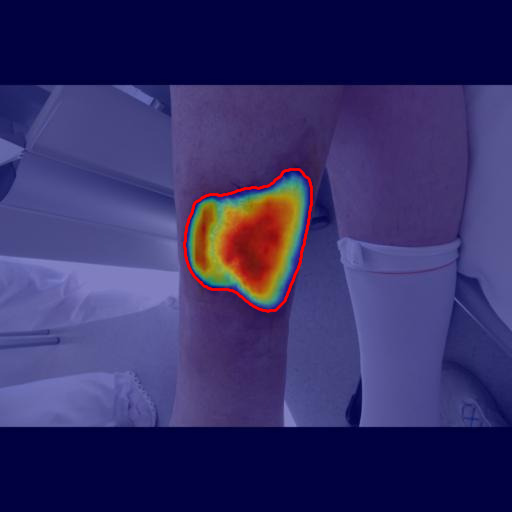}\\[4pt]
                \includegraphics[trim={130 90 145 90}, clip, height=1.5cm, angle=90]{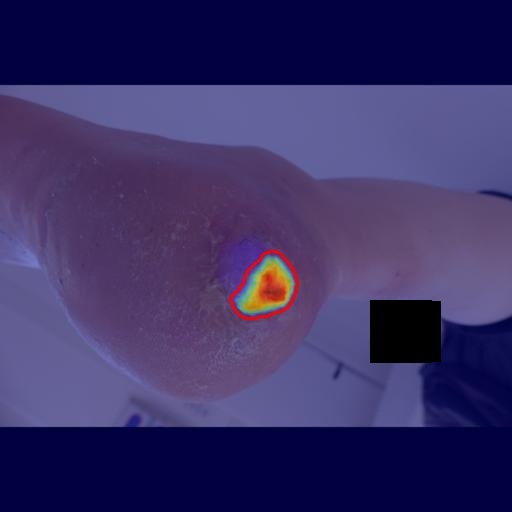}\\[4pt]
                \includegraphics[trim={80 90 120 90}, clip, height=1.5cm, angle=90]{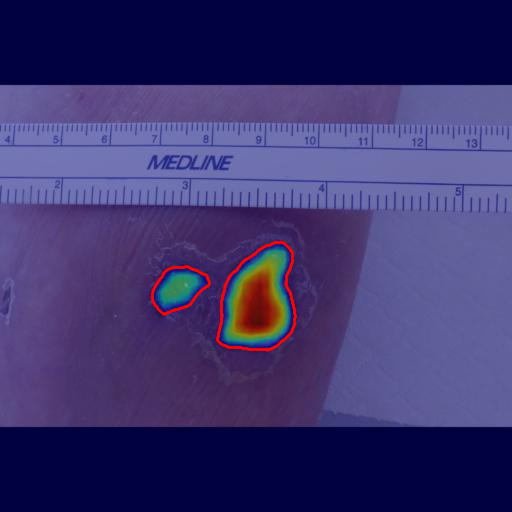}\\[4pt]
                \includegraphics[trim={110 0 110 140}, clip, height=1.5cm, angle=90]{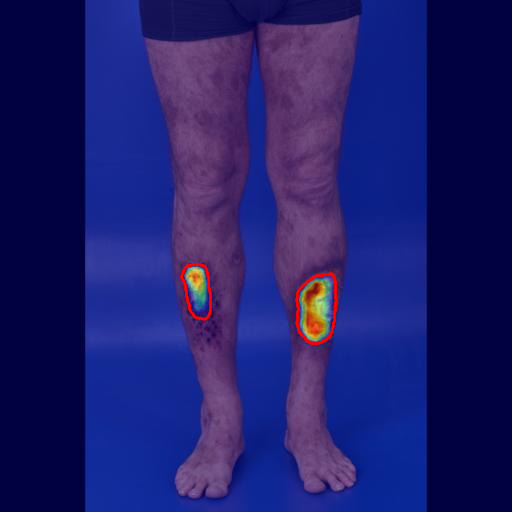}\\[4pt]
                \caption*{\scriptsize SegNeXt}
            \end{subfigure}
            \hspace{-0.6cm}
            \begin{subfigure}{0.165\textwidth}
                \centering
                \includegraphics[trim={0 85 0 85}, clip, width=1.5cm, angle=180]{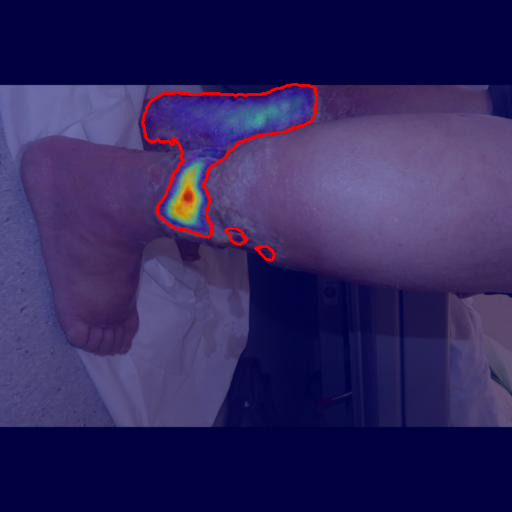}\\[4pt]
                \includegraphics[trim={40 90 40 90}, clip, width=1.5cm, angle=180]{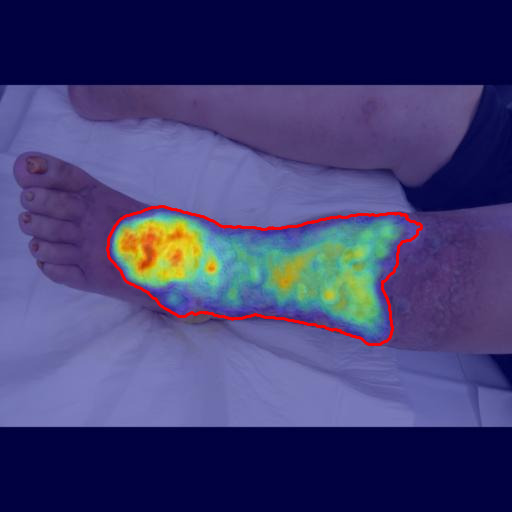}\\[4pt]
                \includegraphics[trim={50 120 90 120}, clip, width=1.5cm, angle=0]{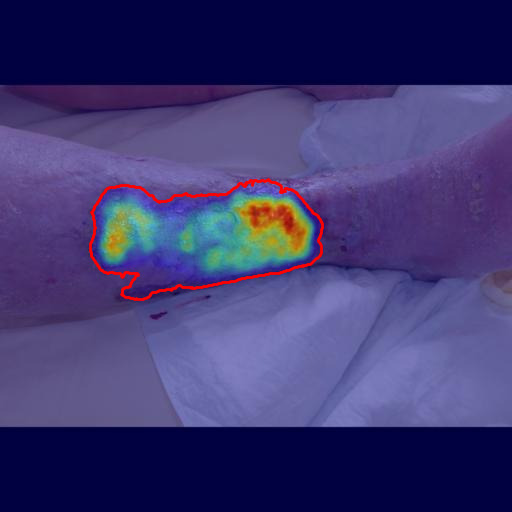}\\[4pt]
                \includegraphics[trim={85 0 85 0}, clip, height=1.5cm, angle=90]{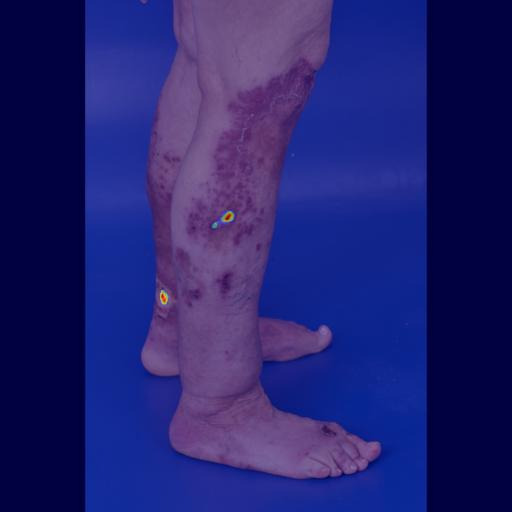}\\[4pt]
                \includegraphics[trim={85 0 85 50}, clip, height=1.5cm, angle=90]{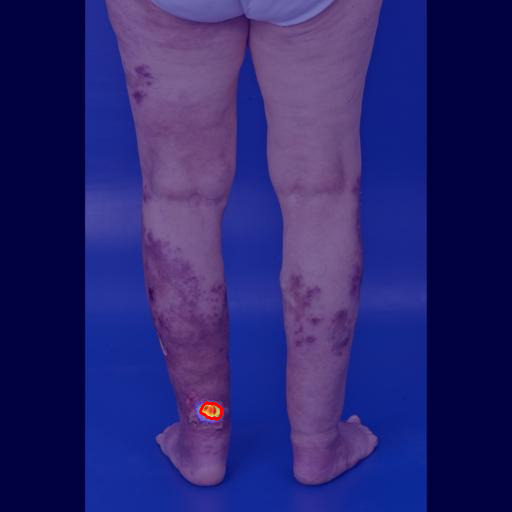}\\[4pt]
                \includegraphics[trim={140 90 130 150}, clip, height=1.5cm, angle=90]{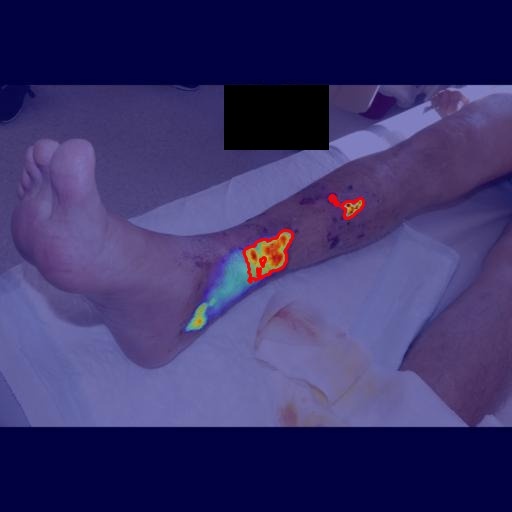}\\[4pt]
                \includegraphics[trim={100 90 100 90}, clip, height=1.5cm, angle=90]{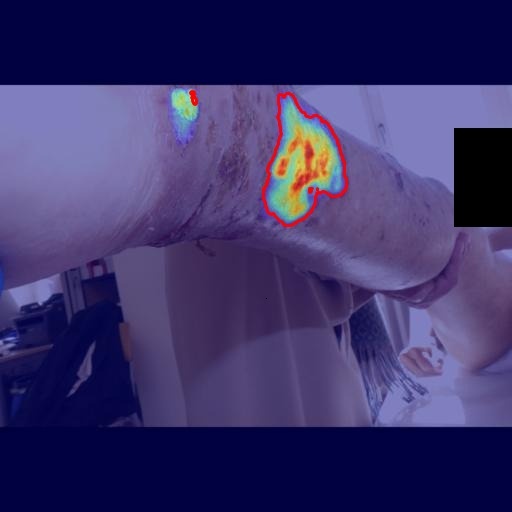}\\[4pt]
                \includegraphics[trim={110 90 100 90}, clip, height=1.5cm, angle=90]{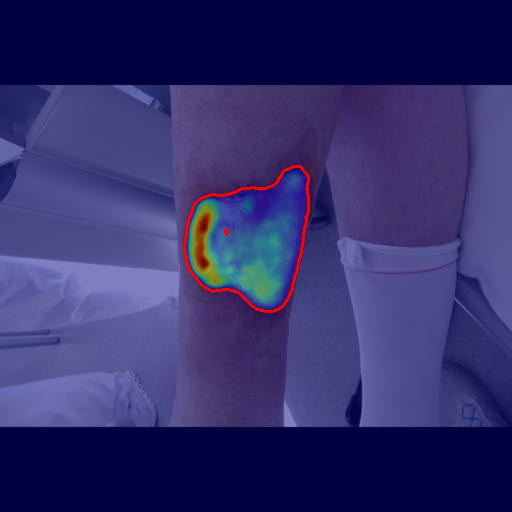}\\[4pt]
                \includegraphics[trim={130 90 145 90}, clip, height=1.5cm, angle=90]{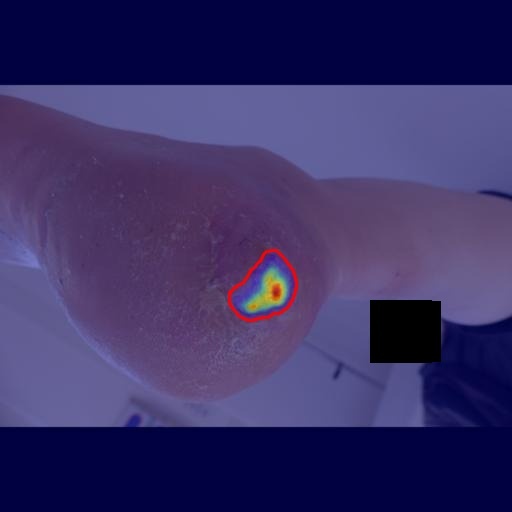}\\[4pt]
                \includegraphics[trim={80 90 120 90}, clip, height=1.5cm, angle=90]{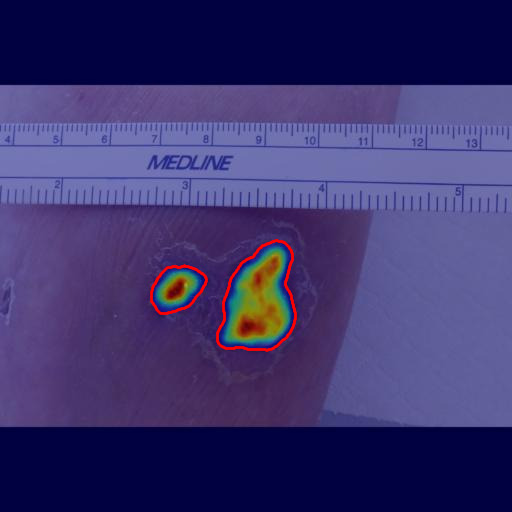}\\[4pt]
                \includegraphics[trim={110 0 110 140}, clip, height=1.5cm, angle=90]{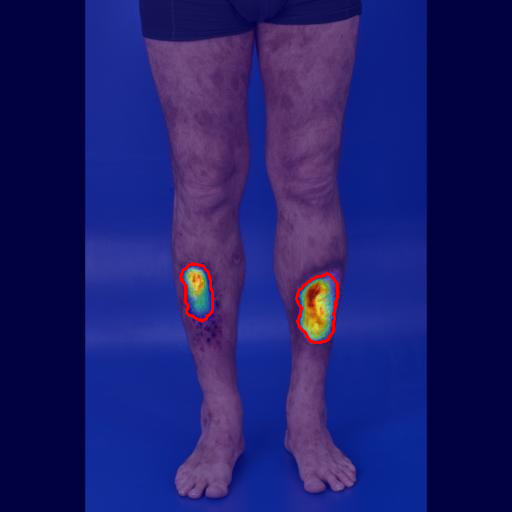}\\[4pt]
                \caption*{\scriptsize FuSegN.}
            \end{subfigure}
            \hspace{-0.6cm}
            \begin{subfigure}{0.165\textwidth}
                \centering
                \includegraphics[trim={0 85 0 85}, clip, width=1.5cm, angle=180]{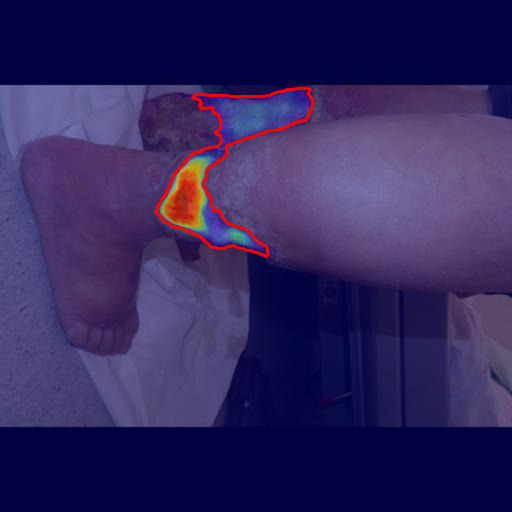}\\[4pt]
                \includegraphics[trim={40 90 40 90}, clip, width=1.5cm, angle=180]{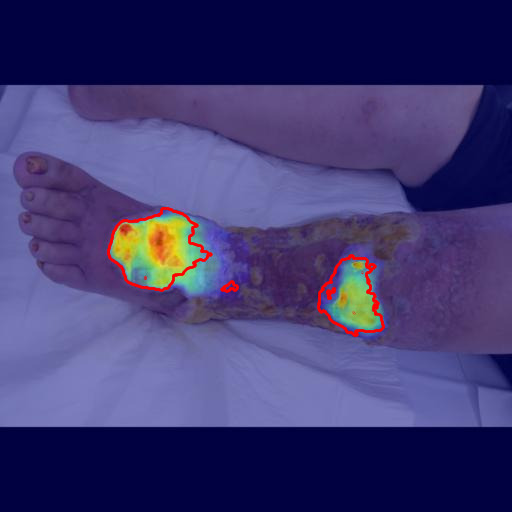}\\[4pt]
                \includegraphics[trim={50 120 90 120}, clip, width=1.5cm, angle=0]{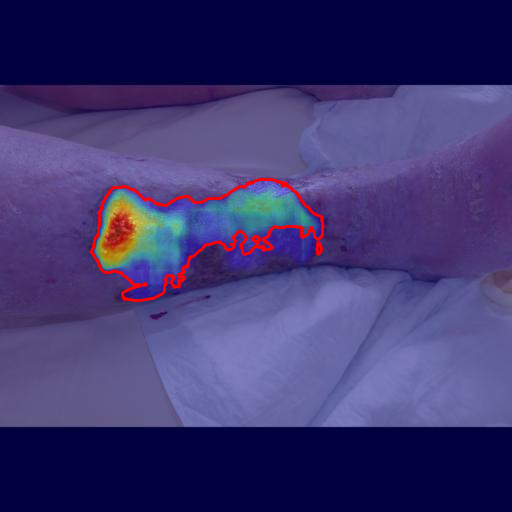}\\[4pt]
                \includegraphics[trim={85 0 85 0}, clip, height=1.5cm, angle=90]{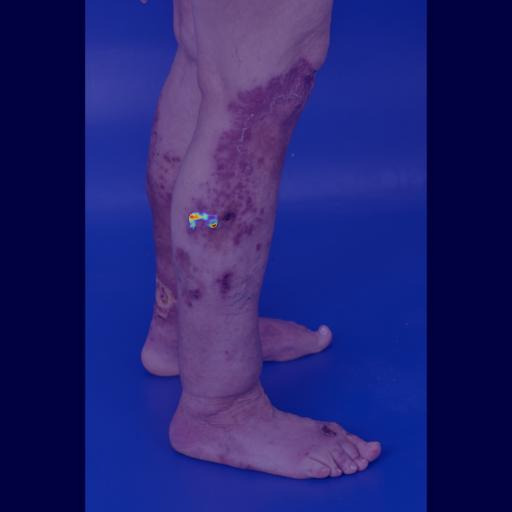}\\[4pt]
                \includegraphics[trim={85 0 85 50}, clip, height=1.5cm, angle=90]{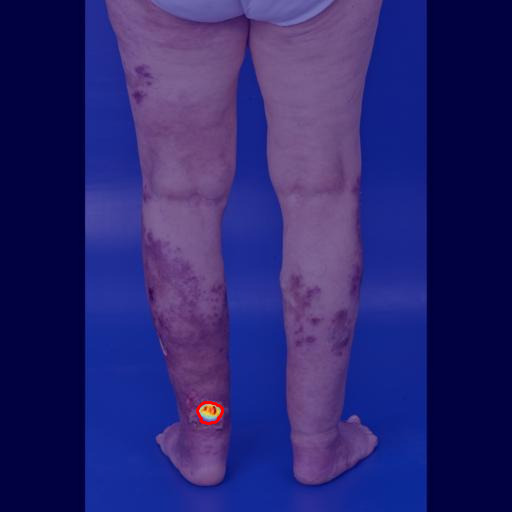}\\[4pt]
                \includegraphics[trim={140 90 130 150}, clip, height=1.5cm, angle=90]{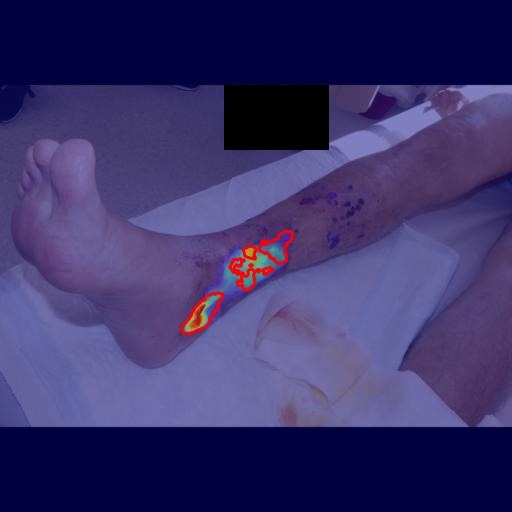}\\[4pt]
                \includegraphics[trim={100 90 100 90}, clip, height=1.5cm, angle=90]{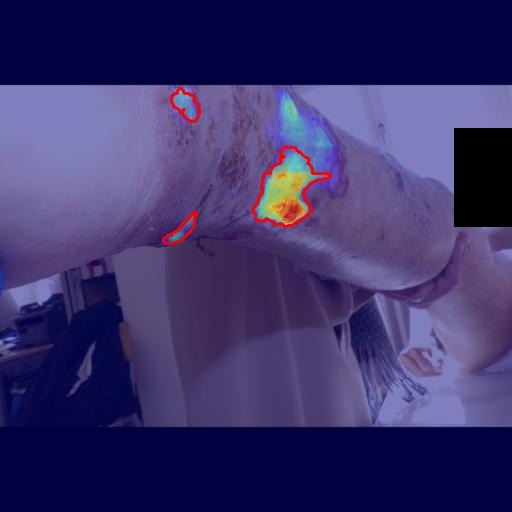}\\[4pt]
                \includegraphics[trim={110 90 100 90}, clip, height=1.5cm, angle=90]{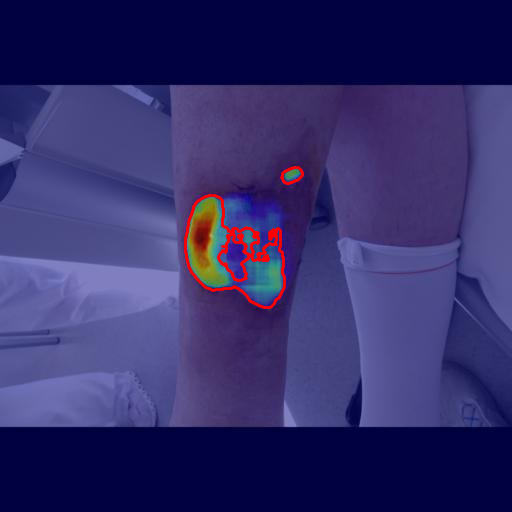}\\[4pt]
                \includegraphics[trim={130 90 145 90}, clip, height=1.5cm, angle=90]{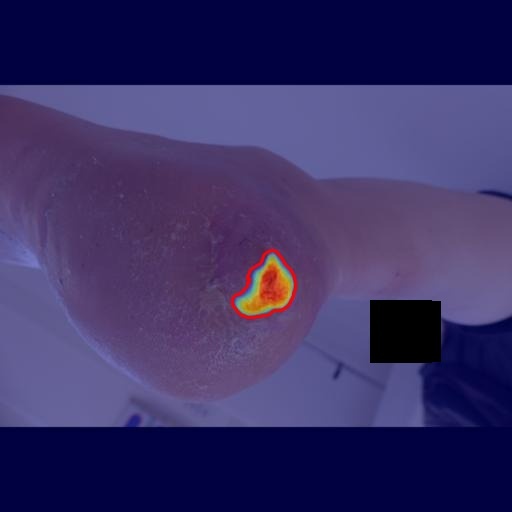}\\[4pt]
                \includegraphics[trim={80 90 120 90}, clip, height=1.5cm, angle=90]{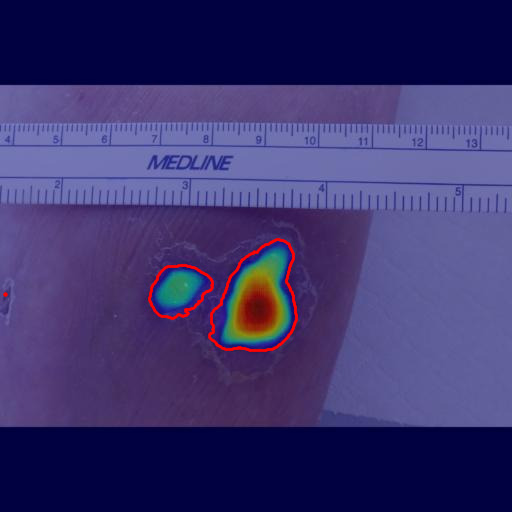}\\[4pt]
                \includegraphics[trim={110 0 110 140}, clip, height=1.5cm, angle=90]{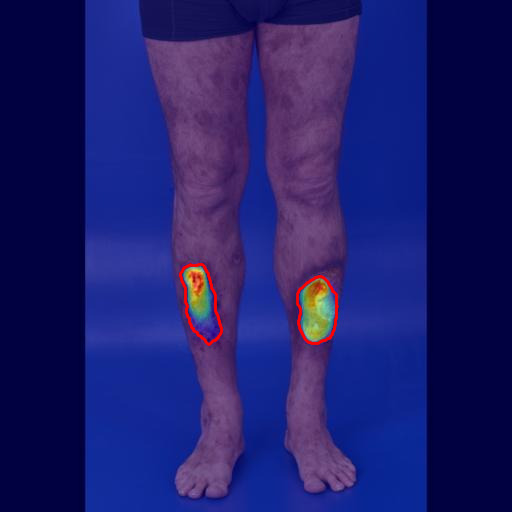}\\[4pt]
                \caption*{\scriptsize U-Net}
            \end{subfigure}
            \hspace{-0.6cm}
            \begin{subfigure}{0.165\textwidth}
                \centering
                \includegraphics[trim={0 85 0 85}, clip, width=1.5cm, angle=180]{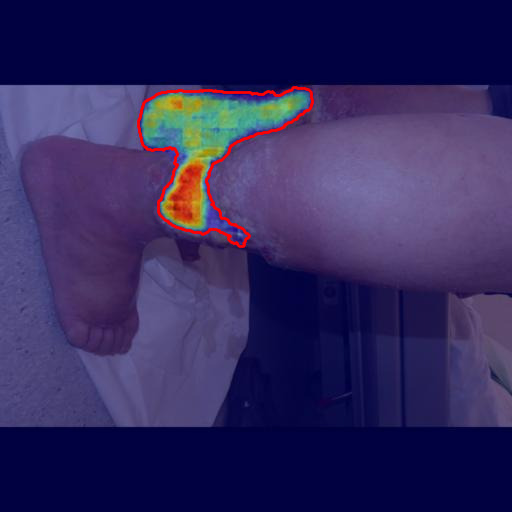}\\[4pt]
                \includegraphics[trim={40 90 40 90}, clip, width=1.5cm, angle=180]{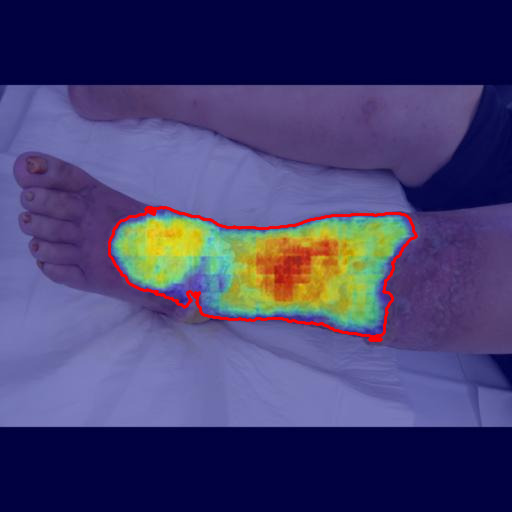}\\[4pt]
                \includegraphics[trim={50 120 90 120}, clip, width=1.5cm, angle=0]{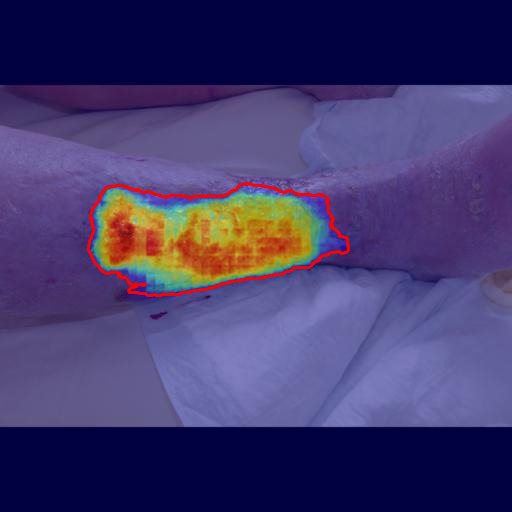}\\[4pt]
                \includegraphics[trim={85 0 85 0}, clip, height=1.5cm, angle=90]{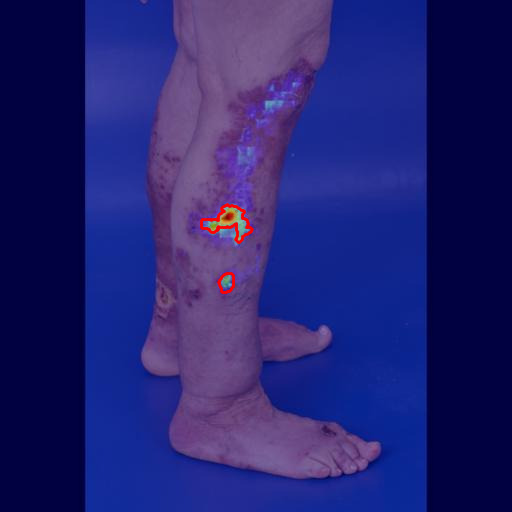}\\[4pt]
                \includegraphics[trim={85 0 85 50}, clip, height=1.5cm, angle=90]{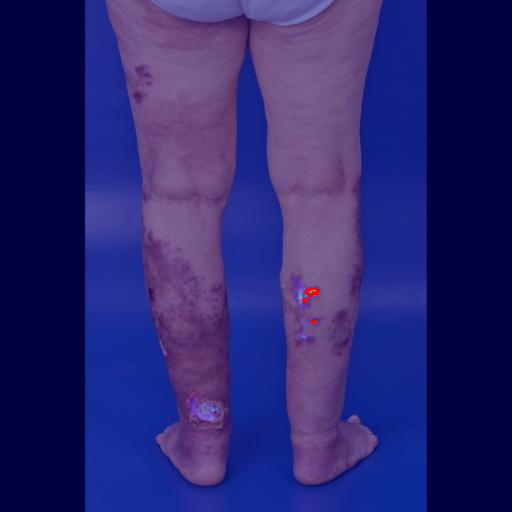}\\[4pt]
                \includegraphics[trim={140 90 130 150}, clip, height=1.5cm, angle=90]{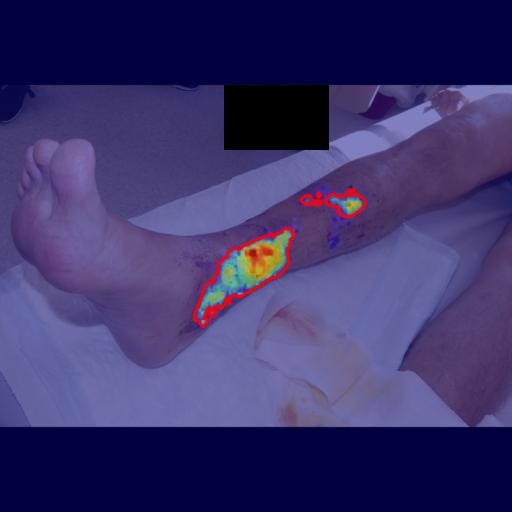}\\[4pt]
                \includegraphics[trim={100 90 100 90}, clip, height=1.5cm, angle=90]{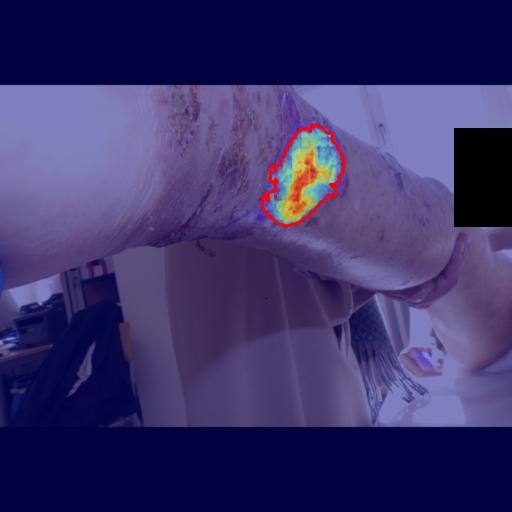}\\[4pt]
                \includegraphics[trim={110 90 100 90}, clip, height=1.5cm, angle=90]{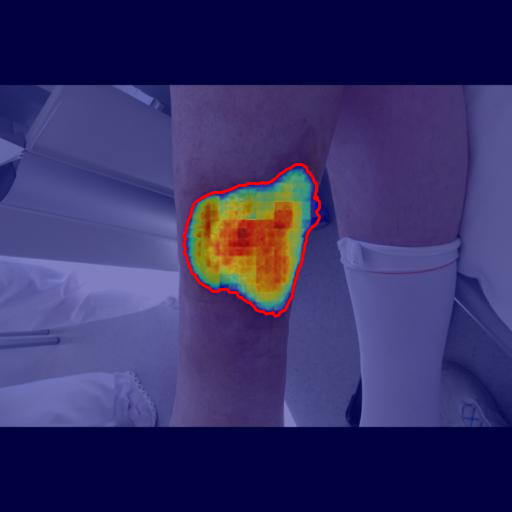}\\[4pt]
                \includegraphics[trim={130 90 145 90}, clip, height=1.5cm, angle=90]{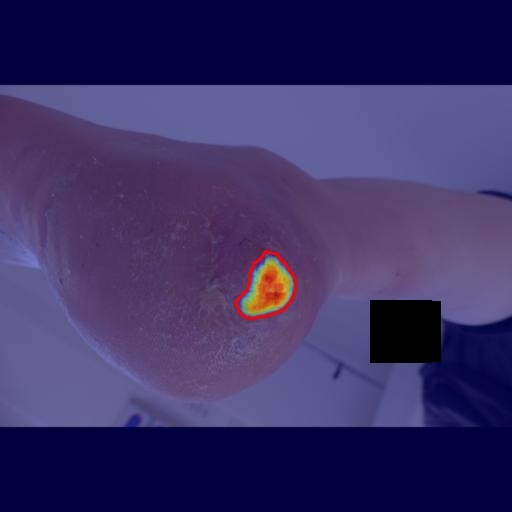}\\[4pt]
                \includegraphics[trim={80 90 120 90}, clip, height=1.5cm, angle=90]{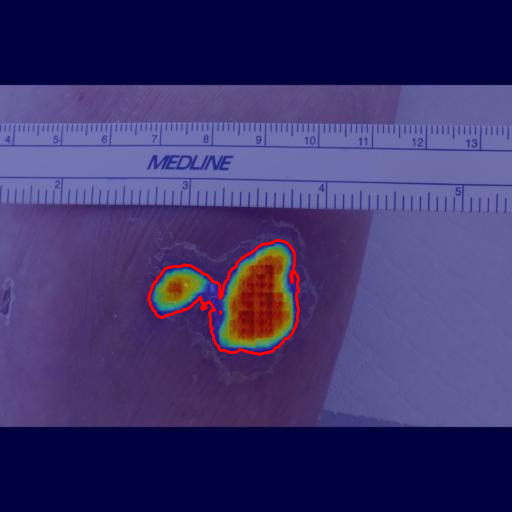}\\[4pt]
                \includegraphics[trim={110 0 110 140}, clip, height=1.5cm, angle=90]{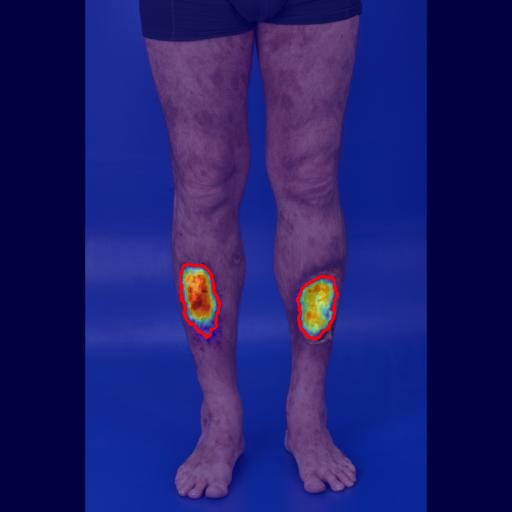}\\[4pt]
                \caption*{\scriptsize MISSForm.}
            \end{subfigure}
            \hspace{-0.6cm}
            \begin{subfigure}{0.165\textwidth}
                \centering
                \includegraphics[trim={0 85 0 85}, clip, width=1.5cm, angle=180]{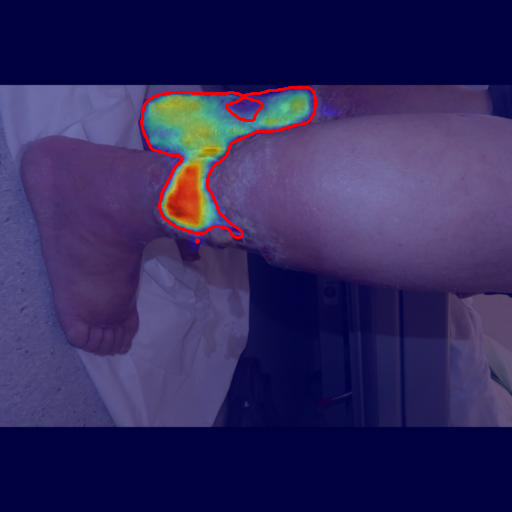}\\[4pt]
                \includegraphics[trim={40 90 40 90}, clip, width=1.5cm, angle=180]{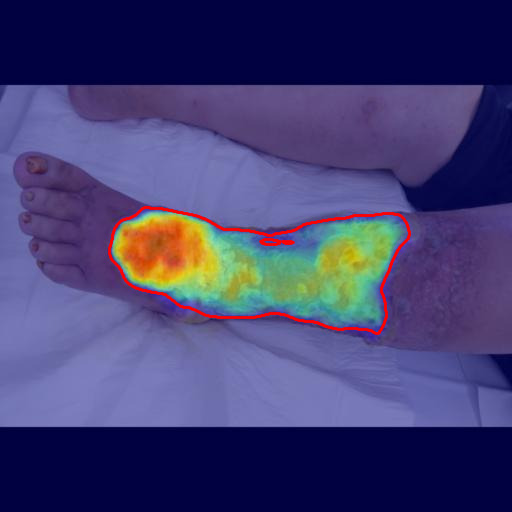}\\[4pt]
                \includegraphics[trim={50 120 90 120}, clip, width=1.5cm, angle=0]{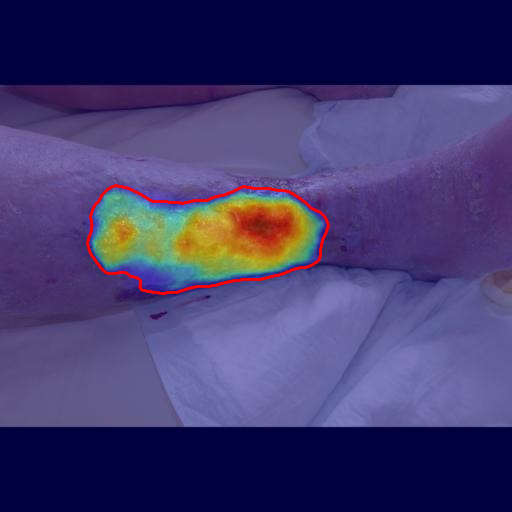}\\[4pt]
                \includegraphics[trim={85 0 85 0}, clip, height=1.5cm, angle=90]{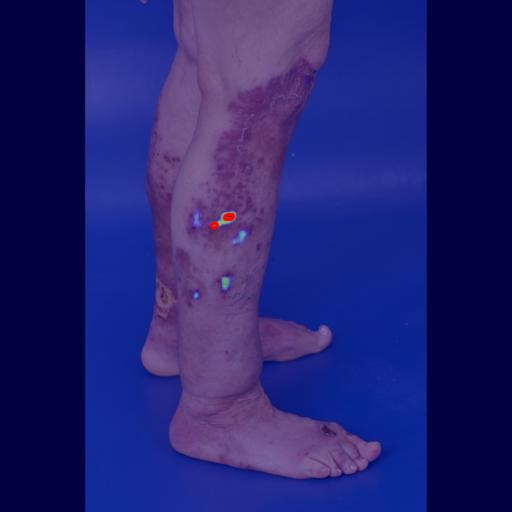}\\[4pt]
                \includegraphics[trim={85 0 85 50}, clip, height=1.5cm, angle=90]{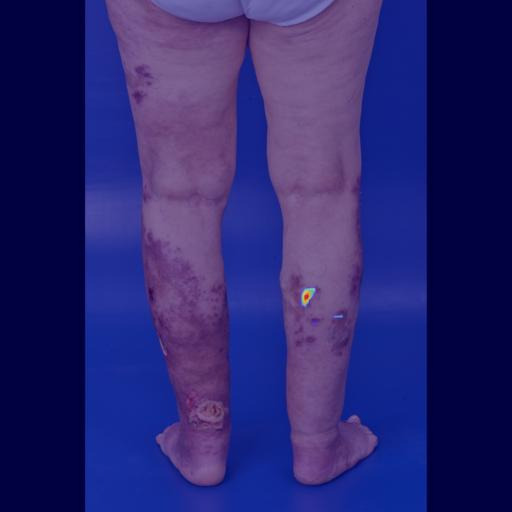}\\[4pt]
                \includegraphics[trim={140 90 130 150}, clip, height=1.5cm, angle=90]{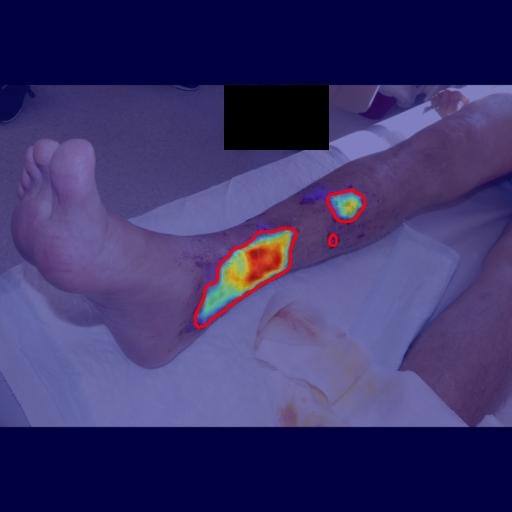}\\[4pt]
                \includegraphics[trim={100 90 100 90}, clip, height=1.5cm, angle=90]{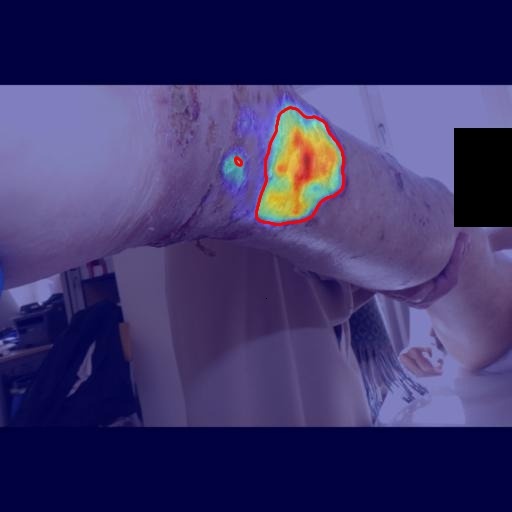}\\[4pt]
                \includegraphics[trim={110 90 100 90}, clip, height=1.5cm, angle=90]{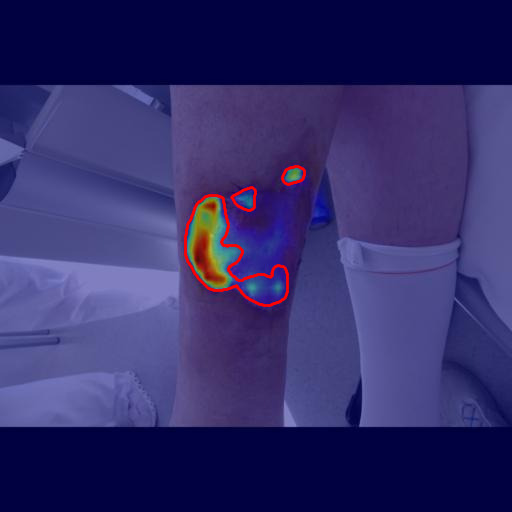}\\[4pt]
                \includegraphics[trim={130 90 145 90}, clip, height=1.5cm, angle=90]{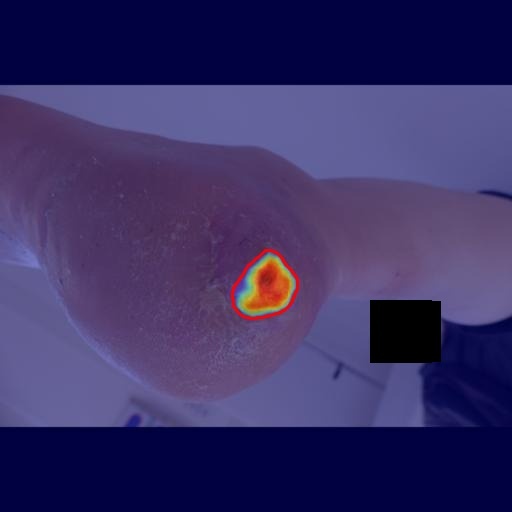}\\[4pt]
                \includegraphics[trim={80 90 120 90}, clip, height=1.5cm, angle=90]{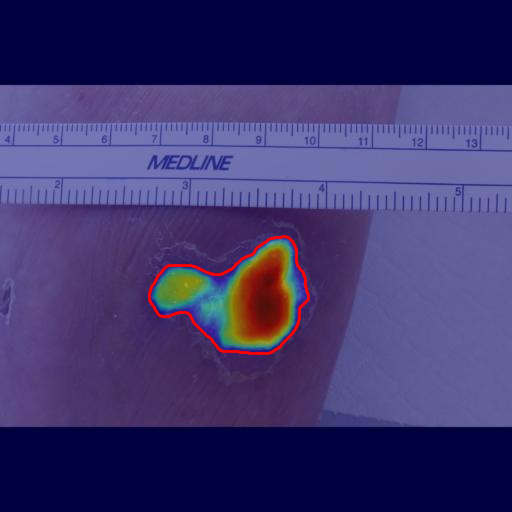}\\[4pt]
                \includegraphics[trim={110 0 110 140}, clip, height=1.5cm, angle=90]{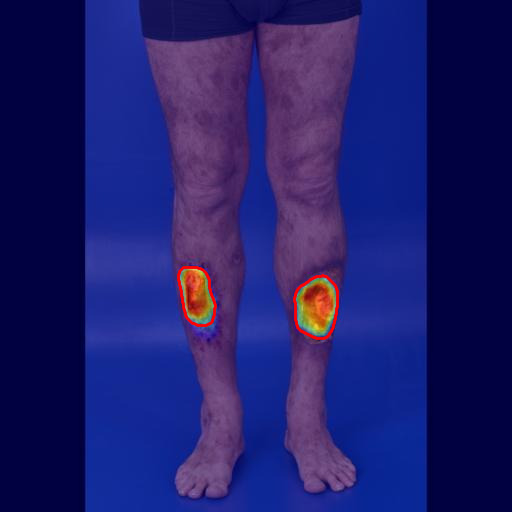}\\[4pt]
                \caption*{\scriptsize HiFormer}
            \end{subfigure}
        \end{adjustbox}
    \end{turn}
    \caption{Grad-CAM for further OOD images, alongside their GT annotation.}
    \label{fig:eval:grad_cam:all_models:further_examples}
\end{figure}
\clearpage
\section{Real-World Deployment}

\subsection{Visualization and Results of Size Retrieval}

\setlength{\tabcolsep}{4pt}
\setlength\LTleft{-1.8cm}
\enlargethispage{2cm} 
\begin{longtable}{cccccccm{1.7cm}}
\captionsetup{justification=raggedright, width=1.3\textwidth}
\caption{Comparison of the mask, size (in mm), and area (in $\text{cm}^2$) predictions for different wounds. If more than one wound is detected, we report the dimension and area of the largest wound in the image.} 
\label{tab:appendix:all_size_retrievals} \\

\toprule
\textbf{ID} & \textbf{\acs{vwconv}} & \textbf{\ac{internimage}} & \textbf{\ac{vwmit}} & \textbf{\ac{segformer}} & \textbf{\ac{transnext}} & \textbf{Ensemble} & \textbf{Experts} \\
\midrule
\endfirsthead

\multicolumn{7}{c}{\textbf{Continued from previous page}} \\
\toprule
\textbf{ID} & \textbf{\ac{vwconv}} & \textbf{\ac{internimage}} & \textbf{\ac{vwmit}} & \textbf{\ac{segformer}} & \textbf{\ac{transnext}} & \textbf{Ensemble} & \textbf{Experts} \\
\midrule
\endhead

\bottomrule
\endlastfoot


\vspace{0.3cm}
1&
\makecell{\includegraphics[width=2cm]{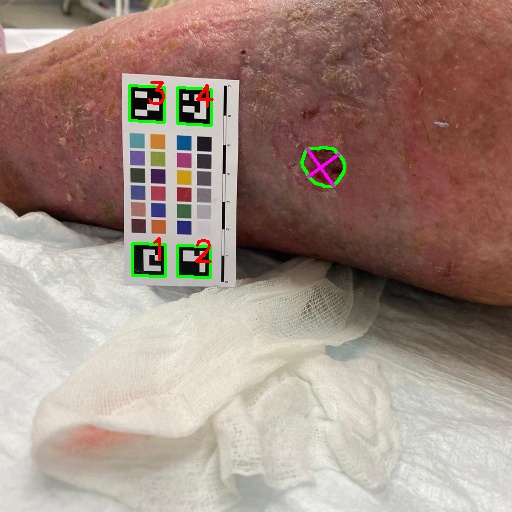}    \\16 x 13\\1.6$ \text{cm}^2$} &
\makecell{\includegraphics[width=2cm]{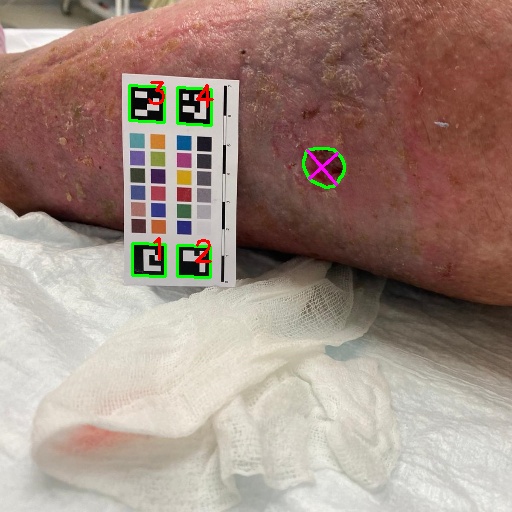}          \\16 x 13\\1.5$ \text{cm}^2$} &
\makecell{\includegraphics[width=2cm]{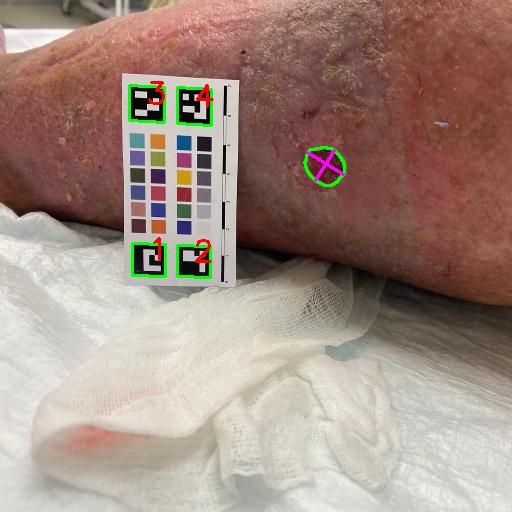}        \\15 x 13\\1.4$ \text{cm}^2$} &
\makecell{\includegraphics[width=2cm]{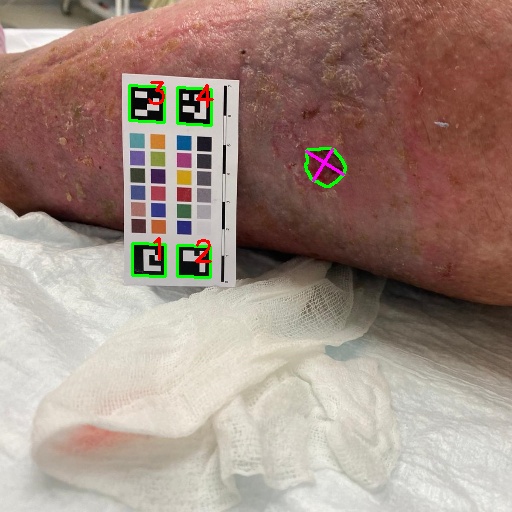}            \\15 x 13\\1.4$ \text{cm}^2$} &
\makecell{\includegraphics[width=2cm]{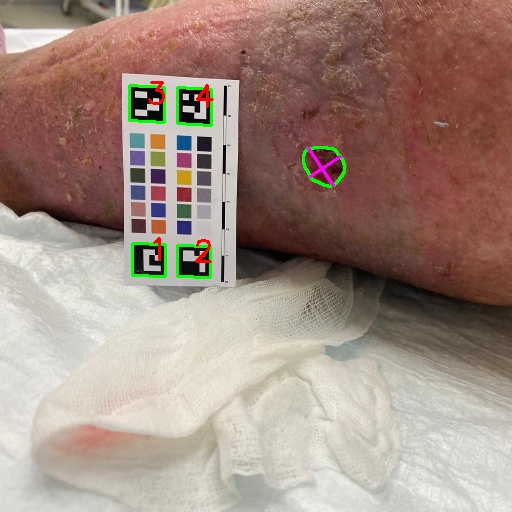}            \\16 x 13\\1.5$ \text{cm}^2$} &
\makecell{\includegraphics[width=2cm]{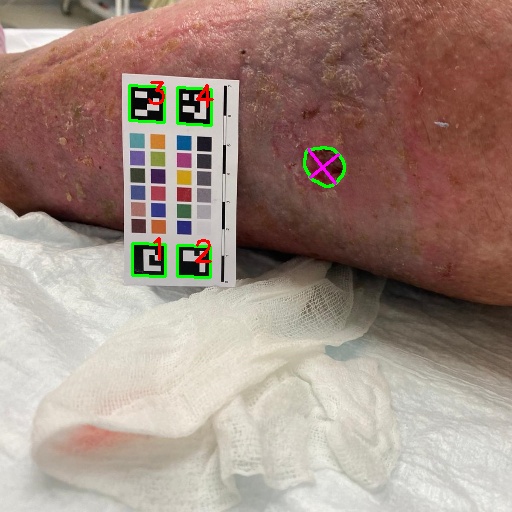}             \\15 x 13\\1.5$ \text{cm}^2$} &
\begin{minipage}[c]{1.7cm}\centering \vspace{-0.5cm}\small 15 x 15\\9 x 8\\13 x 11\\\rule{1.4cm}{0.4pt}\\12.3 x 11.3\strut \end{minipage} \\
\vspace{0.3cm}

2&
\makecell{\includegraphics[width=2cm]{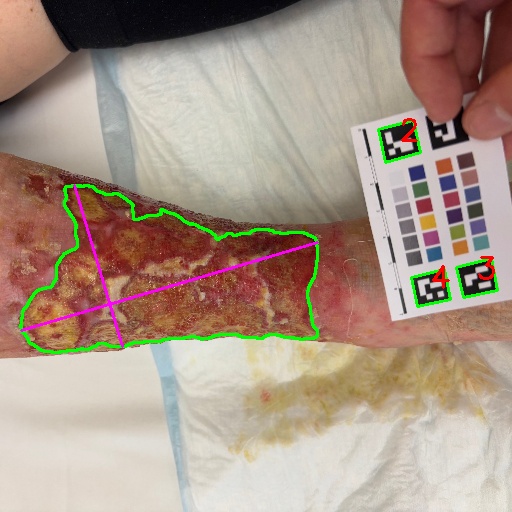}    \\110 x 60\\41.7$ \text{cm}^2$} &
\makecell{\includegraphics[width=2cm]{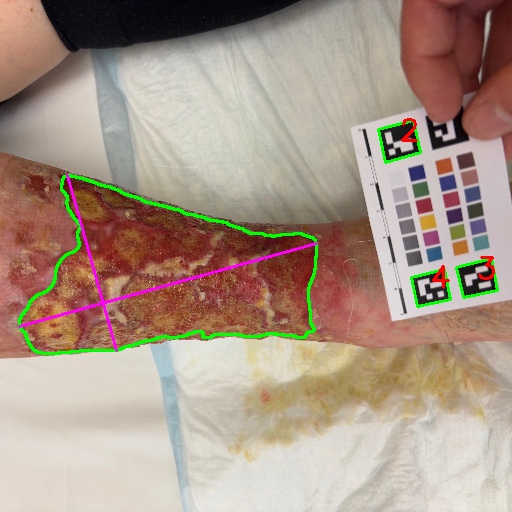}          \\110 x 65\\43.2$ \text{cm}^2$} &
\makecell{\includegraphics[width=2cm]{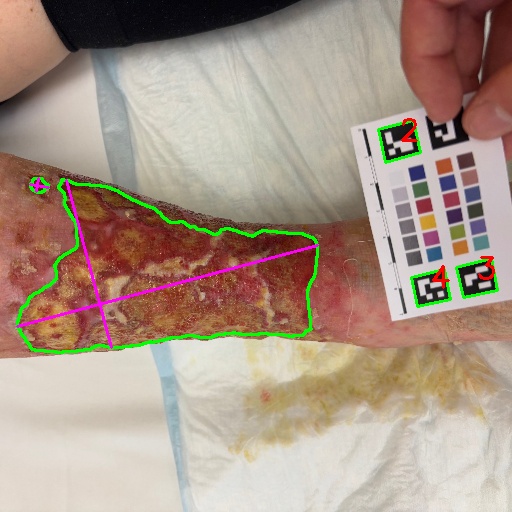}        \\111 x 62\\42.3$ \text{cm}^2$} &
\makecell{\includegraphics[width=2cm]{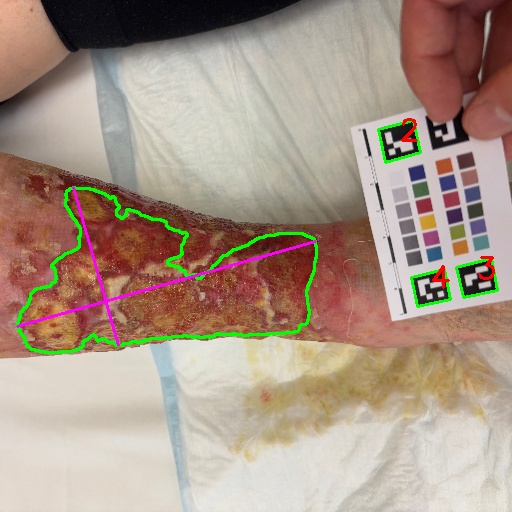}            \\110 x 59\\36.7$ \text{cm}^2$} &
\makecell{\includegraphics[width=2cm]{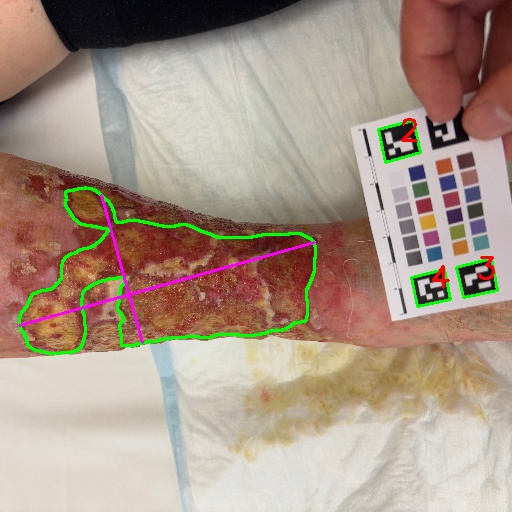}            \\109 x 54\\34.8$ \text{cm}^2$} &
\makecell{\includegraphics[width=2cm]{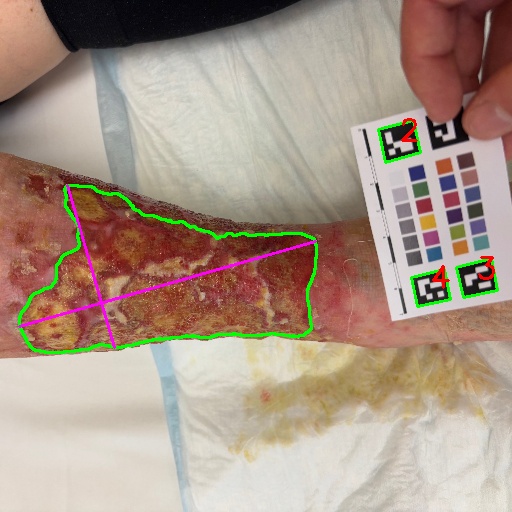}             \\110 x 60\\40.8$ \text{cm}^2$} &
\begin{minipage}[c]{1.7cm}\centering \vspace{-0.5cm}\small 90 x 70\\140 x 120\\75 x 50\\\rule{1.4cm}{0.4pt}\\101.7 x 80.0\strut \end{minipage} \\
\vspace{0.3cm}

3&
\makecell{\includegraphics[width=2cm]{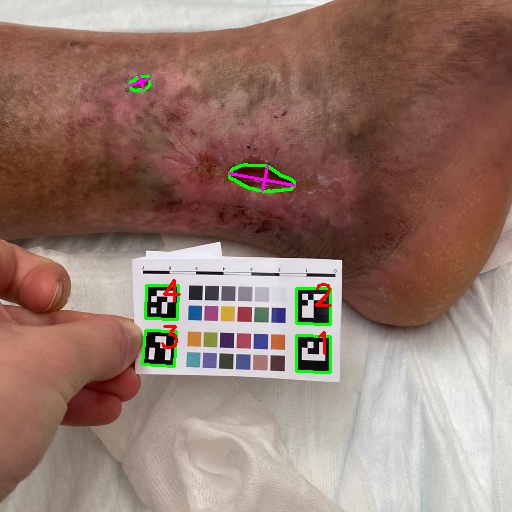}    \\23 x 9 \\1.5$ \text{cm}^2$\\} &
\makecell{\includegraphics[width=2cm]{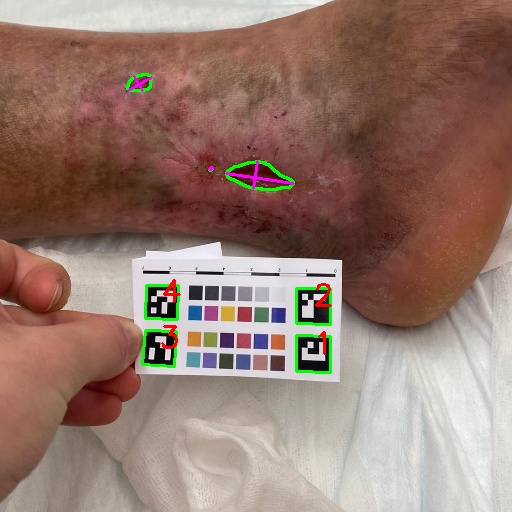}          \\24 x 10\\1.5$ \text{cm}^2$\\} &
\makecell{\includegraphics[width=2cm]{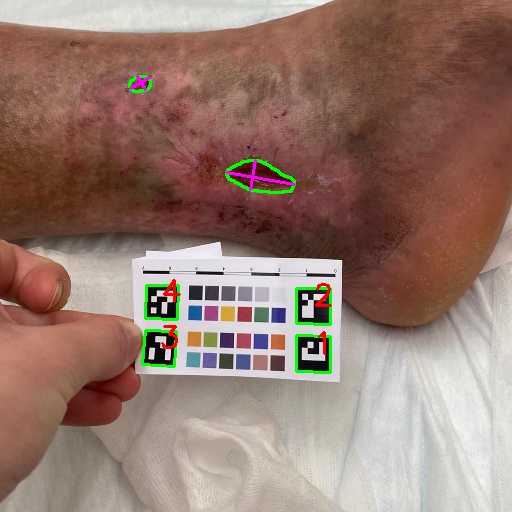}        \\24 x 10\\1.8$ \text{cm}^2$\\} &
\makecell{\includegraphics[width=2cm]{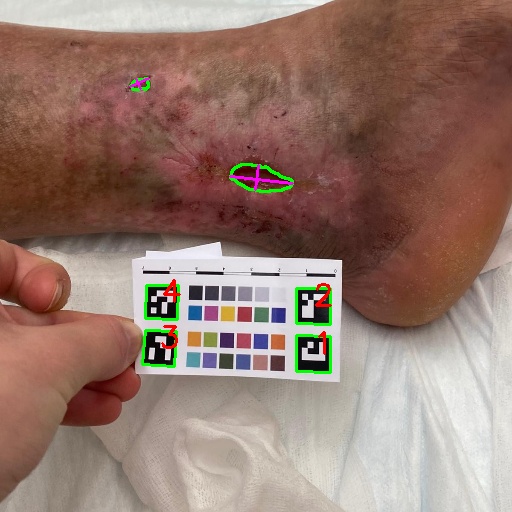}            \\22 x 9 \\1.4$ \text{cm}^2$\\} &
\makecell{\includegraphics[width=2cm]{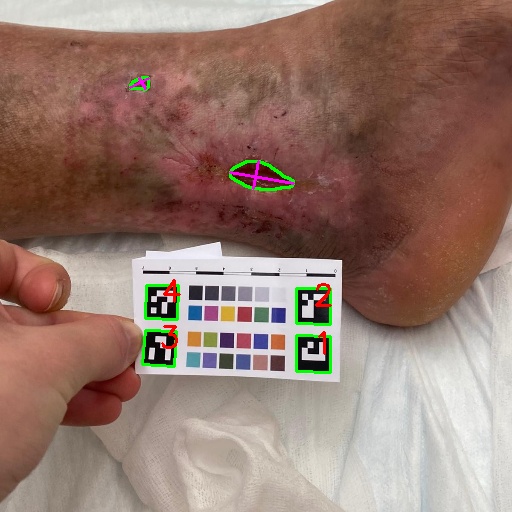}            \\23 x 10\\1.5$ \text{cm}^2$\\} &
\makecell{\includegraphics[width=2cm]{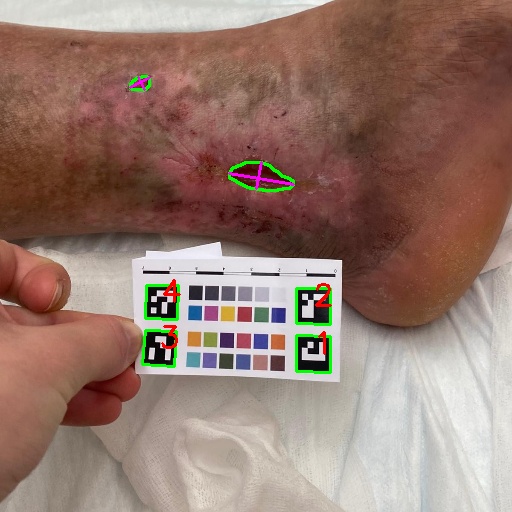}             \\23 x 9 \\1.5$ \text{cm}^2$\\} &
\begin{minipage}[c]{1.7cm}\centering \vspace{-0.5cm}\small 30 x 20\\20 x 7\\30 x 12\\\rule{1.4cm}{0.4pt}\\26.7 x 13.0\strut \end{minipage} \\
\vspace{0.3cm}

4&
\makecell{\includegraphics[width=2cm]{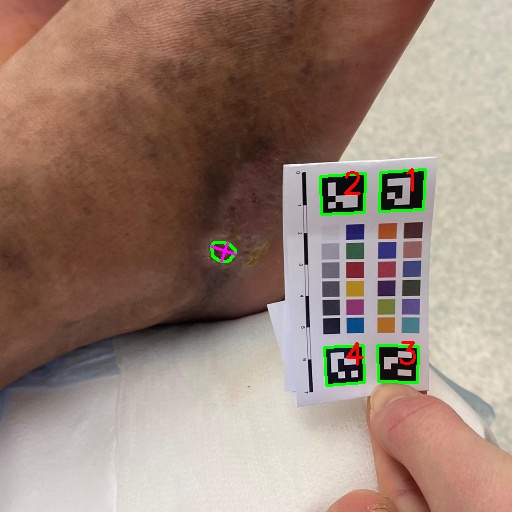}    \\8 x 6 \\0.3 $\text{cm}^2$\\} &
\makecell{\includegraphics[width=2cm]{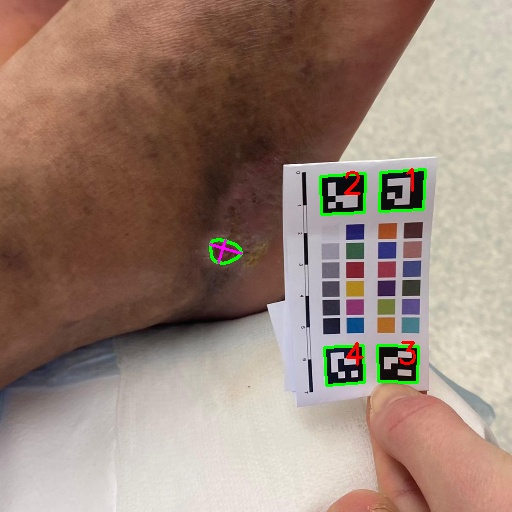}          \\10 x 7\\0.5 $\text{cm}^2$\\} &
\makecell{\includegraphics[width=2cm]{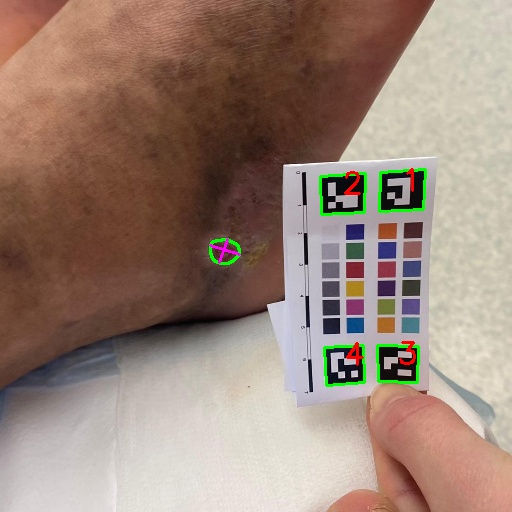}        \\9 x 7 \\0.5 $\text{cm}^2$\\} &
\makecell{\includegraphics[width=2cm]{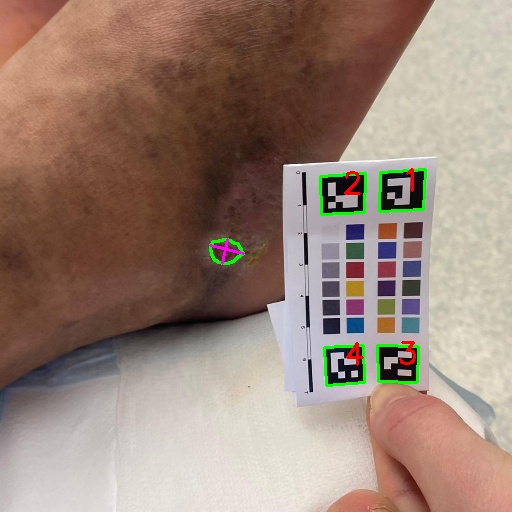}            \\10 x 7\\0.5 $\text{cm}^2$\\} &
\makecell{\includegraphics[width=2cm]{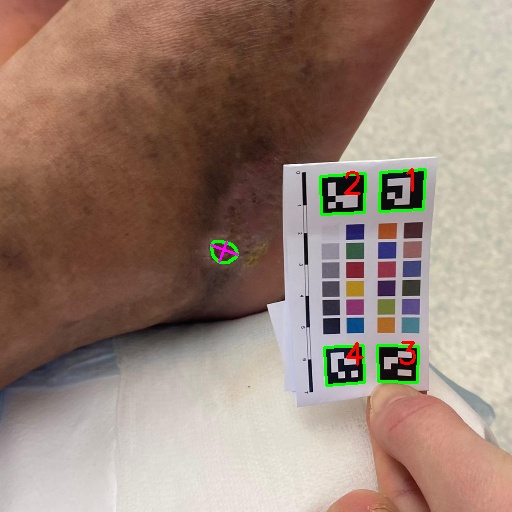}            \\8 x 6 \\0.4 $\text{cm}^2$\\} &
\makecell{\includegraphics[width=2cm]{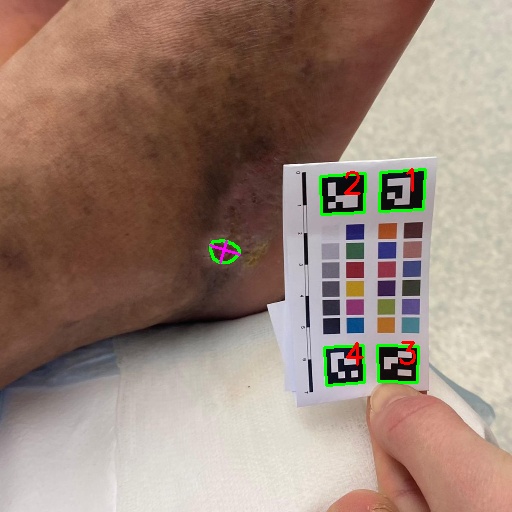}             \\9 x 7 \\0.4 $\text{cm}^2$\\} &
\begin{minipage}[c]{1.7cm}\centering \vspace{-0.5cm}\small 10 x 10\\45 x 25\\8 x 7\\\rule{1.4cm}{0.4pt}\\21.0 x 14.0\strut \end{minipage} \\
\vspace{0.3cm}

5&
\makecell{\includegraphics[width=2cm]{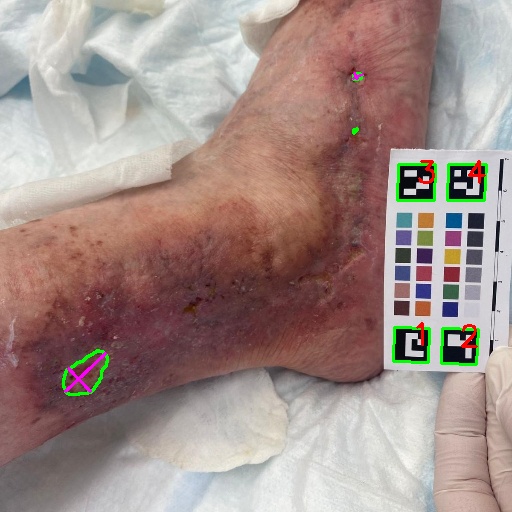}    \\18 x 11\\1.4 $\text{cm}^2$\\} &
\makecell{\includegraphics[width=2cm]{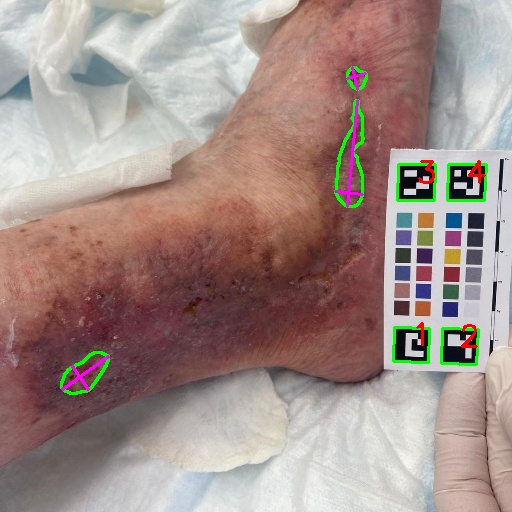}          \\35 x 9 \\1.9 $\text{cm}^2$\\} &
\makecell{\includegraphics[width=2cm]{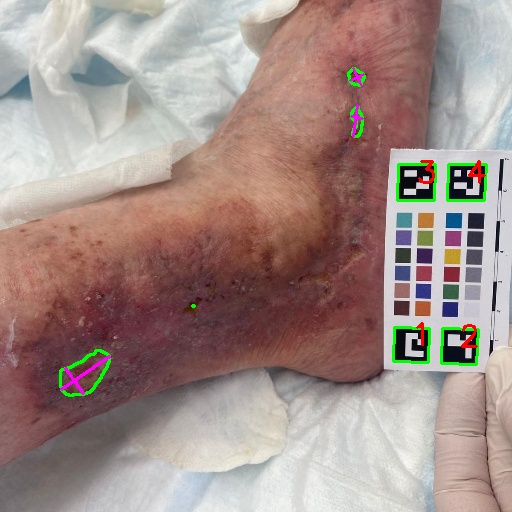}        \\19 x 10\\1.5 $\text{cm}^2$\\} &
\makecell{\includegraphics[width=2cm]{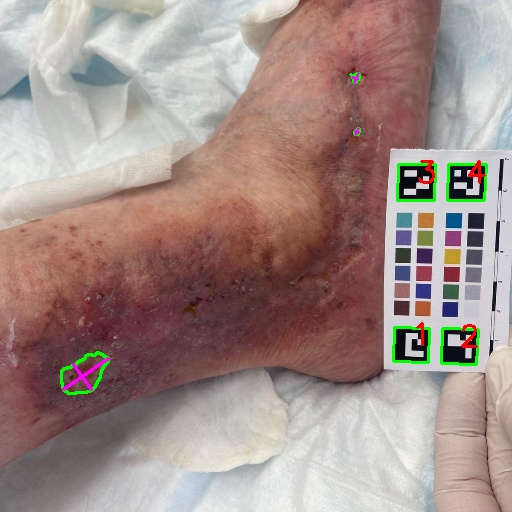}            \\18 x 10\\1.4 $\text{cm}^2$\\} &
\makecell{\includegraphics[width=2cm]{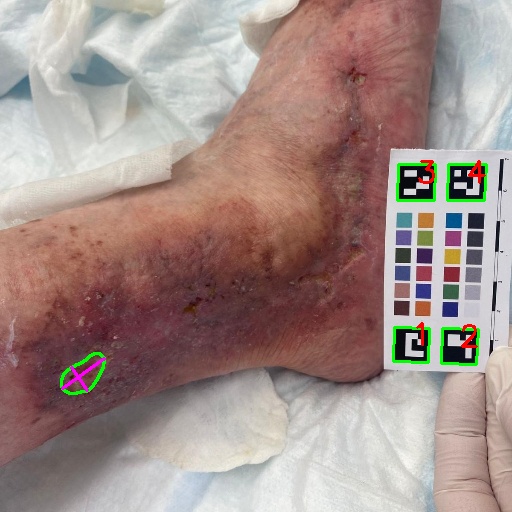}            \\17 x 9 \\1.2 $\text{cm}^2$\\} &
\makecell{\includegraphics[width=2cm]{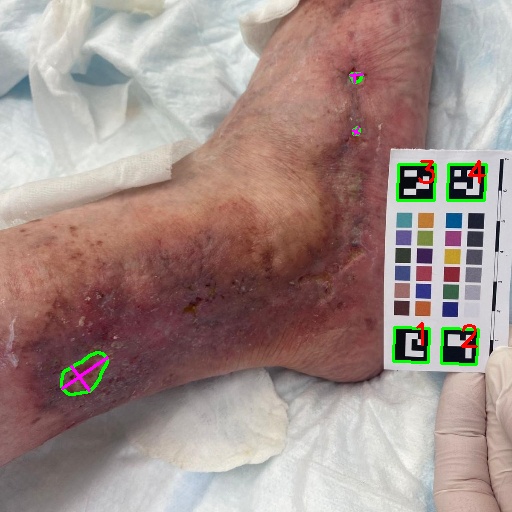}             \\18 x 10\\1.3 $\text{cm}^2$\\} &
\begin{minipage}[c]{1.7cm}\centering \vspace{-0.5cm}\small 50 x 40\\20 x 10\\30 x 15\\\rule{1.4cm}{0.4pt}\\33.3 x 21.7\strut \end{minipage} \\
\vspace{0.3cm}

6&
\makecell{\includegraphics[width=2cm]{images/SizeRetrieval/Bild_06_VWFormerConvNeXtS_mask_annotated.jpg}    \\54 x 40\\12.2 $\text{cm}^2$ \\} &
\makecell{\includegraphics[width=2cm]{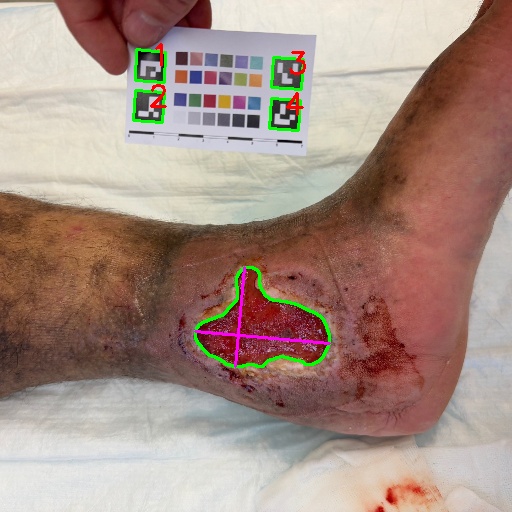}          \\54 x 40\\12.1 $\text{cm}^2$ \\} &
\makecell{\includegraphics[width=2cm]{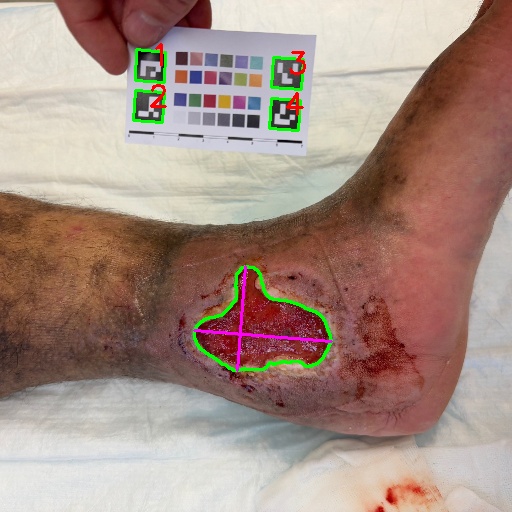}        \\55 x 42\\13.0 $\text{cm}^2$ \\} &
\makecell{\includegraphics[width=2cm]{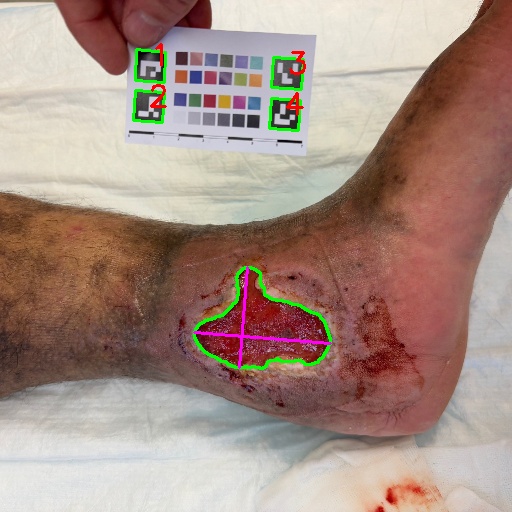}            \\54 x 40\\12.3 $\text{cm}^2$ \\} &
\makecell{\includegraphics[width=2cm]{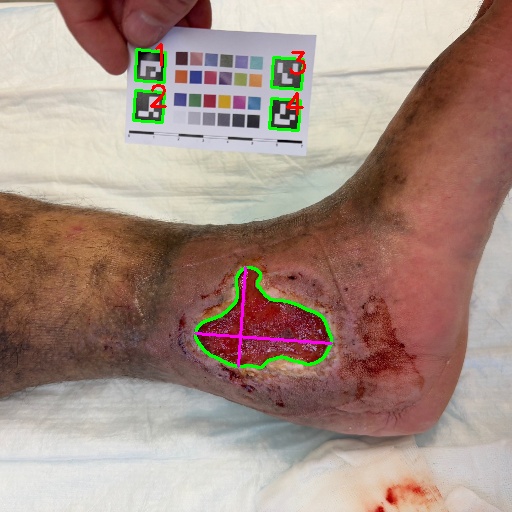}            \\54 x 40\\12.2 $\text{cm}^2$ \\} &
\makecell{\includegraphics[width=2cm]{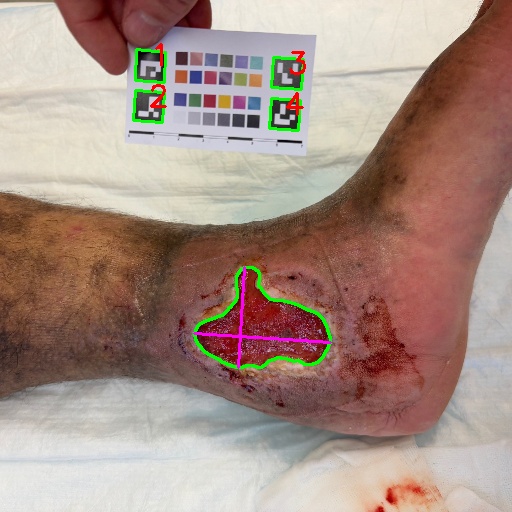}             \\54 x 40\\12.3 $\text{cm}^2$ \\} &
\begin{minipage}[c]{1.7cm}\centering \vspace{-0.5cm}\small  50 x 50\\60 x 50\\40 x 30\\\rule{1.4cm}{0.4pt}\\50.0 x 43.3\strut \end{minipage} \\
\vspace{0.3cm}

7&
\makecell{\includegraphics[width=2cm]{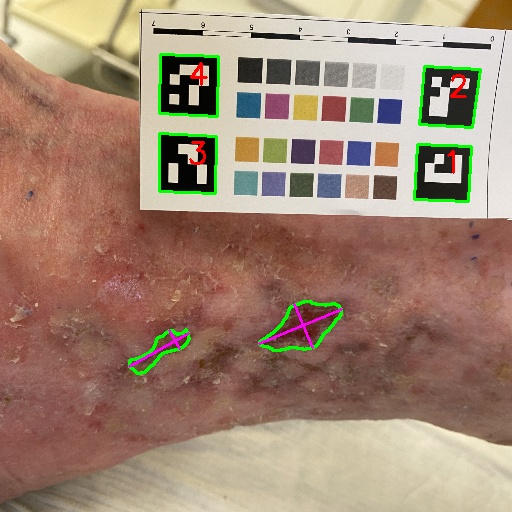}    \\19 x 10\\1.0 $\text{cm}^2$\\} &
\makecell{\includegraphics[width=2cm]{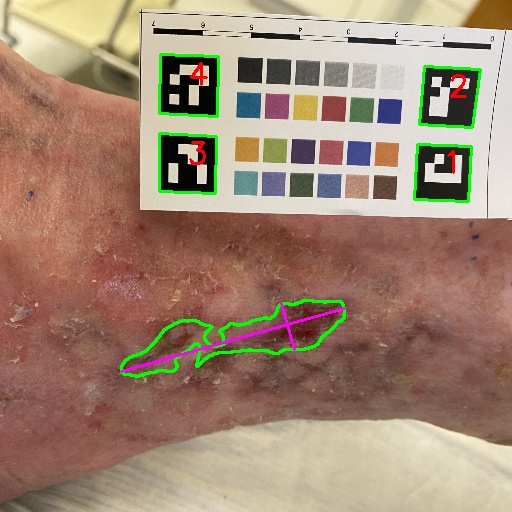}          \\49 x 10\\3.0 $\text{cm}^2$\\} &
\makecell{\includegraphics[width=2cm]{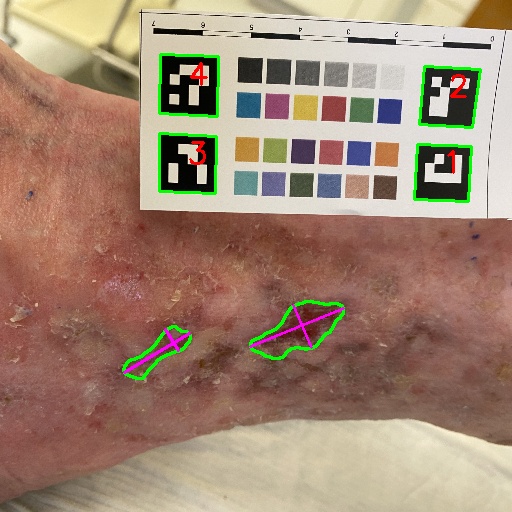}        \\21 x 10\\1.3 $\text{cm}^2$\\} &
\makecell{\includegraphics[width=2cm]{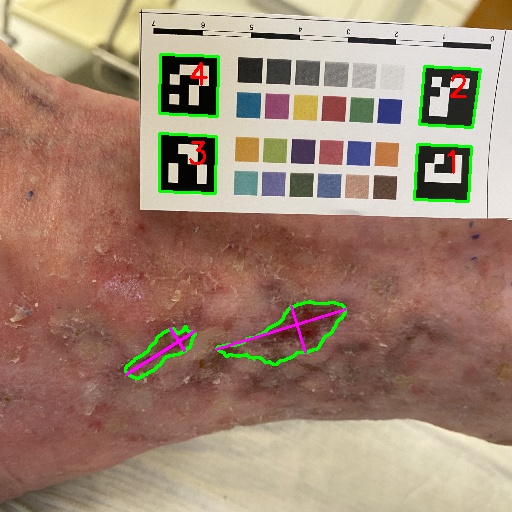}            \\28 x 10\\1.7 $\text{cm}^2$\\} &
\makecell{\includegraphics[width=2cm]{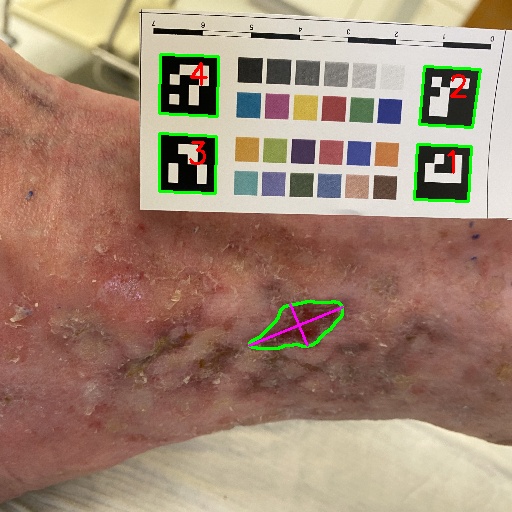}            \\21 x 9 \\1.2 $\text{cm}^2$\\} &
\makecell{\includegraphics[width=2cm]{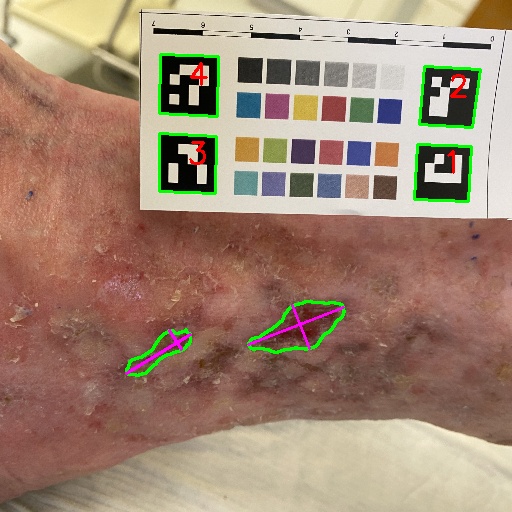}             \\21 x 10\\1.2 $\text{cm}^2$\\} &
\begin{minipage}[c]{1.7cm}\centering \vspace{-0.5cm}\small 40 x 10\\10 x 8\\25 x 5\\\rule{1.4cm}{0.4pt}\\25.0 x 7.7\strut \end{minipage} \\
\vspace{0.3cm}

8&
\makecell{\includegraphics[width=2cm]{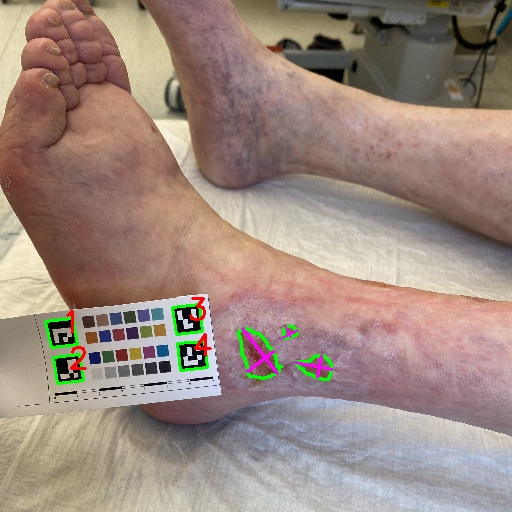}    \\25 x 14\\2.5 $\text{cm}^2$\\} &
\makecell{\includegraphics[width=2cm]{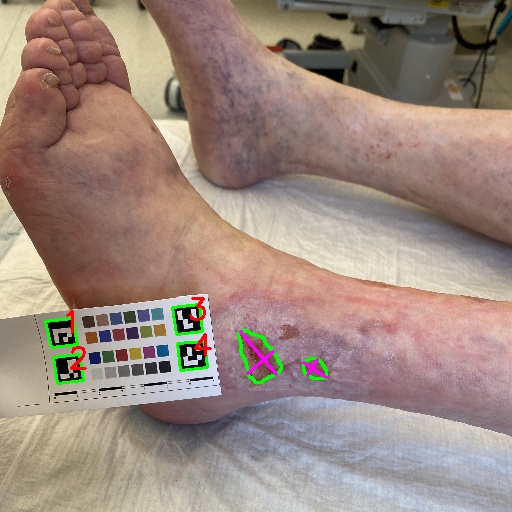}          \\25 x 15\\2.6 $\text{cm}^2$\\} &
\makecell{\includegraphics[width=2cm]{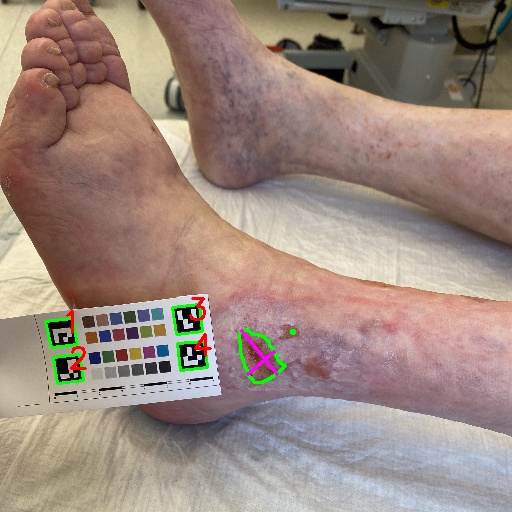}        \\25 x 16\\2.9 $\text{cm}^2$\\} &
\makecell{\includegraphics[width=2cm]{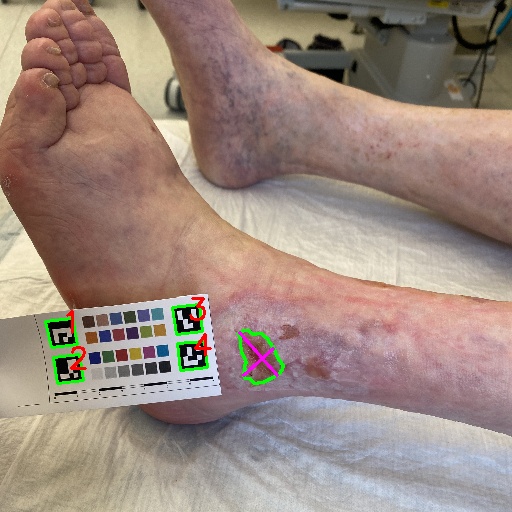}            \\25 x 18\\3.0 $\text{cm}^2$\\} &
\makecell{\includegraphics[width=2cm]{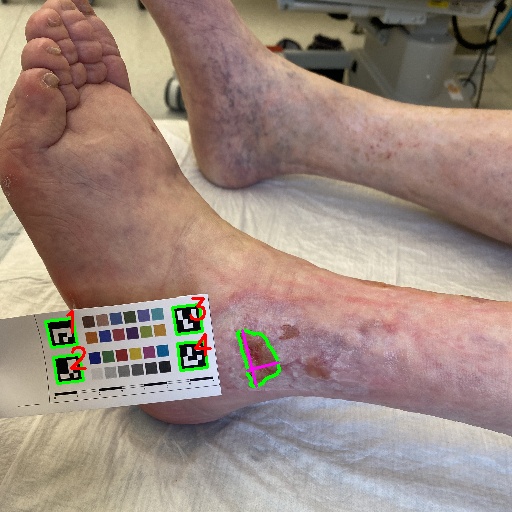}            \\25 x 14\\2.7 $\text{cm}^2$\\} &
\makecell{\includegraphics[width=2cm]{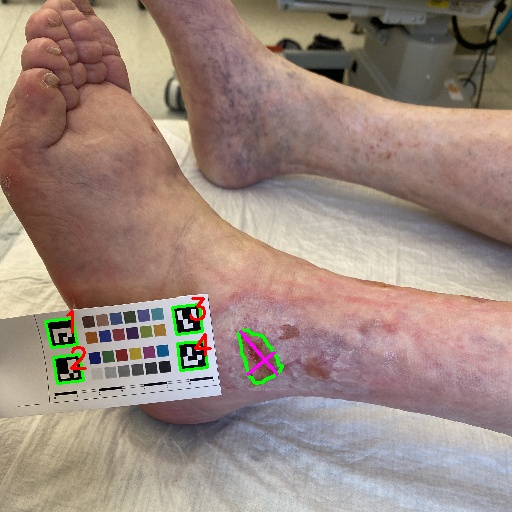}             \\25 x 15\\2.7 $\text{cm}^2$\\} &
\begin{minipage}[c]{1.7cm}\centering \vspace{-0.5cm}\small 45 x 40\\38 x 12\\25 x 13\\\rule{1.4cm}{0.4pt}\\36.0 x 21.7\strut \end{minipage} \\
\vspace{0.3cm}

9&
\makecell{\includegraphics[width=2cm]{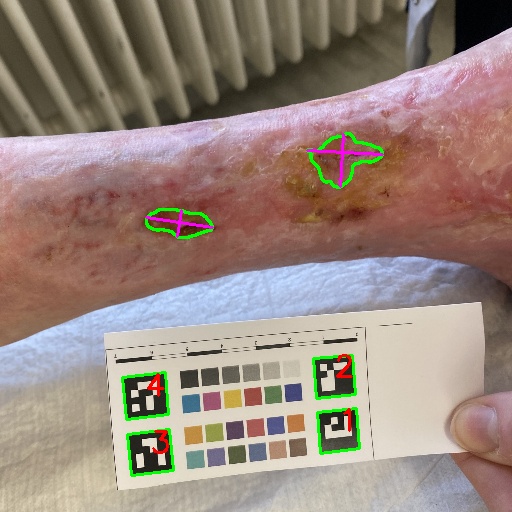}    \\22 x 15\\1.8 $\text{cm}^2$\\} &
\makecell{\includegraphics[width=2cm]{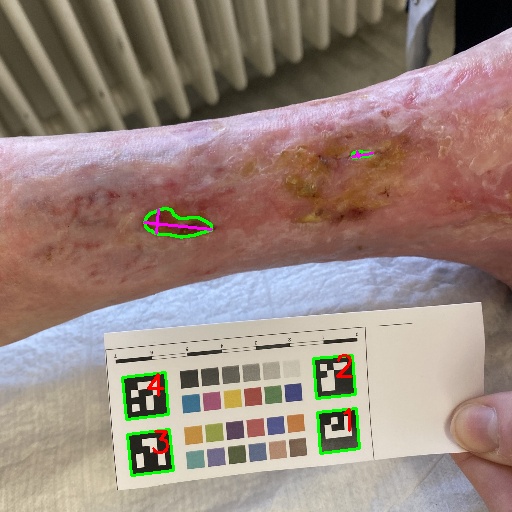}          \\20 x 7 \\1.0 $\text{cm}^2$\\} &
\makecell{\includegraphics[width=2cm]{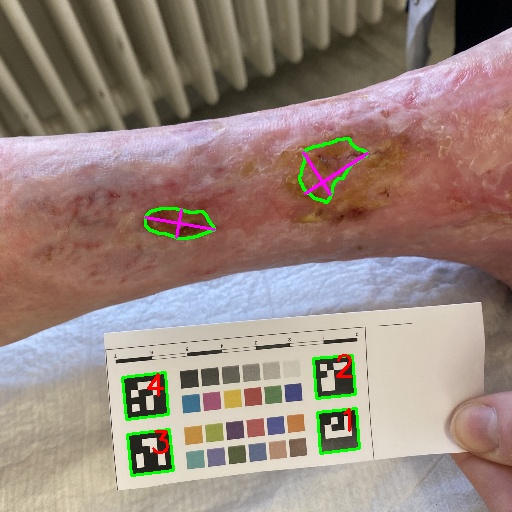}        \\22 x 15\\2.0 $\text{cm}^2$\\} &
\makecell{\includegraphics[width=2cm]{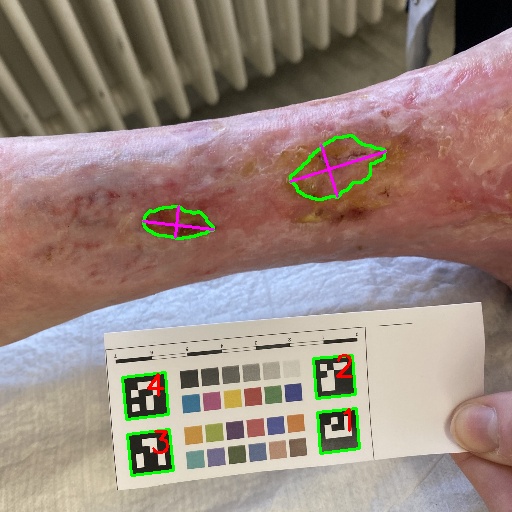}            \\29 x 16\\3.0 $\text{cm}^2$\\} &
\makecell{\includegraphics[width=2cm]{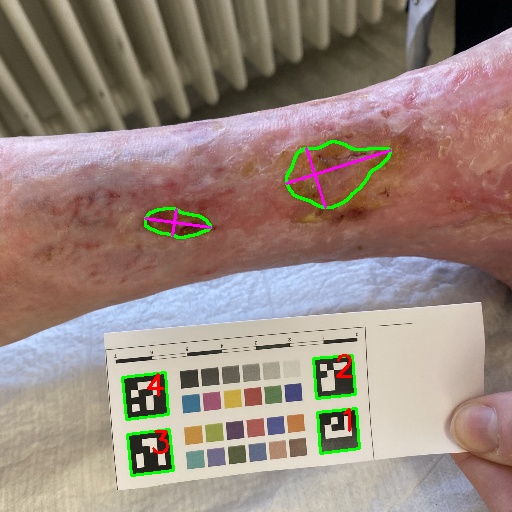}            \\32 x 18\\3.7 $\text{cm}^2$\\} &
\makecell{\includegraphics[width=2cm]{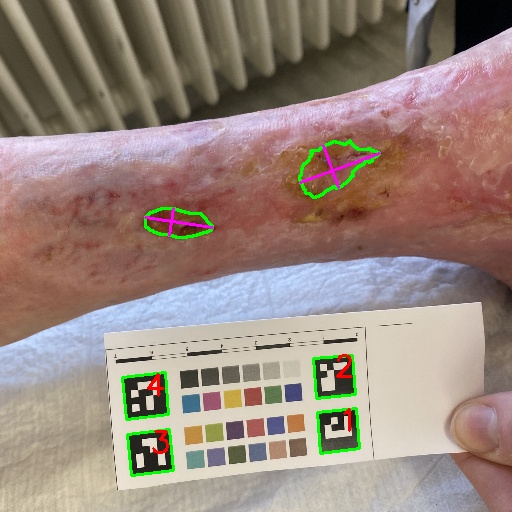}             \\25 x 13\\2.1 $\text{cm}^2$\\} &
\begin{minipage}[c]{1.7cm}\centering \vspace{-0.5cm}\small 90 x 40\\90 x 40\\40 x 30\\\rule{1.4cm}{0.4pt}\\73.3 x 36.7\strut \end{minipage} \\
\vspace{0.3cm}

10&
\makecell{\includegraphics[width=2cm]{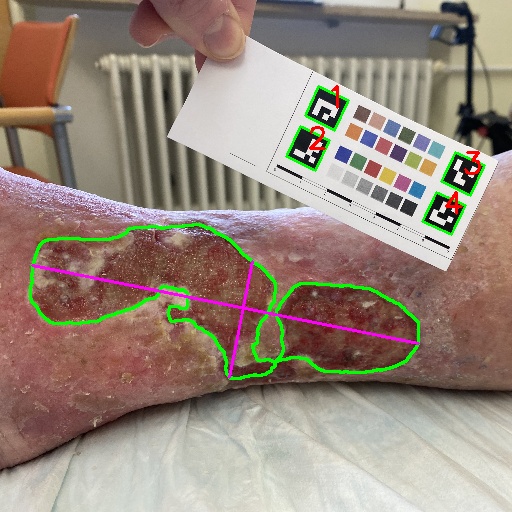}    \\144 x 43\\41.2 $\text{cm}^2$ \\} &
\makecell{\includegraphics[width=2cm]{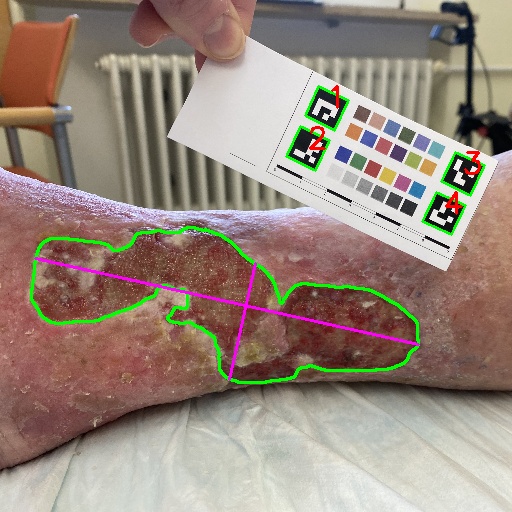}          \\143 x 43\\42.6 $\text{cm}^2$ \\} &
\makecell{\includegraphics[width=2cm]{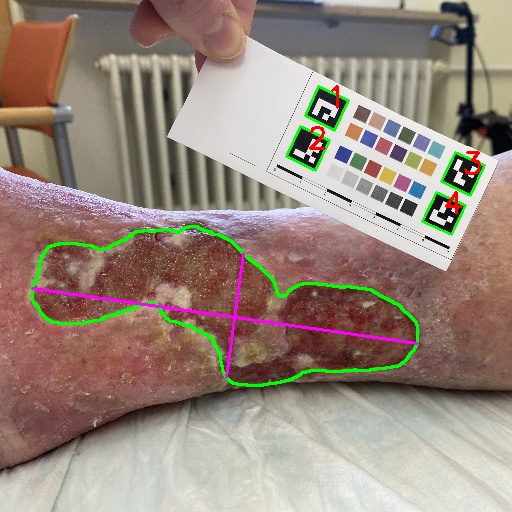}        \\142 x 46\\43.8 $\text{cm}^2$ \\} &
\makecell{\includegraphics[width=2cm]{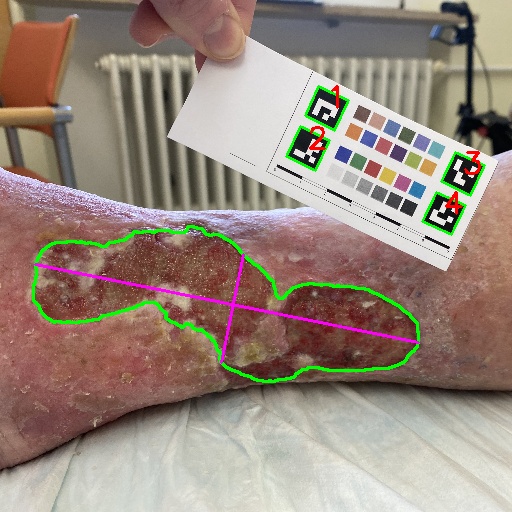}            \\143 x 40\\42.9 $\text{cm}^2$ \\} &
\makecell{\includegraphics[width=2cm]{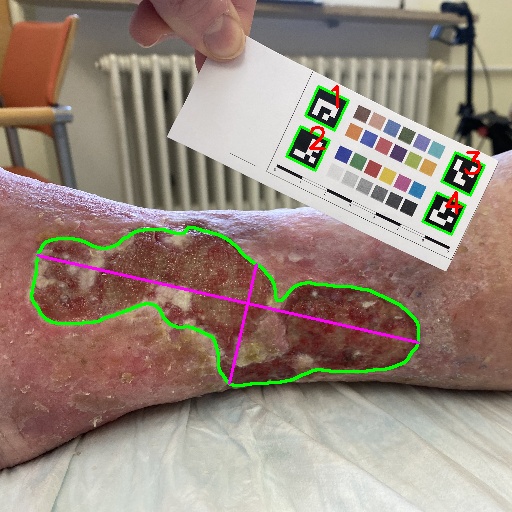}            \\143 x 45\\44.6 $\text{cm}^2$ \\} &
\makecell{\includegraphics[width=2cm]{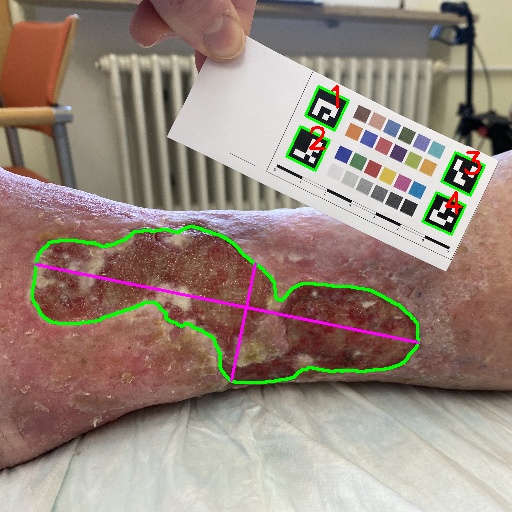}             \\143 x 44\\43.8 $\text{cm}^2$ \\} &
\begin{minipage}[c]{1.7cm}\centering \vspace{-0.5cm}\small 120 x 70\\140 x 50\\100 x 40\\\rule{1.4cm}{0.4pt}\\120.0 x 53.3\strut \end{minipage} \\
\vspace{0.3cm}

11&
\makecell{\includegraphics[width=2cm]{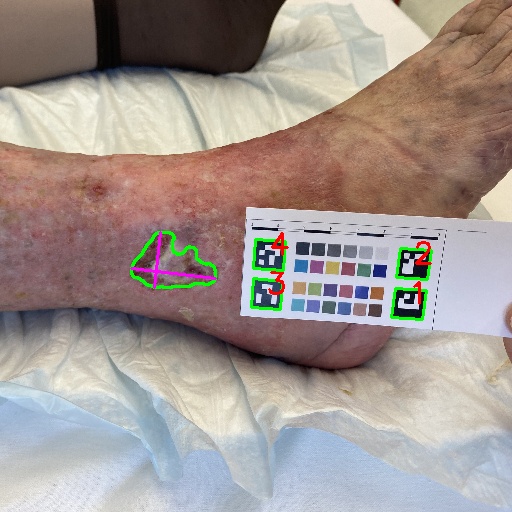}    \\34 x 22\\4.5 $\text{cm}^2$\\} &
\makecell{\includegraphics[width=2cm]{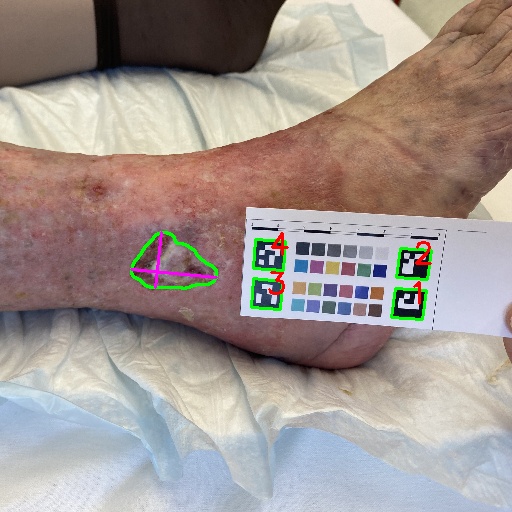}          \\34 x 22\\4.9 $\text{cm}^2$\\} &
\makecell{\includegraphics[width=2cm]{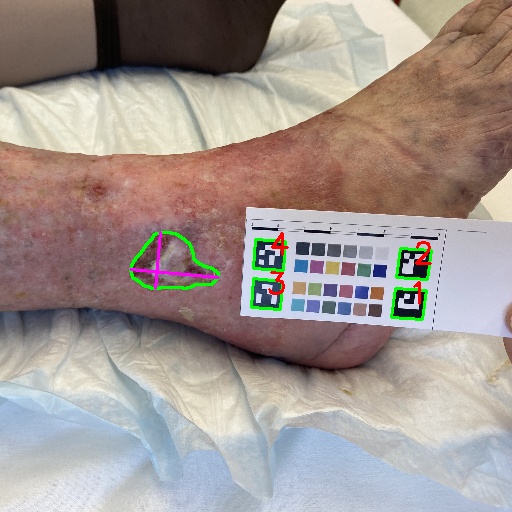}        \\35 x 22\\5.0 $\text{cm}^2$\\} &
\makecell{\includegraphics[width=2cm]{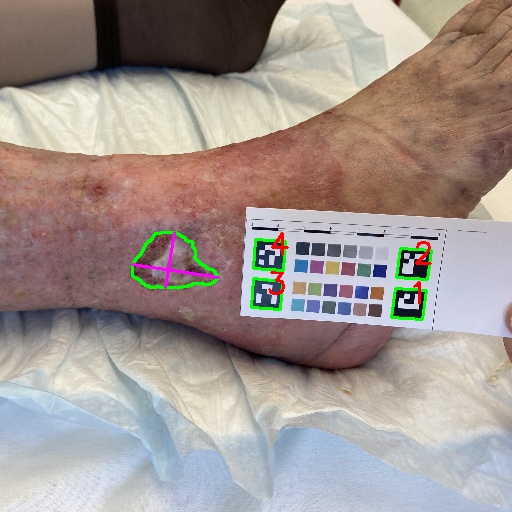}            \\34 x 22\\5.0 $\text{cm}^2$\\} &
\makecell{\includegraphics[width=2cm]{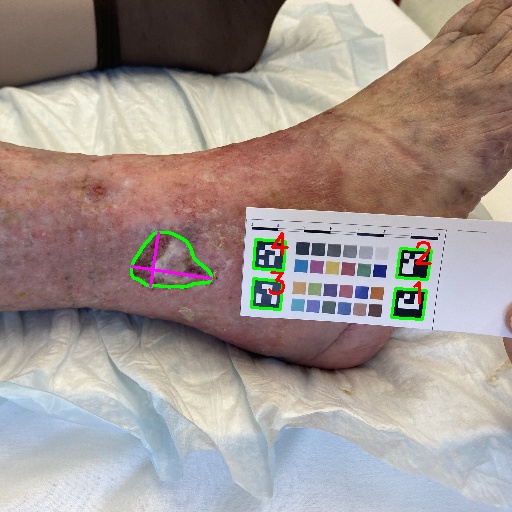}            \\32 x 22\\4.7 $\text{cm}^2$\\} &
\makecell{\includegraphics[width=2cm]{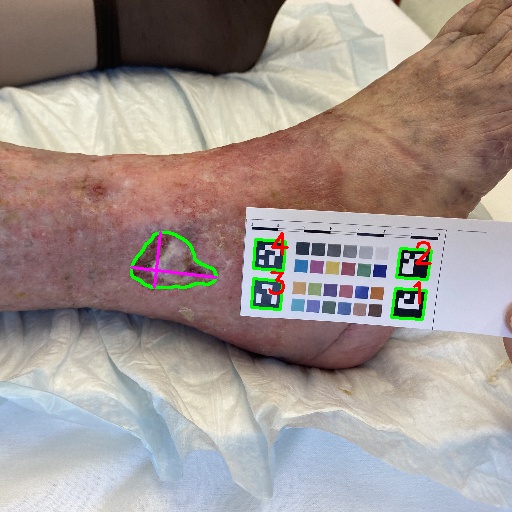}             \\34 x 22\\4.9 $\text{cm}^2$\\} &
\begin{minipage}[c]{1.7cm}\centering \vspace{-0.5cm}\small 40 x 35\\38 x 30\\25 x 18\\\rule{1.4cm}{0.4pt}\\34.3 x 27.7\strut \end{minipage} \\
\vspace{0.3cm}

12&
\makecell{\includegraphics[width=2cm]{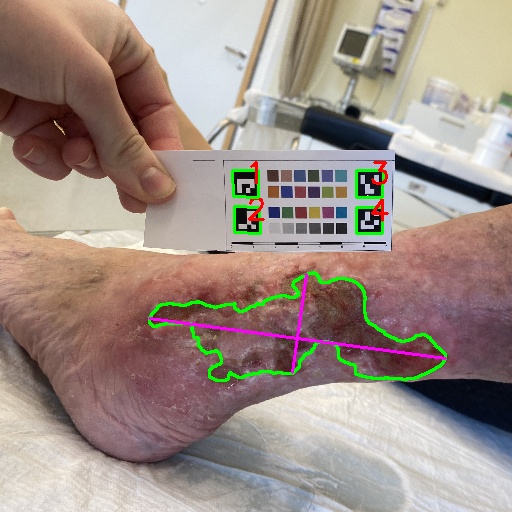}    \\134 x 44\\32.0 $\text{cm}^2$\\} &
\makecell{\includegraphics[width=2cm]{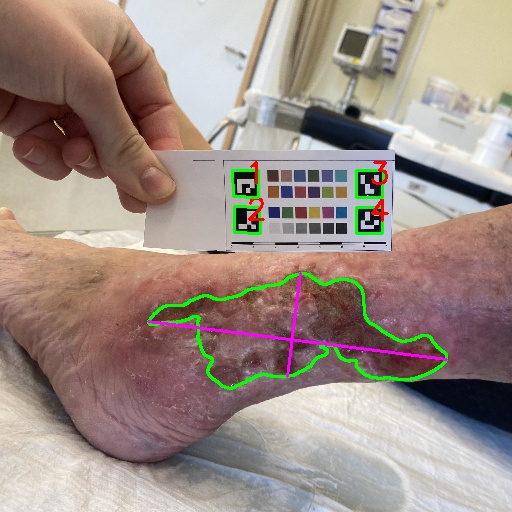}          \\135 x 47\\35.7 $\text{cm}^2$\\} &
\makecell{\includegraphics[width=2cm]{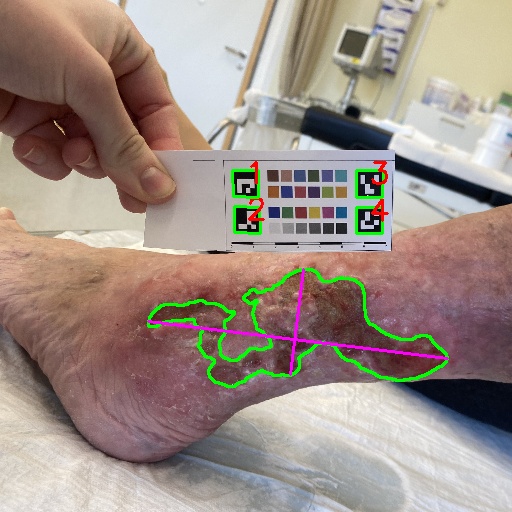}        \\135 x 47\\30.6 $\text{cm}^2$\\} &
\makecell{\includegraphics[width=2cm]{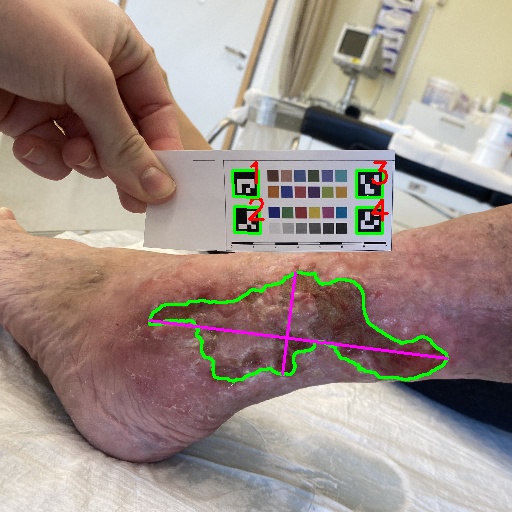}            \\135 x 46\\33.1 $\text{cm}^2$\\} &
\makecell{\includegraphics[width=2cm]{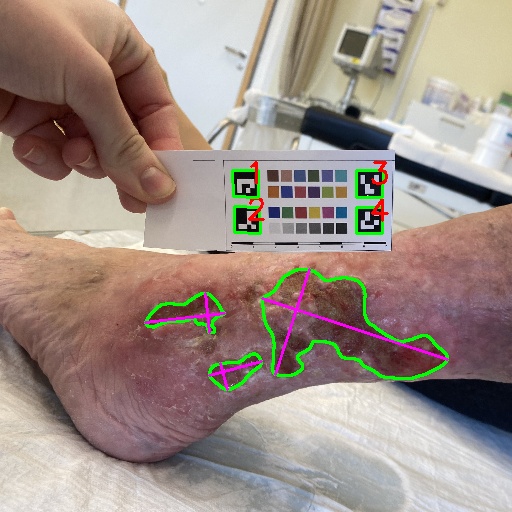}            \\87 x 49 \\21.5 $\text{cm}^2$\\} &
\makecell{\includegraphics[width=2cm]{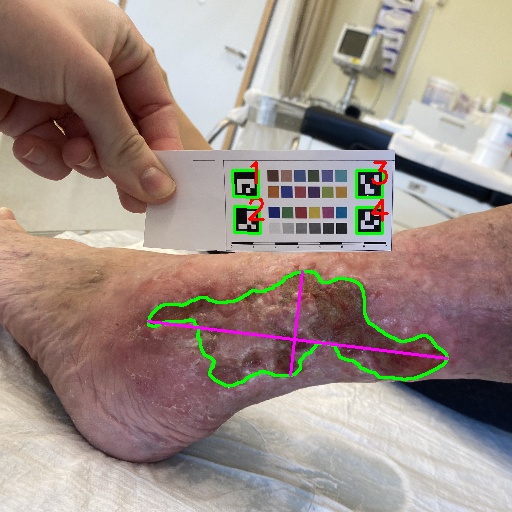}             \\135 x 47\\34.4 $\text{cm}^2$\\} &
\begin{minipage}[c]{1.7cm}\centering \vspace{-0.5cm}\small 130 x 60\\150 x 70\\80 x 45\\\rule{1.4cm}{0.4pt}\\120.0 x 58.3\strut \end{minipage} \\
\vspace{0.3cm}

13&
\makecell{\includegraphics[width=2cm]{images/SizeRetrieval/Bild_13_VWFormerConvNeXtS_mask_annotated.jpg}    \\76 x 26\\15.3 $\text{cm}^2$\\} &
\makecell{\includegraphics[width=2cm]{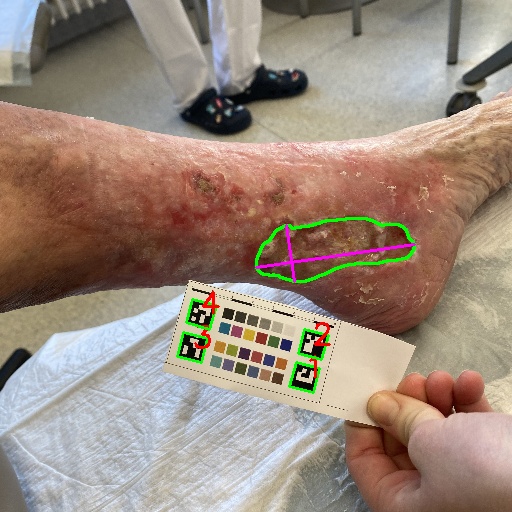}          \\76 x 27\\15.3 $\text{cm}^2$\\} &
\makecell{\includegraphics[width=2cm]{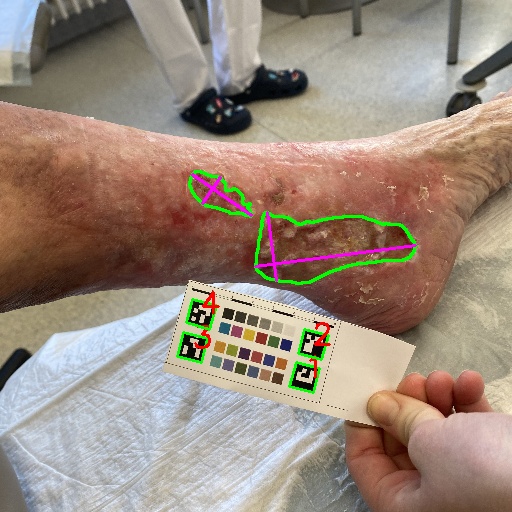}        \\77 x 33\\17.5 $\text{cm}^2$\\} &
\makecell{\includegraphics[width=2cm]{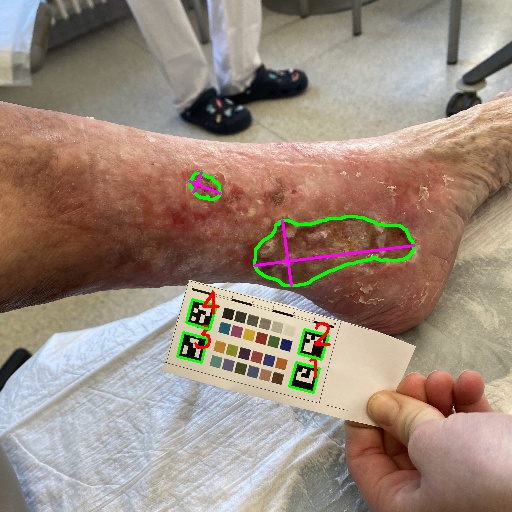}            \\77 x 31\\16.2 $\text{cm}^2$\\} &
\makecell{\includegraphics[width=2cm]{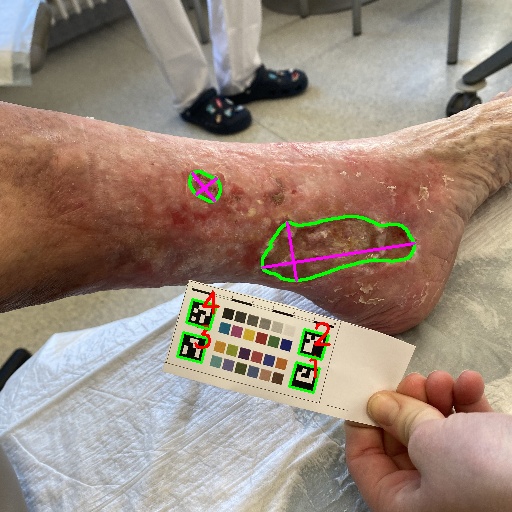}            \\73 x 28\\14.9 $\text{cm}^2$\\} &
\makecell{\includegraphics[width=2cm]{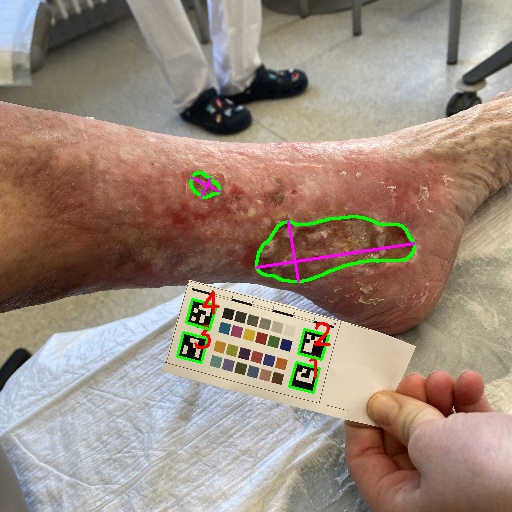}             \\76 x 29\\15.7 $\text{cm}^2$\\} &
\begin{minipage}[c]{1.7cm}\centering \vspace{-0.5cm}\small 110 x 40\\80 x 40\\70 x 30\\\rule{1.4cm}{0.4pt}\\86.7 x 36.7\strut \end{minipage} \\
\vspace{0.3cm}

14&
\makecell{\includegraphics[width=2cm]{images/SizeRetrieval/Bild_14_VWFormerConvNeXtS_mask_annotated.jpg}    \\20 x 16\\2.2 $\text{cm}^2$\\} &
\makecell{\includegraphics[width=2cm]{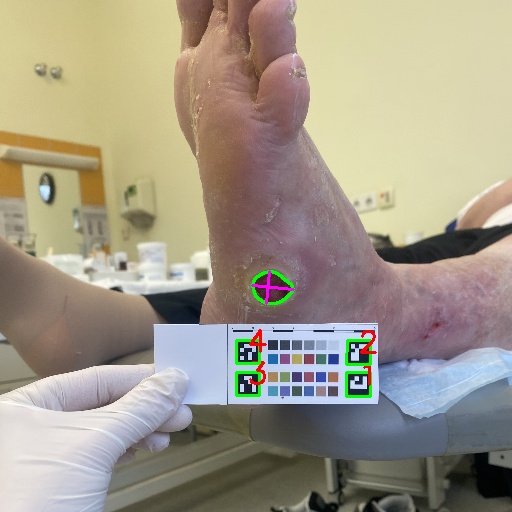}          \\21 x 16\\2.3 $\text{cm}^2$\\} &
\makecell{\includegraphics[width=2cm]{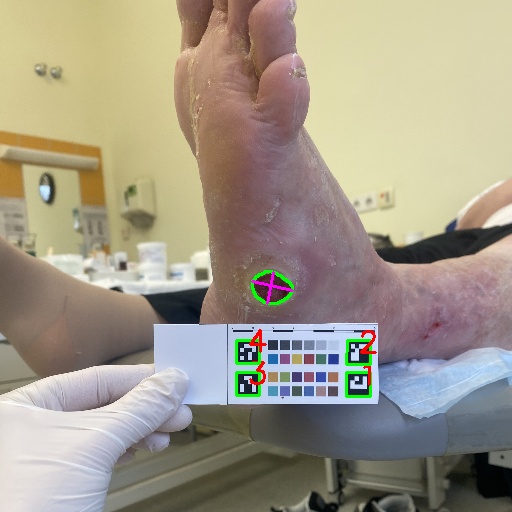}        \\21 x 16\\2.4 $\text{cm}^2$\\} &
\makecell{\includegraphics[width=2cm]{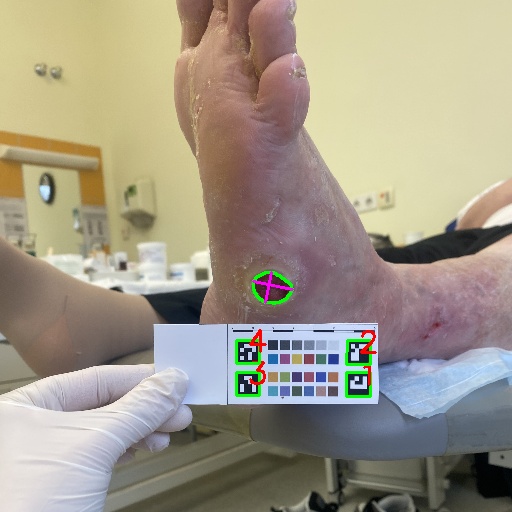}            \\20 x 15\\2.2 $\text{cm}^2$\\} &
\makecell{\includegraphics[width=2cm]{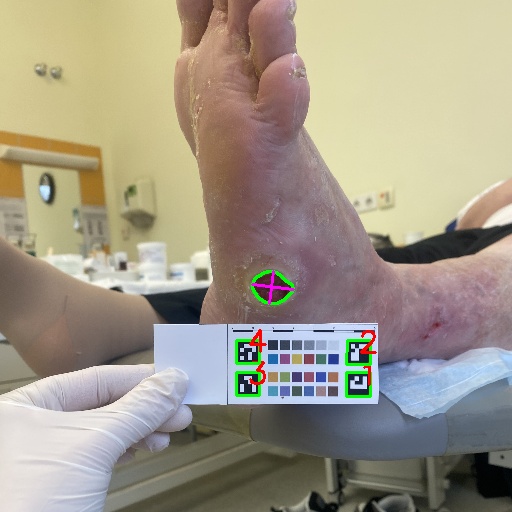}            \\21 x 16\\2.3 $\text{cm}^2$\\} &
\makecell{\includegraphics[width=2cm]{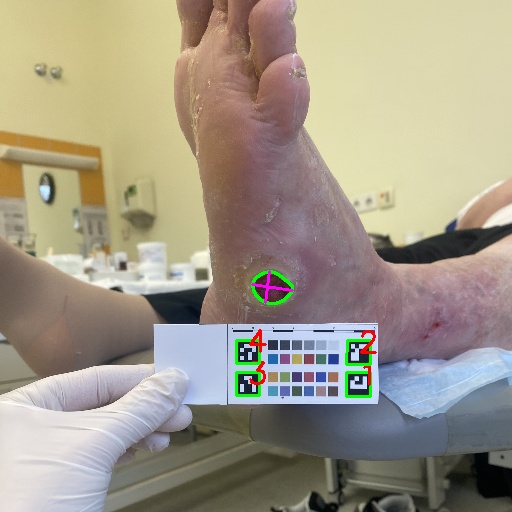}             \\20 x 16\\2.3 $\text{cm}^2$\\} &
\begin{minipage}[c]{1.7cm}\centering \vspace{-0.5cm}\small 20 x 20\\22 x 20\\18 x 14\\\rule{1.4cm}{0.4pt}\\20.0 x 18.0\strut \end{minipage} \\
\vspace{0.3cm}

15&
\makecell{\includegraphics[width=2cm]{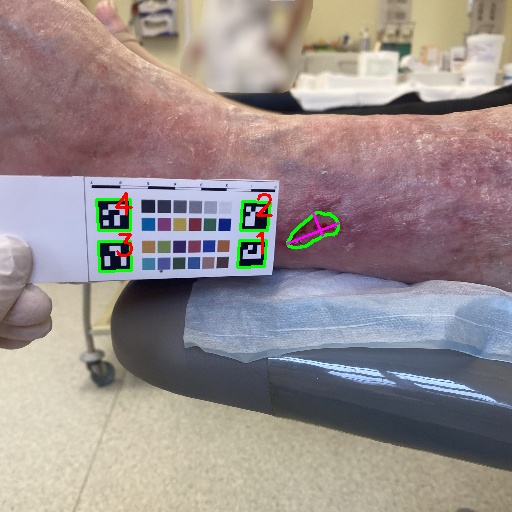}    \\22 x 10\\1.7 $\text{cm}^2$\\} &
\makecell{\includegraphics[width=2cm]{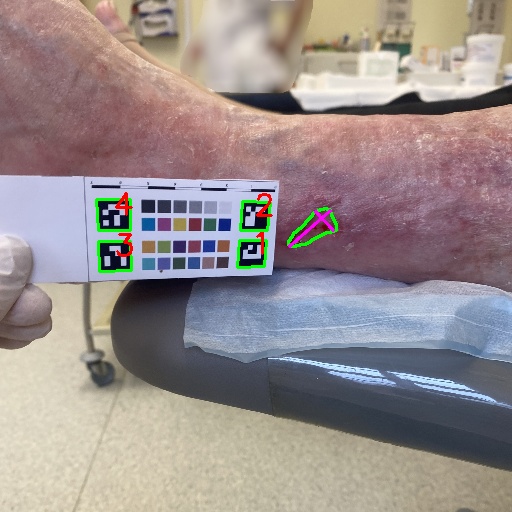}          \\22 x 11\\1.5 $\text{cm}^2$\\} &
\makecell{\includegraphics[width=2cm]{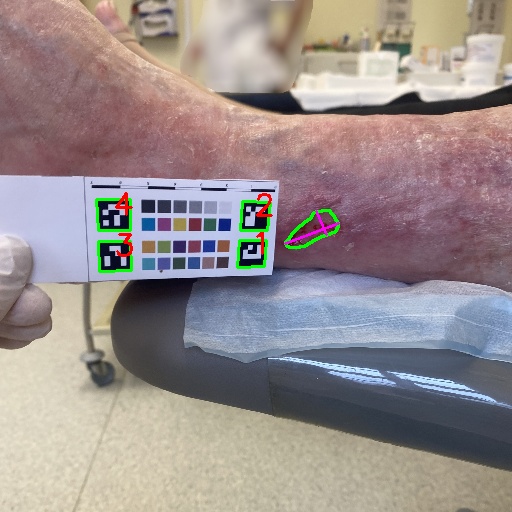}        \\23 x 10\\1.7 $\text{cm}^2$\\} &
\makecell{\includegraphics[width=2cm]{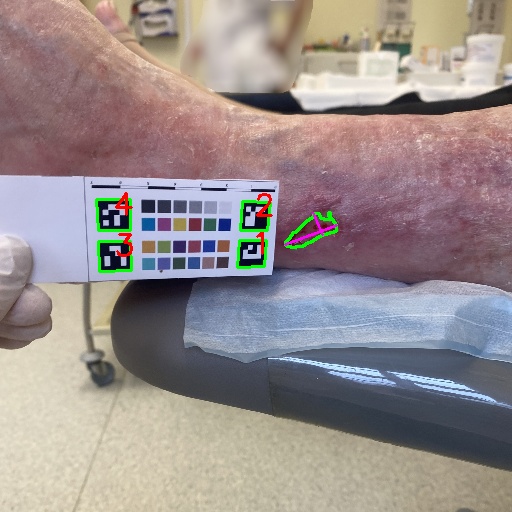}            \\22 x 9 \\1.4 $\text{cm}^2$\\} &
\makecell{\includegraphics[width=2cm]{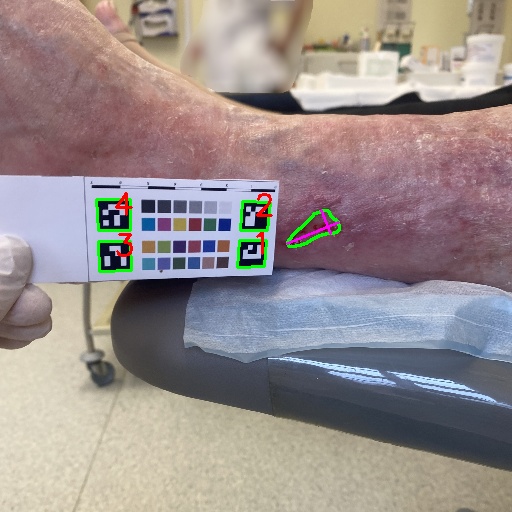}            \\22 x 10\\1.5 $\text{cm}^2$\\} &
\makecell{\includegraphics[width=2cm]{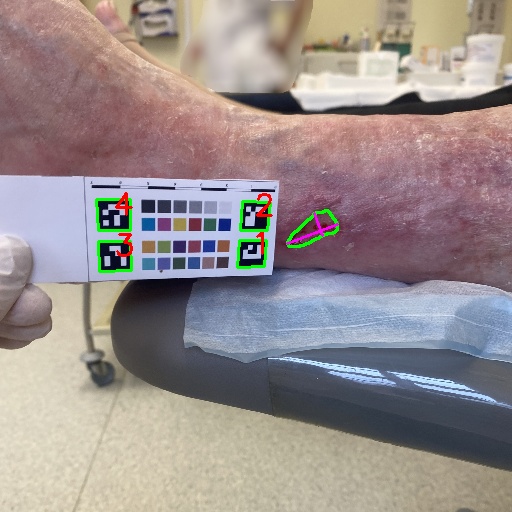}             \\22 x 9 \\1.5 $\text{cm}^2$\\} &
\begin{minipage}[c]{1.7cm}\centering \vspace{-0.5cm}\small 30 x 25\\25 x 10\\30 x 10\\\rule{1.4cm}{0.4pt}\\28.3 x 15.0\strut \end{minipage} \\
\vspace{0.3cm}

16&
\makecell{\includegraphics[width=2cm]{images/SizeRetrieval/Bild_16_VWFormerConvNeXtS_mask_annotated.jpg}    \\7 x 6\\0.3 $\text{cm}^2$\\} &
\makecell{\includegraphics[width=2cm]{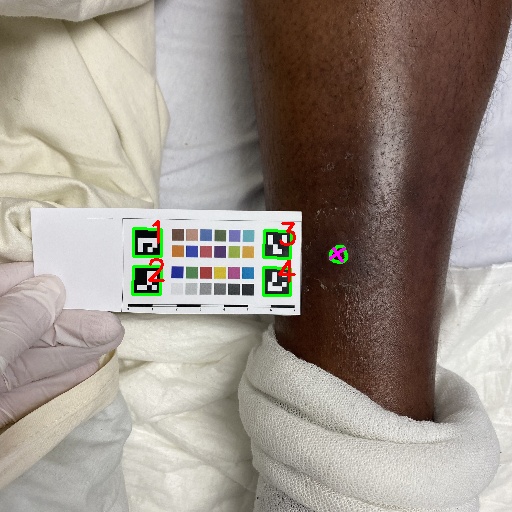}          \\7 x 7\\0.4 $\text{cm}^2$\\} &
\makecell{\includegraphics[width=2cm]{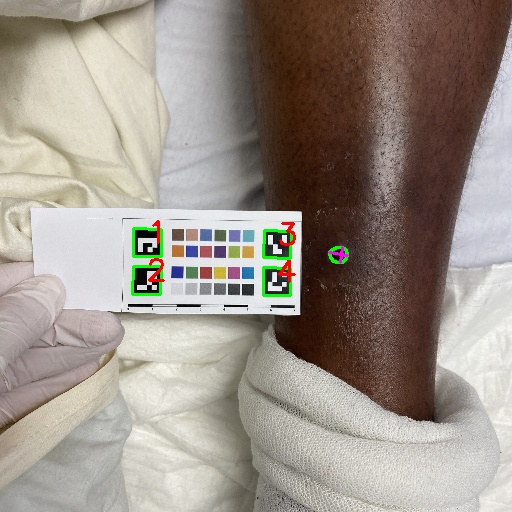}        \\8 x 6\\0.4 $\text{cm}^2$\\} &
\makecell{\includegraphics[width=2cm]{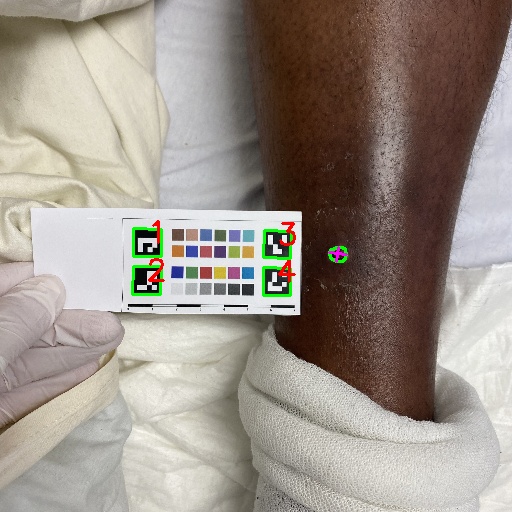}            \\7 x 6\\0.3 $\text{cm}^2$\\} &
\makecell{\includegraphics[width=2cm]{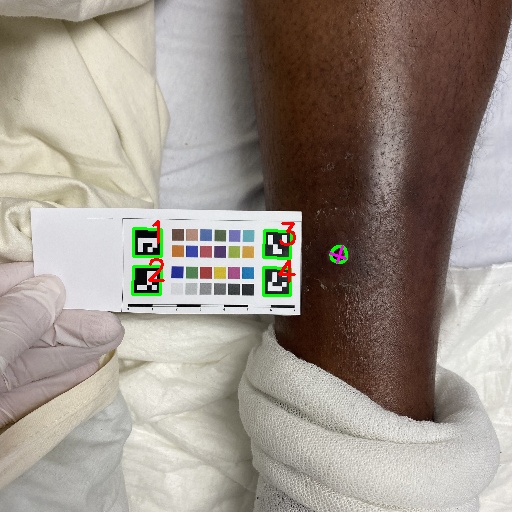}            \\7 x 7\\0.4 $\text{cm}^2$\\} &
\makecell{\includegraphics[width=2cm]{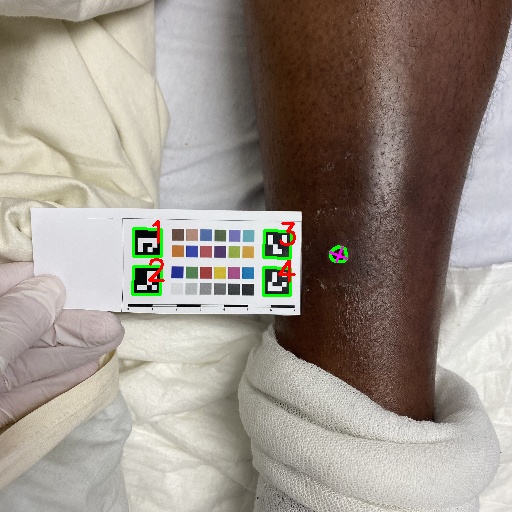}             \\7 x 7\\0.3 $\text{cm}^2$\\} &
\begin{minipage}[c]{1.7cm}\centering \vspace{-0.5cm}\small 5 x 5\\5 x 4\\5 x 5\\\rule{1.4cm}{0.4pt}\\5.0 x 4.7\strut \end{minipage} \\
\vspace{0.3cm}

17&
\makecell{\includegraphics[width=2cm]{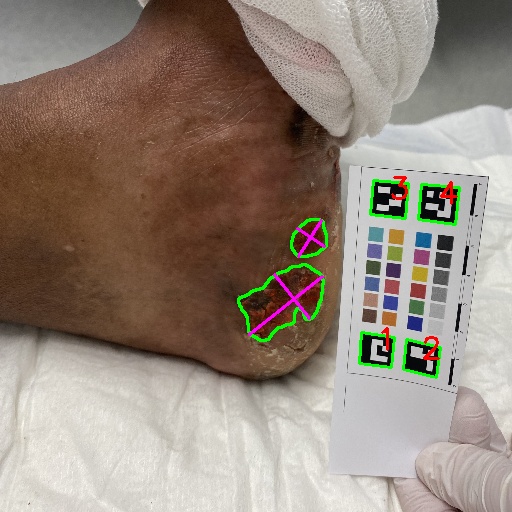}    \\33 x 20\\4.6 $\text{cm}^2$\\} &
\makecell{\includegraphics[width=2cm]{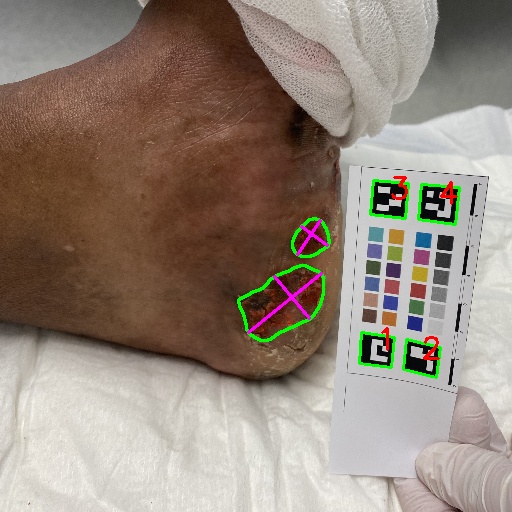}          \\33 x 19\\5.0 $\text{cm}^2$\\} &
\makecell{\includegraphics[width=2cm]{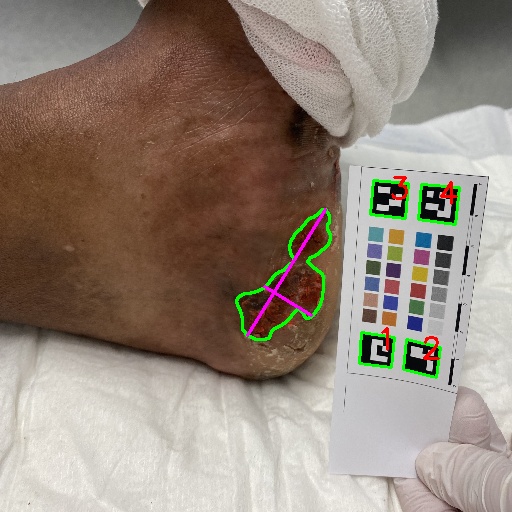}        \\52 x 19\\6.5 $\text{cm}^2$\\} &
\makecell{\includegraphics[width=2cm]{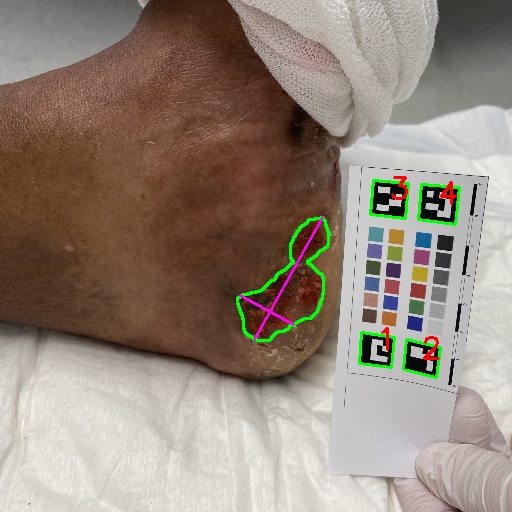}            \\49 x 21\\6.6 $\text{cm}^2$\\} &
\makecell{\includegraphics[width=2cm]{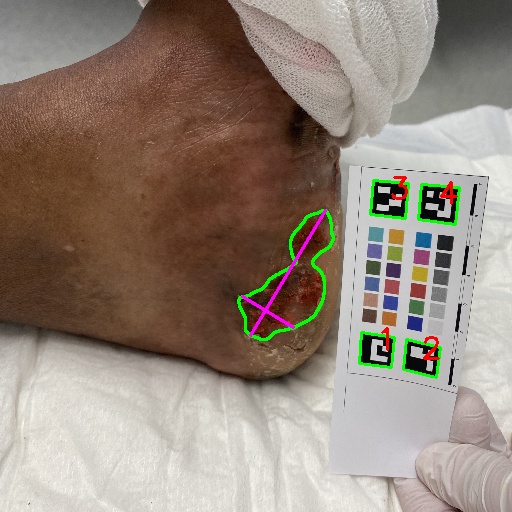}            \\51 x 21\\7.0 $\text{cm}^2$\\} &
\makecell{\includegraphics[width=2cm]{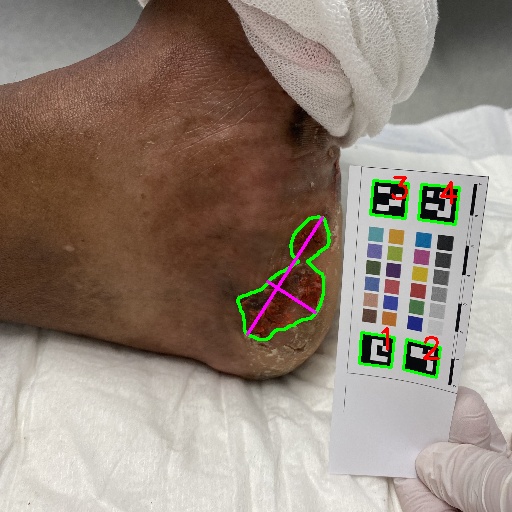}             \\49 x 20\\6.5 $\text{cm}^2$\\} &
\begin{minipage}[c]{1.7cm}\centering \vspace{-0.5cm}\small 30 x 30\\5 x 5\\30 x 20\\\rule{1.4cm}{0.4pt}\\21.7 x 18.3\strut \end{minipage} \\
\vspace{0.3cm}

18&
\makecell{\includegraphics[width=2cm]{images/SizeRetrieval/Bild_18_VWFormerConvNeXtS_mask_annotated.jpg}    \\25 x 13\\2.5 $\text{cm}^2$\\} &
\makecell{\includegraphics[width=2cm]{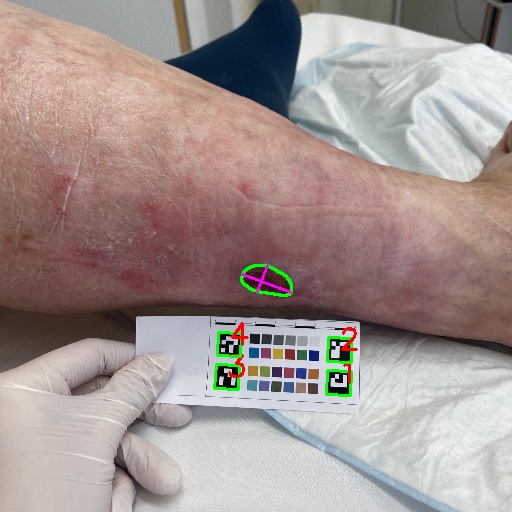}          \\26 x 14\\2.7 $\text{cm}^2$\\} &
\makecell{\includegraphics[width=2cm]{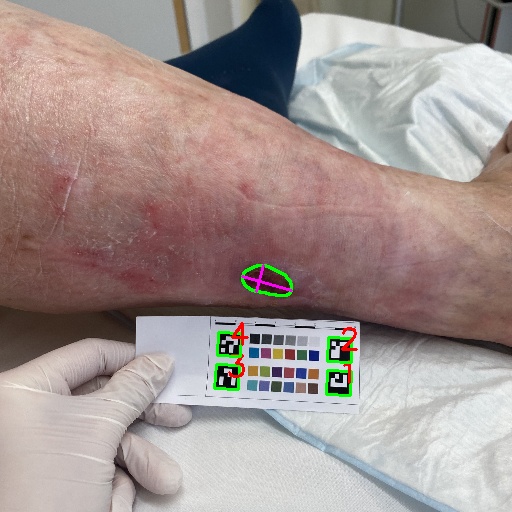}        \\25 x 13\\2.6 $\text{cm}^2$\\} &
\makecell{\includegraphics[width=2cm]{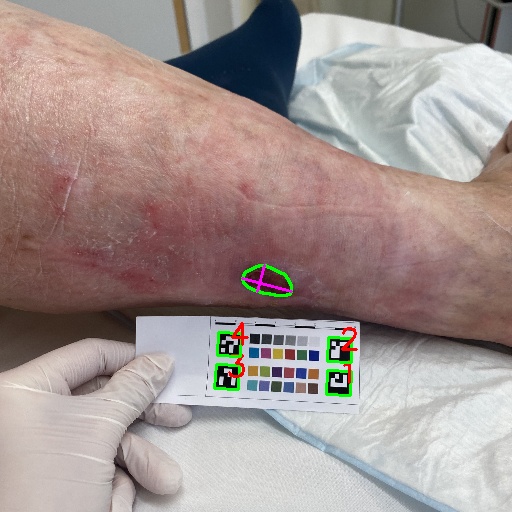}            \\25 x 13\\2.6 $\text{cm}^2$\\} &
\makecell{\includegraphics[width=2cm]{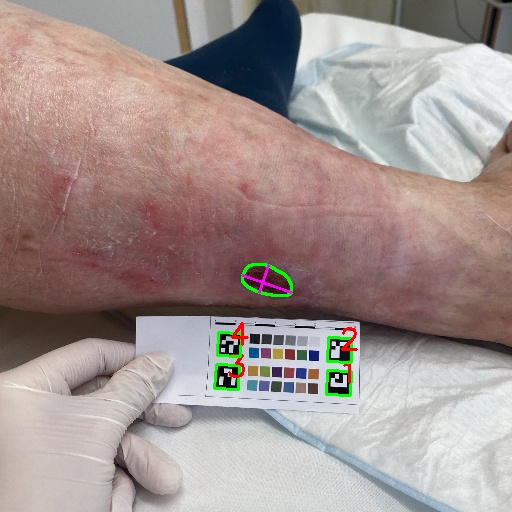}            \\26 x 14\\2.8 $\text{cm}^2$\\} &
\makecell{\includegraphics[width=2cm]{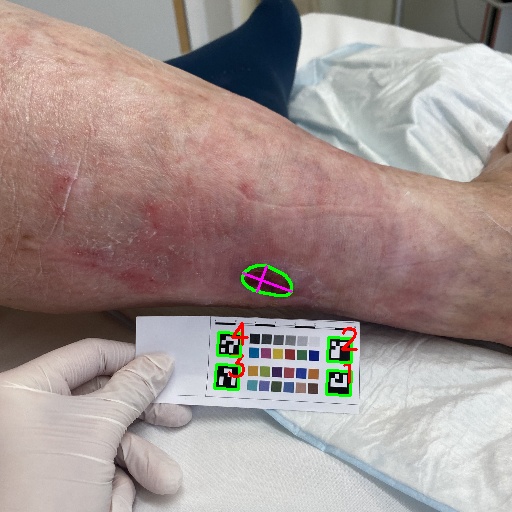}             \\25 x 13\\2.6 $\text{cm}^2$\\} &
\begin{minipage}[c]{1.7cm}\centering \vspace{-0.5cm}\small 30 x 20\\28 x 15\\22 x 15\\\rule{1.4cm}{0.4pt}\\26.7 x 16.7\strut \end{minipage} \\
\vspace{0.3cm}

19&
\makecell{\includegraphics[width=2cm]{images/SizeRetrieval/Bild_19_VWFormerConvNeXtS_mask_annotated.jpg}    \\126 x 49\\50.6 $\text{cm}^2$\\} &
\makecell{\includegraphics[width=2cm]{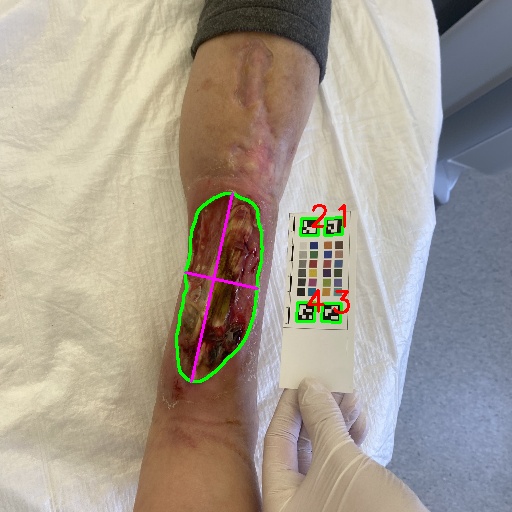}          \\125 x 48\\49.0 $\text{cm}^2$\\} &
\makecell{\includegraphics[width=2cm]{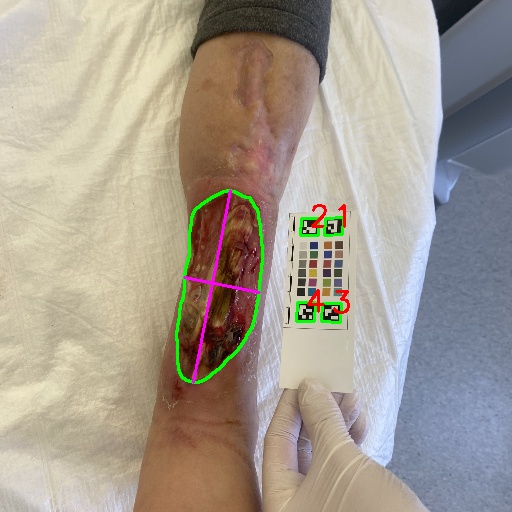}        \\127 x 50\\51.3 $\text{cm}^2$\\} &
\makecell{\includegraphics[width=2cm]{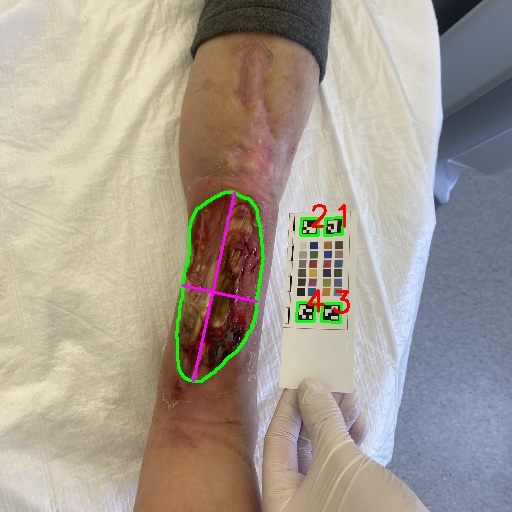}            \\125 x 49\\50.7 $\text{cm}^2$\\} &
\makecell{\includegraphics[width=2cm]{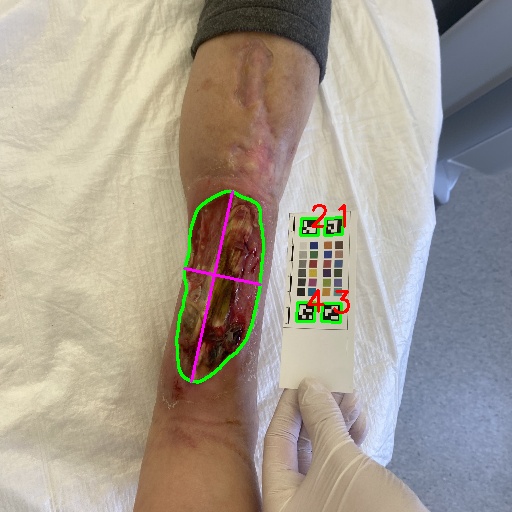}            \\126 x 50\\50.9 $\text{cm}^2$\\} &
\makecell{\includegraphics[width=2cm]{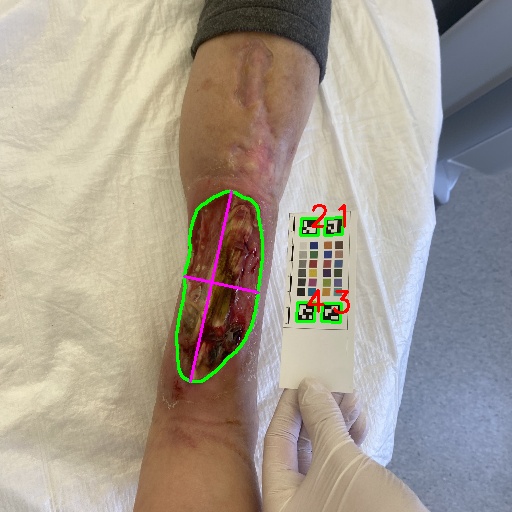}             \\126 x 49\\50.5 $\text{cm}^2$\\} &
\begin{minipage}[c]{1.7cm}\centering \vspace{-0.5cm}\small 120 x 50\\130 x 50\\120 x 45\\\rule{1.4cm}{0.4pt}\\123.3 x 48.3\strut \end{minipage} \\
\vspace{0.3cm}

20&
\makecell{\includegraphics[width=2cm]{images/SizeRetrieval/Bild_20_VWFormerConvNeXtS_mask_annotated.jpg}    \\76 x 36\\18.5 $\text{cm}^2$\\} &
\makecell{\includegraphics[width=2cm]{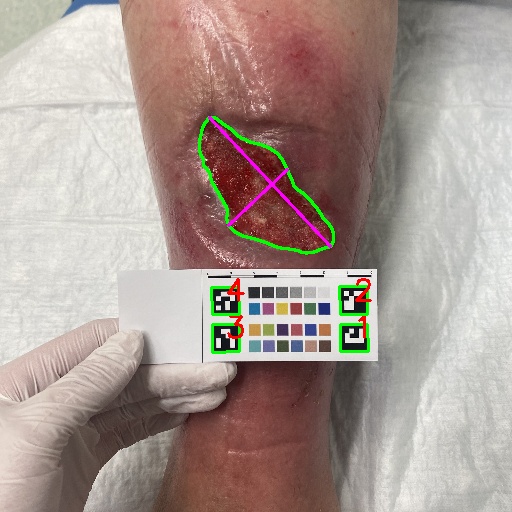}          \\76 x 35\\17.7 $\text{cm}^2$\\} &
\makecell{\includegraphics[width=2cm]{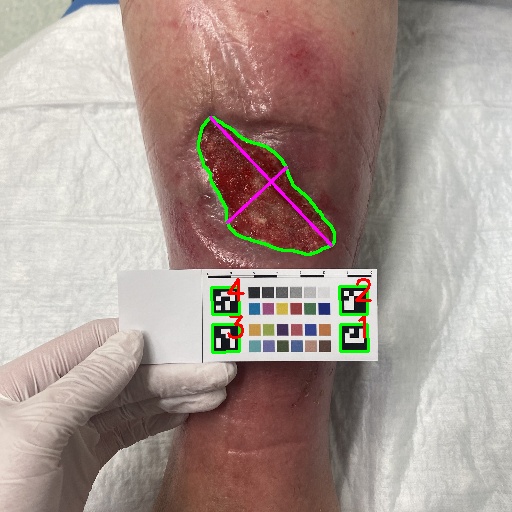}        \\76 x 35\\18.0 $\text{cm}^2$\\} &
\makecell{\includegraphics[width=2cm]{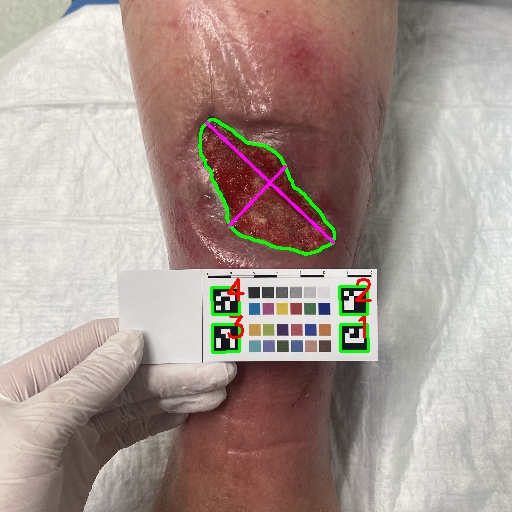}            \\75 x 35\\17.4 $\text{cm}^2$\\} &
\makecell{\includegraphics[width=2cm]{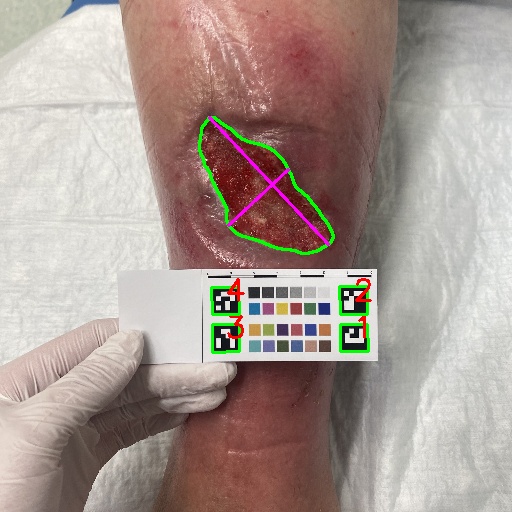}            \\75 x 35\\17.7 $\text{cm}^2$\\} &
\makecell{\includegraphics[width=2cm]{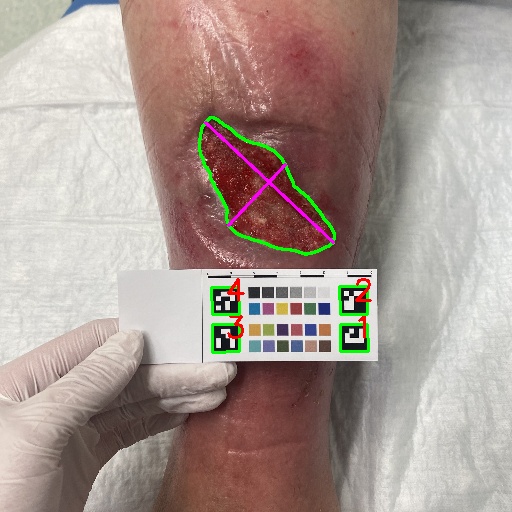}             \\75 x 36\\17.8 $\text{cm}^2$\\} &
\begin{minipage}[c]{1.7cm}\centering \vspace{-0.5cm}\small 55 x 40\\80 x 35\\70 x 35\\\rule{1.4cm}{0.4pt}\\68.3 x 36.7\strut \end{minipage} \\

\end{longtable}

\clearpage

\subsection{Mask Assessment and Size Annotation Tool}

\begin{figure}[ht!]
    \centering
    \begin{subfigure}{.48\textwidth}
        \centering
        \includegraphics[width=\linewidth]{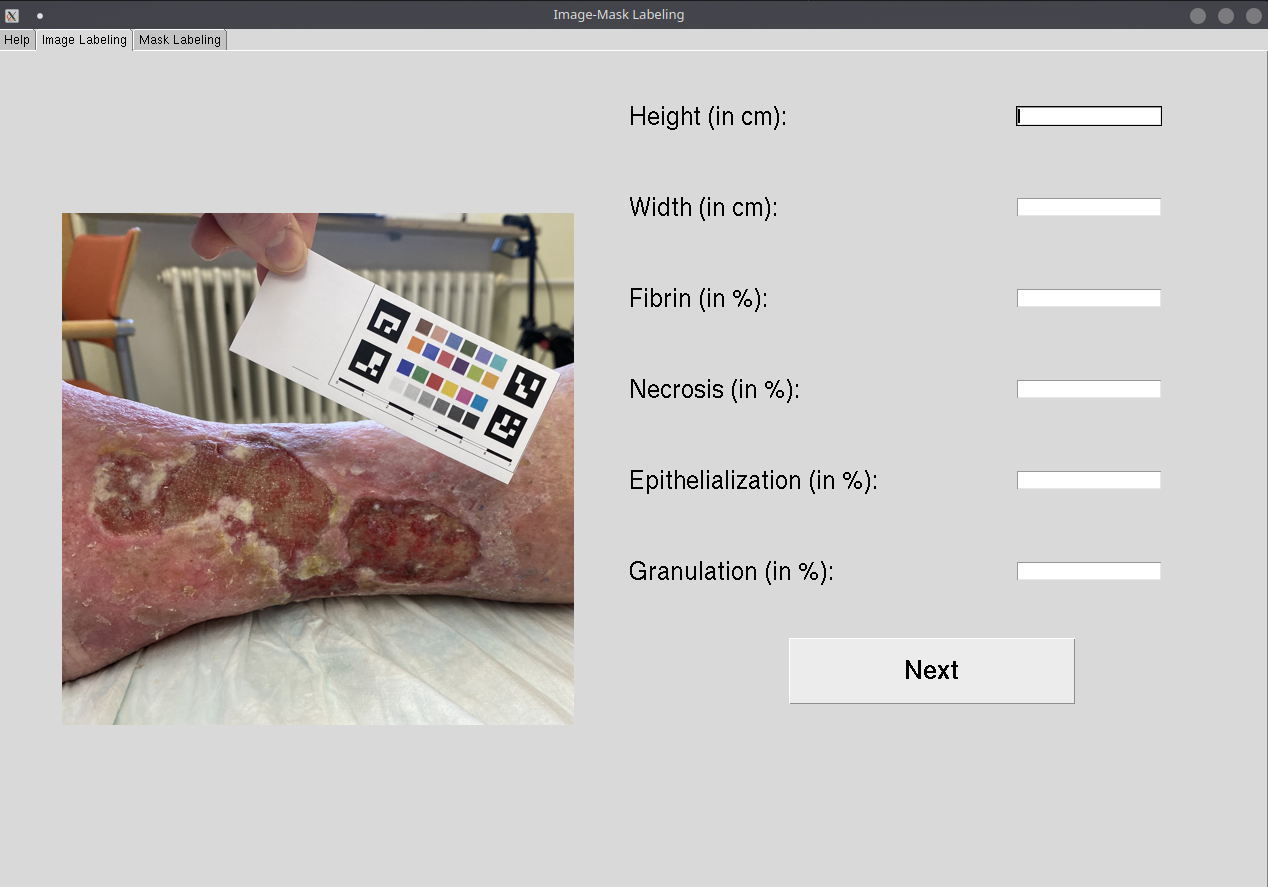}
        \caption{Wound details (incl. size)}
        \label{fig:tool:imageDetails}
    \end{subfigure}%
    \hfill
    \begin{subfigure}{.48\textwidth}
        \centering
        \includegraphics[width=\linewidth]{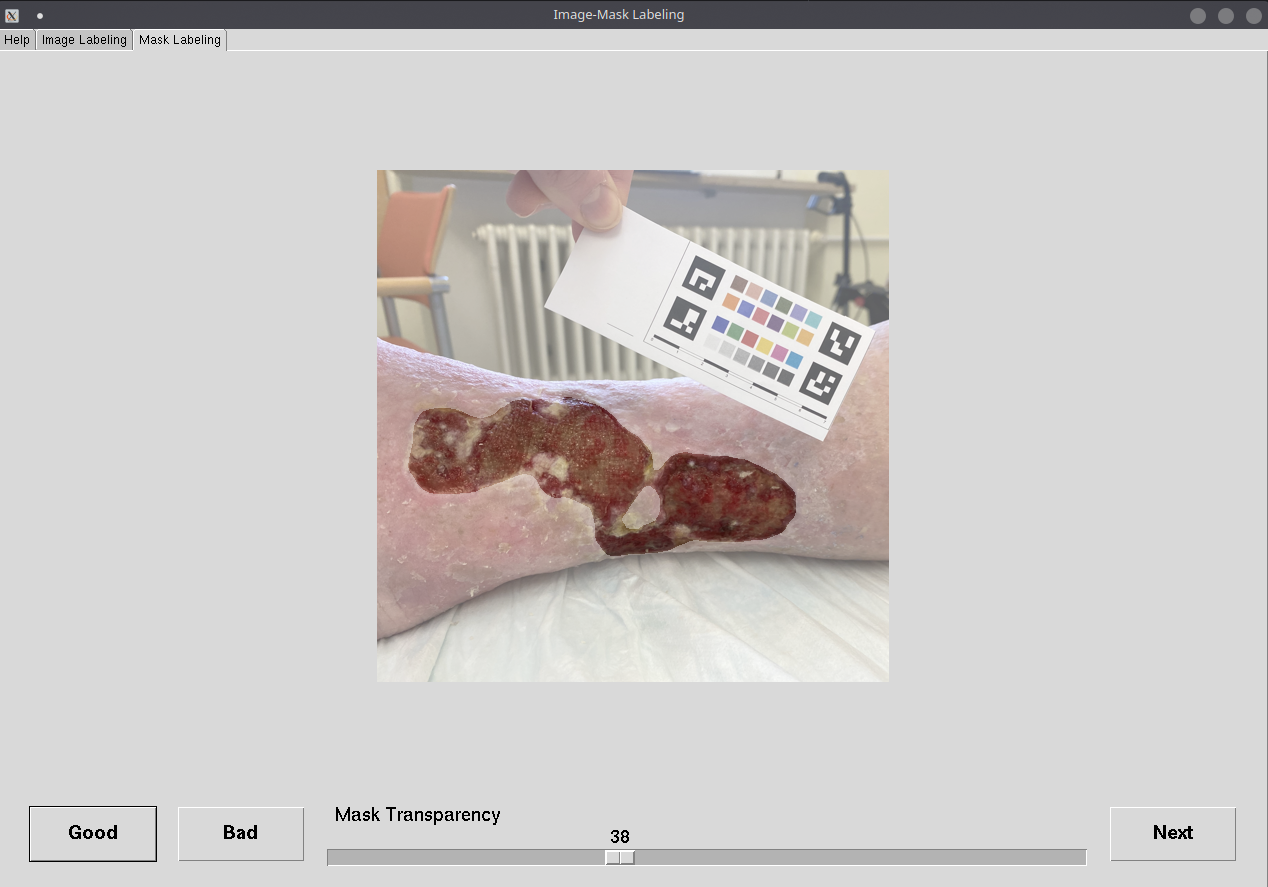}
        \caption{Mask assessment}
        \label{fig:tool:masks}
    \end{subfigure}%
    \\
    \begin{subfigure}{.48\textwidth}
        \centering
        \includegraphics[width=\linewidth]{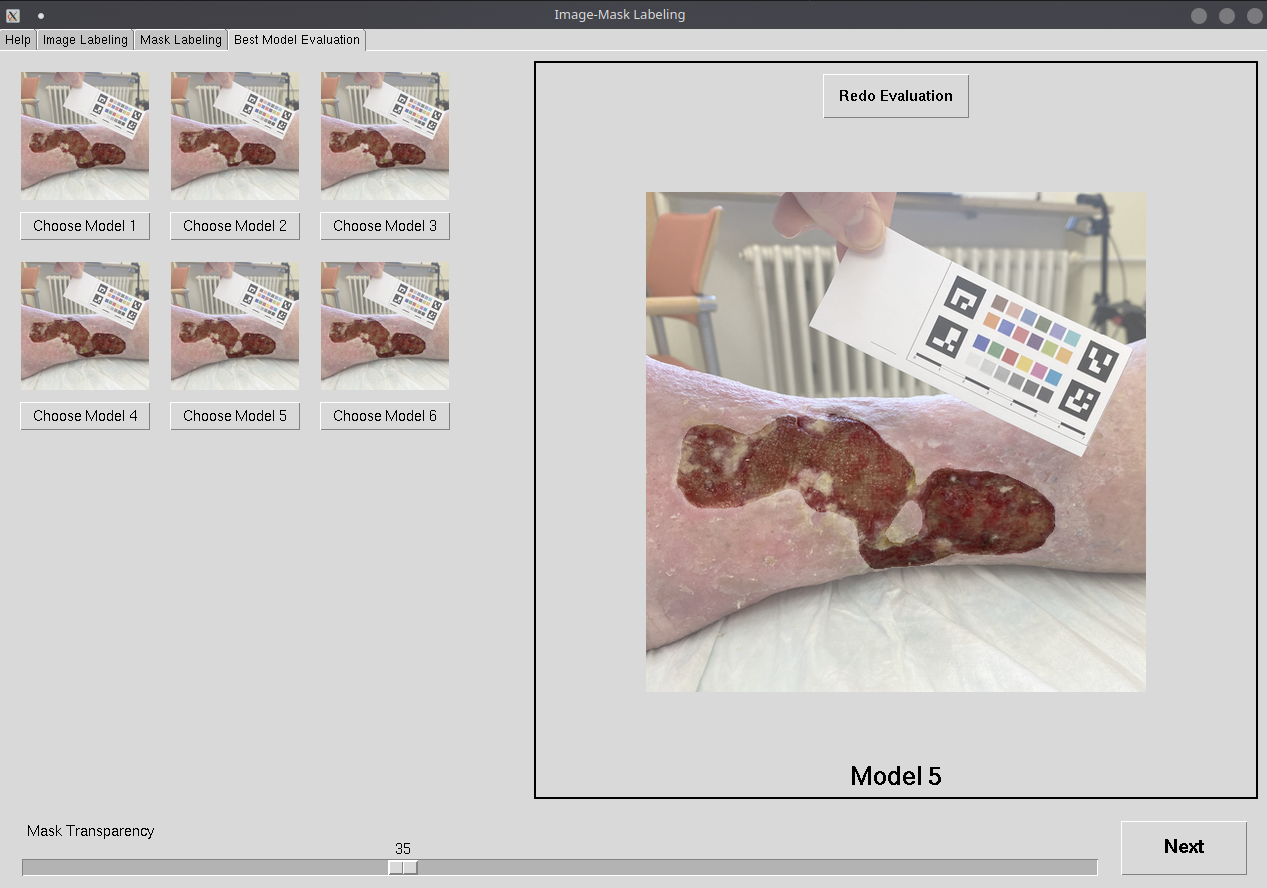}
        \caption{Best mask selection}
        \label{fig:tool:bestMask}
    \end{subfigure}
    \caption{Mask Quality Assessment and Image Annotation Tool}
    \label{fig:tool:overall}
\end{figure}

\bibliographystylesupp{splncs04}

\end{document}